\documentclass[english]{article}

\usepackage{geometry}
\usepackage[T1]{fontenc}
\usepackage[latin9]{inputenc}
\usepackage{bm}
\usepackage{amsmath}
\usepackage{amssymb} 
\usepackage[unicode=true,
 bookmarks=false, 
 breaklinks=false,pdfborder={0 0 1},colorlinks=false]
 {hyperref}
\hypersetup{
 colorlinks,citecolor=blue,filecolor=blue,linkcolor=blue,urlcolor=blue}
\usepackage{xcolor,colortbl}
\definecolor{Gray}{gray}{0.85}
\usepackage{enumitem}
\makeatletter

\usepackage{amsthm}
\usepackage{cite}  
\usepackage{comment}
\usepackage[sort,compress]{natbib}
\usepackage{booktabs,mathtools}
\usepackage{graphicx}
\usepackage[linesnumbered,ruled,vlined]{algorithm2e}
\usepackage{algorithmic}
\usepackage{multirow}
\usepackage{dsfont}
\usepackage{color}
\usepackage{float}
\usepackage{hhline} 

\definecolor{yxc}{RGB}{255,0,0}
\definecolor{yjc}{RGB}{225,0,100}
\definecolor{ytw}{RGB}{255,69,0}
\definecolor{gen}{RGB}{0,0,200}
\definecolor{lxs}{RGB}{138,43,226}
\definecolor{hew}{RGB}{0,47,167}

\definecolor{own_pink}{RGB}{217,25,169}
\definecolor{own_blue}{RGB}{0,100,223}

\definecolor{own_pink}{RGB}{217,25,169}
\definecolor{own_blue}{RGB}{0,100,223}

\allowdisplaybreaks

\DeclareMathOperator{\ind}{\mathds{1}}  

\newcommand{\defn}{\coloneqq}

\newcommand{\rob}{{\mathsf{rob}}}
\newcommand{\TV}{{\mathsf{TV}}}

\newcommand{\infs}{}

\newcommand{\no}{0} 
\newcommand{\ror}{\sigma} 
\newcommand{\unb}{\cU} 

\newcommand{\DRVI}{{\sf DRVI}\xspace}

\newcommand{\that}{\widehat{\cT}}

\newcommand{\pmin}{P}
\newcommand{\pmhat}{\widehat{P}}
\newcommand{\Pv}{\underline{P}}
\newcommand{\Phatv}{\underline{\widehat{P}}}

\newcommand{\e}{{e}}

\newcommand{\tvp}{\mathcal{C}}
\newcommand{\chip}{\mathcal{F}}

\newcommand{\cA}{\mathcal{A}}

\newcommand{\cC}{\mathcal{C}}
\newcommand{\cD}{\mathcal{D}}

\newcommand{\cF}{\mathcal{F}}
\newcommand{\cG}{\mathcal{G}}

\newcommand{\cM}{\mathcal{M}}
\newcommand{\cN}{N}

\newcommand{\cP}{\mathcal{P}}

\newcommand{\cS}{{\mathcal{S}}}
\newcommand{\cT}{{\mathcal{T}}}
\newcommand{\cU}{\mathcal{U}}

\newcommand{\mymid}{\,|\,} 

\usepackage{scalerel,stackengine}
\stackMath
\newcommand\reallywidehat[1]{%
\savestack{\tmpbox}{\stretchto{%
  \scaleto{%
    \scalerel*[\widthof{\ensuremath{#1}}]{\kern-.6pt\bigwedge\kern-.6pt}%
    {\rule[-\textheight/2]{1ex}{\textheight}}
  }{\textheight}%
}{0.5ex}}%
\stackon[1pt]{#1}{\tmpbox}%
}
\newcommand\reallywidecheck[1]{%
\savestack{\tmpbox}{\stretchto{%
  \scaleto{
    \scalerel*[\widthof{\ensuremath{#1}}]{\kern-.6pt\bigwedge\kern-.6pt}%
    {\rule[-\textheight/2]{1ex}{\textheight}}
  }{\textheight}%
}{0.5ex}}%
\stackon[1pt]{#1}{\scalebox{-1}{\tmpbox}}%
}

\usepackage{tikz}
\usepackage{array}
\usetikzlibrary{calc}

\title{The Curious Price of Distributional Robustness \\in Reinforcement Learning with a Generative Model}
 \author{
 	Laixi Shi\thanks{Department of Electrical and Computer
Engineering, Johns Hopkins University, MD, USA.}\\
 	JHU 
 	\and
	Gen Li\thanks{Department of Statistics, The Chinese University of Hong Kong, Hong Kong.} \\ 
	CUHK \\
	\and
	Yuting Wei\thanks{Department of Statistics and Data Science, Wharton School, University of Pennsylvania, Philadelphia, PA 19104, USA.} \\
 UPenn  \\
	\and
	Yuxin Chen\footnotemark[3]  \\
	UPenn\\
	\and
	Matthieu Geist\thanks{Earth Species Project.}\\
	ESP \\
	\and
 	Yuejie Chi\thanks{Department of Statistics and Data Science, Yale University, CT, USA.} \\ 	 
  	Yale
 	}

\date{May 2023; Revised September 2025}

\begin{document}

\theoremstyle{plain} \newtheorem{lemma}{\textbf{Lemma}}
\newtheorem{proposition}{\textbf{Proposition}}
\newtheorem{theorem}{\textbf{Theorem}}
\newtheorem{assumption}{Assumption}
\newtheorem{corollary}{Corollary}
\theoremstyle{remark}\newtheorem{remark}{\textbf{Remark}}

\maketitle

\begin{abstract}
This paper investigates model robustness in reinforcement learning (RL) to reduce the sim-to-real gap in practice. We adopt the framework of distributionally robust Markov decision processes (RMDPs), aimed at learning a  policy that optimizes the worst-case performance when the deployed environment falls within a prescribed uncertainty set around the nominal MDP. 
Despite recent efforts, the sample complexity of RMDPs remained mostly unsettled regardless of the uncertainty set in use. It was unclear if distributional robustness bears any statistical consequences when benchmarked against standard RL.
Assuming access to a generative model that draws samples based on the nominal MDP, we provide a near-optimal characterization of the sample complexity of RMDPs when the uncertainty set is specified via either the total variation (TV) distance or $\chi^2$ divergence. The algorithm studied here is a model-based method called {\em distributionally robust value iteration}, which is shown to be near-optimal for the full range of uncertainty levels. Somewhat surprisingly, our results uncover that RMDPs are not necessarily easier or harder to learn than standard MDPs. The statistical consequence incurred by the robustness requirement depends heavily on the size and shape of the uncertainty set: in the case w.r.t.~the TV distance, the minimax sample complexity of RMDPs is always smaller than that of standard MDPs; in the case w.r.t.~the $\chi^2$ divergence, the sample complexity of RMDPs far exceeds the standard MDP counterpart.
\end{abstract}

\noindent \textbf{Keywords:} distributionally robust RL, robust Markov decision processes, sample complexity,   distributionally robust value iteration, model-based RL

\allowdisplaybreaks

\setcounter{tocdepth}{2}
\tableofcontents

\section{Introduction}

Reinforcement learning (RL) strives to learn desirable sequential decisions based on trial-and-error interactions with an unknown environment. As a fast-growing subfield of artificial intelligence, it has achieved remarkable success in a variety of applications, such as networked systems \citep{qu2022scalable}, trading \citep{park2015adaptive}, operations research \citep{de2003greedy,pan2023adjustable,zhao2021learning}, large language model alignment \citep{koubaa2023gpt,ziegler2019fine}, healthcare \citep{liu2019deep,fatemi2021medical}, robotics and control \citep{kober2013reinforcement,mnih2013playing}. Due to the unprecedented dimensionality of the state-action
space, the issue of data efficiency inevitably lies at the core of modern RL practice. A large portion of recent efforts in RL has been directed towards designing sample-efficient algorithms and understanding the fundamental statistical bottleneck for a diverse range of RL scenarios.

While standard RL has been heavily investigated with enriched understanding, its use can be significantly hampered in practice due to the sim-to-real gap or environment uncertainty \citep{bertsimas2019adaptive}; 
for instance, a policy learned in an ideal, nominal environment might fail catastrophically when the deployed environment is subject to small changes in task objectives or adversarial perturbations \citep{zhang2020robust,klopp2017robust,mahmood2018benchmarking}. 
This challenge is particularly pronounced in many classical operations research problems where decision-making under uncertainty is paramount. Consequently, in addition to maximizing the long-term cumulative reward, robustness emerges as another critical goal for RL. This is especially true in high-stakes applications such as supply chain management (e.g., inventory control under uncertain demand \citep{federgruen1999combined}), revenue management (e.g., dynamic pricing with unknown market behaviors \citep{morales2012revenue,bitran2003overview}), and financial portfolio management \citep{delage2010distributionally}. 
Towards achieving this, distributionally robust RL \citep{iyengar2005robust,nilim2005robust,xu2012distributionally,bauerle2022distributionally}, which leverages insights from distributionally robust optimization and supervised learning \citep{rahimian2019distributionally,gao2020finite,bertsimas2018data,duchi2021learning,blanchet2019quantifying,chen2019distributionally,lam2019recovering}, becomes a natural yet versatile framework;   
 the aim is to learn a policy that performs well even when the deployed environment deviates from the nominal one in the face of environment uncertainty.

Solid theoretical guarantees are essential for deploying distributionally robust RL in high-stakes, real-world applications, where rigorous performance assurance is critical. Towards this, in this paper, we pursue fundamental understanding about whether, and how, the choice of distributional robustness bears statistical implications in learning a desirable policy, 
through the lens of sample complexity.
More concretely, imagine that one has access to a generative model (also called a simulator) that draws samples from a Markov decision processes (MDP) with a nominal transition kernel \citep{kearns1999finite}. Standard RL aims to learn the optimal policy tailored to the nominal kernel, for which the minimax sample complexity limit has been fully settled \citep{azar2013minimax,agarwal2020model,li2020breaking}. 
In contrast, distributionally robust RL seeks to learn a more {\em robust} policy using the same set of samples, 
with the aim of optimizing the worst-case performance when the transition kernel is arbitrarily chosen from some {\em prescribed} uncertainty set around the nominal kernel;  this setting is frequently referred to as robust MDPs (RMDPs).\footnote{While it is straightforward to incorporate additional uncertainty of the reward in our framework, we do not consider it here for simplicity, since the key challenge is to deal with the uncertainty of the transition kernel.} Clearly, the RMDP framework helps ensure that the performance of the learned policy does not fail catastrophically as long as the sim-to-real gap is not overly large. It is then natural to wonder how the robustness consideration impacts data efficiency: 
{\em is there a statistical premium that one needs to pay in quest of additional robustness? }

Compared with standard MDPs, the class of RMDPs encapsulates richer models, 
given that one is allowed to prescribe the shape and size of the uncertainty set.  Oftentimes, the uncertainty set is hand-picked as a small ball surrounding the nominal kernel, 
with the size and shape of the ball specified by some distance-like metric $\rho$ between probability distributions and some uncertainty level $\ror$.
To ensure tractability of solving RMDPs, the uncertainty set is often selected to obey certain structures.  
For instance, a number of prior works assumed that the uncertainty set can be decomposed as a product of independent uncertainty subsets over each state or state-action pair \citep{zhou2021finite,wiesemann2013robust}, dubbed as the $s$- and $(s,a)$-rectangularity, respectively. The current paper adopts the second choice by assuming $(s,a)$-rectangularity for the uncertainty set. An additional challenge with RMDPs arises from distribution shift, where the transition kernel drawn from the uncertainty set can be different from the nominal kernel. 
This challenge leads to complicated nonlinearity and nested optimization in the problem structure not present in standard MDPs.

\subsection{Prior art and open questions}\label{sec:prior-works}

\newcommand{\topsepremove}{\aboverulesep = 0mm \belowrulesep = 0mm} \topsepremove

	\begin{table}[t]
	\begin{center}
\begin{tabular}{c|c|c|c}
\hline
\toprule
	\multirow{3}{*}{Result type} & \multirow{3}{*}{Reference}  &  \multicolumn{2}{c}{Sample complexity}  \tabularnewline
	\cline{3-4}
	& & \multirow{2}{*}{$ 0 < \sigma \lesssim 1-\gamma $} &  \multirow{2}{*}{$1-\gamma\lesssim \sigma <1$}   \tabularnewline
	& & &  \tabularnewline
\hline
\toprule
  \multirow{7}{*}{Upper bound} & \multirow{2}{*}{\citet{yang2021towards}  \vphantom{$\frac{1^{7}}{1^{7^{7}}}$} } & \multicolumn{2}{c}{\multirow{2}{*}{$\frac{S^2A }{  \sigma^2  (1-\gamma)^4 \varepsilon^2} $ }}   \tabularnewline
&  &  \multicolumn{2}{c}{}\tabularnewline
\cline{2-4} &  \multirow{2}{*}{\citet{panaganti2021sample} \vphantom{$\frac{1^{7}}{1^{7^{7}}}$}} & \multicolumn{2}{c}{\multirow{2}{*}{ $\frac{S^2A }{  (1-\gamma)^4 \varepsilon^2} $
} } \tabularnewline
&  &  \multicolumn{2}{c}{}\tabularnewline

\cline{2-4}
& \multirow{2}{*}{  {\bf This paper \vphantom{$\frac{1^{7}}{1^{7^{7}}}$} } }   &  \multirow{2}{*}{$\frac{SA }{  (1-\gamma)^3 \varepsilon^2}$} &   \multirow{2}{*}{  $\frac{SA }{  (1-\gamma)^2 \ror \varepsilon^2}$}   \tabularnewline 
   &    &  & \tabularnewline 
\hline
\toprule
 \multirow{4}{*}{Lower bound} & \multirow{2}{*}{\citet{yang2021towards}   \vphantom{$\frac{1^{7}}{1^{7^{7}}}$} }  &  \multirow{2}{*}{$\frac{SA }{  (1-\gamma)^3 \varepsilon^2}$} &  \multirow{2}{*}{$\frac{SA(1-\gamma)}{ \ror^4 \varepsilon^2}$}   \tabularnewline
    &    &   \tabularnewline 
\cline{2-4}
	& \multirow{2}{*}{  {\bf This paper \vphantom{$\frac{1^{7}}{1^{7^{7}}}$} } }   &  \multirow{2}{*}{$\frac{SA }{  (1-\gamma)^3 \varepsilon^2}$} &   \multirow{2}{*}{  $\frac{SA }{  (1-\gamma)^2 \ror \varepsilon^2}$}   \tabularnewline 
   &    &   \tabularnewline 
\hline 
\toprule
\end{tabular}
	\end{center}
	\caption{Comparisons between our results and prior art for finding an $\varepsilon$-optimal robust policy in infinite-horizon RMDPs with a generative model, where
	 the uncertainty set is measured w.r.t.~the TV distance. Here, $S$, $A$, $\gamma$, and $\ror \in (0, 1)$ are the state space size, the action space size, the discount factor, and the uncertainty level, respectively, and all logarithmic factors are omitted in the table. Our results provide the first matching upper and lower bounds (up to log factors), 
	improving upon all prior results.  \label{tab:tv}  } 
\end{table}

\begin{table}[t]
	\begin{center}
\begin{tabular}{c|c|c|c}
\hline
\toprule
	\multirow{3}{*}{Result type} & \multirow{3}{*}{Reference}  &  \multicolumn{2}{c}{Sample complexity}  \tabularnewline
	\cline{3-4}
	& & \multirow{2}{*}{$ 0 < \sigma \lesssim 1 -\gamma $} &  \multirow{2}{*}{$ \sigma \gtrsim 1-\gamma$}   \tabularnewline
	& & &  \tabularnewline
\hline
\toprule
  \multirow{6}{*}{Upper bound}&   \multirow{2}{*}{ \citet{panaganti2021sample}  \vphantom{$\frac{1^{7}}{1^{7^{7}}}$} } & \multicolumn{2}{c}{\multirow{2}{*}{$\frac{S^2A (1+\sigma)}{  (1-\gamma)^4 \varepsilon^2} $
} } \tabularnewline
&  &  \multicolumn{2}{c}{} \tabularnewline 
\cline{2-4} & \multirow{2}{*}{ \citet{yang2021towards}   \vphantom{$\frac{1^{7}}{1^{7^{7}}}$} }  & \multicolumn{2}{c}{ \multirow{2}{*}{$\frac{S^2A (1+\sigma)^2}{  (\sqrt{1+\sigma} - 1)^2  (1-\gamma)^4 \varepsilon^2}$ }  } \tabularnewline
&  &  \multicolumn{2}{c}{} \tabularnewline 
\cline{2-4}
&  \multirow{2}{*}{ {\bf This paper}  \vphantom{$\frac{1^{7}}{1^{7^{7}}}$} } &    \multicolumn{2}{c}{ \multirow{2}{*}{ $\frac{SA}{(1-\gamma)^3 \varepsilon^2} \left(1+\frac{\max\{\sqrt{\ror} , \ror\} }{1-\gamma} \right)$}  }  \tabularnewline
&  &    \multicolumn{2}{c}{} \tabularnewline 
\hline
\toprule
 \multirow{4}{*}{Lower bound} &  \multirow{2}{*}{\citet{yang2021towards} \vphantom{$\frac{1^{7}}{1^{7^{7}}}$ } }  & \multirow{2}{*}{$\frac{SA}{(1-\gamma)^3 \varepsilon^2}$ }   & \multirow{2}{*}{$ \frac{SA}{(1-\gamma)^2\sigma \varepsilon^2}$}   \tabularnewline
&  &     \tabularnewline 
 \cline{2-4} & \multirow{2}{*}{ {\bf This paper}  \vphantom{$\frac{1^{7}}{1^{7^{7}}}$} } & \multirow{2}{*}{$\frac{SA}{(1-\gamma)^3\varepsilon^2}$}  & \multirow{2}{*}{$\frac{SA \ror }{(1-\gamma)^4\varepsilon^2}$}   \tabularnewline
&  &    \tabularnewline 
\hline 
\toprule
\end{tabular}

	\end{center}
	\caption{Comparisons between our results and prior art on finding an $\varepsilon$-optimal robust policy in infinite-horizon RMDPs with a generative model, where
		 the uncertainty set is measured w.r.t.~the $\chi^2$ divergence. Here, $S$, $A$, $\gamma$, and $\ror \in (0, \infty)$ are the state space size, the action space size, the discount factor, and the uncertainty level, respectively, and all logarithmic factors are omitted in the table.  
		Our results are tight (up to log factors) in all ranges of $\sigma$, except when $(1-\gamma)^2 \lesssim \sigma \lesssim 1$ by no more than a factor of $1/\sqrt{1-\gamma}$.  } 
\label{tab:chi2}

\end{table}

In this paper, we focus attention on RMDPs in the context of $\gamma$-discounted infinite-horizon setting, 
assuming access to a generative model. The uncertainty set considered herein is specified using one of the $f$-divergence metrics: the total variation (TV) distance and the $\chi^2$ divergence. These two choices are motivated by their practical appeals: easy to implement, and already adopted by empirical RL \citep{lee2021optidice,pan2023adjustable}.  

A popular learning approach is model-based, which first estimates the nominal transition kernel using a plug-in estimator based on the collected samples, and then runs a planning algorithm (e.g., a robust variant of value iteration) on top of the estimated kernel. Despite the surge of recent activities, however, existing statistical guarantees for the above paradigm remained highly inadequate, 
as we shall elaborate on momentarily (see Table~\ref{tab:tv} and Table~\ref{tab:chi2} respectively for a summary of existing results). For concreteness, let $S$ be the size of the state space,  $A$ the size of the action space, $\gamma$ the discount factor (so that the effective horizon is $\frac{1}{1-\gamma}$), and $\ror$ the uncertainty level. We are interested in how the sample complexity --- the number of samples needed for an algorithm to output a policy whose robust value function (the worst-case value over all the transition kernels in the uncertainty set) is at most $\varepsilon$ away from the optimal robust one --- scales with all these salient problem parameters.

\begin{itemize}
	\item {\em Large gaps between existing upper and lower bounds.} There remained large gaps between the sample complexity upper and lower bounds established in prior literature, regardless of the divergence metric in use. Specifically, considering the cases using either TV distance or $\chi^2$ divergence, the state-of-the-art upper bounds \citep{panaganti2021sample} scales quadratically with the size $S$ of the state space, while the lower bound \citep{yang2021towards} exhibits only linear scaling with $S$. Moreover, in the $\chi^2$ divergence case, the state-of-the-art upper bound grows linearly with the uncertainty level $\sigma$ when $\ror \gtrsim 1$,\footnote{Let $\mathcal{X} \coloneqq \big( S, A, \frac{1}{1-\gamma}, \sigma, \frac{1}{\varepsilon}, \frac{1}{\delta} \big)$.  
 The notation $f(\mathcal{X}) = O(g(\mathcal{X}))$ or $f(\mathcal{X}) \lesssim g(\mathcal{X})$ indicates that there exists a universal constant $C_1>0$ such that $f\leq C_1 g$,  the notation $f(\mathcal{X}) \gtrsim g(\mathcal{X}) $ indicates that $g(\mathcal{X}) = O(f(\mathcal{X}))$, and the notation $f(\mathcal{X})\asymp g(\mathcal{X}) $ indicates that  $f(\mathcal{X}) \lesssim g(\mathcal{X})$ and  $f(\mathcal{X}) \gtrsim g(\mathcal{X})$ hold simultaneously. Additionally, the notation $\widetilde{O}(\cdot)$ is defined in the same way as ${O}(\cdot)$ except that it hides logarithmic factors. } while the lower bound \citep{yang2021towards} is inversely proportional to $\ror$. These lead to unbounded gaps between the upper and lower bounds as $\ror$ grows. {\em Can we hope to close these gaps for RMDPs?}

\item {\em Benchmarking with standard MDPs.} Perhaps a more pressing issue is that, past works failed to provide an affirmative answer 
	regarding how to benchmark the sample complexity of RMDPs with that of standard MDPs regardless of the chosen shape (determined by $\rho$) or size (determined by $\sigma$) of the uncertainty set, given the large unresolved gaps mentioned above.
		Specifically, existing sample complexity upper (resp.~lower) bounds are all larger (resp.~smaller) than the sample size requirement for standard MDPs. 
		As a consequence, it remains mostly unclear  {\em whether learning RMDPs is harder or easier than learning standard MDPs.}

\end{itemize}

\subsection{Main contributions}

\begin{figure}[H]
	\centering
	\begin{tabular}{cc}
	\includegraphics[width=0.48\linewidth]{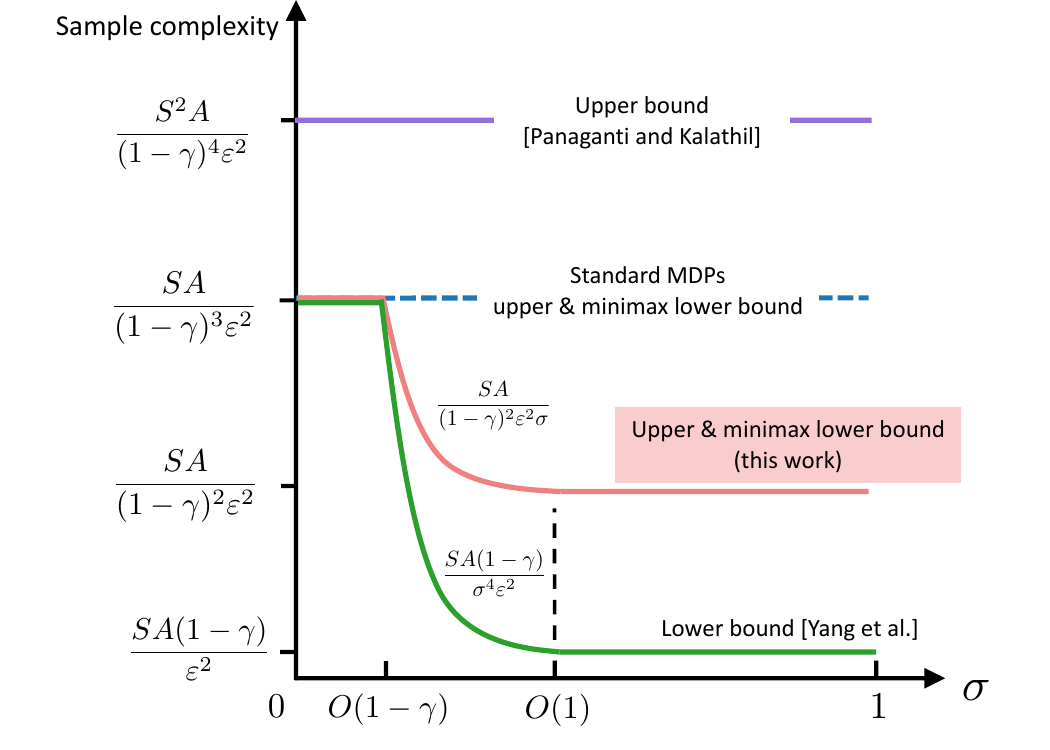} & \includegraphics[width=0.48\linewidth]{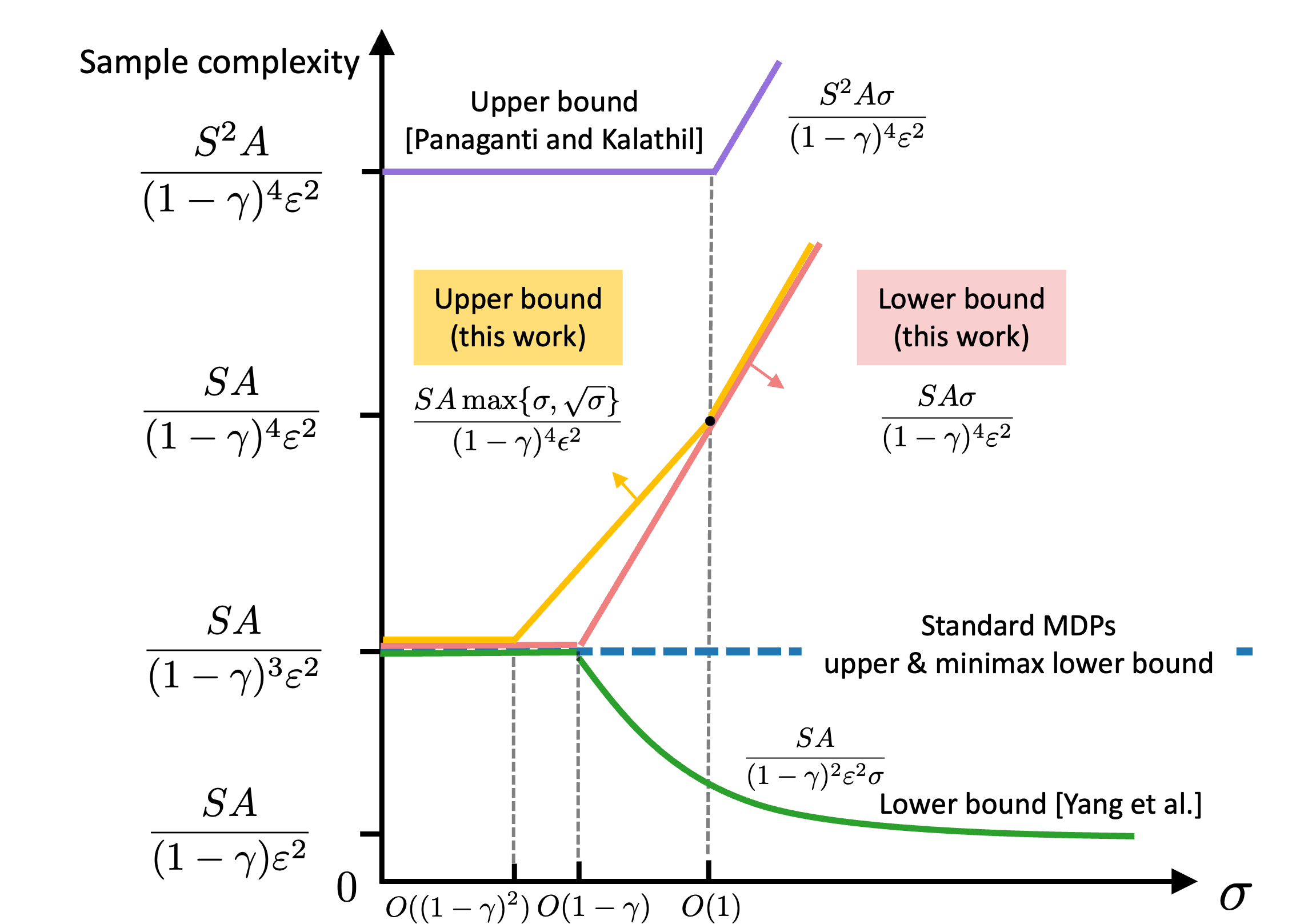} \\
	\hspace{3em} \text{(a) TV distance} & \hspace{2em} \text{(b) $\chi^2$ divergence}
	\end{tabular}
	\caption{Illustrations of the obtained sample complexity upper and lower bounds for  learning RMDPs with comparisons to state-of-the-art and the sample complexity of standard MDPs, where the uncertainty set is specified using the TV distance (a) and the $\chi^2$ divergence (b).  }  
		\label{fig:comopare-to-prior}
	\end{figure}

To address the aforementioned questions, 
this paper develops strengthened sample complexity upper bounds on learning RMDPs with the TV distance and $\chi^2$ divergence in the infinite-horizon setting,
using a model-based approach called distributionally robust value iteration (\DRVI). Improved minimax lower bounds are also developed to help gauge the tightness of our upper bounds  
and enable benchmarking with standard MDPs.
The novel analysis framework developed herein leads to new insights into the interplay between the geometry of uncertainty sets and statistical hardness.
	
\paragraph{Sample complexity of RMDPs under the TV distance.} We summarize our results and compare them with past works in Table~\ref{tab:tv}; see Figure~\ref{fig:comopare-to-prior}(a) for a graphical illustration.

	\begin{itemize}
		\item {\bf Minimax-optimal sample complexity.} We prove that \DRVI reaches $\varepsilon$ accuracy as soon as the sample complexity is on the order of
		$$\widetilde{O}\left(\frac{SA}{(1-\gamma)^2\varepsilon^2} \min \left\{ \frac{1}{1-\gamma}, \frac{1}{\ror}\right\}\right)$$ 
for all $\sigma \in (0,1)$, assuming that $\varepsilon$ is small enough. In addition, a matching minimax lower bound (modulo some logarithmic factor) is established to guarantee the tightness of the upper bound. To the best of our knowledge, this is the {\em first} minimax-optimal sample complexity for RMDPs, which was previously unavailable regardless of the divergence metric and uncertainty level in use and is over the full range of the uncertainty level.
			
		\item {\bf RMDPs are easier to learn than standard MDPs under the TV distance.} Given the sample complexity $\widetilde{O}\left(\frac{SA}{(1-\gamma)^3\varepsilon^2} \right)$ of standard MDPs \citep{li2020breaking}, it can be seen that learning RMDPs under the TV distance is never harder than learning standard MDPs; 
			more concretely, the sample complexity for RMDPs matches that of standard MDPs when $\sigma \lesssim 1-\gamma$, and becomes smaller by a factor of $\sigma/(1-\gamma)$ when $  1-\gamma \lesssim \sigma <1$. 
			Therefore, in this case, distributional robustness comes almost for free, 
			given that we do not need to collect more samples.

	\end{itemize}

\paragraph{ Sample complexity of RMDPs under the $\chi^2$ divergence.} We summarize our results and provide comparisons with prior works in Table~\ref{tab:chi2}; see Figure~\ref{fig:comopare-to-prior}(b) for an illustration.

	\begin{itemize}
		\item {\bf Near-optimal sample complexity.} We demonstrate that \DRVI yields $\varepsilon$ accuracy as soon as the sample complexity is on the order of
		$$\widetilde{O}\left(\frac{SA}{(1-\gamma)^3 \varepsilon^2} \left(1+\frac{\max\{\sqrt{\ror} , \ror\}}{1-\gamma} \right)\right)$$ 
for all $\sigma \in (0,\infty)$. This result is the first sample complexity in this setting that scales linearly in the size  $S$ of the state space and is strictly tighter than existing bounds; 
			in other words, our theory breaks the quadratic scaling bottleneck that was present in prior works \citep{panaganti2021sample,yang2021towards}. We have also developed a nearly matched lower bound that is optimized by leveraging the geometry of the uncertainty set under different ranges of $\ror$ on the order of
			$$  \widetilde{\Omega} \left(\frac{SA}{(1-\gamma)^3 \varepsilon^2} \left(1+\frac{ \ror }{1-\gamma} \right)\right) . $$
			By comparing the upper and lower bounds, our theory is tight in  all ranges of $\sigma$, except  by at most a factor of $\sqrt{\frac{1}{1-\gamma}}$ when $(1-\gamma)^2 \lesssim \sigma \lesssim 1$. This verifies the near-optimality of \DRVI's sample complexity across a broad range of uncertainty levels, and significantly improves upon prior results (as there exists an unbounded gap between prior upper and lower bounds as $\sigma \rightarrow \infty$).

		\item  {\bf RMDPs are harder to learn than standard MDPs under the $\chi^2$ divergence.} In contrast to the case with the TV distance, our results show that RMDPs under the $\chi^2$ divergence are never easier to learn than standard MDPs. When $\sigma$ is relatively small ($ \sigma \lesssim (1-\gamma)^2$), the sample size requirement of RMDPs shown by the matched upper and lower bound is 
			on the order of $\frac{SA}{(1-\gamma)^3 \varepsilon^2} $ (up to log factor), 
			matching the sample complexity of standard MDPs. When the uncertainty level $ \sigma \gtrsim 1-\gamma$,  the lower bound surpasses the sample complexity of standard MDPs, highlighting the difficulty of solving RMDPs under the $\chi^2$ divergence compared to standard MDPs. Notably, the upper and lower bounds match each other on the order of $\frac{SA\sigma}{(1-\gamma)^4}$ when $ \sigma \gtrsim 1$, exhibiting linear growth with respect to $\sigma$.

	\end{itemize}

In sum, our sample complexity bounds not only strengthen the prior art in the development of both upper and lower bounds, but also unveil that the additional robustness consideration affects the sample complexity in a somewhat surprising manner. As it turns out, RMDPs are not necessarily harder nor easier to learn than standard MDPs; 
the conclusion is far more nuanced and highly dependent on both the size and shape of the uncertainty set: difference choices of the uncertainty set can lead to dramatically different sample size requirements.  
This constitutes a curious phenomenon that has not been elucidated in prior analyses, providing insights to the field of implementing distributionally robust formulations in RL.

\paragraph{Technical novelty.} 
Our upper bound analysis is driven by several key technical innovations.
\begin{itemize}
	\item {\em Tailored error controls cognizant to the uncertainty levels.} Due to the distributional robustness formulation, it necessitates the management of nonlinear interactions with the worst-case transition kernel induced by any fixed policy $\pi$, which makes existing techniques for standard MDPs vacuous.  We provide new error decompositions that are more salient to the uncertainty level and tighter characterizations on the dynamic range (i.e., span) of the robust value functions, both of which are crucial to improved sample complexities.
	 
\item {\em Overcome the quadratic scaling with respect to $S$.} To overcome the quadratic scaling bottleneck with respect to the state space size $S$, we decouple the statistical dependency across the iterates of the robust value iteration using tailored leave-one-out arguments \citep{agarwal2020model,li2024settling,shi2024distributionally} that have not been
introduced to this setting previously. 
\end{itemize} 
Turning to the lower bound, we develop new hard instances that differ from those used for standard MDPs \citep{gheshlaghi2013minimax,li2024settling}, guided by the following innovations.
\begin{itemize}
\item {\em Asymmetric structures of reward allocation.} In contrast to standard MDPs, the bootstrapping in robust MDPs is asymmetric over all states, since the worst-case transition probability puts more weights on the states with lower values. In response, we develop new hard instances by setting larger rewards on the state with action-invariant transitions to enable tighter lower bounds.  
\item {\em Construction of $\sigma$-dependent hard instances to address nonlinearity.} Unlike standard RL, which requires only a single hard instance (i.e., $\sigma = 0$), distributionally robust RL demands a tailored hard instance for each value of the uncertainty level $\sigma$. To achieve tight lower bounds, the nonlinearity from the robust formulation necessitates precise characterization of the worst-case transition distribution to maximize the gap from the nominal one at varying uncertainty levels. This yields a new series of hard instances, improving upon prior work \citep{yang2021towards}, which used a single instance for all uncertainty levels.
\end{itemize} 
 
\paragraph{Extension: offline RL with uniform coverage.}  Last but not least, we extend our analysis framework to accommodate a widely studied offline setting with uniform data coverage \citep{zhou2021finite,yang2021towards} in Section~\ref{sec:offline-main}. In particular, given a historical dataset with minimal coverage probability $\mu_{\min}$ over the state-action space (see Assumption~\ref{assumption-offline}), we provide sample complexity results for both cases with TV distance or $\chi^2$ divergence, where in effect the dependency with the size of the state-action space $SA$ is replaced by $1/\mu_{\min}$. The sample complexity upper bounds significantly improve upon  prior art \citep{yang2021towards} by a factor of $\frac{S}{(1-\gamma)^2}$ (resp.~at least $S(1+\ror)$) when the uncertainty set is measured by the TV distance (resp.~the $\chi^2$ divergence). 

\paragraph{Notation and paper organization.} 
Throughout this paper, we denote by $\Delta(\cS)$ the probability simplex over a set $\cS$ and $x = \big[x(s,a)\big]_{(s,a)\in\cS\times\cA}\in \mathbb{R}^{SA}$ (resp.~$x = \big[x(s)\big]_{s\in\cS}\in \mathbb{R}^{S}$) as any vector that constitutes certain values for each state-action pair (resp.~state). In addition, we denote by $x \circ y=\big[x(s) \cdot y(s) \big]_{s\in\cS}$ the 
 Hadamard product of any two vectors $x, y\in\mathbb{R}^S$. 

The remainder of this paper is structured as follows. Section~\ref{sec:formulation} presents the background about discounted infinite-horizon standard MDPs and formulates distributionally robust MDPs. In Section~\ref{sec:algorithm}, a model-based approach is introduced, tailored to both the TV distance and the $\chi^2$ divergence. 
Upper and lower bounds on the sample complexity are developed in Section~\ref{sec:theory}, covering both divergence metrics. Section~\ref{sec:analysis} provides an outline of our analysis, and Section~\ref{sec:numerical} provides numerical experiments to corroborate our theory. Section~\ref{sec:offline-main} further extends the findings to the offline RL setting with uniform data coverage.
We then summarize several additional related works in Section~\ref{sec:related} and conclude the main paper with further discussions in Section~\ref{sec:discussion}. 
The proof details are deferred to the appendix.

\section{Problem formulation}\label{sec:formulation}

In this section, we formulate distributionally robust Markov decision processes (RMDPs) in the discounted infinite-horizon setting, 
introduce the sampling mechanism, and describe our goal.

\paragraph{Standard MDPs.} To begin, we first introduce the standard Markov decision processes (MDPs), which facilitate the understanding of RMDPs. A discounted infinite-horizon MDP is represented by 
$\mathcal{M}= \big(\mathcal{S},\mathcal{A}, \gamma,P, r \big)$, where $\mathcal{S} =\{1,\cdots, S\}$ and $\mathcal{A}= \{1,\cdots,A\}$ are the finite state and action spaces, respectively, $\gamma\in[0,1)$ is the discounted factor, $P: \cS \times \cA \rightarrow \Delta (\cS) $ denotes the probability transition kernel, and $r: \cS \times \cA \rightarrow [0,1]$ is the immediate reward function which is assumed to be deterministic. A policy is denoted by $\pi:\mathcal{S} \rightarrow \Delta(\mathcal{A})$, which specifies the action selection probability over the action space in any state. When the policy is deterministic, we overload the notation and refer to $\pi(s)$ as the action selected by policy $\pi$ in state $s$. To characterize the cumulative reward, the value function $V^{\pi,P}$ for any policy $\pi$ under the transition kernel $P$ is defined by
\begin{align}
	\label{eq:def_V}
	\forall s\in\cS:\qquad V^{\pi,P}(s ) &\defn  \mathbb{E}_{\pi, P} 
	\left[  \sum_{t=0}^{\infty} \gamma^t r\big(s_{t}, a_t \big) \,\Big|\, s_{0}=s \right]  , 
\end{align}
where the expectation is taken over the randomness of the trajectory $\{s_t, a_t\}_{t=0}^\infty$ generated by executing policy $\pi$ under the transition kernel $P$, namely, $a_t\sim \pi(\cdot \mymid s_t)$ and $s_{t+1} \sim P(\cdot \mymid s_t, a_t )$ for all $t\geq 0$. Similarly, the Q-function $Q^{\pi,P}$ associated with any policy $\pi$ under the transition kernel $P$ is defined as
\begin{align} 
	\label{eq:def_Qh}
	\forall (s,a)\in \cS \times \cA:\qquad Q^{\pi, P}(s,a ) & \defn \mathbb{E}_{\pi, P} 
	\left[  \sum_{t=0}^{\infty} \gamma^t r\big(s_{t}, a_t \big) \,\Big|\, s_{0}=s, a_0=a \right],
	\end{align}
where the expectation is again taken over the randomness of the trajectory under policy $\pi$.

\paragraph{Distributionally robust MDPs.}
We now introduce the distributionally robust MDP (RMDP) tailored to the discounted infinite-horizon setting, 
denoted by $\cM_{\mathsf{rob}} = \{\cS,\cA, \gamma, \cU_\rho^{\ror}(P^\no), r\}$, where $\cS, \cA, \gamma, r$ are identical to those in the standard MDP. A key distinction from the standard MDP is that: rather than assuming a fixed transition kernel $P$, it allows the transition kernel to be chosen arbitrarily from a prescribed uncertainty set  $\cU_\rho^{\ror}(P^\no)$ centered around a {\em nominal} kernel $P^\no: \cS\times\cA \rightarrow \Delta(\cS)$, where the uncertainty set is specified using some distance metric $\rho$ of radius $\ror>0$. 
In particular, given the nominal transition kernel $P^\no$ and some uncertainty level $\ror$, the uncertainty set---with the divergence metric $\rho: \Delta(\cS) \times \Delta(\cS) \rightarrow \mathbb{R}^+$---is specified as
\begin{align}\label{eq:general-infinite-P}
	\cU_\rho^{\ror}(P^\no) &\defn \otimes \; \cU_\rho^{\ror}(P^{\no}_{s,a})\qquad \text{with}\quad
	\cU_\rho^{\ror}(P^\no_{s,a}) \defn \left\{ P_{s,a} \in \Delta (\cS): \rho \left(P_{s,a}, P^0_{s,a}\right) \leq \ror \right\},
\end{align}
where we denote a vector of the transition kernel $P$ or $P^{\no}$ at state-action pair $(s,a)$ respectively as
\begin{align}\label{eq:defn-P-sa}
	P_{s,a} \defn P(\cdot \mymid s,a) \in \mathbb{R}^{1\times S}, \qquad P_{s,a}^\no \defn P^\no(\cdot \mymid s,a) \in \mathbb{R}^{1\times S}.
\end{align}
In other words, the uncertainty is imposed in a decoupled manner for each state-action pair, obeying the so-called $(s,a)$-rectangularity \citep{zhou2021finite,wiesemann2013robust}.

In RMDPs, we are interested in the worst-case performance of a policy $\pi$ over all the possible transition kernels in the uncertainty set. This is measured by the {\em robust value function} $V^{\pi, \ror} $ and the {\em robust Q-function} $Q^{\pi,\ror}$ in $\cM_\rob$, defined respectively as
\begin{align} \label{eq:robust-value-def}
	\forall (s,a)\in \cS \times \cA:\quad  V^{\pi,\ror}(s) &\defn \inf_{P\in \unb_\rho^{\ror}(P^{\no})} V^{\pi,P} (s), \qquad Q^{\pi,\ror}(s,a) \defn \inf_{P\in \unb_\rho^{\ror}(P^{\no})} Q^{\pi,P}(s,a).
\end{align}

\paragraph{Optimal robust policy and robust Bellman operator.}
As a generalization of properties of standard MDPs, it is well-known that there exists at least one
deterministic policy that maximizes the robust value function (resp.~robust Q-function) simultaneously for all states (resp.~state-action pairs) \citep{iyengar2005robust,nilim2005robust,wiesemann2013robust}. Therefore, we denote the {\em optimal robust value function} (resp.~{\em optimal robust Q-function}) as $V^{\star,\ror}$ (resp.~$Q^{\star,\ror}$), and the optimal robust policy as $\pi^\star$, which satisfy
\begin{subequations} \label{eq:mdp-value-Q}
\begin{align}
	\forall s \in \cS: \quad &V^{\star,\ror}(s) \defn V^{\pi^\star,\ror}(s) = \max_\pi V^{\pi,\ror}(s), \\
	\forall (s,a) \in \cS \times \cA: \quad &Q^{\star,\ror}(s,a) \defn Q^{\pi^\star,\ror}(s,a) = \max_\pi Q^{\pi,\ror}(s,a).
\end{align}
\end{subequations}
A key machinery in RMDPs is a generalization of Bellman's optimality principle, 
encapsulated in the following {\em robust Bellman consistency equation} (resp.~{\em robust Bellman optimality equation}):
\begin{subequations}\label{eq:bellman-equ-infinite-all}
\begin{align}
	\forall (s,a)\in \cS\times \cA: \quad &Q^{\pi,\ror}(s,a) = r(s,a) + \gamma\inf_{ \cP \in \unb^{\ror}_\rho(P^{\no}_{s,a})} \cP V^{\pi,\ror}, \label{eq:bellman-equ-pi-infinite}\\
	\forall (s,a)\in \cS\times \cA: \quad &Q^{\star,\ror}(s,a) = r(s,a) + \gamma\inf_{\cP\in \unb^{\ror}_\rho(P^{\no}_{s,a})} \cP V^{\star,\ror}. \label{eq:bellman-equ-star-infinite}
\end{align}
\end{subequations}
The robust Bellman operator \citep{iyengar2005robust,nilim2005robust} is denoted by $\cT^\ror(\cdot): \mathbb{R}^{SA} \rightarrow \mathbb{R}^{SA}$ and defined as follows: 
\begin{align}\label{eq:robust_bellman}
	\forall (s,a)\in \cS\times \cA :\quad \cT^\ror(Q)(s,a) \defn r(s,a) + \gamma \inf_{ \cP \in \unb^{\ror}_\rho(P^{\no}_{s,a})} \cP V,  \quad \text{with}  \quad  V(s) \defn \max_a Q(s,a).
\end{align}
Given that $Q^{\star,\ror}$ is the unique fixed point of $\cT^\ror$,
one can recover the optimal robust value function and Q-function using a procedure termed {\em distributionally robust value iteration} (\DRVI). 
Generalizing the standard value iteration, \DRVI starts from some given initialization and recursively applies the robust Bellman operator until convergence. 
As has been shown previously, this procedure converges rapidly due to the $\gamma$-contraction property of $\cT^\ror$  w.r.t.~the $\ell_\infty$ norm \citep{iyengar2005robust,nilim2005robust}.

\paragraph{Specification of the divergence $\rho$.}
We consider two popular choices of the uncertainty set measured in terms of two different $f$-divergence metric: the total variation distance and the $\chi^2$ divergence, 
given respectively by \citep{tsybakov2009introduction}
\begin{align}
\rho_{\TV}\left(P_{s,a}, P^0_{s,a}\right) & \defn \frac{1}{2} \left\| P_{s,a} - P^0_{s,a}\right\|_1 =  \frac{1}{2} \sum_{s'\in \cS} P^{\no}(s' \mymid s,a)\left| 1 - \frac{P(s' \mymid s,a)}{P^{\no}(s' \mymid s,a)}\right|, \label{eq:tv-distance} \\
\rho_{\chi^2} \left(P_{s,a}, P^0_{s,a}\right) & \defn \sum_{s'\in \cS} P^{\no}(s' \mymid s,a)\left(1 - \frac{P(s' \mymid s,a)}{P^{\no}(s' \mymid s,a)}\right)^2. \label{eq:chi-squared-distance}
\end{align}
Note that $\rho_{\TV}\left(P_{s,a}, P^0_{s,a}\right)\in [0,1]$ and $\rho_{\chi^2} \left(P_{s,a}, P^0_{s,a}\right) \in [0,\infty)$ in general. As we shall see shortly, these two choices of divergence metrics result in drastically different messages when it comes to sample complexities.

\paragraph{Applications for different divergence $\rho$.}
The choice of uncertainty set --- its shape and size --- should be determined by the specific application to avoid models that are either over-conservative or insufficiently expressive. Below, we provide some example applications for both the TV distance and the $\chi^2$ divergence.
\begin{itemize}
		\item The TV distance is most appropriate when the main concern is the possibility of rare, non-local perturbations that fundamentally alter system dynamics in ways not reflected in the nominal model. For example, supply chain  disruptions \citep{amico2024adapting} are typically abrupt---a key supplier may suddenly become unavailable due to bankruptcy, natural disaster, or geopolitical events. In online 3D bin packing \citep{pan2023adjustable}, the arrival of hard sequences of items can lead to out-of-distribution states that the nominal model assigns negligible probability to. Robust RL with the TV distance allows modeling such worst-case transitions by enabling probability mass to move to previously unobserved states, thus providing robustness against these rare but catastrophic events.  

    \item The $\chi^2$ divergence is well-suited when the support of the distribution is stable but the probabilities within that support may shift. For instance, in finance and marketing, the set of possible outcomes (e.g., stock movements or consumer choices) is generally known and stable. Uncertainty arises primarily from changes in the likelihood of these outcomes, driven by market conditions or consumer preferences. Here, $\chi^2$ divergence (as well as the Wasserstein or KL divergence) is appropriate, as it models reweighting of existing probabilities without introducing new states.
\end{itemize}

\paragraph{Sampling mechanism: a generative model.}
Following \citet{zhou2021finite,panaganti2021sample}, we assume access to a generative model or a simulator \citep{kearns1999finite}, which allows us to collect $N$ independent samples for each state-action pair 
generated based on the {\em nominal} kernel $P^{\no}$:
\begin{align}
	\forall (s,a)\in \cS\times\cA, \qquad s_{i, s,a} \overset{i.i.d}{\sim} P^\no(\cdot \mymid s,a), \qquad i = 1, 2,\cdots, N.
\end{align}
The total sample size is, therefore, $NSA$. 

\paragraph{Goal.}
Given the collected samples, the task is to learn the robust optimal policy for the RMDP --- w.r.t.~some prescribed uncertainty set $\cU^\ror(P^\no)$ around the nominal kernel --- using as few samples as possible. Specifically, given some target accuracy level $\varepsilon>0$, the goal is to seek an $\varepsilon$-optimal robust policy $\widehat{\pi}$ obeying
\begin{align}
	\forall s\in \cS:\quad V^{\star, \sigma}(s) - V^{\widehat{\pi}, \sigma}(s) \leq \varepsilon.
\end{align}

\section{Model-based algorithm: distributionally robust value iteration}\label{sec:algorithm}

We consider a model-based approach tailored to RMDPs, which first constructs an empirical nominal transition kernel based on the collected samples, and then applies distributionally robust value iteration (DRVI) to compute an optimal robust policy.

\paragraph{Empirical nominal kernel.} The empirical nominal transition kernel $\widehat{P}^\no \in \mathbb{R}^{SA\times S}$ can be constructed on the basis of the empirical frequency of state transitions, i.e.,
\begin{align}
	\forall (s,a)\in \cS\times \cA:\quad \widehat{P}^0(s'\mymid s,a) \defn  \frac{1}{N} \sum\limits_{i=1}^N \mathds{1} \big\{ s_{i,s,a} = s' \big\},
	\label{eq:empirical-P-infinite}
\end{align}
which leads to an empirical RMDP $\widehat{\cM}_{\mathsf{rob}} = \{\cS,\cA, \gamma, \cU^{\ror}_\rho(\widehat{P}^\no), r\}$. Analogously, we can define the corresponding robust value function (resp.~robust Q-function) of policy $\pi$ in $\widehat{\cM}_{\mathsf{rob}}$ as
$\widehat{V}^{\pi,\ror}$ (resp.~$\widehat{Q}^{\pi,\ror}$) (cf. \eqref{eq:mdp-value-Q}). In addition, we denote the corresponding {\em optimal robust policy} as $\widehat{\pi}^{\star}$ and the {\em optimal robust value function} (resp.~{\em optimal robust Q-function}) as $\widehat{V}^{\star, \ror}$ (resp.~$\widehat{Q}^{\star, \ror}$) (cf.~\eqref{eq:bellman-equ-infinite-all}), which satisfies the robust Bellman optimality equation:
\begin{align}
	\forall (s,a)\in \cS\times \cA: \quad &\widehat{Q}^{\star,\ror}(s,a) = r(s,a) + \gamma\inf_{\cP\in \unb^{\ror}_\rho(\widehat{P}^{\no}_{s,a})} \cP\widehat{V}^{\star,\ror}. \label{eq:bellman-equ-star-infinite-est}
\end{align}

Equipped with $\widehat{P}^\no$, we can define the empirical robust Bellman operator $\that^\ror$ as
\begin{align}\label{eq:robust_bellman_empirical}
	\forall (s,a)\in &\cS\times \cA :\quad \that^\ror(Q)(s,a) \defn r(s,a) + \gamma \inf_{ \cP \in \unb^{\ror}_\rho(\widehat{P}^{\no}_{s,a})} \cP V,  \quad \text{with}  \quad  V(s) \defn \max_a Q(s,a).
\end{align}

\paragraph{\DRVI: distributionally robust value iteration.} To compute the fixed point of $\that^\ror$, we introduce distributionally robust value iteration (\DRVI), which is  
 summarized in Algorithm~\ref{alg:cvi-dro-infinite}. From an initialization $\widehat{Q}_0 = 0$, the update rule at the $t$-th ($t\geq 1$) iteration can be formulated as:
\begin{align}
\forall (s,a)\in & \cS\times \cA :\quad \widehat{Q}_t(s,a) = \that^\ror \big(\widehat{Q}_{t-1} \big)(s,a) = r(s,a) + \gamma \inf_{ \cP \in \unb_\rho^{\ror}(\widehat{P}^{\no}_{s,a})} \cP \widehat{V}_{t-1}, \label{eq:vi-iteration}
\end{align}
where $\widehat{V}_{t-1}(s) = \max_a \widehat{Q}_{t-1}(s,a)$ for all $s \in \cS$. However, directly solving \eqref{eq:vi-iteration} is computationally expensive
since it involves optimization over an $S$-dimensional probability simplex at each iteration, especially when the dimension of the state space $\cS$ is large. Fortunately, in view of strong duality \citep{iyengar2005robust}, \eqref{eq:vi-iteration} can be equivalently solved using its dual problem, which concerns optimizing a {\em scalar} dual variable and thus can be solved efficiently. In what follows, we shall illustrate this for the two choices of the divergence $\rho$ of interest (cf.~\eqref{eq:tv-distance} and \eqref{eq:chi-squared-distance}). Before continuing, for any $V\in \mathbb{R}^S$, we denote $[V]_{\alpha}$ as its clipped version by some non-negative value $\alpha$, namely,
\begin{align}
	[V]_{\alpha}(s) \defn \begin{cases} \alpha, & \text{if } V(s) > \alpha, \\
V(s), & \text{otherwise.}
\end{cases} \label{eq:V-alpha-defn}
\end{align}
 
\begin{itemize}
\item TV distance, where the uncertainty set is $\unb^{\ror}_\rho(\widehat{P}^{\no}_{s,a}) \defn \cU^{\ror}_{\mathsf{TV}}(\widehat{P}^{\no}_{s,a}) \defn  \cU^{\ror}_{\rho_{\mathsf{TV}}}(\widehat{P}^{\no}_{s,a})$ w.r.t. the TV distance $\rho = \rho_{\mathsf{TV}}$ defined in \eqref{eq:tv-distance}. In particular, we have the following lemma due to strong duality, which is a direct consequence of \citet[Lemma~4.3]{iyengar2005robust}.
\begin{lemma}[Strong duality for TV]\label{lemma:tv-dual-form}
Consider any probability vector  $P\in\Delta(\cS)$, any fixed uncertainty level $\ror$ and the uncertainty set $\unb^{\ror}(P) \defn \cU^{\ror}_{\mathsf{TV}}(P)$. For any vector $V\in \mathbb{R}^S$ obeying $  V \geq  {0}$, recalling the definition of $[V]_\alpha$ in \eqref{eq:V-alpha-defn}, one has
\begin{align}
	\inf_{ \cP \in \unb^{\ror}(P)} \cP V &= \max_{\alpha\in [\min_s V(s), \max_s V(s)]} \left\{P \left[V\right]_{\alpha} - \ror \left(\alpha - \min_{s'}\left[V\right]_{\alpha}(s') \right)\right\} \label{eq:vi-l1norm}.
\end{align}
\end{lemma}
In view of the above lemma, the following dual update rule is equivalent to \eqref{eq:vi-iteration} in \DRVI:
\begin{align}
	\widehat{Q}_{t}(s,a) = r(s,a) + \gamma \max_{\alpha\in\left[\min_s \widehat{V}_{t-1}(s), \max_s \widehat{V}_{t-1}(s)\right]} \left\{\widehat{P}^{\no}_{s,a} \left[\widehat{V}_{t-1}\right]_{\alpha} - \ror \left(\alpha - \min_{s'}\left[\widehat{V}_{t-1}\right]_{\alpha}(s') \right)\right\}. \label{eq:tv-operator-equal}
\end{align}

\item $\chi^2$ divergence, where the uncertainty set is $\unb^{\ror}_\rho(\widehat{P}^{\no}_{s,a}) \defn \cU^{\ror}_{\chi^2}(\widehat{P}^{\no}_{s,a}) \defn \cU^{\ror}_{\rho_{\chi^2}}(\widehat{P}^{\no}_{s,a})$ w.r.t. the $\chi^2$ divergence $\rho = \rho_{\chi^2}$ defined in \eqref{eq:chi-squared-distance}. We introduce the following lemma which directly follows from \citep[Lemma~4.2]{iyengar2005robust}.
\begin{lemma}[Strong duality for $\chi^2$]\label{lem:dual-vi-l2-norm}
Consider any probability vector $P\in\Delta(\cS)$, any fixed uncertainty level $\ror$ and the uncertainty set $\unb^{\ror}(P) \defn \cU^{\ror}_{\chi^2}(P)$. For any vector $V\in\mathbb{R}^S$ obeying $V\geq  {0}$,   one has
	\begin{align}
		\inf_{ \cP \in \unb^{\ror}(P)} \cP V = \max_{\alpha\in [\min_s V(s), \max_s V(s)] } \left\{ P [V]_\alpha - \sqrt{\sigma \mathsf{Var}_{P}\left([V]_\alpha\right) } \right\},
	\end{align}
where $\mathsf{Var}_{P}\left( \cdot \right)$ is defined as \eqref{eq:defn-variance}.
\end{lemma}
In view of the above lemma, the update rule \eqref{eq:vi-iteration} in \DRVI can be equivalently written as: 
\begin{align}
\widehat{Q}_{t}(s,a) = r(s,a) + \gamma \max_{\alpha\in\left[\min_s \widehat{V}_{t-1}(s), \max_s \widehat{V}_{t-1}(s)\right]} \left\{ \widehat{P}^{\no}_{s,a} \left[\widehat{V}_{t-1} \right]_\alpha - \sqrt{\sigma \mathsf{Var}_{\widehat{P}^{\no}_{s,a}}\left( \left[\widehat{V}_{t-1} \right]_\alpha \right) } \right\}. \label{eq:chi2-operator-equal}
\end{align}

\end{itemize}

 The proofs of Lemma~\ref{lemma:tv-dual-form} and  Lemma~\ref{lem:dual-vi-l2-norm} are provided in Appendix~\ref{sec:robust_bellman_properties-proof}. To complete the description, we output the greedy policy of the final Q-estimate $\widehat{Q}_T$ as the final policy $\widehat{\pi}$, namely,
\begin{align}
	 \forall s \in \cS: \quad \widehat{\pi}(s) = \arg\max_a \widehat{Q}_T(s,a).
\end{align}
Encouragingly, the iterates $\big\{\widehat{Q}_t \big\}_{t\geq0}$ of \DRVI converge linearly to the fixed point $\widehat{Q}^{\star,\ror}$, 
owing to the appealing $\gamma$-contraction property of $\that^\ror$. 

\begin{algorithm}[t]
\DontPrintSemicolon
	\textbf{input:} empirical nominal transition kernel $\widehat{P}^{\no}$; reward function $r$; uncertainty level $\ror$; number of iterations $T$. \\ 
	\textbf{initialization:} $\widehat{Q}_0(s,a)= 0$, $\widehat{V}_0(s)=0$ for all $(s,a) \in \cS\times \cA$. \\
   \For{$t = 1,2,\cdots, T$}
	{
		
		\For{$s\in \cS, a\in \cA$}{
			Set $\widehat{Q}_t(s, a)$ according to \eqref{eq:vi-iteration};

		}
		\For{$s\in \cS$}{
			Set $\widehat{V}_t(s) = \max_a \widehat{Q}_t(s, a)$;
		}
	}

	\textbf{output:} $\widehat{Q}_T$, $\widehat{V}_T$ and $\widehat{\pi}$ obeying $\widehat{\pi}(s) \defn \arg\max_a \widehat{Q}_T(s,a)$.
	\caption{Distributionally robust value iteration (\DRVI) for infinite-horizon RMDPs.}
 \label{alg:cvi-dro-infinite}
\end{algorithm}

\section{Theoretical guarantees: sample complexity analyses}\label{sec:theory}

We now present our main results, which concern the sample complexities of learning RMDPs when the uncertainty set is specified using the TV distance or the $\chi^2$ divergence. Somewhat surprisingly, different choices of the uncertainty set can lead to dramatically different consequences in the sample size requirement.

\subsection{The case of TV distance: RMDPs are easier to learn than standard MDPs}\label{sec:upper-L1}

We start with the case where the uncertainty set is measured via the TV distance.
The following theorem, whose proof is deferred to Section~\ref{proof:thm:l1-upper-bound}, develops an upper bound on the sample complexity of \DRVI in order to return an $\varepsilon$-optimal robust policy. 
The key challenge of the analysis lies in careful control of the robust value function $V^{\pi,\ror}$ as a function of the uncertainty level $\ror$.

\begin{theorem}[Upper bound under TV distance]\label{thm:l1-upper-bound} 
	Let the uncertainty set be $\unb_\rho^\ror(\cdot) = \cU^{\ror}_{\mathsf{TV}}(\cdot)$, as specified by the TV distance \eqref{eq:tv-distance}. Consider any discount factor $\gamma \in \left[\frac{1}{4},1 \right)$, 
 uncertainty level $\ror\in (0,1)$, and $\delta \in (0,1)$. 
	Let $\widehat{\pi}$ be the output policy of Algorithm~\ref{alg:cvi-dro-infinite} after $T = C_1 \log \big( \frac{N}{1-\gamma}\big)$ iterations. 
	Then with probability at least $1-\delta$, one has
\begin{align}
	\forall s\in\cS: \quad V^{\star, \ror}(s) - V^{\widehat{\pi}, \ror}(s) \leq \varepsilon
\end{align}
for any $\varepsilon \in \left(0, \sqrt{1/\max\{1-\gamma, \ror\}} \right]$,
as long as the total number of samples obeys
\begin{align}
	NSA \geq  \frac{C_2 SA}{ (1-\gamma)^2 \max\{1-\gamma, \ror\} \varepsilon^2}\log\left(\frac{SAN}{(1-\gamma)\delta}\right).
\end{align}
Here, $C_1, C_2>0$ are some large enough universal constants.  
\end{theorem}

\begin{remark}\label{remark:tv-upper}
Note that Theorem~\ref{thm:l1-upper-bound} is not only valid when invoking Algorithm~\ref{alg:cvi-dro-infinite}. 
In fact, the theorem holds for any oracle planning algorithm (designed based on the empirical transitions $\widehat{P}^\no$) whose output policy $\widehat{\pi}$ obeys 
\begin{equation}\label{eq:oracle_requirement}
\big\| \widehat{V}^{\star,\ror} - \widehat{V}^{\widehat{\pi},\ror} \big\|_\infty \leq O\left(\frac{(1-\gamma)^2 }{N} \log\left(\frac{SAN}{(1-\gamma)\delta}\right)  \right). 
\end{equation}
\end{remark}

Before discussing the implications of Theorem~\ref{thm:l1-upper-bound}, we present a matching minimax lower bound that confirms the tightness and optimality of the upper bound, which in turn pins down the sample complexity requirement for learning RMDPs with TV distance. The proof is based on constructing new hard instances inspired by the asymmetric structure of RMDPs, with the details postponed to Section~\ref{proof:thm:l1-lower-bound}.

\begin{theorem}[Lower bound under TV distance]\label{thm:l1-lower-bound}
Consider any tuple $(S, A, \gamma, \ror, \varepsilon)$ obeying $\ror\in (0, 1 - c_0]$ with $0 <c_0 \leq \frac{1}{8}$ being any small enough positive constant, $\gamma\in\left[ \frac{1}{2}, 1\right)$, and $\varepsilon \in \big(0, \frac{c_0}{256(1-\gamma)} \big]$. We can construct a collection of infinite-horizon RMDPs $\cM_0, \cM_1$ defined by the uncertainty set $\unb_\rho^\ror(\cdot) = \cU^{\ror}_{\mathsf{TV}}(\cdot)$, an initial state distribution $\varphi$, and a dataset with $N$ independent samples for each state-action pair over the nominal transition kernel (for $\cM_0$ and $\cM_1$ respectively), such that
	\[
	\inf_{\widehat{\pi}}\max\left\{ \mathbb{P}_{0}\big( V^{\star,\ror}(\varphi)-V^{\widehat{\pi}, \ror}(\varphi)>\varepsilon\big), \,
	\mathbb{P}_{1}\big( V^{\star,\ror }(\varphi) - V^{\widehat{\pi},\ror}(\varphi) >\varepsilon\big)\right\} \geq\frac{1}{8},
\]
provided that
\[NSA \leq \frac{c_0   SA \log 2  }{ 8192 (1-\gamma)^2 \max\{ 1 -\gamma, \sigma\}\varepsilon^2}.\]
Here, the infimum is taken over all estimators $\widehat{\pi}$, and $\mathbb{P}_{0}$ (resp.~$\mathbb{P}_{1}$) denotes the probability
when the RMDP is $\mathcal{M}_{0}$ (resp.~$\mathcal{M}_{1}$).
\end{theorem}
 
Below, we interpret the above theorems and highlight several key implications about the sample complexity requirements for learning RMDPs for the case w.r.t.~the TV distance.

\paragraph{Near minimax-optimal sample complexity.} Theorem~\ref{thm:l1-upper-bound} shows that the total number of samples required for \DRVI (or any oracle planning algorithm claimed in Remark~\ref{remark:tv-upper}) to yield $\varepsilon$-accuracy is
\begin{align}\label{eq:tv-final-samples}
\widetilde{O} \left(\frac{SA}{ (1-\gamma)^2 \max\{1-\gamma, \ror\} \varepsilon^2} \right).
\end{align}
Taken together with the minimax lower bound asserted by Theorem~\ref{thm:l1-lower-bound}, this confirms the near optimality of the sample complexity (up to some logarithmic factor) almost over the full range of the uncertainty level $\ror$. 
Importantly, this sample complexity scales linearly with the size of the state-action space, 
and is inversely proportional to $\sigma$ in the regime where $\sigma \gtrsim 1-\gamma$.

\paragraph{RMDPs is easier than standard MDPs with TV distance.} Recall that the sample complexity requirement for learning standard MDPs with a generative model is \citep{gheshlaghi2013minimax,agarwal2020model,li2020breaking}
\begin{equation}\label{eq:standard-final-samples}
 \widetilde{O} \left(\frac{SA}{ (1-\gamma)^3 \varepsilon^2} \right)
\end{equation}
in order to yield $\varepsilon$ accuracy. 
Comparing this with the sample complexity requirement in \eqref{eq:tv-final-samples} for RMDPs under the TV distance, we confirm that the latter is at least as easy as --- if not easier than --- standard MDPs. In particular, when $\sigma \lesssim 1-\gamma$ is small, the sample complexity of RMDPs is the same as that of standard MDPs as in \eqref{eq:standard-final-samples}, which is as anticipated since the RMDP reduces to the standard MDP when $\sigma=0$. On the other hand, when $1-\gamma \lesssim \sigma < 1 $, the sample complexity of RMDPs simplifies to
\begin{align}\label{eq:tv-large-sigma}
\widetilde{O} \left(\frac{SA}{ (1-\gamma)^2   \ror  \varepsilon^2} \right) ,
\end{align}
which is smaller than that of standard MDPs by a factor of $\sigma/(1-\gamma)$.

\paragraph{Comparison with state-of-the-art bounds.}
For the upper bound, our results (cf.~Theorem~\ref{thm:l1-upper-bound}) significantly improves over the prior art $\widetilde{O} \left(\frac{S^2A }{(1-\gamma)^4 \varepsilon^2}\right)$ of \citet{panaganti2021sample} by at least a factor of $\frac{S}{1-\gamma}$ and even $\frac{S}{(1-\gamma)^2}$ when the uncertainty level $1-\gamma \lesssim \sigma < 1 $ is large.
Turning to the lower bound side, \citet{yang2021towards} developed a lower bound for RMDPs under the TV distance, which scales as
$$ \widetilde{\Omega} \left(\frac{SA(1-\gamma)}{  \varepsilon^2}\min\left\{ \frac{1}{(1-\gamma)^4}  ,\frac{1}{  \ror^4}  \right\} \right) .$$
Clearly, this is worse than ours by a factor of
$ \frac{\sigma^3}{(1-\gamma)^3}\in \big( 1, \frac{1}{(1-\gamma)^3} \big)$
in the regime where $1-\gamma \lesssim \sigma < 1 $.

\subsection{The case of $\chi^2$ divergence: RMDPs are harder than standard MDPs}

We now switch attention to the case when the uncertainty set is measured via the $\chi^2$ divergence. 
The theorem below presents an upper bound on the sample complexity for this case, whose proof is deferred to Appendix~\ref{proof:thm:chi2-upper-bound}.

\begin{theorem}[Upper bound under $\chi^2$ divergence]\label{thm:l2-upper-bound}
	Let the uncertainty set be $\unb_\rho^\ror(\cdot) = \cU^{\ror}_{\chi^2}(\cdot)$, as specified using the $\chi^2$ divergence \eqref{eq:chi-squared-distance}. Consider any uncertainty level $\ror\in (0, \infty)$, $\gamma\in [1/4,1)$ and $\delta \in(0,1)$. With probability at least $1-\delta$, the output policy $\widehat{\pi}$ from Algorithm~\ref{alg:cvi-dro-infinite} with at most $T = c_1 \log \big( \frac{N}{1-\gamma}\big)$ iterations yields
\begin{align}
	\forall s\in\cS: \quad V^{\star, \ror}(s) - V^{\widehat{\pi}, \ror}(s) \leq \varepsilon
\end{align}
for any $\varepsilon\in \big(0,\frac{1}{1-\gamma} \big]$, as long as the total number of  samples obeying
\begin{align}
	NSA \geq \frac{c_2SA  }{(1-\gamma)^3 \varepsilon^2} \left(1+\frac{\max\{\sqrt{\ror}, \ror \} }{1-\gamma} \right) \log\left(\frac{SA N}{\delta}\right) .
\end{align}
Here, $c_1, c_2>0$ are some large enough universal constants.  
\end{theorem}

\begin{remark}\label{remark:chi2-upper}
Akin to Remark~\ref{remark:tv-upper}, the sample complexity derived in Theorem~\ref{thm:l2-upper-bound} continues to hold for any oracle planning algorithm that outputs a policy $\widehat{\pi}$ obeying $\big\| \widehat{V}^{\star,\ror} - \widehat{V}^{\widehat{\pi},\ror} \big\|_\infty \leq O\Big(\frac{ \log(\frac{SAN}{(1-\gamma)\delta}) }{N^2} \Big)$.
\end{remark}

In addition, in order to gauge the tightness of Theorem~\ref{thm:l2-upper-bound} and understand the minimal sample complexity requirement under the $\chi^2$ divergence, we further develop a minimax lower bound as follows; the proof is deferred to Appendix~\ref{proof:thm:chi2-lower-bound}. 

\begin{theorem}[Lower bound under $\chi^2$ divergence]\label{thm:chi2-lower-bound}
Consider any $(S, A, \gamma, \ror, \varepsilon)$ obeying  $\gamma\in [\frac{3}{4}, 1)$, $\ror \in (0,\infty)$, and
\begin{align}
\varepsilon &\leq  \frac{c_3}{(1-\gamma)},
\end{align}
for some small universal constant $c_3>0$.
Then we can construct two infinite-horizon RMDPs $\cM_0, \cM_1$ defined by the uncertainty set $\unb_\rho^\ror(\cdot) = \cU^{\ror}_{\chi^2}(\cdot)$, an initial state distribution $\varphi$, and a dataset with $N$ independent samples per $(s,a)$ pair over the nominal transition kernel (for $\cM_0$ and $\cM_1$ respectively), such that
\begin{align}
	\inf_{\widehat{\pi}}\max\left\{ \mathbb{P}_{0}\big( V^{\star,\ror}(\varphi)-V^{\widehat{\pi}, \ror}(\varphi)>\varepsilon\big), \,
	\mathbb{P}_{1}\big( V^{\star,\ror }(\varphi) - V^{\widehat{\pi},\ror}(\varphi) >\varepsilon\big)\right\} \geq\frac{1}{8}, \label{eq:lower-bound-chi2-all}
\end{align}
provided that the total number of samples
\begin{align} \label{eq:final-chi2-lower-bound}
NSA \leq \frac{c_4SA}{(1-\gamma)^3  \varepsilon^2}\left(1+\frac{\ror}{1-\gamma} \right) 
\end{align}
for some universal constant $c_4>0$.
\end{theorem}

We are now positioned to single out several key implications of the above theorems.

\paragraph{Nearly tight sample complexity.} In order to achieve $\varepsilon$-accuracy for RMDPs under the $\chi^2$ divergence, Theorem~\ref{thm:l2-upper-bound} asserts that a total number of samples on the order of
\begin{align}\label{eq:chi2-final-samples}
 \widetilde{O} \left(\frac{SA}{(1-\gamma)^3 \varepsilon^2} \left(1+\frac{ \max\{\sqrt{\ror},  \ror \}}{1-\gamma} \right)\right).
\end{align}
is sufficient for \DRVI (or any other oracle planning algorithm as discussed in Remark~\ref{remark:chi2-upper}). Taking this together with the minimax lower bound in Theorem~\ref{thm:chi2-lower-bound} confirms that the sample complexity is optimal over a broad range of the uncertainty level $\sigma$ (when $\sigma \lesssim (1-\gamma)^2$ and $\sigma\gtrsim 1$). In particular,
\begin{itemize}
\item when $\sigma \lesssim (1-\gamma)^2$, our sample complexity upper bound $\widetilde{O} \left( \frac{SA }{  (1-\gamma)^3 \varepsilon^2} \right)$ is sharp and matches the minimax lower bound; 
\item  when $(1-\gamma)^2 \lesssim \sigma \lesssim 1$, the upper bound for our sample complexity is on the order of $\widetilde{O} \left( \frac{SA \sqrt{\sigma}}{  (1-\gamma)^4 \varepsilon^2} \right)$, which is near-optimal, up to a factor of at most $\sqrt{\frac{1}{1-\gamma}}$;
\item  when $\sigma\gtrsim 1$, our sample complexity upper bound $\widetilde{O} \left( \frac{SA \sigma}{  (1-\gamma)^4 \varepsilon^2} \right)$ again matches the minimax lower bound.
\end{itemize}

\paragraph{RMDPs are harder to learn than standard MDPs with $\chi^2$ divergence.} The minimax lower bound developed in Theorem~\ref{thm:chi2-lower-bound} reveals an opposite behavior in the sample size requirement when the uncertainty set is measured via the $\chi^2$ divergence, compared to the TV distance, across the entire range of the uncertainty level $\ror\in(0,\infty)$. 
Several distinct regimes are worth highlighting, demonstrating that learning robust MDPs (RMDPs) under $\chi^2$ divergence is never easier than learning standard MDPs:
\begin{itemize}
\item When $\sigma \lesssim 1-\gamma$, the lower bound reduces to
\begin{align}
\widetilde{\Omega} \left( \frac{SA }{  (1-\gamma)^3 \varepsilon^2} \right),
\end{align}
which matches the sample complexity of standard MDPs.
\item When $\sigma \gtrsim 1-\gamma$, the lower bound is  on the order of 
\begin{align}
\widetilde{\Omega} \left( \frac{SA \sigma}{  (1-\gamma)^4 \varepsilon^2} \right),
\end{align}
which is consistently {\em greater} than the sample complexity of standard MDPs.
\end{itemize}

\paragraph{Comparison with state-of-the-art bounds.} Our upper bound significantly improves over the prior art $\widetilde{O} \left(\frac{S^2A(1+\ror) }{(1-\gamma)^4 \varepsilon^2}\right)$ of \citet{panaganti2021sample} by at least a factor of $S$, and provides the {\em first} finite-sample complexity that scales {\em linearly} with respect to  $S$ for discounted infinite-horizon RMDPs, which typically exhibit more complicated statistical dependencies than the finite-horizon counterpart. On the other hand, \citet{yang2021towards} established a lower bound on the order of $\widetilde{\Omega}\left(\frac{SA}{(1-\gamma)^2\sigma \varepsilon^2}\right)$ when $\sigma \gtrsim 1-\gamma$, which is always smaller than the requirement of standard MDPs, and diminishes when $\sigma$ grows. Consequently, the gap between the upper and lower bounds in \citet{yang2021towards} makes it challenging to justify whether robust MDPs (RMDPs) are harder than standard MDPs under  $\chi^2$ divergence, as well as to determine the linear scaling of sample size with respect to $\sigma$ as $\sigma$ grows. This work not only improves both the upper and lower bounds but also provides nearly matching results to rigorously demonstrate that, under $\chi^2$ divergence, RMDPs are harder than standard MDPs across the entire range of uncertainty levels $\varepsilon$, Furthermore, when $\sigma \gtrsim 1$, the matched lower and upper bound establishes that the sample size requirement scales linearly with $\sigma$ as $\sigma$ grows. We note that this phenomenon is consistent with prior results on other distributionally robust optimization problems~\citep{duchi2021learning}, which also establish a lower bound with linear dependence on $\sigma$.

\subsection{Why are the TV and $\chi^2$ cases drastically different}
Our results demonstrate that the sample complexity of robust RL depends fundamentally on the size and structure of the uncertainty set. Consider, for a moment, how the estimation error propagates through one iteration of the value iteration from a fixed value function $V$ (cf. \eqref{eq:vi-iteration}), when the nomination kernel $P^0$ is replaced by its plug-in estimate $\widehat{P}^0$, leading to the two terms for standard RL and robust RL, respectively:
	 \begin{align*}
    \text{Standard RL:}  \quad  \delta_{\text{standard}} &= \Big| P^0 {V} -\widehat{P}^0 {V} \Big|  , \\
    \text{Robust RL:}  \quad  \delta_{\text{robust}} &= \Big|\inf_{\mathcal{P}\in \mathcal{U}^\sigma_\rho\left(P^0 \right)} \mathcal{P} V  - \inf_{\mathcal{P} \in \mathcal{U}^\sigma_\rho\left(\widehat{P}^0 \right)} \mathcal{P} V \Big| .
  \end{align*}
Notably, in standard RL, the estimation error is linear in the estimation gap $P^0 - \widehat{P}^0$, whereas in robust RL, the error term involves a nonlinear inner optimization over the uncertainty set, often without a closed form.
\begin{itemize}
	\item For the TV distance uncertainty set, the error term $\delta_{\text{robust}} \approx  \Big| P^0 {V} -\widehat{P}^0 {V} \Big|  $ is approximately linear with respect to $P^0 - \widehat{P}^0$ after converting to the dual formulation \eqref{eq:vi-l1norm}, similar to standard RL. This is due to the homogeneous shape across all states and almost instance-independent structure (w.r.t. uncertainty set center) of the TV uncertainty set. Crucially, our analysis shows that the range of the robust value function of robust RL for a fixed policy is smaller than that of standard RL when $\sigma \gtrsim 1-\gamma$. This favorable low-variance statistical property means robust RL may require less data in this regime.
	\item  For the $\chi^2$ divergence uncertainty set,   the error term $\delta_{\text{robust}}$ becomes highly nonlinear, significantly amplifying the nominal transition estimation error $P^0 - \widehat{P}^0$. This effect stems from  the heterogeneous structure of the $\chi^2$ uncertainty set, which varies across states and is strongly dependent on the specific center of the set ($P^0$ or $\widehat{P}^0$). Consequently, the sample complexity is higher than that of standard RL.
	\end{itemize}

\section{Analysis for Distributionally robust MDPs}
\label{sec:analysis}

In this section, we present the principal and critical technical steps underlying our main results for both the TV and $\chi^2$ divergence cases, together with intuitive insights and comparisons to their non-robust (standard) MDP counterparts. The full proofs are deferred to the appendix.

\subsection{Preliminaries of the analysis}

\subsubsection{Additional notations} 
For convenience, we introduce the notation $[T]\coloneqq \{1,\cdots,T\}$ for any positive integer $T>0$. Moreover, for any two vectors $x=[x_i]_{1\leq i\leq n}$ and $y=[y_i]_{1\leq i\leq n}$, the notation $ {x}\leq {y}$ (resp.~$ {x}\geq {y}$) means
$x_{i}\leq y_{i}$ (resp.~$x_{i}\geq y_{i}$) for all $1\leq i\leq n$. And for any vecvor $x$, we overload the notation by letting $x \circ x = \big[x(s,a)^2\big]_{(s,a)\in\cS\times\cA}$ (resp.~$x\circ x= \big[x(s)^2\big]_{s\in\cS}$). With slight abuse of notation, we denote ${0}$ (resp.~${1}$) as the all-zero (resp.~all-one) vector, and drop the subscript $\rho$ to write $ \cU^\ror(\cdot) = \cU^\ror_\rho(\cdot)$ whenever the argument holds for all divergence $\rho$. 

\paragraph{Matrix notation.} 
To continue, we recall or introduce some additional matrix notation that is useful throughout the analysis. 
\begin{itemize}

	\item $P^\no \in \mathbb{R}^{SA\times S}$: the matrix of the nominal transition kernel with $P^\no_{s,a}$ as the $(s,a)$-th row.
	\item $\widehat{P}^\no \in \mathbb{R}^{SA\times S}$: the matrix of the estimated nomimal transition kernel with $\widehat{P}^\no_{s,a}$ as the $(s,a)$-th row.
 \item $r \in \mathbb{R}^{SA}$: a vector representing the reward function $r$ (so that $r_{(s,a)} =r(s,a)$ for all $(s,a)\in \cS\times \cA$). 
\item $r_{\pi} \in \mathbb{R}^{S}$: a reward vector restricted to the actions chosen by the policy $\pi$, namely,  $r_{\pi}(s) = r(s,\pi(s))$ for all $s\in \cS$ (or simply, $r_{\pi}=\Pi^{\pi}r$). 

 	\item  $\mathrm{Var}_{P}(V) \in \mathbb{R}^{SA}$: for any transition kernel $P \in \mathbb{R}^{SA \times S}$ and vector $V \in \mathbb{R}^S$, we denote the $(s,a)$-th row of $\mathrm{Var}_{P}(V)$ as
\begin{align}\label{eq:defn-variance-vector}
	\mathsf{Var}_{P}(s,a) \defn \mathrm{Var}_{P_{s,a}}(V),
\end{align} 
where
\begin{align}\label{eq:defn-variance}
  \mathrm{Var}_{P_{s,a}}(V) \defn P_{s,a} (V \circ V)-  (P_{s,a} V  ) \circ  (P_{s,a} V  ).
\end{align}

	\item $\pmin^{V} \in \mathbb{R}^{SA\times S}$, $\pmhat^{V} \in \mathbb{R}^{SA\times S}$: the matrices representing the probability transition kernel in the uncertainty set that leads to the worst-case value for any vector $V\in\mathbb{R}^S$. We denote $\pmin_{s,a}^{V}$ (resp.~$\pmhat_{s,a}^{V}$) as the $(s,a)$-th row of the transition matrix $\pmin^{V}$ (resp.~$\widehat{\pmin}^{V}$). In truth, the $(s,a)$-th rows of these transition matrices are defined as
	\begin{subequations}\label{eq:inf-p-special}
\begin{align}
	\pmin_{s,a}^{V} &= \mathrm{argmin}_{\cP\in \unb^{\ror}(P^{\no}_{s,a})} \cP V, \qquad \text{and} \qquad  \pmhat_{s,a}^{V} = \mathrm{argmin}_{\cP\in \unb^{\ror}(\widehat{P}^{\no}_{s,a})} \cP V.
\end{align}
Furthermore, we make use of the following short-hand notation:
\begin{align}
	\pmin_{s,a}^{\pi, V} &:= \pmin_{s,a}^{V^{\pi, \ror}}= \mathrm{argmin}_{\cP\in \unb^{\ror}(P^{\no}_{s,a})} \cP V^{\pi, \ror}, \qquad  \pmin_{s,a}^{\pi, \widehat{V}}:=\pmin_{s,a}^{\widehat{V}^{\pi, \ror}}  = \mathrm{argmin}_{\cP\in \unb^{\ror}(P^{\no}_{s,a})} \cP \widehat{V}^{\pi, \ror},   \\
	 \pmhat_{s,a}^{\pi, V} &:= \pmhat_{s,a}^{V^{\pi, \ror}} = \mathrm{argmin}_{P\in \unb^{\ror}(\widehat{P}^{\no}_{s,a})} P V^{\pi, \ror}, \qquad \pmhat_{s,a}^{\pi, \widehat{V}}:=\pmhat_{s,a}^{\widehat{V}^{\pi, \ror}}  = \mathrm{argmin}_{P\in \unb^{\ror}(\widehat{P}^{\no}_{s,a})} P \widehat{V}^{\pi, \ror}. 
\end{align}
\end{subequations}
The corresponding probability transition matrices are denoted by  $\pmin^{\pi, V} \in \mathbb{R}^{SA\times S}$, $\pmin^{\pi, \widehat{V}} \in \mathbb{R}^{SA\times S}$,  $\pmhat^{\pi, V} \in \mathbb{R}^{SA\times S}$ and $\pmhat^{\pi, \widehat{V}} \in \mathbb{R}^{SA\times S}$, respectively.

\item $P^\pi \in \mathbb{R}^{S\times S}$, $\widehat{P}^\pi \in \mathbb{R}^{S\times S}$, $\Pv^{\pi,V}\in \mathbb{R}^{S\times S}$, $\Pv^{\pi, \widehat{V}}\in \mathbb{R}^{S\times S}$, $\Phatv^{\pi, V} \in \mathbb{R}^{S\times S}$ and $\Phatv^{\pi, \widehat{V}} \in \mathbb{R}^{S\times S}$: six {\em square} probability  transition matrices w.r.t. policy $\pi$ over the states, namely  
	\begin{align}
	\label{eqn:ppivq}
		&P^\pi \defn \Pi^\pi P^\no, \qquad \widehat{P}^\pi \defn  \Pi^\pi \widehat{P}^\no, \qquad  \Pv^{\pi,V} \defn \Pi^{\pi}\pmin^{\pi, V}, \qquad \Pv^{\pi, \widehat{V}} \defn \Pi^{\pi}\pmin^{\pi, \widehat{V}}, \nonumber \\
		&\Phatv^{\pi, V}  \defn \Pi^{\pi} \pmhat^{\pi, V} , \qquad \text{and} \qquad \Phatv^{\pi, \widehat{V}} \defn \Pi^{\pi} \pmhat^{\pi, \widehat{V}}.
	\end{align}
Here, $\Pi^{\pi} \in \{0,1\}^{S \times SA}$ is a projection matrix associated with a given deterministic policy $\pi$ taking the following form
\begin{align}
\label{eqn:bigpi}
	\Pi^{\pi} = {\scriptsize
	\begin{pmatrix}
		\e_{\pi(1)}^{\top} &        {0}^{\top}     &  \cdots &  {0}^{\top} \\
		        {0}^{\top}     & \e_{\pi(2)}^\top &  \cdots &  {0}^{\top} \\
			  \vdots  &        \vdots    & \ddots & \vdots \\	
		         {0}^{\top}    &      {0}^{\top}        &    \cdots     & \e_{\pi(S)}^{\top}
	\end{pmatrix}  },
\end{align}
where $\e_{\pi(1)}^{\top},\e_{\pi(2)}^{\top},\ldots,\e_{\pi(S)}^{\top}\in\mathbb{R}^{A}$ are standard basis vectors.	
We denote $P^\pi_s$ as the $s$-th row of the transition matrix $P^\pi$; similar quantities can be defined for the other matrices as well.

\end{itemize}

\subsubsection{Facts of the robust Bellman operator and the empirical robust MDP}
\label{sec:robust_bellman_properties}

\paragraph{$\gamma$-contraction of the robust Bellman operator.} It is worth noting that the  robust Bellman operator (cf.~\eqref{eq:robust_bellman}) shares the nice $\gamma$-contraction property of the standard Bellman operator, stated as below.

\begin{lemma}[$\gamma$-Contraction]\citep[Theorem~3.2]{iyengar2005robust}\label{lem:contration-of-T}
	For any $\gamma\in  [ 0, 1 )$, the robust Bellman operator $\cT^\ror(\cdot)$ (cf.~\eqref{eq:robust_bellman}) is a $\gamma$-contraction w.r.t. $\|\cdot \|_\infty$. Namely, for any $Q_1, Q_2 \in \mathbb{R}^{SA}$ s.t. $Q_1(s,a), Q_2(s,a) \in \big[0, \frac{1}{1-\gamma}\big]$ for all $(s,a) \in \cS \times \cA$, one has
	\begin{align}\label{eq:lemma-contraction}
		\left\| \cT^\ror(Q_1) - \cT^\ror(Q_2) \right\|_\infty \leq \gamma \left\| Q_1 -Q_2\right\|_\infty.
	\end{align}
Additionally, $Q^{\star,\ror}$ is the unique fixed point of $\cT^\ror(\cdot)$ obeying $0\leq  {Q}^{\star,\ror}(s,a) \leq \frac{1}{1-\gamma}$ for all $(s,a)\in \cS\times \cA$.
\end{lemma}

\paragraph{Bellman equations of the empirical robust MDP $\widehat{\cM}_{\mathsf{rob}}$.}
To begin with, recall that the empirical robust MDP $\widehat{\cM}_{\mathsf{rob}} = \{\cS,\cA,  \gamma,\cU^{\ror}(\widehat{P}^\no), r\}$ based on the estimated nominal distribution $\widehat{P}^\no$ constructed in \eqref{eq:empirical-P-infinite} and its corresponding robust value function (resp.~robust Q-function) $\widehat{V}^{\pi,\ror}$ (resp.~$\widehat{Q}^{\pi,\ror}$).

Note that $\widehat{Q}^{\star,\ror}$ is the unique fixed point of $\that^\ror(\cdot)$ (see Lemma~\ref{lem:contration-of-T}), the empirical robust Bellman operator constructed using $\widehat{P}^\no$. Moreover, similar to \eqref{eq:bellman-equ-infinite-all}, for $\widehat{\cM}_{\mathsf{rob}}$, the Bellman's optimality principle gives the following {\em robust Bellman consistency equation} (resp.~{\em robust Bellman optimality equation}):
\begin{subequations}
\begin{align}
	\forall (s,a)\in \cS\times \cA: \quad &\widehat{Q}^{\pi,\ror}(s,a) = r(s,a) + \gamma\inf_{\cP\in \unb^{\ror}(\widehat{P}^{\no}_{s,a})} \cP \widehat{V}^{\pi,\ror}, \label{eq:bellman-equ-pi-infinite-estimate}\\
	\forall (s,a)\in \cS\times \cA: \quad &\widehat{Q}^{\star,\ror}(s,a) = r(s,a) + \gamma\inf_{\cP \in \unb^{\ror}(\widehat{P}^{\no}_{s,a})} \cP \widehat{V}^{\star,\ror}. \label{eq:bellman-equ-star-infinite-estimate}
\end{align}
\end{subequations}

With these in mind, combined with the matrix notation (introduced at the beginning of Section~\ref{sec:analysis}), for any policy $\pi$, we can write the robust Bellman consistency equations as
\begin{align}
	Q^{\pi,\sigma} = r + \gamma \inf_{\cP \in \cU^{\sigma}(P^0)} \cP V^{\pi, \ror} \quad \text{and} \quad \widehat{Q}^{\pi,\sigma} = r + \gamma \inf_{\cP \in \cU^{\sigma}(\widehat{P}^\no)} \cP \widehat{V}^{\pi, \ror}, 
\end{align}
which leads to
\begin{align}
	V^{\pi,\sigma} &= r_\pi + \gamma \Pi^\pi \inf_{\cP \in \cU^{\sigma}(P^0)} \cP V^{\pi, \ror} \overset{\mathrm{(i)}}{=} r_\pi + \gamma \Pv^{\pi,V} V^{\pi,\sigma}, \nonumber \\
	\widehat{V}^{\pi,\sigma} &= r_\pi + \gamma \Pi^\pi \inf_{\cP \in \cU^{\sigma}(\widehat{P}^\no)} \cP \widehat{V}^{\pi, \ror} \overset{\mathrm{(ii)}}{=} r_\pi + \gamma \Phatv^{\pi, \widehat{V}}  \widehat{V}^{\pi,\sigma}, \label{eq:r-bellman-matrix}
\end{align}
where (i) and (ii) holds by the definitions in \eqref{eqn:bigpi}, \eqref{eq:inf-p-special} and \eqref{eqn:ppivq}.

Encouragingly, the above property of the robust Bellman operator ensures the fast convergence of \DRVI.
We collect this consequence in the following lemma, whose proof is postponed to Appendix~\ref{proof:lem:infinite-converge}.
\begin{lemma}\label{lem:infinite-converge}
	Let $\widehat{Q}_0 =0$. The iterates $\{\widehat{Q}_t\}, \{\widehat{V}_t\}$ of \DRVI (cf.~Algorithm~\ref{alg:cvi-dro-infinite}) obey
	\begin{align} \label{eq:converge1}
		\forall  t\geq 0: \quad \big\| \widehat{Q}_t -\widehat{Q}^{\star,\ror} \big\|_\infty \leq \frac{\gamma^t}{1-\gamma} \qquad \text{and} \qquad \big\| \widehat{V}_t -\widehat{V}^{\star,\ror} \big\|_\infty \leq \frac{\gamma^t}{1-\gamma}.
	\end{align}
	Furthermore,  the output policy $\widehat{\pi}$ obeys 
	\begin{align}  \label{eq:V-result}
	\big\| \widehat{V}^{\star,\ror} - \widehat{V}^{\widehat{\pi},\ror} \big\|_\infty &\leq \frac{2\gamma \varepsilon_{\mathsf{opt}}}{1-\gamma}, \qquad \mbox{where}\quad 
\big\|\widehat{V}^{\star,\ror} - \widehat{V}_{T-1} \big\|_\infty  =:\varepsilon_{\mathsf{opt}}.
\end{align}

\end{lemma}

\subsection{Proof outline of the upper bounds: Theorem~\ref{thm:l1-upper-bound} and \ref{thm:l2-upper-bound}}
\label{proof:thm:general-upper-bound}

In the following, we present the  proof outline for the upper bounds, which applies to both the TV distance and $\chi^2$ divergence cases, focusing on the error decomposition that applies to both uncertainty sets.

Before proceeding, applying Lemma~\ref{lem:infinite-converge} yields that for any $\varepsilon_{\mathsf{opt}} > 0$, as long as $T \geq \log(\frac{1}{(1-\gamma) \varepsilon_{\mathsf{opt}}}) $, one has
\begin{align} \label{eq:opt-error}
\big\| \widehat{V}^{\star,\ror} - \widehat{V}^{\widehat{\pi},\ror} \big\|_\infty \leq \frac{2\gamma \varepsilon_{\mathsf{opt}} }{1 -\gamma},
\end{align}
allowing us to justify the more general statement in Remark~\ref{remark:tv-upper}.
To control the performance gap $\left\|V^{\star, \ror}- V^{\widehat{\pi}, \ror} \right\|_\infty$, the proof is divided into several key steps.

Recall the optimal robust policy $\pi^\star$ w.r.t. $\cM_{\mathsf{rob}}$ and the optimal robust policy $\widehat{\pi}^\star$, the optimal robust value function $\widehat{V}^{\star,\ror}$ (resp.~robust value function $\widehat{Q}^{\pi,\ror}$) w.r.t. $\widehat{\cM}_{\mathsf{rob}}$. The term of interest $V^{\star, \ror} - V^{\widehat{\pi}, \ror}$ can be decomposed as
\begin{align}
	V^{\star, \ror} - V^{\widehat{\pi}, \ror} 
	&= \left(V^{\pi^\star, \ror} - \widehat{V}^{\pi^\star, \ror}\right) +  \left(\widehat{V}^{\pi^\star, \ror} - \widehat{V}^{\widehat{\pi}^\star, \ror} \right) + \left(\widehat{V}^{\widehat{\pi}^\star, \ror} - \widehat{V}^{\widehat{\pi}, \ror}  \right) + \left(\widehat{V}^{\widehat{\pi}, \ror} - V^{\widehat{\pi}, \ror}\right) \nonumber \\
	&\overset{\mathrm{(i)}}{\leq} \left(V^{\pi^\star, \ror} - \widehat{V}^{\pi^\star, \ror}\right) + \left(\widehat{V}^{\widehat{\pi}^\star, \ror} - \widehat{V}^{\widehat{\pi}, \ror}  \right) + \left(\widehat{V}^{\widehat{\pi}, \ror} - V^{\widehat{\pi}, \ror}\right)  \nonumber \\
	&\overset{\mathrm{(ii)}}{\leq} \left(V^{\pi^\star, \ror} - \widehat{V}^{\pi^\star, \ror}\right) + \frac{2\gamma \varepsilon_{\mathsf{opt}} }{1 -\gamma}1 + \left(\widehat{V}^{\widehat{\pi}, \ror} - V^{\widehat{\pi}, \ror}\right) \label{eq:l1-decompose}
\end{align}
where (i) holds by  $\widehat{V}^{\pi^\star, \ror} - \widehat{V}^{\widehat{\pi}^\star, \ror} \leq 0$ since $\widehat{\pi}^\star$ is the robust optimal policy for $\widehat{\cM}_{\mathsf{rob}}$, and (ii) comes from the fact in \eqref{eq:opt-error}.

To control the two important terms in \eqref{eq:l1-decompose}, we first consider a more general term $\widehat{V}^{\pi, \ror } - V^{\pi, \ror}$ for any policy $\pi$. Towards this, plugging in \eqref{eq:r-bellman-matrix} yields
\begin{align}
\widehat{V}^{\pi, \ror } - V^{\pi, \ror} & =  r_{\pi} + \gamma \Phatv^{\pi, \widehat{V}} \widehat{V}^{\pi, \ror } - \big( r_\pi + \gamma  \Pv^{\pi, V} V^{\pi, \ror } \big) \nonumber \\
&  = \Big(\gamma \Phatv^{\pi, \widehat{V}} \widehat{V}^{\pi, \ror } - \gamma \Pv^{\pi, \widehat{V} } \widehat{V}^{\pi, \ror } \Big)   + \Big(\gamma \Pv^{\pi, \widehat{V} } \widehat{V}^{\pi, \ror }  - \gamma  \Pv^{\pi, V} V^{\pi, \ror }\Big) \nonumber \\
& \overset{\mathrm{(i)}}{\leq} \gamma \Big(\Pv^{\pi, V} \widehat{V}^{\pi, \ror } - \Pv^{\pi, V} V^{\pi, \ror}  \Big)  +  \Big(\gamma \Phatv^{\pi, \widehat{V}} \widehat{V}^{\pi, \ror } - \gamma \Pv^{\pi, \widehat{V} } \widehat{V}^{\pi, \ror } \Big) ,\nonumber 
\end{align}
where (i) holds by observing
\begin{align*}
	  \Pv^{\pi, \widehat{V} } \widehat{V}^{\pi, \ror }     
	&\leq \Pv^{\pi, V} \widehat{V}^{\pi, \ror }  
\end{align*}
due to the optimality of $  \Pv^{\pi, \widehat{V} } $ (cf.~\eqref{eq:inf-p-special}). Rearranging terms leads to
\begin{align}
\widehat{V}^{\pi, \ror } - V^{\pi, \ror} &\leq \gamma \left(I - \gamma \Pv^{\pi, V} \right)^{-1} \Big( \Phatv^{\pi, \widehat{V}} \widehat{V}^{\pi, \ror } - \Pv^{\pi, \widehat{V} } \widehat{V}^{\pi, \ror } \Big) .
\label{eq:Vl-Vhat-l-perturbation1}
\end{align}

Similarly, we can also deduce
\begin{align}
\widehat{V}^{\pi, \ror } - V^{\pi, \ror} & =  r_{\pi} + \gamma \Phatv^{\pi, \widehat{V}} \widehat{V}^{\pi, \ror } - \left( r_{\pi} + \gamma  \Pv^{\pi, V} V^{\pi, \ror } \right) \nonumber \\
&  = \Big(\gamma \Phatv^{\pi, \widehat{V}} \widehat{V}^{\pi, \ror } - \gamma \Pv^{\pi, \widehat{V} } \widehat{V}^{\pi, \ror } \Big)   + \left(\gamma \Pv^{\pi, \widehat{V} } \widehat{V}^{\pi, \ror }  - \gamma  \Pv^{\pi, V} V^{\pi, \ror }\right) \nonumber \\
& \geq \gamma \left(\Pv^{\pi, \widehat{V}} \widehat{V}^{\pi, \ror } - \Pv^{\pi, \widehat{V}} V^{\pi, \ror}  \right)  + \Big(\gamma \Phatv^{\pi, \widehat{V}} \widehat{V}^{\pi, \ror } - \gamma \Pv^{\pi, \widehat{V} } \widehat{V}^{\pi, \ror } \Big)  \nonumber \\
& \geq  \gamma \left(I - \gamma \Pv^{\pi, \widehat{V}}\right)^{-1} \Big( \Phatv^{\pi, \widehat{V}} \widehat{V}^{\pi, \ror } - \Pv^{\pi, \widehat{V} } \widehat{V}^{\pi, \ror } \Big).
\label{eq:Vl-Vhat-l-perturbation2}
\end{align}

Combining \eqref{eq:Vl-Vhat-l-perturbation1} and \eqref{eq:Vl-Vhat-l-perturbation2}, we arrive at
\begin{align}\label{eq:V-bound-two-side}
	\big\|\widehat{V}^{\pi, \ror } - V^{\pi, \ror}\big\|_\infty &\leq \gamma \max \Big\{ 
	\Big\|\left(I - \gamma \Pv^{\pi, V} \right)^{-1} \Big( \Phatv^{\pi, \widehat{V}} \widehat{V}^{\pi, \ror } - \Pv^{\pi, \widehat{V} } \widehat{V}^{\pi, \ror } \Big) \Big\|_\infty, \nonumber\\
	&\qquad  \Big\| \left(I - \gamma \Pv^{\pi, \widehat{V}}\right)^{-1}  \Big( \Phatv^{\pi, \widehat{V}} \widehat{V}^{\pi, \ror } - \Pv^{\pi, \widehat{V} } \widehat{V}^{\pi, \ror }  \Big)  \Big\|_\infty \Big\}.
\end{align}
By decomposing the error in a symmetric way, we can similarly obtain
\begin{align}
\label{eq:V-bound-two-side2}
\big\|\widehat{V}^{\pi, \ror } - V^{\pi, \ror}\big\|_\infty &\leq \gamma \max \Big\{ 
	\Big\|\left(I - \gamma \Phatv^{\pi, V} \right)^{-1} \Big( \Phatv^{\pi, V} V^{\pi, \ror } - \Pv^{\pi, V } V^{\pi, \ror } \Big) \Big\|_\infty, \nonumber\\
	&\qquad \Big\| \Big(I - \gamma \Phatv^{\pi, \widehat{V}}\Big)^{-1} \Big( \Phatv^{\pi, V} V^{\pi, \ror } - \Pv^{\pi, V } V^{\pi, \ror } \Big) \Big\|_\infty \Big\}.
\end{align}

With the above facts in mind, we are ready to control the two terms $\big\|\widehat{V}^{\pi^\star, \ror } - V^{ \pi^\star, \ror} \big\|_\infty$ and $\big\|\widehat{V}^{\widehat{\pi}, \ror } - V^{\widehat{\pi}, \ror} \big\|_\infty$ in \eqref{eq:l1-decompose} separately.
More specifically, taking $\pi = \pi^\star$, applying \eqref{eq:V-bound-two-side2} leads to
\begin{align} 
\big\|\widehat{V}^{\pi^\star, \ror } - V^{ \pi^\star, \ror}\big\|_\infty &\leq \gamma \max \Big\{ 
\Big\| \Big(I - \gamma \Phatv^{\pi^\star, V}\Big)^{-1} \Big( \Phatv^{\pi^\star, V} V^{\pi^\star, \ror } - \Pv^{\pi^\star, V } V^{\pi^\star, \ror } \Big) \Big\|_\infty, \nonumber\\
	&\qquad  \Big\|\Big(I - \gamma \Phatv^{\pi^\star, \widehat{V}} \Big)^{-1} \Big( \Phatv^{\pi^\star, V} V^{\pi^\star, \ror } - \Pv^{\pi^\star, V } V^{\pi^\star, \ror } \Big) \Big\|_\infty \Big\}. \label{eq:tv-v-star-two-terms}
\end{align}
Similarly, taking $\pi = \widehat{\pi}$, applying \eqref{eq:V-bound-two-side} leads to
\begin{align} 
\big\|\widehat{V}^{\widehat{\pi}, \ror } - V^{\widehat{\pi}, \ror}\big\|_\infty &\leq \gamma \max\Big\{ 
	\Big\|\left(I - \gamma \Pv^{\widehat{\pi}, \widehat{V}} \right)^{-1} \Big( \Phatv^{\widehat{\pi}, \widehat{V}} \widehat{V}^{\widehat{\pi}, \ror } - \Pv^{\widehat{\pi}, \widehat{V} } \widehat{V}^{\widehat{\pi}, \ror } \Big) \Big\|_\infty, \nonumber\\
	&\qquad  \Big\| \left(I - \gamma \Pv^{\widehat{\pi}, V}\right)^{-1} \Big( \Phatv^{\widehat{\pi}, \widehat{V}} \widehat{V}^{\widehat{\pi}, \ror } - \Pv^{\widehat{\pi}, \widehat{V} } \widehat{V}^{\widehat{\pi}, \ror } \Big) \Big\|_\infty \Big\}. \label{eq:tv-v-pihat-two-terms}
\end{align}
With this error decomposition in hand, for each uncertainty sets (e.g., total variation distance or $\chi^2$ divergence), the two deviation terms $\big\|\widehat{V}^{\pi^\star, \ror } - V^{ \pi^\star, \ror} \big\|_\infty$ and $\big\|\widehat{V}^{\widehat{\pi}, \ror } - V^{\widehat{\pi}, \ror} \big\|_\infty$ must be controlled in a set-specific manner to yield the tight sample-complexity upper bounds (cf.~Theorem~\ref{thm:l1-upper-bound} and Theorem~\ref{thm:l2-upper-bound}). The detailed and tailored controlling are postponed to the appendix.

\begin{remark}[asymmetry in error decomposition] In robust RL, each value function estimation gap ($\big\|\widehat{V}^{\pi^\star, \ror } - V^{ \pi^\star, \ror}\big\|_\infty$ or $\big\|\widehat{V}^{\widehat{\pi}, \ror } - V^{\widehat{\pi}, \ror}\big\|_\infty$) is governed by two distinct terms derived from the upper and lower bounds (see \eqref{eq:tv-v-star-two-terms} and \eqref{eq:tv-v-pihat-two-terms}), respectively. These terms cannot be simplified into a single, identical expression due to the varying worst-case transition kernels associated with different value functions. Consequently, in robust RL, there are four distinct and critical terms that need to be managed, which are  significantly more complicated than standard RL.
\end{remark}

\subsection{Construction of hard instances for the lower bounds: Theorem~\ref{thm:l1-lower-bound} and \ref{thm:chi2-lower-bound}} \label{proof:thm:l1-lower-bound}

\begin{figure}[h]
	\centering
	
	\includegraphics[width=0.8\linewidth]{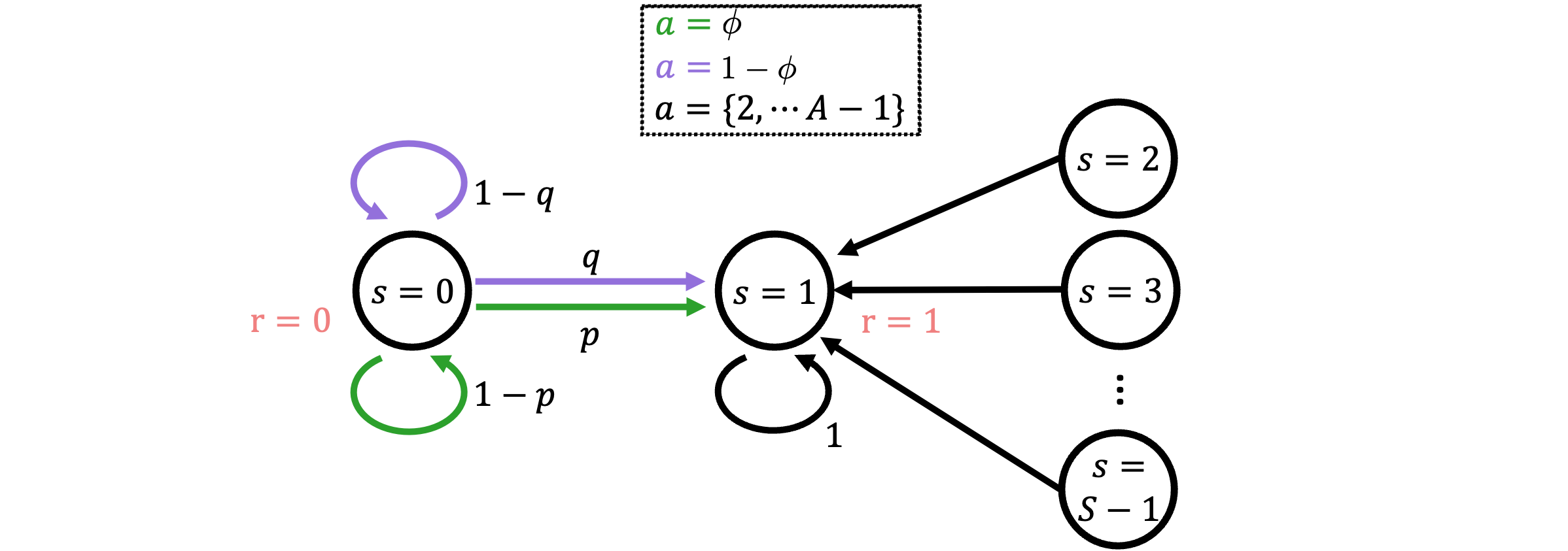} 
	\caption{The constructed hard robust MDP instance for the lower bound. }  
		\label{fig:lower-bound-instance}
	\end{figure}

To achieve a tight lower bound for robust MDPs, we construct new hard instances (illustrated in Figure~\ref{fig:lower-bound-instance}) that are different from those for standard MDPs \citep{gheshlaghi2013minimax}.

\paragraph{Construction of two hard MDPs.} In the following, we present the hard instances constructed for both the TV distance and $\chi^2$ divergence cases, which tackle the aforementioned challenges. Suppose there are two standard MDPs defined as below, illustrated in Figure.~\ref{fig:lower-bound-instance}:
\begin{align*}
   \left\{ \cM_\phi=
    \left(\mathcal{S}, \mathcal{A}, P^{\phi}, r, \gamma \right) 
    \mymid \phi = \{0,1\}
    \right\}.
\end{align*}
Given any state $s\in \{2,3,\cdots, S-1\}$, the corresponding action space are $\mathcal{A} = \{0, 1, 2, \cdots, A-1\}$. For states $s=0$ or $s=1$, the action space is only $\cA' =\{0,1\}$. For any $\phi\in\{0,1\}$, the transition kernel $P^\phi$ of the constructed MDP $\cM_\phi$ is defined as
\begin{align} \label{eq:Ph-construction-lower-infinite}
P^{\phi}(s^{\prime} \mymid s, a) = \left\{ \begin{array}{lll}
         p\mathds{1}(s^{\prime} = 1) + (1-p)\mathds{1}(s^{\prime} = 0)  & \text{if} & (s, a) = (0, \phi) \\
         q\mathds{1}(s^{\prime} = 1) + (1-q)\mathds{1}(s^{\prime} = 0) & \text{if} & (s, a) = (0, 1-\phi) \\
         \mathds{1}(s^{\prime} = 1) & \text{if}     & s \geq 1
                \end{array}\right.,
\end{align}
where $p$ and $q$ are set to satisfy
\begin{align}\label{eq:p-q-defn-infinite}
   0 \leq p \leq 1 \quad \text{ and } \quad 0\leq q = p - \Delta
\end{align}
for some $p$ and $\Delta>0$ that shall be tailored for different uncertainty sets and uncertainty level $\sigma$ to address nonlinearity and diversity.

The above transition kernel $P^{\phi}$ implies that state $1$ is an absorbing state, namely, the MDP will always stay after it arrives at $1$.
Then, we define the reward function as
\begin{align}
r(s, a) = \left\{ \begin{array}{lll}
         1 & \text{if } s = 1 \\
         0 & \text{otherwise}  \ 
                \end{array}\right. .
        \label{eq:rh-construction-lower-bound-infinite}
\end{align}

Additionally, we choose the following initial state distribution:
\begin{align}
    \varphi(s) = 
    \begin{cases} 1, \quad &\text{if }s=0 \\
        0, &\text{otherwise }
    \end{cases}.
    \label{eq:rho-defn-infinite-LB}
\end{align}

Here, the constructed two instances are set with different probability transition from state $0$ with reward $0$ but not state $1$ (with action-invariant transition distribution) with reward $1$ (which were used in standard MDPs \citep{li2024settling}), addressing the challenges of reward allocation and yielding a larger gap between the robust value functions of the two instances.

\paragraph{Proof outline of the lower bounds.}
We outline the proof procedure while deferring the formal proof in the appendix: 1) {\em Construction of hard instances.} As introduced above, we first construct two robust MDPs, $\mathcal{M}_0$ and $\mathcal{M}_1$, that are nearly identical. Their nominal transition probabilities differ only slightly at a single state (state $0$), carefully tailored to the uncertainty level $\sigma$ to create a challenging scenario for different $\sigma$ and any learning algorithm. 2) {\em Transferring estimation  to hypothesis testing.} We then demonstrate that any algorithm achieving a small sub-optimality gap must implicitly identify the true underlying MDP. A near-optimal policy must select the correct action, which effectively serves as a guess for whether the data was generated from $\mathcal{M}_0$ or $\mathcal{M}_1$. 3) {\em Invoking an information-theoretic argument.} Finally, we leverage Le Cam's approach \citep{lecam1973convergence,tsybakov2009introduction}   to show that distinguishing between $\mathcal{M}_0$ and $\mathcal{M}_1$ is statistically impossible if the number of samples is below the threshold stated in the theorem. The KL divergence between the two data-generating distributions is too small to permit reliable identification, thus any algorithm will fail in at least one case. This proves that a certain number of samples is necessary, establishing the minimax lower bound.

\section{Numerical experiments}\label{sec:numerical}

\begin{figure}[H]
	\centering
	\includegraphics[width=0.35\linewidth]{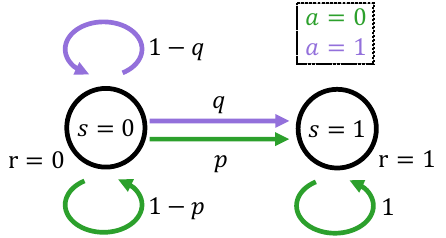} 
	\caption{Illustrations of the considered MDP.}  
		\label{fig:MDP_eg}
	\end{figure}

\begin{figure}[H]
	\centering
	\begin{tabular}{cc}
\includegraphics[width=0.4\linewidth]{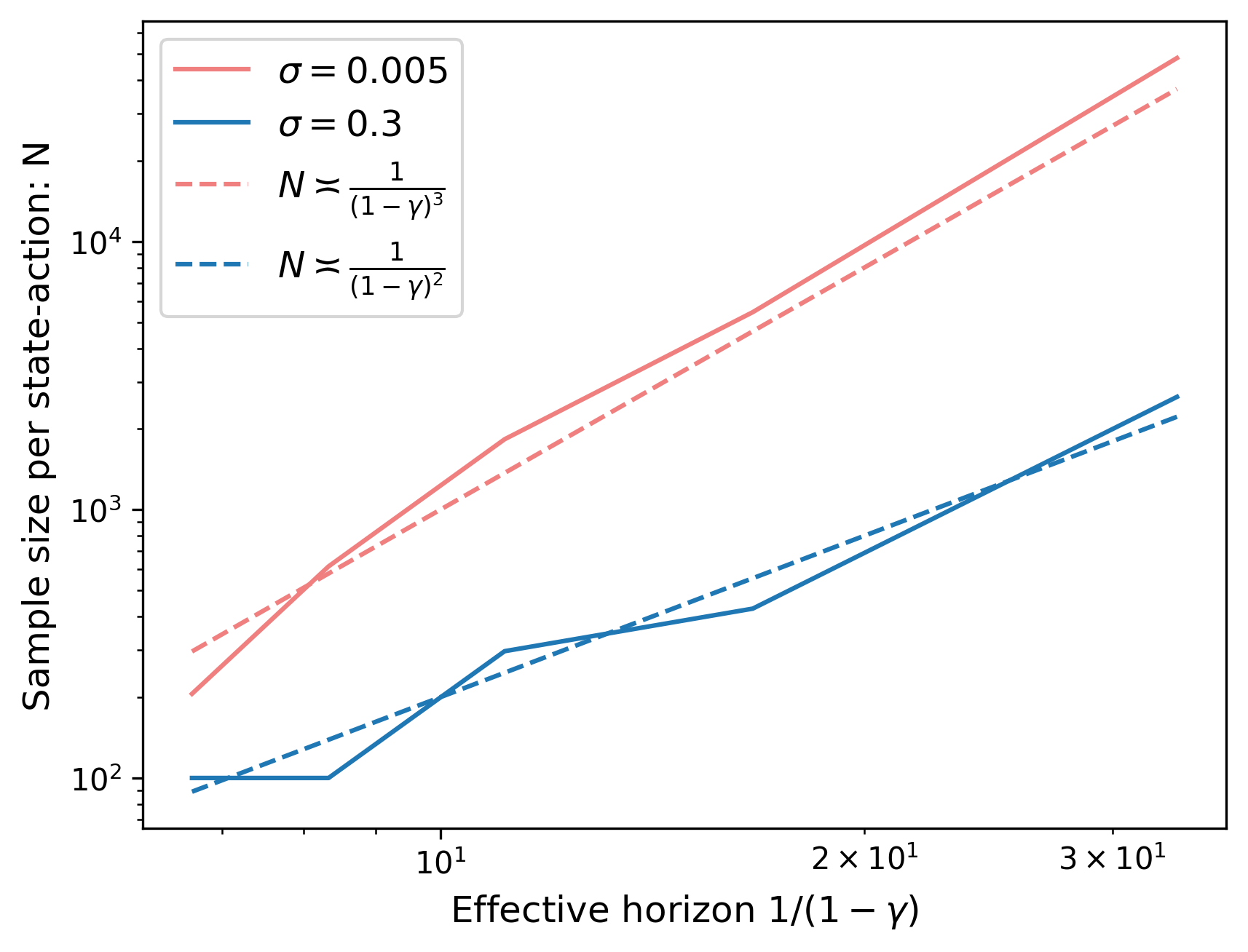} &  \includegraphics[width=0.405\linewidth]{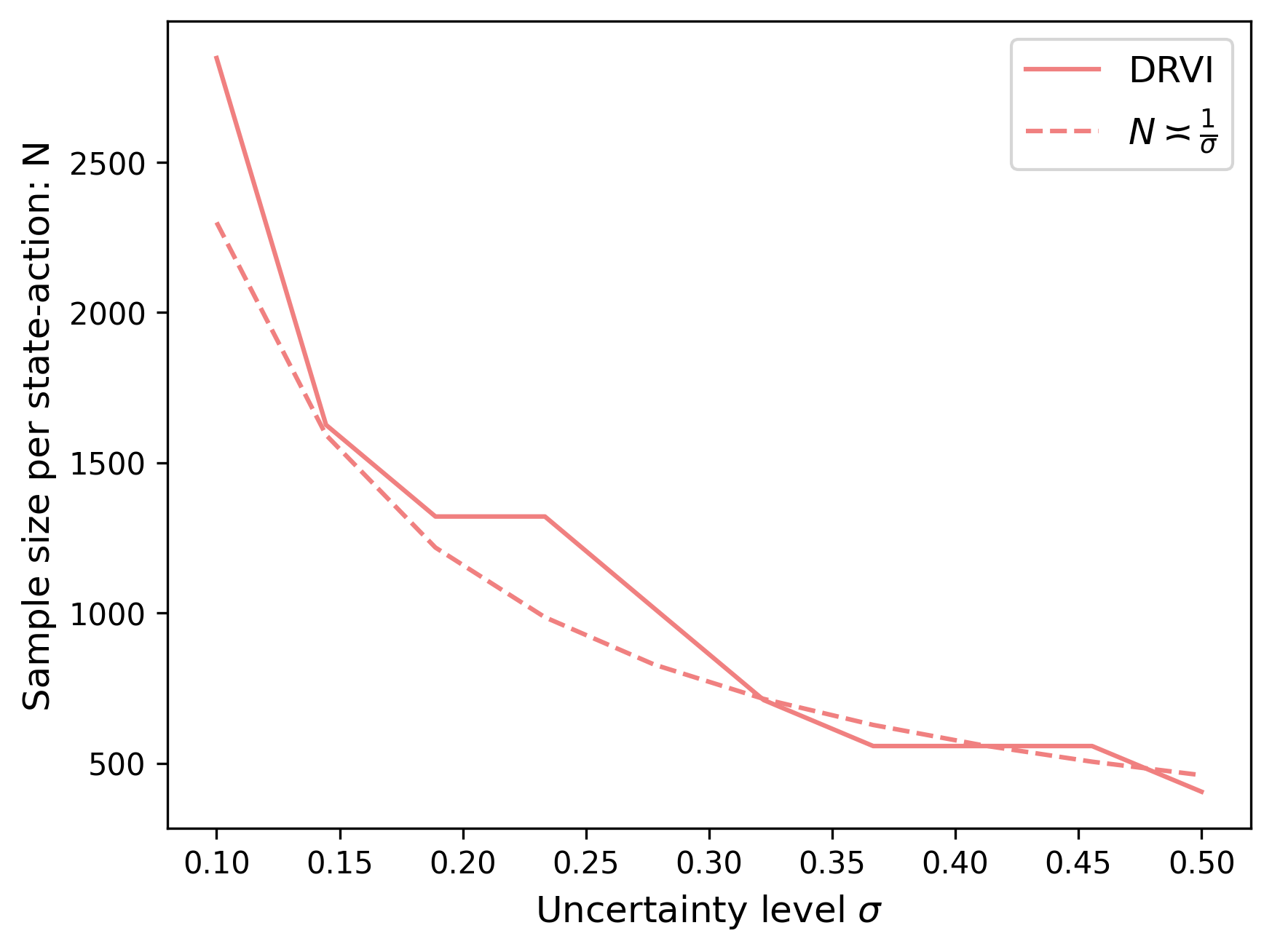} \\
 (a) TV distance, w.r.t. $1/(1-\gamma)$  &  (b) TV distance, w.r.t. $\sigma$   \\
 \includegraphics[width=0.4\linewidth]{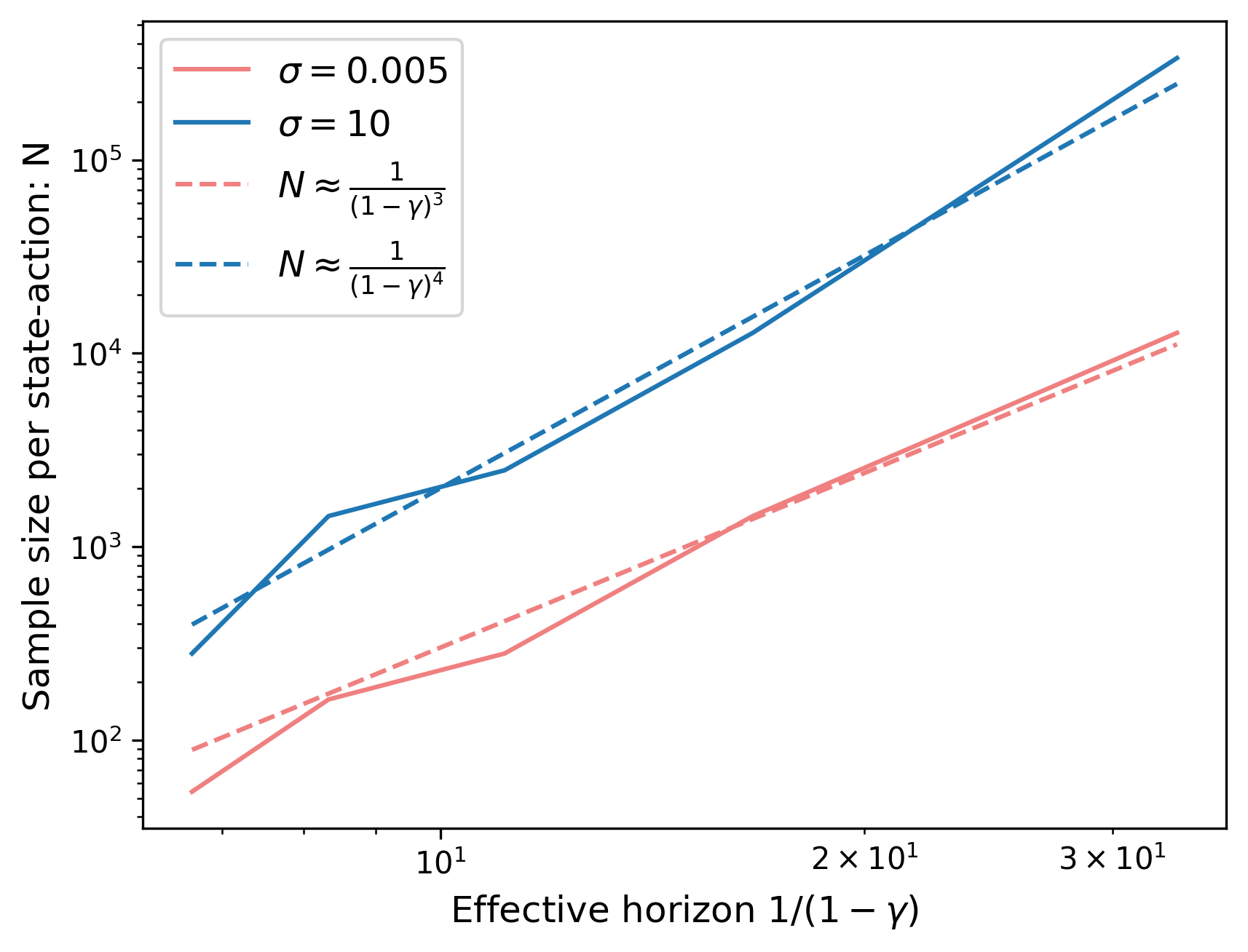} & \includegraphics[width=0.405\linewidth]{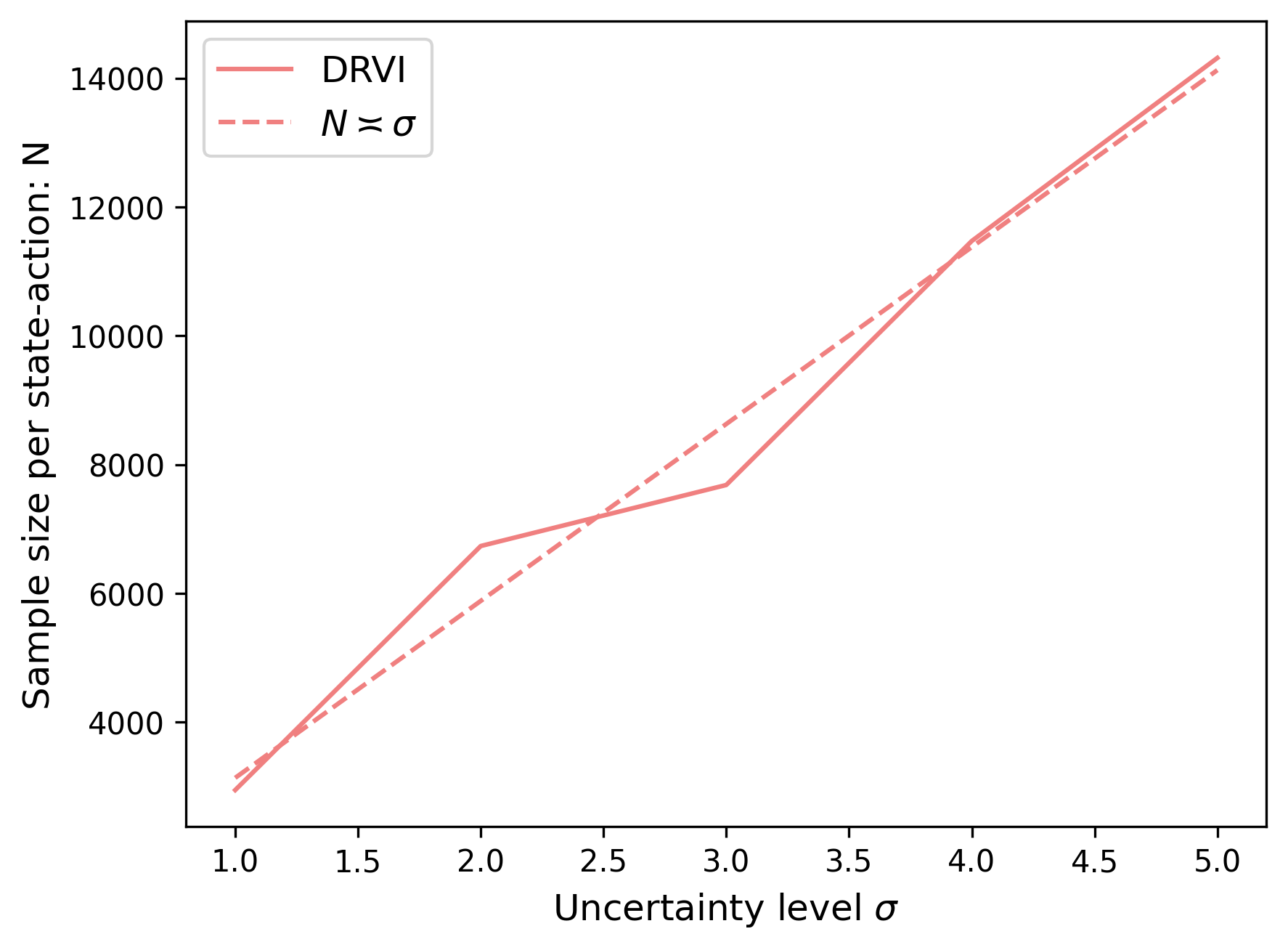}\\
 (c) $\chi^2$ divergence, w.r.t. $1/(1-\gamma)$  & (d) $\chi^2$ divergence, w.r.t. $\sigma$
	\end{tabular}
	\caption{Sample complexity of \DRVI with an uncertainty set under the TV distance (a-b) and the $\chi^2$ divergence (c-d), with respect to the effective horizon $1/(1-\gamma)$ and $\sigma$.}  
		\label{fig:numerical}
	\end{figure}

To further corroborate the theoretical results in Section~\ref{sec:theory}, we conduct numerical experiments to validate the sample complexity in both TV and $\chi^2$ divergence cases, shown in Figure~\ref{fig:numerical}. Specifically, we evaluate \DRVI (Algorithm~\ref{alg:cvi-dro-infinite}) over a specific robust MDP $\mathcal{M}_\phi= \{\cS,\cA, \gamma, \cU_\rho^{\ror}(P^\no), r\}$ (illustrated in Figure~\ref{fig:MDP_eg}) when the uncertainty level $\sigma$ varies. Here, $\mathcal{S} = \{0, 1\}$, $\mathcal{A} = \{0, 1\}$, the nominal transition kernel $P^0$ obeys $P^0(1|1,0) = P^0(1|1,1) =1, P^0(1|0,0) =p, P^0(1|0,1) =q$, and the reward $r(0,0) = r(0,1) =0, r(1,0) = r(1,1) = 1$. In all experiments, we generate $N$ samples per state-action pair from the generated model. For each point $(N, \gamma, \sigma)$, we conduct 200 Monte Carlo simulations and claim $N$ successfully attain $\varepsilon$ accuracy if the accuracy is achieved at least $190$ times.

\paragraph{Results.} For both cases, we examine the dependency with respect to the effective horizon $\frac{1}{1-\gamma}$ and uncertainty level $\sigma$, which dominate the distinction between the sample complexity of robust RL and that of standard RL (see Figure~\ref{fig:comopare-to-prior}). We fix $\varepsilon= 0.13$ (a randomly chosen small value) for both cases.  For the TV case, we set $p = 1.05 \max(1-\gamma, \sigma)$, and $q = p - 16(1-\gamma)\max(1-\gamma, \sigma)\varepsilon$ inspired by the proof of our lower bound (cf.~Theorem~\ref{thm:l1-lower-bound}).   Figure~\ref{fig:numerical}(a) shows that as $\frac{1}{1-\gamma}$ varies, the numerical sample complexity per state-action pair $N$ scales on the order of $\frac{1}{(1-\gamma)^3}$ when the uncertainty level is small ($\sigma = 0.005$), while on the order of  $\frac{1}{(1-\gamma)^2}$ when the uncertainty level is large ($\sigma = 0.3$). The results match the derived sample requirement dependency w.r.t. $\frac{1}{1-\gamma}$, namely $\widetilde{O} \left(\frac{1}{ (1-\gamma)^2 \max\{1-\gamma, \ror\}} \right)$ (see Figure \ref{fig:comopare-to-prior}(a)). On the other end, with a fixed value of $\gamma = 0.95$, Figure~\ref{fig:numerical}(b) shows that the numerical sample complexity scales on the order of $1/\sigma$, again matching our theory.

For the case of $\chi^2$ divergence, inspired by the proof of our lower bound (cf.~Theorem~\ref{thm:chi2-lower-bound}), we set 
\begin{align}
\begin{cases}
q = 1- \gamma, \quad p = \min\{q + 40\varepsilon (1-\gamma)^2, 1\}   & \text{if} \quad \sigma \leq 1 -\gamma, \\
q = \frac{\sigma}{1+\sigma}, \quad p = \min \{ q + \frac{130 \varepsilon(1-\gamma)^2}{1+\sigma} , 1 \} & \text{otherwise}. 
\end{cases}
\end{align}
Figure~\ref{fig:numerical}(c) shows that as $\frac{1}{1-\gamma}$ varies, the numerical sample complexity per state-action pair $N$ scales on the order of $\frac{1}{(1-\gamma)^3}$ when the uncertainty level is small ($\sigma = 0.005$), while on the order of  $\frac{1}{(1-\gamma)^4}$ when the uncertainty level is large ($\sigma = 10$). The results match the derived upper bound of sample requirement dependency w.r.t. $\frac{1}{1-\gamma}$, i.e., $\widetilde{O} \left(\frac{SA}{(1-\gamma)^3 \varepsilon^2} \left(1+\frac{ \max\{\sqrt{\ror},  \ror \}}{1-\gamma} \right)\right)$ (see Figure \ref{fig:comopare-to-prior}(b)). In addition, with a fixed value of $\gamma = 0.95$, Figure~\ref{fig:numerical}(d) demonstrates that the numerical sample complexity increases linearly w.r.t. the uncertainty level $\sigma$ when $\sigma$ is large (in the range $(1,\infty)$), which matches our theoretical results.


\section{Offline distributionally robust RL with uniform coverage}\label{sec:offline-main}
In this section, we extend our theoretical analysis to broader sampling mechanism scenarios with offline datasets. We first specify the offline settings as below.

\paragraph{Offline/batch dataset.}
Suppose that we observe a batch/historical dataset $\cD^{\mathsf{b}} = \{(s_i, a_i, r_i, s_i')\}_{1\leq i \leq N_{\mathsf{b}}}$ consisting of $N_{\mathsf{b}}$ sample transitions generated independently. Specifically, the state-action pair $(s_i,a_i)$ is drawn from some behavior distribution $\mu^{\mathsf{b}} \in \Delta(\cS\times \cA)$, followed by a next state $s_i'$ drawn over the nominal transition kernel $P^\no$, i.e.,
\begin{align}
\label{eq:infinite-batch-set-generation}
	(s_i, a_i) \overset{\text{i.i.d.}}{\sim} \mu^{\mathsf{b}}  \quad \text{and} \quad s_i' \overset{\text{i.i.d.}}{\sim} P^\no(\cdot \mymid s_i, a_i), \qquad 1\leq i \leq N_{\mathsf{b}}.
\end{align}
We consider uniform coverage historical dataset that is widely studied in offline settings for both standard RL and robust RL \citep{liao2022batch,chen2019information,jin2020provably,zhou2021finite,yang2021towards}, specified in the following assumption.
\begin{assumption}\label{assumption-offline}
Suppose the historical dataset $\cD^{\mathsf{b}}$ obeys
\begin{align}
	\mu_{\min} \defn \min_{(s,a) \in \cS \times \cA} \mu^{\mathsf{b}}(s,a) >0.
\end{align}
\end{assumption}

Armed with the above dataset $\cD^{\mathsf{b}}$, the empirical nominal transition kernel $\widehat{P}^\no \in \mathbb{R}^{SA\times S}$ can be constructed through \eqref{eq:empirical-P-infinite} analogously. Then in such offline setting, we introduce the sample complexity upper bounds for \DRVI and information-theoretical lower bounds in the cases of TV or $\chi^2$ divergence respectively. The proof of the following  corollaries are postponed to Appendix~\ref{proof:corollary}.

\subsection{The case of TV distance}
With above historical dataset $\cD^{\mathsf{b}}$ in hand, we achieve the following corollary implied by Theorem~\ref{thm:l1-upper-bound}.

\begin{corollary}[Upper bound under TV distance]\label{thm:l1-upper-bound-offline} 
	Let the uncertainty set be $\unb_\rho^\ror(\cdot) = \cU^{\ror}_{\mathsf{TV}}(\cdot)$ defined in \eqref{eq:tv-distance}, and $C_3, C_4>0$ be some large enough universal constants. Consider any discount factor $\gamma \in \left[\frac{1}{4},1 \right)$, 
 uncertainty level $\ror\in (0,1)$, and $\delta \in (0,1)$. 
	Let $\widehat{\pi}$ be the output policy of Algorithm~\ref{alg:cvi-dro-infinite} after $T = C_3 \log \big(\frac{N_{\mathsf{b}}}{1-\gamma}\big)$ iterations, based on a dataset $\cD^{\mathsf{b}}$ satisfying Assumption~\ref{assumption-offline}. 
	Then with probability at least $1-\delta$, one has
\begin{align}
	\forall s\in\cS: \quad V^{\star, \ror}(s) - V^{\widehat{\pi}, \ror}(s) \leq \varepsilon
\end{align}
for any $\varepsilon \in \left(0, \sqrt{1/\max\{1-\gamma, \ror\}} \right]$,
as long as the total number of samples obeys
\begin{align}
	N_{\mathsf{b}} \geq  \frac{C_4 }{\mu_{\min} (1-\gamma)^2 \max\{1-\gamma, \ror\} \varepsilon^2}\log\left(\frac{N_{\mathsf{b}}SA}{(1-\gamma)\delta}\right).
\end{align}
\end{corollary}

We also derive a lower bound in the offline setting by adapting Theorem~\ref{thm:l1-lower-bound}.
\begin{corollary}[Lower bound under TV distance]\label{thm:l1-lower-bound-offline} 
Let the uncertainty set be $\unb_\rho^\ror(\cdot) = \cU^{\ror}_{\mathsf{TV}}(\cdot)$ defined in \eqref{eq:tv-distance}. Consider any tuple $(S, \gamma, \ror, \varepsilon, \mu_{\min})$ that obeys $\mu_{\min} >0$, $\ror\in (0, 1 - c_0]$ with $0 <c_0 \leq \frac{1}{8}$ being any small enough positive constant, $\gamma\in\left[ \frac{1}{2}, 1\right)$, and $\varepsilon \in \big(0, \frac{c_0}{256(1-\gamma)} \big]$. We can construct two infinite-horizon RMDPs $\cM_0, \cM_1$, an initial state distribution $\varphi$, and a dataset with $N_{\mathsf{b}}$ samples satisfying Assumption~\ref{assumption-offline} (for $\cM_0$ and $\cM_1$ respectively) such that
	\[
	\inf_{\widehat{\pi}}\max\left\{ \mathbb{P}_{0}\big( V^{\star,\ror}(\varphi)-V^{\widehat{\pi}, \ror}(\varphi)>\varepsilon\big), \,
	\mathbb{P}_{1}\big( V^{\star,\ror }(\varphi) - V^{\widehat{\pi},\ror}(\varphi) >\varepsilon\big)\right\} \geq\frac{1}{8},
\]
provided that
\[N_{\mathsf{b}} \leq \frac{c_0    \log 2  }{  8192 \mu_{\min}(1-\gamma)^2 \max\{ 1 -\gamma, \sigma\}\varepsilon^2}.\]
Here, the infimum is taken over all estimators $\widehat{\pi}$, and $\mathbb{P}_{0}$ (resp.~$\mathbb{P}_{1}$) denotes the probability
when the RMDP is $\mathcal{M}_{0}$ (resp.~$\mathcal{M}_{1}$).
\end{corollary}
 
\paragraph{Discussions.} In the offline setting with uniform coverage dataset (cf.~Assumption~\ref{assumption-offline}), 
Corollary~\ref{thm:l1-upper-bound} shows that \DRVI algorithm can find an $\varepsilon$-optimal policy with the following sample complexity
\begin{align}
\widetilde{O} \left(\frac{1}{ \mu_{\min}(1-\gamma)^2 \max\{1-\gamma, \ror\} \varepsilon^2} \right),
\end{align}
which is near minimax optimal with respect to all salient parameters (up to logarithmic factors) almost over the full range of the uncertainty level $\ror$, verified by the lower bound in Corollary~\ref{thm:l1-lower-bound}. Our sample complexity upper bound (Corollary~\ref{thm:l1-upper-bound}) significantly improves over the prior art $\widetilde{O}\left(\frac{S(2+\ror)^2}{ \mu_{\min} \ror^2 (1-\gamma)^4 \varepsilon^2} \right)$ \citep{yang2021towards} by at least a factor of $\frac{S}{(1-\gamma)^2}$, and even more than $\frac{S}{(1-\gamma)^3}$ when the uncertainty level $0 < \ror \lesssim 1- \gamma$ is small.

\subsection{The case of $\chi^2$ divergence} 
With uncertainty sets measured by the $\chi^2$ divergence, we obtain the following upper bounds for \DRVI and information-theoretical lower bounds, adapted from Theorem~\ref{thm:l2-upper-bound} and Theorem~\ref{thm:chi2-lower-bound} respectively.
\begin{corollary}[Upper bound under $\chi^2$ divergence]\label{thm:l2-upper-bound-offline}
	Let the uncertainty set be $\unb_\rho^\ror(\cdot) = \cU^{\ror}_{\chi^2}(\cdot)$ specified by the $\chi^2$ divergence (cf.~\eqref{eq:chi-squared-distance}), and $c_1, c_2>0$ be some large enough universal constants. Consider any uncertainty level $\ror\in (0, \infty)$, $\gamma\in [1/4,1)$ and $\delta \in(0,1)$. Given a  dataset $\cD^{\mathsf{b}}$ satisfying Assumption~\ref{assumption-offline}, with probability at least $1-\delta$, the output policy $\widehat{\pi}$ from Algorithm~\ref{alg:cvi-dro-infinite} with at most $T = c_1 \log \big( \frac{N_{\mathsf{b}}}{1-\gamma}\big)$ iterations yields
\begin{align}
	\forall s\in\cS: \quad V^{\star, \ror}(s) - V^{\widehat{\pi}, \ror}(s) \leq \varepsilon
\end{align}
for any $\varepsilon\in \big(0,\frac{1}{1-\gamma} \big]$, as long as the total number of  samples obeying
\begin{align}
	N_{\mathsf{b}} \geq \frac{c_2 \left(1+\frac{\sqrt{\ror} + \ror}{1-\gamma} \right) }{\mu_{\min}(1-\gamma)^3 \varepsilon^2} \log\left(\frac{N_\mathsf{b}}{\mu_{\min}\delta}\right).
\end{align} 
\end{corollary}

\begin{corollary}[Lower bound under $\chi^2$ divergence]\label{thm:chi2-lower-bound-offline}
Let the uncertainty set be $\unb_\rho^\ror(\cdot) = \cU^{\ror}_{\chi^2}(\cdot)$, and $c_3, c_4>0$ be some universal constants. Consider any tuple $(S,\gamma, \ror, \varepsilon, \mu_{\min})$ obeying $\mu_{\min} >0$, $\gamma\in [\frac{3}{4}, 1)$, $\ror \in (0,\infty)$, and
\begin{align}
\varepsilon &\leq \frac{c_3}{(1-\gamma)}.
\end{align}
Then we can construct two infinite-horizon RMDPs $\cM_0, \cM_1$, an initial state distribution $\varphi$, and a dataset with $N_{\mathsf{b}}$  independent samples satisfying Assumption~\ref{assumption-offline} over the nominal transition kernel (for $\cM_0$ and $\cM_1$ respectively), such that
\begin{align}
	\inf_{\widehat{\pi}}\max\left\{ \mathbb{P}_{0}\big( V^{\star,\ror}(\varphi)-V^{\widehat{\pi}, \ror}(\varphi)>\varepsilon\big), \,
	\mathbb{P}_{1}\big( V^{\star,\ror }(\varphi) - V^{\widehat{\pi},\ror}(\varphi) >\varepsilon\big)\right\} \geq\frac{1}{8}, \label{eq:lower-bound-chi2-all}
\end{align}
provided that the total number of samples
\begin{align} \label{eq:final-chi2-lower-bound}
N_{\mathsf{b}}\leq \frac{c_4}{ \mu_{\min}(1-\gamma)^3  \varepsilon^2} \left(1+\frac{\sigma}{1-\gamma}\right).
\end{align}
\end{corollary}

\paragraph{Discussions.}
Corollary~\ref{thm:l2-upper-bound-offline} indicates that in the offline setting with uniform coverage dataset (cf.~Assumption~\ref{assumption-offline}), \DRVI can achieve $\varepsilon$-accuracy for RMDPs under the $\chi^2$ divergence with a total number of samples on the order of
\begin{align}\label{eq:chi2-final-samples-offline}
 \widetilde{O} \left(\frac{1+\frac{\sqrt{\ror} + \ror}{1-\gamma}  }{\mu_{\min}(1-\gamma)^3 \varepsilon^2}\right).
\end{align}
The above upper bound is relatively tight, since it matches the lower bound derived in Corollary~\ref{thm:chi2-lower-bound-offline} when the uncertainty level $\ror \lesssim  (1-\gamma)^2$ and $\ror \gtrsim 1$. As the uncertainty level increases and $\ror \gtrsim 1$, the matching sample complexity upper and lower bounds are on the order of $\frac{\ror }{   \mu_{\min}  (1-\gamma)^4\varepsilon^2}$, accurately reflecting the linear dependency on $\ror$. In addition, it significantly improves upon the prior art $\widetilde{O}\left(\frac{S(1+\ror)^2}{ \mu_{\min} (\sqrt{1+\ror}-1)^2 (1-\gamma)^4 \varepsilon^2} \right)$ \citep{yang2021towards} by at least a factor of $S(1+\ror)$.

\section{Other related works}\label{sec:related}

This section briefly discusses a small sample of other related works. 
We limit our discussions primarily to provable RL algorithms in the tabular setting with finite state and action spaces, which are most related to the current paper.

\paragraph{Finite-sample guarantees for standard RL.}
A surge of recent research has utilized the toolkit from high-dimensional probability/statistics to investigate the performance of standard RL algorithms in non-asymptotic settings. There has been a considerable amount of research into non-asymptotic sample analysis of standard RL for a variety of settings; partial examples include, but are not limited to, the works via probably approximately correct (PAC) bounds for the generative model setting \citep{kearns1999finite,beck2012error,li2022minimax,chen2020finite,azar2013minimax,sidford2018near,agarwal2020model,li2023q,li2020breaking,wainwright2019stochastic} and the offline setting \citep{liao2022batch,chen2019information,rashidinejad2021bridging,xie2021policy,yin2021near_double,shi2022pessimistic,li2024settling,jin2021pessimism,yan2022efficacy,woo2024federated,uehara2022review}, as well as the online setting via both regret-based and PAC-based analyses \citep{jin2018q,bai2019provably,li2021breaking,zhang2020almost,dong2019q,jin2020reward,li2023minimax,jafarnia2020model,yang2021q,woo2023blessing}.

\paragraph{Robustness in RL.}
While standard RL has achieved remarkable success, 
current RL algorithms still have significant drawbacks in that the learned policy could be completely off if the deployed environment is subject to perturbation, model mismatch, or other structural changes. To address these challenges, an emerging line of works begin to address robustness of RL algorithms with respect to the uncertainty or perturbation over different components of MDPs --- state, action, reward, and the transition kernel; see \citet{moos2022robust} for a recent review. Besides the framework of distributionally robust MDPs (RMDPs) \citep{iyengar2005robust} adopted by this work, to promote robustness in RL, there exist various other works including but not limited to \citet{zhang2020robust,zhang2021robust,han2022solution,qiaoben2021strategically,sun2021exploring,xiong2022defending} investigating the robustness w.r.t. state uncertainty, where the agent's policy is chosen based on a perturbed observation generated from the state by adding restricted noise or adversarial attack. Besides, \citet{tessler2019action,tan2020robustifying} considered the robustness w.r.t. the uncertainty of the action, namely, the action is possibly distorted by an adversarial agent abruptly or smoothly, and \citet{ding2023seeing} tackles robustness against spurious correlations..

\paragraph{Distributionally robust RL.}  
Rooted in the literature of distributionally robust optimization, which has primarily been investigated in the context of supervised learning \citep{rahimian2019distributionally,gao2020finite,bertsimas2018data,duchi2021learning,blanchet2019quantifying}, distributionally robust dynamic programming and RMDPs have  attracted considerable attention recently \citep{iyengar2005robust,xu2012distributionally,wolff2012robust,kaufman2013robust,ho2018fast,smirnova2019distributionally,ho2021partial,goyal2022robust,derman2020distributional,tamar2014scaling,badrinath2021robust}. In the context of RMDPs, both empirical and theoretical studies have been widely conducted, although most prior theoretical analyses focus on planning with an exact knowledge of the uncertainty set \citep{iyengar2005robust,xu2012distributionally,tamar2014scaling}, or are asymptotic in nature \citep{roy2017reinforcement}.

Resorting to the tools of high-dimensional statistics, various recent works begin to shift attention to understand the finite-sample performance of provable robust RL algorithms, under diverse data generating mechanisms and forms of the uncertainty set over the transition kernel. Besides the infinite-horizon setting, finite-sample complexity bounds for RMDPs with the TV distance and the $\chi^2$ divergence are also developed for the finite-horizon setting in \citet{xu2023improved,dong2022online,lu2024distributionally}.
In addition, many other forms of uncertainty sets have been considered. For example,
\citet{wang2021online} considered a R-contamination uncertain set and proposed a provable robust Q-learning algorithm for the online setting with similar guarantees as standard MDPs. The KL divergence is another popular choice widely considered, where \citet{yang2021towards,panaganti2021sample,zhou2021finite,shi2022distributionally,xu2023improved,wang2023finite,blanchet2023double,liu2022distributionally,wang2023sample,liang2023single,wang2023robust} investigated the sample complexity of both model-based and model-free algorithms under the simulator, offline settings, or single-trajectory setting.
\citet{xu2023improved} considered a variety of uncertainty sets including one associated with Wasserstein distance. \citet{badrinath2021robust,ramesh2023distributionally,panaganti2022robust,ma2022distributionally,he2024sample,liu2024minimax,liu2024distributionally} considered  function approximation settings. Moreover, various other related issues have been explored such as the difference of various uncertainty types \citep{wang2023foundation}, the iteration complexity of the policy-based methods \citep{li2022first,kumar2023policy,li2023first}, the case when the uncertainty level is instance-dependent small enough \citep{clavier2023towards}, regularization-based robust RL \citep{yang2023avoiding,zhang2023regularized}, and distributionally robust optimization for offline RL \citep{panaganti2023bridging}.

\section{Discussions}\label{sec:discussion}

This work has developed improved sample complexity bounds for learning RMDPs when the uncertainty set is measured via the TV distance or the $\chi^2$ divergence, 
assuming availability of a generative model. Our results have not only strengthened the prior art in both the upper and lower bounds, but have also unlocked curious insights into how the quest for distributional robustness impacts the sample complexity. 
As a key takeaway of this paper, RMDPs are not necessarily harder nor easier to learn than standard MDPs, 
as the answer depends --- in a rather subtle manner --- on the specific choice of  the uncertainty set. 
For the case w.r.t.~the TV distance, 
we have settled the minimax sample complexity for RMDPs, which is never larger than that required to learn standard MDPs. Regarding the case w.r.t.~the $\chi^2$ divergence, we have uncovered that learning RMDPs can oftentimes be provably harder than the standard MDP counterpart.  
All in all, our findings help raise awareness that the choice of the uncertainty set not only represents a preference in robustness, but also exerts fundamental influences upon the intrinsic statistical complexity.

Moving forward, our work opens up numerous avenues for future studies, and we point out a few below. 

\begin{itemize}

\item {\em Extensions to the finite-horizon and multi-agent settings.} It is likely that our current analysis framework can be extended to tackle finite-horizon and multi-agent RMDPs \citep{shi2024sample,shi2024breaking}, which would help complete our understanding for the tabular cases.


\item {\em A unified theory for other families of uncertainty sets.} Our work raises an interesting question concerning how the geometry of the uncertainty sets intervenes the sample complexity. Characterizing the tight sample complexity for RMDPs under a more general family of uncertainty sets --- such as using $\ell_p$ distance or $f$-divergence, as well as $s$-rectangular sets --- would be highly desirable; see some recent developments in \citet{clavier2024near,li2025near}, for example.

\item {\em Instance-dependent sample complexity analyses.} We note that we focus on understanding the minimax-optimal sample complexity of RMDPs, which might be rather pessimistic. When consider a given MDP, the feasible and reasonable magnitude of the uncertainty level $\sigma$ is limited by a certain {\em instance-dependent} range. It will be desirable to study instance-dependent sample complexity of RMDPs, which might shed more light on guiding the practice. 

\item {\em Adaptation to function approximation.} Last but not least, it will be of great interest to study the interaction between distributional robustness and function approximation, in models such as linear MDPs and beyond \citep{blanchet2023double,wang2025sample}.

\end{itemize}

\section*{Acknowledgement}
 
The work of L. Shi and Y. Chi is supported in part by the grants ONR N00014-19-1-2404, NSF CCF-2106778, DMS-2134080, and CNS-2148212. 
L. Shi is also gratefully supported by the Leo Finzi Memorial Fellowship, Wei Shen and Xuehong Zhang Presidential Fellowship, and
Liang Ji-Dian Graduate Fellowship at Carnegie Mellon University, and the Resnick Institute and Computing, Data, and Society
Postdoctoral Fellowship at California Institute of Technology. G. Li is supported in part by the Chinese University of Hong Kong Direct Grant for Research and the Hong Kong Research Grants Council ECS 2191363. The work of Y.~Wei is supported in part by the NSF grants DMS-2147546/2015447, CAREER award DMS-2143215, CCF-2106778, and the Google Research Scholar Award. The work of Y.~Chen is supported in part by the Alfred P.~Sloan Research Fellowship, the Google Research Scholar Award, the AFOSR grant FA9550-22-1-0198, 
the ONR grant N00014-22-1-2354,  and the NSF grants CCF-2221009 and CCF-1907661.
The authors also acknowledge Mengdi Xu, Zuxin Liu and He Wang for valuable discussions. 

\bibliography{../bibfileRL,../bibfileDRO}
\bibliographystyle{apalike}


\appendix


\section{Proof of the preliminaries}\label{sec:robust_bellman_properties-proof}

Before moving forward, let us introduce some additional definitions and facts that will be useful throughout the appendix.

\paragraph{Kullback-Leibler (KL) divergence.}  First, for any two distributions $P$ and $Q$, we denote by $\mathsf{KL}(P \parallel Q)$ the Kullback-Leibler (KL) divergence of $P$ and $Q$. 
Letting $\mathsf{Ber}(p)$ be the Bernoulli distribution with mean $p$,  
we also introduce 
\begin{align}
	\mathsf{KL}(p \parallel q) \coloneqq
	p\log\frac{p}{q}+(1-p)\log\frac{1-p}{1-q}  
	\quad \text{and} \quad 
	\chi^2 (p \parallel q) \coloneqq \frac{(p-q)^2}{q} +  \frac{(p-q)^2}{1-q} = \frac{(p-q)^2}{q(1-q)},
	\label{eq:defn-KL-bernoulli}
\end{align}
which represent respectively the KL divergence and the $\chi^2$ divergence of $\mathsf{Ber}(p)$ from $\mathsf{Ber}(q)$ \citep{tsybakov2009introduction}.

The following lemma bounds the Lipschitz constant of the variance function.
\begin{lemma}\label{eq:tv-auxiliary-lemma}
Consider any $  {0} \leq V_1, V_2 \leq \frac{1}{1-\gamma} $ obeying $\|V_1 - V_2\|_\infty \leq x$ and any probability vector $P\in\Delta(\cS)$ (here $\Delta(\cS)$ represents the simplex over the state space $\cS$), one has
\begin{align}
 \left| \mathrm{Var}_P(V_1) - \mathrm{Var}_P(V_2) \right| \leq \frac{2x}{(1-\gamma)}.
\end{align}
\end{lemma}
{\em Proof of Lemma~\ref{eq:tv-auxiliary-lemma}:} It is immediate to check that
\begin{align}
\left| \mathrm{Var}_P(V_1) - \mathrm{Var}_P(V_2) \right| & = \left| P ( V_1 \circ V_1 )  - (P V_1) \circ (PV_1) - P (V_2 \circ V_2)  + (P V_2) \circ (P V_2)\right| \notag \\
& \leq \left| P \big( V_1 \circ V_1 - V_2 \circ V_2 \big) \right| + \left| (P V_1 + P V_2) P(V_1 - V_2)\right| \notag \\
& \leq  2\|V_1 + V_2\|_\infty \|V_1 - V_2\|_\infty \leq \frac{2x}{(1-\gamma)}.
\end{align}
where the penultimate inequality holds by the triangle inequality.

\subsection{Proof of Lemma~\ref{lemma:tv-dual-form} and Lemma~\ref{lem:dual-vi-l2-norm}}

\paragraph{Proof of Lemma~\ref{lemma:tv-dual-form}.}
To begin with, applying \citep[Lemma~4.3]{iyengar2005robust}, the term of interest obeys 
\begin{align}
	\inf_{ \cP \in \unb^{\ror}(P)} \cP V  &= \max_{\mu \in \mathbb{R}^S, \mu\geq 0} \left\{P \left(V - \mu\right) - \ror\left(\max_{s'}\left\{V(s') - \mu(s')\right\} - \min_{s'}\left\{V(s') - \mu(s')\right\} \right)\right\}, \label{eq:vi-l1norm-raw}
\end{align}
where $\mu(s')$ represents the $s'$-th entry of $\mu\in\mathbb{R}^S$. Denoting $\mu^\star$ as the optimal dual solution, taking $\alpha = \max_{s'}\left\{V(s') - \mu^\star(s')\right\}$, it is easily verified that $\mu^\star$ obeys
\begin{align}
\mu^\star(s) = \begin{cases} V(s) - \alpha, & \text{if } V(s) > \alpha \\
0, & \text{otherwise}.
\end{cases} \label{eq:l1-dual-transfer}
\end{align}

Therefore, \eqref{eq:vi-l1norm-raw} can be solved by optimizing $\alpha$ as below \citep[Lemma~4.3]{iyengar2005robust}:
\begin{align}
	\inf_{ \cP \in \unb^{\ror}(P)} \cP V = \max_{\alpha\in\left[\min_s V(s), \max_s V(s)\right]}  \left\{ P \left[V\right]_{\alpha} - \ror \left(\alpha - \min_{s'}\left[V\right]_{\alpha}(s') \right) \right\}.
	 \label{eq:l1-dual-vi-alpha}
\end{align}

\paragraph{Proof of Lemma~\ref{lem:dual-vi-l2-norm}.}
Due to strong duality \citep[Lemma~4.2]{iyengar2005robust}, it holds that 
\begin{align}
	\inf_{ \cP \in \unb^{\ror}(P)} \cP V &= \max_{\mu \in \mathbb{R}^S, \mu\geq 0} \left\{P \left(V -\mu \right) - \sqrt{\sigma \mathsf{Var}_{P}\left(V -\mu \right) } \right\} , \label{eq:vi-l2norm}
\end{align}
and the optimal $\mu^\star$ obeys
\begin{align}
\mu^\star(s) = \begin{cases} V(s) - \alpha, & \text{if } V (s) > \alpha \\
0, & \text{otherwise}.
\end{cases} \label{eq:l2-dual-transfer}
\end{align}
for some $\alpha \in [\min_s V(s), \max_s V(s)]$.
As a result, solving \eqref{eq:vi-l2norm} is equivalent to optimizing the scalar $\alpha$ as below:
\begin{align}
\inf_{ \cP \in \unb^{\ror}(P)} \cP V = \max_{\alpha\in [\min_s V(s), \max_s V(s)]} \left\{ P [V]_\alpha - \sqrt{\sigma \mathsf{Var}_{P}\left([V]_\alpha\right) } \right\}.
\end{align}

\subsection{Proof of Lemma~\ref{lem:infinite-converge}}\label{proof:lem:infinite-converge}

Applying the $\gamma$-contraction property in Lemma~\ref{lem:contration-of-T} directly yields that for any $t\geq0$,
\begin{align}
	\| \widehat{Q}_t -\widehat{Q}^{\star,\ror}\|_\infty = \big\| \that^\ror(\widehat{Q}_{t-1}) - \that^\ror(\widehat{Q}^{\star,\ror}) \big\|_\infty & \leq \gamma \big\| \widehat{Q}_{t-1} -\widehat{Q}^{\star,\ror} \big\|_\infty \nonumber \\
	& \leq \cdots \leq \gamma^t \big\| \widehat{Q}_0 -\widehat{Q}^{\star,\ror} \big\|_\infty = \gamma^t \big\| \widehat{Q}^{\star,\ror} \big\|_\infty \leq \frac{\gamma^t}{1-\gamma}, \nonumber
\end{align}
where the last inequality holds by the fact $\| \widehat{Q}^{\star,\ror}\|_\infty \leq \frac{1}{1-\gamma}$ (see Lemma~\ref{lem:contration-of-T}).
In addition,
\begin{align*}
	\big\| \widehat{V}_t -\widehat{V}^{\star,\ror} \big\|_\infty & = \max_{s\in \cS} \Big\| \max_{a\in\cA} \widehat{Q}_t (s,a) - \max_{a\in\cA}\widehat{Q}^{\star,\ror} (s,a) \Big\|_\infty \leq \big\| \widehat{Q}_t -\widehat{Q}^{\star,\ror} \big\|_\infty \leq \frac{\gamma^t}{1-\gamma}, 
\end{align*}
where the penultimate inequality holds by the maximum operator is $1$-Lipschitz. This completes the proof of \eqref{eq:converge1}.

We now move to establish \eqref{eq:V-result}. 
Note that 
there exists at least one state $s_0\in \cS$ that is associated with the maximum of the value gap, i.e.,
\begin{align*}
 \big\|\widehat{V}^{\star,\ror} - \widehat{V}^{\widehat{\pi},\ror} \big\|_\infty = \widehat{V}^{\star,\ror}(s_0) - \widehat{V}^{\widehat{\pi},\ror}(s_0) \geq  \widehat{V}^{\star,\ror}(s) - \widehat{V}^{\widehat{\pi},\ror}(s), \qquad \forall s\in\cS.
\end{align*}
Recall $\widehat{\pi}^\star$ is the optimal robust policy for the empirical RMDP $\widehat{\cM}_{\mathsf{rob}}$. For convenience, we denote $a_1 = \widehat{\pi}^\star(s_0)$ and $a_2 = \widehat{\pi}(s_0)$. Then, since $\widehat{\pi}$ is the greedy policy w.r.t. $\widehat{Q}_T$, one has
\begin{align}
	 r(s_0, a_1) + \gamma\inf_{\cP\in \unb^{\ror}(\widehat{P}^{\no}_{s_0,a_1})} \cP \widehat{V}_{T-1} = \widehat{Q}_T (s_0, a_1)  \leq \widehat{Q}_T (s_0, a_2) = r(s_0, a_2) + \gamma\inf_{ \cP\in \unb^{\ror}(\widehat{P}^{\no}_{s_0,a_2})} \cP \widehat{V}_{T-1} .
\end{align}
Recalling the notation in \eqref{eq:inf-p-special}, the above fact and \eqref{eq:V-result} altogether yield
\begin{align}
r(s_0, a_1) + \gamma \pmhat_{s_0,a_1}^{\widehat{V}_{T-1}} \left(\widehat{V}^{\star,\ror} -\varepsilon_{\mathsf{opt}}   {1} \right) &\leq r(s_0, a_1) + \gamma\pmhat_{s_0,a_1}^{\widehat{V}_{T-1}}  \widehat{V}_{T-1} \nonumber \\
&\leq r(s_0, a_2) + \gamma\inf_{\cP\in \unb^{\ror}(\widehat{P}^{\no}_{s_0,a_2})} \cP \widehat{V}_{T-1} \nonumber \\
&\overset{\mathrm{(i)}}{\leq} r(s_0, a_2) + \gamma \widehat{P}_{s_0, a_2}^{\widehat{V}^{\widehat{\pi}, \ror}} \widehat{V}_{T-1} \nonumber \\
&\leq r(s_0, a_2) + \gamma \widehat{P}_{s_0, a_2}^{\widehat{V}^{\widehat{\pi}, \ror}} \left(\widehat{V}^{\star, \ror}  + \varepsilon_{\mathsf{opt}}  {1} \right), \label{eq:reward-gap}
\end{align}
where (i) follows from the optimality criteria. The term of interest can be controlled as
\begin{align}\label{eq:value_function-gap}
&\big\|\widehat{V}^{\star,\ror} - \widehat{V}^{\widehat{\pi},\ror} \big\|_\infty \notag \\
&= \widehat{V}^{\star,\ror}(s_0) - \widehat{V}^{\widehat{\pi},\ror}(s_0)  \nonumber \\
&= r(s_0, a_1) + \gamma  \inf_{\cP\in \unb^{\ror}(\widehat{P}^{\no}_{s_0,a_1})} \cP \widehat{V}^{\star,\ror} - \bigg( r(s_0,a_2) + \gamma \inf_{\cP\in \unb^{\ror}(\widehat{P}^{\no}_{s_0,a_2})} \cP \widehat{V}^{\widehat{\pi},\ror} \bigg) \nonumber \\
&= r(s_0, a_1) - r(s_0,a_2) + \gamma\bigg( \inf_{\cP\in \unb^{\ror}(\widehat{P}^{\no}_{s_0,a_1})} \cP \widehat{V}^{\star,\ror} - \inf_{\cP\in \unb^{\ror}(\widehat{P}^{\no}_{s_0,a_2})} \cP  \widehat{V}^{\widehat{\pi},\ror}   \bigg) \nonumber \\
&\overset{\mathrm{(i)}}{\leq} 2\gamma \varepsilon_{\mathsf{opt}} + \gamma \bigg( \pmhat_{s_0, a_2}^{\widehat{V}^{\widehat{\pi}, \ror}} \widehat{V}^{\star, \ror}  - \pmhat_{s_0,a_1}^{\widehat{V}_{T-1}} \widehat{V}^{\star, \ror} +  \inf_{\cP\in \unb^{\ror}(\widehat{P}^{\no}_{s_0,a_1})} \cP \widehat{V}^{\star,\ror} - \inf_{\cP\in \unb^{\ror}(\widehat{P}^{\no}_{s_0,a_2})} \cP  \widehat{V}^{\widehat{\pi},\ror}  \bigg) \nonumber \\
&= 2\gamma \varepsilon_{\mathsf{opt}} + \gamma\bigg( \pmhat_{s_0, a_2}^{\widehat{V}^{\widehat{\pi}, \ror}} \widehat{V}^{\star, \ror}  - \inf_{\cP\in \unb^{\ror}(\widehat{P}^{\no}_{s_0,a_2})} \cP  \widehat{V}^{\widehat{\pi},\ror} \bigg) + \gamma \bigg( \inf_{\cP\in \unb^{\ror}(\widehat{P}^{\no}_{s_0,a_1})} \cP \widehat{V}^{\star,\ror} - \pmhat_{s_0,a_1}^{\widehat{V}_{T-1}} \widehat{V}^{\star, \ror}\bigg) \nonumber \\
&\overset{\mathrm{(ii)}}{\leq} 2\gamma \varepsilon_{\mathsf{opt}} + \gamma \pmhat_{s_0, a_2}^{\widehat{V}^{\widehat{\pi}, \ror}} \big( \widehat{V}^{\star, \ror} -  \widehat{V}^{\widehat{\pi},\ror} \big) + \gamma\bigg( \pmhat_{s_0,a_1}^{\widehat{V}_{T-1}} \widehat{V}^{\star,\ror} - \pmhat_{s_0,a_1}^{\widehat{V}_{T-1}} \widehat{V}^{\star, \ror} \bigg) \nonumber \\
&\leq 2\gamma \varepsilon_{\mathsf{opt}} + \gamma \big\| \widehat{V}^{\star, \ror} -  \widehat{V}^{\widehat{\pi},\ror} \big\|_\infty ,
\end{align}
where $\mathrm{(i)}$ holds by plugging in \eqref{eq:reward-gap}, and (ii) follows from $\inf_{\cP\in \unb^{\ror}(\widehat{P}^{\no}_{s_0,a_1})} \cP \widehat{V}^{\star,\ror} \le \cP\widehat{V}^{\star,\ror}$ for any $\cP\in \unb^{\ror}(\widehat{P}^{\no}_{s_0,a_1})$.
Rearranging \eqref{eq:value_function-gap} leads to
$$ \big\|\widehat{V}^{\star,\ror} - \widehat{V}^{\widehat{\pi},\ror} \big\|_\infty \leq \frac{2\gamma \varepsilon_{\mathsf{opt}}}{1-\gamma}  .$$


\section{Proof of the upper bound with TV distance: Theorem~\ref{thm:l1-upper-bound}}\label{proof:thm:l1-all}

\subsection{Technical lemmas}

We begin with a key lemma that is new and distinguishes robust MDPs with TV distance from standard MDPs , which plays a critical role in obtaining the sample complexity upper bound in Theorem~\ref{thm:l1-upper-bound}. This lemma concerns the dynamic range of the robust value function $V^{\pi,\ror}$ (cf.~\eqref{eq:robust-value-def}) for any fixed policy $\pi$, which produces tighter control than that in standard MDP (cf.~$\frac{1}{1-\gamma}$) when $\sigma$ is large. The proof is deferred to Appendix~\ref{proof:lemma:tv-key-value-range}.
	\begin{lemma}\label{lemma:tv-key-value-range} 
	 For any nominal transition kernel $P\in \mathbb{R}^{SA\times S}$, any fixed uncertainty level $\ror$, and any policy $\pi$, its corresponding robust value function $V^{\pi,\ror}$ (cf.~\eqref{eq:robust-value-def}) satisfies 
	\begin{align*}
		\max_{s\in\cS}V^{\pi,\ror}(s) - \min_{s\in \cS} V^{\pi,\ror}(s) \leq \frac{1}{\gamma \max\{ 1-\gamma, \sigma\}}.
	\end{align*}
	\end{lemma}
	 
With the above lemma in hand, we introduce the following lemma that is useful throughout this section, whose proof is postponed in Appendix~\ref{proof:lemma:tv-key1}.
\begin{lemma}\label{lemma:tv-key1}
Consider an MDP with  transition kernel matrix $P$ and reward function  $  {0} \leq r \leq 1$. For any policy $\pi$ and its associated state transition matrix $P_\pi \defn \Pi^\pi P$ and value function $0\leq V^{\pi, P}\leq \frac{1}{1-\gamma}$ (cf.~\eqref{eq:def_V}), one has 
\begin{align*}
\left(I - \gamma P_\pi \right)^{-1} \sqrt{\mathrm{Var}_{P_\pi}(V^{\pi, P} )} \leq \sqrt{\frac{8 (\max_s V^{\pi, P}(s)- \min_s V^{\pi, P}(s))  }{\gamma^2(1-\gamma)^2 }} 1.
\end{align*}
\end{lemma}
\begin{remark}
Compared to the results in \cite[Lemma 11]{li2020breaking}, Lemma~\ref{lemma:tv-key1} provides an improved upper bound, expressed in terms of $\max_s V^{\pi, P}(s) - \min_s V^{\pi, P}(s)$ rather than $\|V^{\pi, P}\|_\infty$.
\end{remark}

\subsection{Proof of Theorem~\ref{thm:l1-upper-bound}}
\label{proof:thm:l1-upper-bound}

Throughout this section, for any transition kernel $P$, the uncertainty set is taken as (see \eqref{eq:tv-distance})
\begin{align}\label{eq:consider-tv}
\unb^{\ror}(P) \defn \cU^{\ror}_{\mathsf{TV}}(P) = \otimes \; \cU^{\ror}_{\mathsf{TV}}(P_{s,a}),\qquad &\cU^{\ror}_{\mathsf{TV}}(P_{s,a}) \defn \Big\{ P'_{s,a} \in \Delta (\cS): \frac{1}{2}\left\| P'_{s,a} - P_{s,a}\right\|_1 \leq \ror \Big\}.
\end{align}

To control the two main terms in \eqref{eq:l1-decompose}, respectively, we first recall \eqref{eq:tv-v-star-two-terms} which holds for any uncertainty set:
\begin{align} 
\big\|\widehat{V}^{\pi^\star, \ror } - V^{ \pi^\star, \ror}\big\|_\infty &\leq \gamma \max\Big\{ 
\Big\| \Big(I - \gamma \Phatv^{\pi^\star, \widehat{V}}\Big)^{-1} \Big( \Phatv^{\pi^\star, V} V^{\pi^\star, \ror } - \Pv^{\pi^\star, V } V^{\pi^\star, \ror } \Big) \Big\|_\infty, \nonumber\\
	&\qquad\qquad  \Big\|\Big(I - \gamma \Phatv^{\pi^\star, V} \Big)^{-1} \Big( \Phatv^{\pi^\star, V} V^{\pi^\star, \ror } - \Pv^{\pi^\star, V } V^{\pi^\star, \ror } \Big) \Big\|_\infty \Big\}. 
\end{align}

\subsubsection{Controlling $\|\widehat{V}^{\pi^\star, \ror } - V^{ \pi^\star, \ror}\|_\infty$.}
We shall focus on controlling the two terms on the right hand side of the above results separately.

\paragraph{Step 1: controlling  $\|\widehat{V}^{\pi^\star, \ror } - V^{ \pi^\star, \ror}\|_\infty$: bounding the first term  in \eqref{eq:tv-v-star-two-terms}.}

To control the two terms in \eqref{eq:tv-v-star-two-terms}, we first introduce the following lemma whose proof is postponed to Appendix~\ref{proof:lemma:tv-dro-b-bound-star}. 

\begin{lemma}\label{lemma:tv-dro-b-bound-star}
Consider any $\delta \in (0,1)$. Setting $N \geq\log(\frac{18SAN}{\delta})$, with probability at least $1- \delta$, one has  
\begin{align}\label{eq:tv-dro-b-bound-star}
	\Big|\Phatv^{\pi^\star, V} V^{\pi^\star, \ror } - \Pv^{\pi^\star, V } V^{\pi^\star, \ror } \Big| &\leq 2 \sqrt{\frac{\log(\frac{18SAN}{\delta})}{N}} \sqrt{\mathrm{Var}_{P^{\pi^\star}}(V^{\star, \ror })} + \frac{\log(\frac{ 18SAN}{\delta})}{N(1-\gamma)} 1 \notag \\
	&\leq 3\sqrt{\frac{\log(\frac{18SA N}{\delta})}{(1-\gamma)^2N}} 1,
\end{align}
where $\mathrm{Var}_{P^{\pi^\star}}(V^{\star, \ror })$ is defined in \eqref{eq:defn-variance-vector}.
\end{lemma}
Armed with the above lemma, now we control the first term on the right hand side of \eqref{eq:tv-v-star-two-terms} as follows:
\begin{align}
&\Big(I - \gamma \Phatv^{\pi^\star, V} \Big)^{-1} \Big( \Phatv^{\pi^\star, V} V^{\pi^\star, \ror } - \Pv^{\pi^\star, V } V^{\pi^\star, \ror } \Big) \nonumber \\
& \overset{\mathrm{(i)}}{\leq} \Big(I - \gamma \Phatv^{\pi^\star, V} \Big)^{-1} \Big\| \Phatv^{\pi^\star, V} V^{\pi^\star, \ror } - \Pv^{\pi^\star, V } V^{\pi^\star, \ror } \Big\|_{\infty} \nonumber \\
& \overset{\mathrm{(ii)}}{\leq} \Big(I - \gamma \Phatv^{\pi^\star, V} \Big)^{-1} \bigg(2\sqrt{\frac{\log(\frac{18SAN}{\delta})}{N}} \sqrt{\mathrm{Var}_{P^{\pi^\star}}(V^{\star, \ror })} +  \frac{\log(\frac{18SAN}{\delta})}{N(1-\gamma)} 1 \bigg) \nonumber \\
& \leq \frac{\log(\frac{18SAN}{\delta})}{N(1-\gamma)}  \Big(I - \gamma \Phatv^{\pi^\star, V} \Big)^{-1} 1  + \underbrace{2\sqrt{\frac{\log(\frac{18SAN}{\delta})}{N}} \Big(I - \gamma \Phatv^{\pi^\star, V} \Big)^{-1} \sqrt{\mathrm{Var}_{\Phatv^{\pi^\star, V} }(V^{\star, \ror })}}_{=: \tvp_1}  \nonumber \\
& \quad + \underbrace{ 2\sqrt{\frac{\log(\frac{18SAN}{\delta})}{N}} \Big(I - \gamma \Phatv^{\pi^\star, V} \Big)^{-1} \sqrt{\left| \mathrm{Var}_{\widehat{P}^{\pi^\star}}(V^{\star, \ror }) - \mathrm{Var}_{\Phatv^{\pi^\star, V} }(V^{\star, \ror }) \right|} }_{=: \tvp_2} \nonumber \\
& \quad + \underbrace{2\sqrt{\frac{\log(\frac{18SAN}{\delta})}{N}} \Big(I - \gamma \Phatv^{\pi^\star, V} \Big)^{-1}   \Big(\sqrt{\mathrm{Var}_{P^{\pi^\star}}(V^{\star, \ror })} - \sqrt{\mathrm{Var}_{\widehat{P}^{\pi^\star}}(V^{\star, \ror })} \Big)}_{=: \tvp_3}, \label{eq:tv-v-star-3-main}
\end{align}
where (i) holds by $\Big(I - \gamma \Phatv^{\pi^\star, V} \Big)^{-1} \geq 0$, (ii) follows from Lemma~\ref{lemma:tv-dro-b-bound-star}, and the last inequality arise from 
\begin{align*}
	&\sqrt{\mathrm{Var}_{P^{\pi^\star}}(V^{\star, \ror })}  = \left(\sqrt{\mathrm{Var}_{P^{\pi^\star}}(V^{\star, \ror })} - \sqrt{\mathrm{Var}_{\widehat{P}^{\pi^\star}}(V^{\star, \ror })} \right) + \sqrt{\mathrm{Var}_{\widehat{P}^{\pi^\star}}(V^{\star, \ror })}  \notag \\
	& \leq \left(\sqrt{\mathrm{Var}_{P^{\pi^\star}}(V^{\star, \ror })} - \sqrt{\mathrm{Var}_{\widehat{P}^{\pi^\star}}(V^{\star, \ror })} \right) + \sqrt{\left| \mathrm{Var}_{\widehat{P}^{\pi^\star}}(V^{\star, \ror }) - \mathrm{Var}_{\Phatv^{\pi^\star, V} }(V^{\star, \ror }) \right|} + \sqrt{\mathrm{Var}_{\Phatv^{\pi^\star, V} }(V^{\star, \ror })} 
\end{align*}
by applying the triangle inequality.

To continue, observing that each row of $\Phatv^{\pi^\star, V}$ is a probability distribution obeying that the sum is $1$, we arrive at
\begin{align}\label{eq:tv-first-C0}
	\Big(I - \gamma \Phatv^{\pi^\star, V} \Big)^{-1} 1 = \Big(I + \sum_{t=1}^\infty \gamma^t \Big(\Phatv^{\pi^\star, V}\Big)^t \Big)  1 = \frac{1}{ 1-\gamma} 1. 
\end{align}
Armed with this fact, we shall control the other three terms $\tvp_1, \tvp_2, \tvp_3$ in \eqref{eq:tv-v-star-3-main} separately.

\begin{itemize}
	\item Consider $\tvp_1$. We first introduce the following lemma, whose proof is postponed to Appendix~\ref{proof:lemma:tv-key0}.
	\begin{lemma}\label{lemma:tv-key0}
Consider any $\delta \in (0,1)$. With probability at least  $1-\delta$, one has
\begin{align*}
&\Big(I - \gamma \Phatv^{\pi^\star, V} \Big)^{-1}\sqrt{\mathrm{Var}_{\Phatv^{\pi^\star, V} }(V^{\star, \ror })}\leq 4\sqrt{\frac{\left(1+\sqrt{\frac{ \log(\frac{18SAN}{\delta})}{(1-\gamma)^2N}} \right)  }{\gamma^3 (1-\gamma)^2  \max\{1-\gamma, \ror\}}} 1 \leq  4\sqrt{\frac{\left(1+\sqrt{\frac{\log(\frac{18SAN}{\delta})}{(1-\gamma)^2N}} \right) }{\gamma^3 (1-\gamma)^3 }} 1.
\end{align*}
\end{lemma}
Applying Lemma~\ref{lemma:tv-key0} and inserting back to \eqref{eq:tv-v-star-3-main} leads to
\begin{align}
\tvp_1 & =  2 \sqrt{\frac{\log(\frac{18SAN}{\delta})}{N}} \Big(I - \gamma \Phatv^{\pi^\star, V} \Big)^{-1} \sqrt{\mathrm{Var}_{\Phatv^{\pi^\star, V} }(V^{\star, \ror })} \nonumber \\
& \leq 8\sqrt{\frac{ \log(\frac{18SAN}{\delta})}{\gamma^3 (1-\gamma)^2 \max\{1-\gamma, \ror\} N} \bigg(1+\sqrt{\frac{ \log(\frac{18SAN}{\delta})}{(1-\gamma)^2N}} \bigg)} 1. \label{eq:tv-first-C1}
\end{align}

\item Consider $\tvp_2$. First, denote $V' \defn V^{\star,\ror} - \min_{s'\in\cS} V^{\star,\ror}(s') 1 $, by Lemma~\ref{lemma:tv-key-value-range}, it follows that 
\begin{align}
	 {0}\leq V' \leq  \frac{1}{\gamma \max\{ 1-\gamma, \sigma\}} 1  .
\end{align}
Then, we have for all $(s,a)\in \cS\times \cA$, and $P_{s,a}\in \Delta(\cS)$, and $\widetilde{P}_{s,a} \in \cU^{\ror}(P_{s,a})$:
\begin{align}
  \big|\mathsf{Var}_{\widetilde{P}_{s,a}}(V^{\star,\ror}) - \mathsf{Var}_{P_{s,a}}(V^{\star,\ror}) \big|  & =  \big|\mathsf{Var}_{\widetilde{P}_{s,a}}(V') - \mathsf{Var}_{P_{s,a}}(V') \big|  \notag \\
&   \leq  \big\|\widetilde{P}_{s,a} - P_{s,a} \big\|_1 \big\|V'\big\|_{\infty}^2 \notag \\
& \leq     \frac{2\sigma }{\gamma^2 (\max\{ 1-\gamma, \sigma\})^2}   \leq  \frac{2  }{\gamma^2 \max\{ 1-\gamma, \sigma\} }  . \label{eq:vmax-sigma-big}
\end{align}

Applying the above relation we obtain 
\begin{align}
	\tvp_2 & = 2 \sqrt{\frac{\log(\frac{18SAN}{\delta})}{N}} \Big(I - \gamma \Phatv^{\pi^\star, V} \Big)^{-1} \sqrt{\left| \mathrm{Var}_{\widehat{P}^{\pi^\star}}(V^{\star, \ror }) - \mathrm{Var}_{\Phatv^{\pi^\star, V} }(V^{\star, \ror }) \right|}   \notag \\
	& = 2 \sqrt{\frac{\log(\frac{18SAN}{\delta})}{N}} \Big(I - \gamma \Phatv^{\pi^\star, V} \Big)^{-1}  \sqrt{\left| \Pi^{\pi^\star} \left( \mathrm{Var}_{\widehat{P}^\no }(V^{\star, \ror }) - \mathrm{Var}_{\widehat{P}^{\pi^\star, V} }(V^{\star, \ror }) \right)\right| }   \notag \\
	& \leq 2\sqrt{\frac{\log(\frac{18SAN}{\delta})}{N}} \Big(I - \gamma \Phatv^{\pi^\star, V} \Big)^{-1}  \sqrt{\left\| \mathrm{Var}_{\widehat{P}^\no }(V^{\star, \ror }) - \mathrm{Var}_{\widehat{P}^{\pi^\star, V} }(V^{\star, \ror }) \right\|_\infty 1 }   \notag \\
	& \leq 2\sqrt{\frac{\log(\frac{18SAN}{\delta})}{N}} \Big(I - \gamma \Phatv^{\pi^\star, V} \Big)^{-1} \sqrt{\frac{2}{\gamma^2 \max\{1-\gamma, \ror\}}} 1 = 2\sqrt{\frac{ 2\log(\frac{18SAN}{\delta})}{\gamma^2 (1-\gamma)^2 \max\{1-\gamma, \ror\}  N} } 1, \label{eq:tv-first-C2}
\end{align}
where the last equality uses $  \Big(I - \gamma \Phatv^{\pi^\star, V} \Big)^{-1}  1  = \frac{1}{1-\gamma}$ (cf.~\eqref{eq:tv-first-C0}).

\item Consider $\tvp_3$. The following lemma plays an important role.
\begin{lemma}\citep[Lemma~6]{panaganti2021sample}\label{lemma:tv-key2}
Consider any $\delta \in (0,1)$. For any fixed policy $\pi$ and fixed value vector $V \in\mathbb{R}^S$, one has with probability at least $1-\delta$,
\begin{align*}
 \left|\sqrt{\mathrm{Var}_{\widehat{P}^{\pi}}(V)} -  \sqrt{\mathrm{Var}_{P^{\pi}}(V)} \right| \leq \sqrt{\frac{2\|V\|_{\infty}^2\log(\frac{2SA}{\delta})}{N}} 1.
\end{align*}
\end{lemma}

Applying Lemma~\ref{lemma:tv-key2} with $\pi = \pi^\star$ and $V = V^{\star,\ror}$ leads to
\begin{align*}
	\sqrt{\mathrm{Var}_{P^{\pi^\star}}(V^{\star, \ror })} - \sqrt{\mathrm{Var}_{\widehat{P}^{\pi^\star}}(V^{\star, \ror })} \leq \sqrt{\frac{2\|V^{\star,\ror}\|_{\infty}^2\log(\frac{2SA}{\delta})}{N}} 1 ,
\end{align*}
which can be plugged in \eqref{eq:tv-v-star-3-main} to verify
\begin{align}
\tvp_3 & = 2 \sqrt{\frac{\log(\frac{18SAN}{\delta})}{N}} \Big(I - \gamma \Phatv^{\pi^\star, V} \Big)^{-1} \left(\sqrt{\mathrm{Var}_{P^{\pi^\star}}(V^{\star, \ror })} - \sqrt{\mathrm{Var}_{\widehat{P}^{\pi^\star}}(V^{\star, \ror })} \right)  \notag\\
& \leq \frac{4}{(1-\gamma)} \frac{\log(\frac{SAN}{\delta}) \|V^{\star,\ror}\|_{\infty} }{N} 1 \leq \frac{4\log(\frac{18SAN}{\delta}) }{(1-\gamma)^2N} 1, \label{eq:tv-first-C3}
\end{align}
where the last line uses $  \Big(I - \gamma \Phatv^{\pi^\star, V} \Big)^{-1}  1  = \frac{1}{1-\gamma}$ (cf.~\eqref{eq:tv-first-C0}).

\end{itemize}

Finally, inserting the results of $\tvp_1$ in \eqref{eq:tv-first-C1}, $\tvp_2$ in \eqref{eq:tv-first-C2}, $\tvp_3$ in \eqref{eq:tv-first-C3}, and \eqref{eq:tv-first-C0} back into \eqref{eq:tv-v-star-3-main} gives  
\begin{align}
&\Big(I - \gamma \Phatv^{\pi^\star, V} \Big)^{-1} \Big( \Phatv^{\pi^\star, V} V^{\pi^\star, \ror } - \Pv^{\pi^\star, V } V^{\pi^\star, \ror } \Big) \leq 8\sqrt{\frac{ \log(\frac{18SAN}{\delta})}{\gamma^3 (1-\gamma)^2 \max\{1-\gamma, \ror\} N} \bigg(1+\sqrt{\frac{ \log(\frac{18SAN}{\delta})}{(1-\gamma)^2N}} \bigg)} 1  \notag \\
&\quad + 2\sqrt{\frac{ 2\log(\frac{18SAN}{\delta})}{\gamma^2 (1-\gamma)^2 \max\{1-\gamma, \ror\}  N} } 1 +  \frac{4\log(\frac{18SAN}{\delta}) }{(1-\gamma)^2N} 1 + \frac{\log(\frac{18SAN}{\delta})}{N(1-\gamma)^2}  1 \notag \\
& \leq 10\sqrt{\frac{ 2\log(\frac{18SAN}{\delta})}{\gamma^3 (1-\gamma)^2 \max\{1-\gamma, \ror\} N} \bigg(1+\sqrt{\frac{\log(\frac{SAN}{\delta})}{(1-\gamma)^2N}} \bigg)} 1 + \frac{5\log(\frac{18SAN}{\delta}) }{(1-\gamma)^2N} 1 \notag \\
&  \leq 160 \sqrt{\frac{ \log(\frac{18SAN}{\delta})}{ (1-\gamma)^2 \max\{1-\gamma, \ror\} N} } 1 + \frac{5\log(\frac{18SAN}{\delta}) }{(1-\gamma)^2N} 1,  \label{eq:tv-v-star-first-finish}
\end{align}
where the last inequality holds by the fact $\gamma \geq \frac{1}{4}$ and letting $N \geq \frac{\log(\frac{SAN}{\delta})}{(1-\gamma)^2}$.

\paragraph{Step 2: controlling $\|\widehat{V}^{\pi^\star, \ror } - V^{ \pi^\star, \ror}\|_\infty$: bounding the second term in \eqref{eq:tv-v-star-two-terms}.}

To proceed, applying Lemma~\ref{lemma:tv-dro-b-bound-star} on the second term of the right hand side of \eqref{eq:tv-v-star-two-terms} leads to
\begin{align}
&\Big(I - \gamma \Phatv^{\pi^\star, \widehat{V}} \Big)^{-1} \Big( \Phatv^{\pi^\star, V} V^{\pi^\star, \ror } - \Pv^{\pi^\star, V } V^{\pi^\star, \ror } \Big) \nonumber \\
& \leq 2\Big(I - \gamma \Phatv^{\pi^\star, \widehat{V}} \Big)^{-1}  \bigg( \sqrt{\frac{\log(\frac{18SAN}{\delta})}{N}} \sqrt{\mathrm{Var}_{P^{\pi^\star}}(V^{\star, \ror })} +  \frac{\log(\frac{18SAN}{\delta})}{N(1-\gamma)} 1\bigg) \nonumber \\
& \leq  \frac{2\log(\frac{18SAN}{\delta})}{N(1-\gamma)} \Big(I - \gamma \Phatv^{\pi^\star, \widehat{V}}  \Big)^{-1} 1  + \underbrace{ 2\sqrt{\frac{\log(\frac{18SAN}{\delta})}{N}} \Big(I - \gamma \Phatv^{\pi^\star, \widehat{V}} \Big)^{-1} \sqrt{\mathrm{Var}_{\Phatv^{\pi^\star, \widehat{V}}  }(\widehat{V}^{\pi^\star, \ror })}}_{=: \cC_4}  \nonumber \\
& \quad + \underbrace{ 2\sqrt{\frac{\log(\frac{18SAN}{\delta})}{N}} \Big(I - \gamma \Phatv^{\pi^\star, \widehat{V}}  \Big)^{-1} \left(\sqrt{ \mathrm{Var}_{\Phatv^{\pi^\star, \widehat{V}}  }(V^{\pi^\star, \ror }- \widehat{V}^{\pi^\star, \ror }) } \right)}_{=: \cC_5} \nonumber \\
& \quad + \underbrace{ 2\sqrt{\frac{\log(\frac{18SAN}{\delta})}{N}} \Big(I - \gamma \Phatv^{\pi^\star, \widehat{V}}  \Big)^{-1}\left(\sqrt{\left| \mathrm{Var}_{\widehat{P}^{\pi^\star}}(V^{\star, \ror }) - \mathrm{Var}_{\Phatv^{\pi^\star, \widehat{V}} }(V^{\star, \ror }) \right|} \right)}_{ =: \cC_6} \nonumber \\
& \quad + \underbrace{ 2\sqrt{\frac{\log(\frac{18SAN}{\delta})}{N}} \Big(I - \gamma \Phatv^{\pi^\star, \widehat{V}}  \Big)^{-1} \left(\sqrt{\mathrm{Var}_{P^{\pi^\star}}(V^{\star, \ror })} - \sqrt{\mathrm{Var}_{\widehat{P}^{\pi^\star}}(V^{\star, \ror })} \right)}_{=: \cC_7}. \label{eq:tv-v-star-second}
\end{align} 
We now bound the above terms separately.
\begin{itemize}
\item Applying Lemma~\ref{lemma:tv-key1} with $P = \widehat{P}^{\pi^\star, \widehat{V}}$, $\pi = \pi^\star$ and taking  $V = \widehat{V}^{\pi^\star,\ror}$ which obeys  $\widehat{V}^{\pi^\star,\ror} = r_{\pi^\star} + \gamma \Phatv^{\pi^\star, \widehat{V}} \widehat{V}^{\pi^\star,\ror}$, and in view of \eqref{eq:tv-first-C0},
 the term $\cC_4$ in \eqref{eq:tv-v-star-second} can be controlled as follows:
\begin{align}
\cC_4 &= 2\sqrt{\frac{\log(\frac{18SAN}{\delta})}{N}} \Big(I - \gamma \Phatv^{\pi^\star, \widehat{V}} \Big)^{-1} \sqrt{\mathrm{Var}_{\Phatv^{\pi^\star, \widehat{V}}  }(\widehat{V}^{\pi^\star, \ror })} \notag\\
&\leq 2\sqrt{\frac{\log(\frac{18SAN}{\delta})}{N}} \sqrt{\frac{8 (\max_s \widehat{V}^{\pi^\star, \ror }(s)- \min_s \widehat{V}^{\pi^\star, \ror }(s))  }{\gamma^2(1-\gamma)^2 }} 1 \notag \\
&\leq  8\sqrt{\frac{\log(\frac{18SAN}{\delta})   }{\gamma^3(1-\gamma)^2 \max\{1-\gamma, \ror\} N }} 1 ,\label{eq:tv-first-C4}
\end{align}
where the last inequality holds by applying Lemma~\ref{lemma:tv-key-value-range}.

\item To continue, considering $\cC_5$, we directly observe that (in view of \eqref{eq:tv-first-C0})
\begin{align}
\cC_5 &= 2\sqrt{\frac{\log(\frac{18SAN}{\delta})}{N}} \Big(I - \gamma \Phatv^{\pi^\star, \widehat{V}}  \Big)^{-1}  \sqrt{ \mathrm{Var}_{\Phatv^{\pi^\star, \widehat{V}}  }(V^{\pi^\star, \ror } - \widehat{V}^{\pi^\star, \ror }) }   \notag \\
& \leq 2\sqrt{\frac{\log(\frac{18SAN}{\delta})}{(1-\gamma)^2N}} \left\|V^{\star,\ror} - \widehat{V}^{\pi^\star, \ror } \right\|_\infty 1. \label{eq:tv-first-C5}
\end{align}

\item Then, it is easily verified that $\cC_6$ can be controlled similarly as \eqref{eq:tv-first-C2} as follows:
\begin{align}
\cC_6 & \leq 2\sqrt{\frac{2\log(\frac{18SAN}{\delta})}{\gamma^2 (1-\gamma)^2 \max\{1-\gamma, \ror\}  N} } 1. \label{eq:tv-first-C6}
\end{align}

\item Similarly, $\cC_7$ can be controlled the same as \eqref{eq:tv-first-C3} shown below:
\begin{align}
\tvp_7  & \leq \frac{4\log(\frac{18SAN}{\delta}) }{(1-\gamma)^2N} 1. \label{eq:tv-first-C7}
\end{align}
\end{itemize}

Combining the results in \eqref{eq:tv-first-C4}, \eqref{eq:tv-first-C5}, \eqref{eq:tv-first-C6}, and \eqref{eq:tv-first-C7} and inserting back to \eqref{eq:tv-v-star-second} leads to
\begin{align}
&\Big(I - \gamma \Phatv^{\pi^\star, \widehat{V}} \Big)^{-1} \Big( \Phatv^{\pi^\star, V} V^{\pi^\star, \ror } - \Pv^{\pi^\star, V } V^{\pi^\star, \ror } \Big)  \leq 8\sqrt{\frac{\log(\frac{18SAN}{\delta})   }{\gamma^3(1-\gamma)^2 \max\{1-\gamma, \ror\} N }} 1  \notag \\
& \qquad \qquad+ 2\sqrt{\frac{\log(\frac{18SAN}{\delta})}{(1-\gamma)^2N}} \left\|V^{\star,\ror} - \widehat{V}^{\pi^\star, \ror } \right\|_\infty 1 + 2\sqrt{\frac{2\log(\frac{18SAN}{\delta})}{\gamma^2 (1-\gamma)^2 \max\{1-\gamma, \ror\}  N} } 1 + \frac{4\log(\frac{18SAN}{\delta}) }{(1-\gamma)^2N} 1 \notag \\
& \leq 80\sqrt{\frac{\log(\frac{18SAN}{\delta})   }{(1-\gamma)^2 \max\{1-\gamma, \ror\} N }} 1 + 2\sqrt{\frac{\log(\frac{18SAN}{\delta})}{(1-\gamma)^2N}} \left\|V^{\star,\ror} - \widehat{V}^{\pi^\star, \ror } \right\|_\infty 1 + \frac{4\log(\frac{18SAN}{\delta}) }{(1-\gamma)^2N} 1, \label{eq:tv-v-star-second-finish}
\end{align}
where the last inequality follows from the assumption $\gamma \geq \frac{1}{4}$.

Finally, inserting \eqref{eq:tv-v-star-first-finish} and \eqref{eq:tv-v-star-second-finish} back to \eqref{eq:tv-v-star-two-terms} yields
\begin{align}
&\left\|\widehat{V}^{\pi^\star, \ror } - V^{ \pi^\star, \ror}\right\|_\infty \notag \\
 &\leq \max\Bigg\{160 \sqrt{\frac{ \log(\frac{18SAN}{\delta})}{ (1-\gamma)^2 \max\{1-\gamma, \ror\} N} } + \frac{5\log(\frac{18SAN}{\delta}) }{(1-\gamma)^2N}, \notag \\
& \qquad\quad 80\sqrt{\frac{\log(\frac{18SAN}{\delta})   }{(1-\gamma)^2 \max\{1-\gamma, \ror\} N }}  + 2\sqrt{\frac{\log(\frac{18SAN}{\delta})}{(1-\gamma)^2N}} \left\|V^{\star,\ror} - \widehat{V}^{\pi^\star, \ror } \right\|_\infty  + \frac{4\log(\frac{18SAN}{\delta}) }{(1-\gamma)^2N}   \Bigg\} \notag \\
& \leq 160\sqrt{\frac{ \log(\frac{18SAN}{\delta})}{ (1-\gamma)^2 \max\{1-\gamma, \ror\} N} } + \frac{8\log(\frac{18SAN}{\delta}) }{(1-\gamma)^2N},\label{eq:tv-pi-star-upper-final}
\end{align}
where the last inequality holds by taking $N \geq \frac{16\log(\frac{SAN}{\delta})}{(1-\gamma)^2}$.

\subsubsection{Controlling $\|\widehat{V}^{\widehat{\pi}, \ror } - V^{ \widehat{\pi}, \ror}\|_\infty$}\label{proof:lemma2-body}

Recall the bound in \eqref{eq:tv-v-pihat-two-terms} which holds for any uncertainty set:
\begin{align} 
\big\|\widehat{V}^{\widehat{\pi}, \ror } - V^{\widehat{\pi}, \ror}\big\|_\infty &\leq \gamma \max\Big\{ 
	\Big\| \Big(I - \gamma \Pv^{\widehat{\pi}, V} \Big)^{-1} \Big( \Phatv^{\widehat{\pi}, \widehat{V}} \widehat{V}^{\widehat{\pi}, \ror } - \Pv^{\widehat{\pi}, \widehat{V} } \widehat{V}^{\widehat{\pi}, \ror } \Big) \Big\|_\infty, \nonumber\\
	&\qquad\qquad \Big\| \Big(I - \gamma \Pv^{\widehat{\pi}, \widehat{V}} \Big)^{-1} \Big( \Phatv^{\widehat{\pi}, \widehat{V}} \widehat{V}^{\widehat{\pi}, \ror } - \Pv^{\widehat{\pi}, \widehat{V} } \widehat{V}^{\widehat{\pi}, \ror } \Big) \Big\|_\infty \Big\}.
\end{align}

\paragraph{Step 3: controlling $\|\widehat{V}^{\widehat{\pi}, \ror } - V^{ \widehat{\pi}, \ror}\|_\infty$: bounding the first term in \eqref{eq:tv-v-pihat-two-terms}.}  To begin with, we introduce the following lemma which controls the main term on the right hand side of \eqref{eq:tv-v-pihat-two-terms}, which is proved in Appendix~\ref{proof:lemma:dro-b-bound-infinite-loo-l1}.
\begin{lemma}\label{lemma:dro-b-bound-infinite-loo-l1}
Consider any $\delta \in (0,1)$. Taking $N \geq \log \left(\frac{54SAN^2}{(1-\gamma)\delta} \right)$, with probability at least $1- \delta$, one has 
\begin{align}\label{eq:dro-b-bound-infinite}
	\Big| \Phatv^{\widehat{\pi}, \widehat{V}} \widehat{V}^{\widehat{\pi}, \ror } - \Pv^{\widehat{\pi}, \widehat{V} } \widehat{V}^{\widehat{\pi}, \ror }  \Big| & \leq 2\sqrt{\frac{\log(\frac{54SAN^2} {(1-\gamma)\delta})}{N}} \sqrt{\mathrm{Var}_{P^{\no}_{s,a}}(\widehat{V}^{\star,\ror})} 1 +  \frac{8\log(\frac{54SAN^2 }{(1-\gamma)\delta})}{N(1-\gamma)} 1 + \frac{2\gamma \varepsilon_{\mathsf{opt}} }{1 -\gamma} 1\notag\\
& \leq 10\sqrt{\frac{\log(\frac{54SAN^2} {(1-\gamma)\delta})}{(1 -\gamma)^2 N}} 1 + \frac{2\gamma \varepsilon_{\mathsf{opt}} }{1 -\gamma} 1.
\end{align}
\end{lemma}
 
With Lemma~\ref{lemma:dro-b-bound-infinite-loo-l1} in hand, we have
\begin{align}
	& \Big(I - \gamma \Pv^{\widehat{\pi}, \widehat{V}} \Big)^{-1} \Big( \Phatv^{\widehat{\pi}, \widehat{V}} \widehat{V}^{\widehat{\pi}, \ror } - \Pv^{\widehat{\pi}, \widehat{V} } \widehat{V}^{\widehat{\pi}, \ror } \Big) \nonumber \\
	&  \overset{\mathrm{(i)}}{\leq} \left(I - \gamma \Pv^{\widehat{\pi}, \widehat{V}} \right)^{-1} \left| \Phatv^{\widehat{\pi}, \widehat{V}} \widehat{V}^{\widehat{\pi}, \ror } - \Pv^{\widehat{\pi}, \widehat{V} } \widehat{V}^{\widehat{\pi}, \ror } \right| \nonumber \\
	&\leq  2\sqrt{\frac{\log(\frac{54SAN^2} {(1-\gamma)\delta})}{N}} \left(I - \gamma \Pv^{\widehat{\pi}, \widehat{V}} \right)^{-1}  \sqrt{\mathrm{Var}_{P^{\widehat{\pi} }}(\widehat{V}^{\star, \ror })} + \left(I - \gamma \pmin_Q^{\widehat{\pi}, V^{\widehat{\pi}}}\right)^{-1} \Bigg( \frac{8\log(\frac{54SAN^2 }{(1-\gamma)\delta})}{N(1-\gamma)}  + \frac{2\gamma \varepsilon_{\mathsf{opt}} }{1 -\gamma} \Bigg) 1 \notag \\
	& \overset{\mathrm{(ii)}}{\leq} \Bigg( \frac{8\log(\frac{54SAN^2 }{(1-\gamma)\delta})}{N(1-\gamma)^2}  + \frac{2\gamma \varepsilon_{\mathsf{opt}} }{(1 -\gamma)^2}  \Bigg) 1 +  \underbrace{ 2\sqrt{\frac{\log(\frac{54SAN^2} {(1-\gamma)\delta})}{N}} \left(I - \gamma \Pv^{\widehat{\pi}, \widehat{V}} \right)^{-1}  \sqrt{\mathrm{Var}_{\Pv^{\widehat{\pi}, \widehat{V}}}(\widehat{V}^{\widehat{\pi}, \ror })} }_{=: \cD_1 } \notag \\
	& \quad + \underbrace{ 2\sqrt{\frac{\log(\frac{54SAN^2} {(1-\gamma)\delta})}{N}} \left(I - \gamma \Pv^{\widehat{\pi}, \widehat{V}} \right)^{-1}  \sqrt{ \left|\mathrm{Var}_{\Pv^{\widehat{\pi}, \widehat{V}}}(\widehat{V}^{\star, \ror }) - \mathrm{Var}_{\Pv^{\widehat{\pi}, \widehat{V}}}(\widehat{V}^{\widehat{\pi}, \ror }) \right| } }_{=:  \cD_2 } \notag \\
	& \quad + \underbrace{ 2\sqrt{\frac{\log(\frac{54SAN^2} {(1-\gamma)\delta})}{N}} \left(I - \gamma \Pv^{\widehat{\pi}, \widehat{V}} \right)^{-1}  \sqrt{ \left|\mathrm{Var}_{P^{\widehat{\pi}}}(\widehat{V}^{\star, \ror }) - \mathrm{Var}_{\Pv^{\widehat{\pi}, \widehat{V}}}(\widehat{V}^{\star, \ror }) \right| } }_{=: \cD_3 }, \label{eq:tv-v-hat-3-main}
\end{align}
where (i) and (ii) hold by the fact that each row of $(1-\gamma)\left(I - \gamma \Pv^{\widehat{\pi}, \widehat{V}} \right)^{-1}$ is a probability vector that falls into $\Delta(\cS)$.

The remainder of the proof will focus on controlling the three terms in \eqref{eq:tv-v-hat-3-main} separately.

\begin{itemize}
\item For $\cD_1$, we introduce the following lemma, whose proof is postponed to \ref{proof:lemma:tv-vhat-key1}.
\begin{lemma}\label{lemma:tv-vhat-key1}
Consider any $\delta \in (0,1)$. Taking $ N \geq\frac{\log(\frac{54SAN^2}{(1-\gamma)\delta})}{(1-\gamma)^2}$ and $\varepsilon_{\mathsf{opt}}\leq \frac{1-\gamma}{\gamma}$, one has with probability at least $1-\delta$,
\begin{align*}
\Big(I - \gamma \Pv^{\widehat{\pi}, \widehat{V}} \Big)^{-1}  \sqrt{\mathrm{Var}_{\Pv^{\widehat{\pi}, \widehat{V}}}(\widehat{V}^{\widehat{\pi}, \ror })} &\leq 6\sqrt{\frac{1 }{\gamma^3 (1-\gamma)^2 \max\{1-\gamma, \ror\} }} 1 \leq  6\sqrt{\frac{1}{(1-\gamma)^3 \gamma^2}} 1.
\end{align*}
\end{lemma}

Applying Lemma~\ref{lemma:tv-vhat-key1} and \eqref{eq:tv-first-C0} to \eqref{eq:tv-v-hat-3-main} leads to
\begin{align}
\cD_1 &= 2\sqrt{\frac{\log(\frac{54SAN^2}{(1-\gamma)\delta})}{N}} \Big(I - \gamma \Pv^{\widehat{\pi}, \widehat{V}} \Big)^{-1}  \sqrt{\mathrm{Var}_{\Pv^{\widehat{\pi}, \widehat{V}}}(\widehat{V}^{\widehat{\pi}, \ror })} \nonumber \\
& \leq 12 \sqrt{\frac{\log(\frac{54SAN^2}{(1-\gamma)\delta})}{\gamma^3 (1-\gamma)^2 \max\{1-\gamma, \ror\} N}} 1. \label{eq:tv-first-D1}
\end{align}

\item Applying Lemma~\ref{eq:tv-auxiliary-lemma} with $\|\widehat{V}^{\star, \ror } - \widehat{V}^{\widehat{\pi}, \ror } \|_\infty \leq \frac{2\gamma \varepsilon_{\mathsf{opt}} }{1 -\gamma}$ and \eqref{eq:tv-first-C0}, $\cD_2$ can be controlled as 
\begin{align}
	\cD_2 &= 2\sqrt{\frac{\log(\frac{54SAN^2}{(1-\gamma)\delta})}{N}} \Big(I - \gamma \Pv^{\widehat{\pi}, \widehat{V}} \Big)^{-1}  \sqrt{ \left|\mathrm{Var}_{\Pv^{\widehat{\pi}, \widehat{V}}}(\widehat{V}^{\star, \ror }) - \mathrm{Var}_{\Pv^{\widehat{\pi}, \widehat{V}}}(\widehat{V}^{\widehat{\pi}, \ror }) \right| } \notag \\
	&\leq 4\sqrt{\frac{\log(\frac{54SAN^2}{(1-\gamma)\delta})}{N}} \Big(I - \gamma \Pv^{\widehat{\pi}, \widehat{V}} \Big)^{-1} \frac{\sqrt{\gamma \varepsilon_{\mathsf{opt}}} }{1-\gamma} \leq 4\sqrt{\frac{\gamma \varepsilon_{\mathsf{opt}} \log(\frac{54SAN^2}{(1-\gamma)\delta})}{(1-\gamma)^4N}} 1. \label{eq:tv-first-D2}
\end{align}

\item $\cD_3$ can be controlled similar to $\cC_2$ in \eqref{eq:tv-first-C2} as follows:
\begin{align}
	\cD_3 & = 2\sqrt{\frac{\log(\frac{54SAN^2}{(1-\gamma)\delta})}{N}} \left(I - \gamma \Pv^{\widehat{\pi}, \widehat{V}} \right)^{-1}  \sqrt{ \left|\mathrm{Var}_{P^{\widehat{\pi}}}(\widehat{V}^{\star, \ror }) - \mathrm{Var}_{\Pv^{\widehat{\pi}, \widehat{V}}}(\widehat{V}^{\star, \ror }) \right| } \notag \\
	& \leq 4\sqrt{\frac{\log(\frac{54SAN^2}{(1-\gamma)\delta})}{N}} \left(I - \gamma \Pv^{\widehat{\pi}, \widehat{V}} \right)^{-1} \sqrt{\frac{1}{\gamma^2 \max\{1-\gamma, \ror\}}} 1 \leq 4\sqrt{\frac{\log(\frac{54SAN^2}{(1-\gamma)\delta})}{\gamma^2 (1-\gamma)^2 \max\{1-\gamma, \ror\}  N} } 1 \label{eq:tv-first-D3}
\end{align}
\end{itemize}
Finally, summing up the results in \eqref{eq:tv-first-D1},  \eqref{eq:tv-first-D2}, and \eqref{eq:tv-first-D3} and inserting them back to \eqref{eq:tv-v-hat-3-main} yields: taking $ N \geq\frac{\log(\frac{54SAN^2}{(1-\gamma)\delta})}{(1-\gamma)^2}$ and $\varepsilon_{\mathsf{opt}}\leq \frac{1-\gamma}{\gamma}$, with probability at least $1-\delta$,
\begin{align}
&\left(I - \gamma \Pv^{\widehat{\pi}, \widehat{V}} \right)^{-1} \left( \Phatv^{\widehat{\pi}, \widehat{V}} \widehat{V}^{\widehat{\pi}, \ror } - \Pv^{\widehat{\pi}, \widehat{V} } \widehat{V}^{\widehat{\pi}, \ror } \right) \leq  \left( \frac{8\log(\frac{54SAN^2 }{(1-\gamma)\delta})}{N(1-\gamma)^2}  + \frac{2\gamma \varepsilon_{\mathsf{opt}} }{(1 - \gamma)^2}  \right) 1 \notag \\
& \quad + 12 \sqrt{\frac{\log(\frac{54SAN^2}{(1-\gamma)\delta})}{\gamma^3 (1-\gamma)^2 \max\{1-\gamma, \ror\} N}} 1 +  4\sqrt{\frac{\gamma \varepsilon_{\mathsf{opt}} \log(\frac{54SAN^2}{(1-\gamma)\delta})}{(1-\gamma)^4N}} 1  + 4\sqrt{\frac{\log(\frac{54SAN^2}{(1-\gamma)\delta})}{\gamma^2 (1-\gamma)^2 \max\{1-\gamma, \ror\}  N} } 1 \notag \\
&\leq 16 \sqrt{\frac{\log(\frac{54SAN^2}{(1-\gamma)\delta})}{\gamma^3 (1-\gamma)^2 \max\{1-\gamma, \ror\} N}} 1 + \frac{14\log(\frac{54SAN^2 }{(1-\gamma)\delta})}{N(1-\gamma)^2}  1, \label{eq:tv-v-hat-first-finish}
\end{align}
where the last inequality holds by taking $\varepsilon_{\mathsf{opt}} \leq \min \left\{\frac{1-\gamma}{\gamma}, \frac{\log(\frac{54SAN^2}{(1-\gamma)\delta})}{\gamma N} \right\} = \frac{\log(\frac{54SAN^2}{(1-\gamma)\delta})}{\gamma N}$.

\paragraph{Step 4: controlling $\|\widehat{V}^{\widehat{\pi}, \ror } - V^{ \widehat{\pi}, \ror}\|_\infty$: bounding the second term in \eqref{eq:tv-v-pihat-two-terms}.}
Towards this, applying Lemma~\ref{lemma:dro-b-bound-infinite-loo-l1} leads to
\begin{align}
	&\Big(I - \gamma \Pv^{\widehat{\pi}, V} \Big)^{-1} \Big( \Phatv^{\widehat{\pi}, \widehat{V}} \widehat{V}^{\widehat{\pi}, \ror } - \Pv^{\widehat{\pi}, \widehat{V} } \widehat{V}^{\widehat{\pi}, \ror } \Big)  \leq \Big(I - \gamma \Pv^{\widehat{\pi}, V} \Big)^{-1} \Big| \Phatv^{\widehat{\pi}, \widehat{V}} \widehat{V}^{\widehat{\pi}, \ror } - \Pv^{\widehat{\pi}, \widehat{V} } \widehat{V}^{\widehat{\pi}, \ror } \Big| \nonumber \\
	&\leq  2\sqrt{\frac{\log(\frac{54SAN^2}{(1-\gamma)\delta})}{N}} \Big(I - \gamma \Pv^{\widehat{\pi}, V} \Big)^{-1}  \sqrt{\mathrm{Var}_{P^{\widehat{\pi} }}(\widehat{V}^{\star, \ror })} + \Big(I - \gamma \Pv^{\widehat{\pi}, V} \Big)^{-1} \Bigg( \frac{8\log(\frac{54SAN^2 }{(1-\gamma)\delta})}{N(1-\gamma)}  + \frac{2\gamma \varepsilon_{\mathsf{opt}} }{1 -\gamma} \Bigg) 1 \notag \\
	& \leq \Bigg( \frac{8\log(\frac{54SAN^2 }{(1-\gamma)\delta})}{N(1-\gamma)^2} + \frac{2\gamma \varepsilon_{\mathsf{opt}} }{(1 -\gamma)^2} \Bigg) 1 +  \underbrace{ 2\sqrt{\frac{\log(\frac{54SAN^2}{(1-\gamma)\delta})}{N}} \left(I - \gamma \Pv^{\widehat{\pi}, V} \right)^{-1}  \sqrt{\mathrm{Var}_{\Pv^{\widehat{\pi}, V} }(V^{\widehat{\pi}, \ror })} }_{ =: \cD_4 } \notag \\
	& \quad + \underbrace{ 2\sqrt{\frac{\log(\frac{54SAN^2}{(1-\gamma)\delta})}{N}} \Big(I - \gamma \Pv^{\widehat{\pi}, V} \Big)^{-1}  \sqrt{  \mathrm{Var}_{\Pv^{\widehat{\pi}, V} }(\widehat{V}^{\widehat{\pi}, \ror } -V^{\widehat{\pi}, \ror }) } }_{ =: \cD_5 } \notag \\
	& \quad + \underbrace{2 \sqrt{\frac{\log(\frac{54SAN^2}{(1-\gamma)\delta})}{N}} \Big(I - \gamma \Pv^{\widehat{\pi}, \widehat{V}} \Big)^{-1}  \sqrt{ \left|\mathrm{Var}_{\Pv^{\widehat{\pi}, V} }(\widehat{V}^{\star, \ror }) - \mathrm{Var}_{\Pv^{\widehat{\pi}, V} }(\widehat{V}^{\widehat{\pi}, \ror }) \right| } }_{ =: \cD_6 } \notag \\
	& \quad + \underbrace{ 2 \sqrt{\frac{\log(\frac{54SAN^2}{(1-\gamma)\delta})}{N}} \Big(I - \gamma \Pv^{\widehat{\pi}, \widehat{V}} \Big)^{-1}  \sqrt{ \left|\mathrm{Var}_{P^{\widehat{\pi}} }(\widehat{V}^{\star, \ror })  - \mathrm{Var}_{\Pv^{\widehat{\pi}, V} }(\widehat{V}^{\star, \ror })  \right| } }_{ =: \cD_7 }.  \label{eq:tv-v-hat-3-main2}
\end{align}
We shall bound each of the terms separately.
\begin{itemize}
\item Applying Lemma~\ref{lemma:tv-key1} with $P = \Pv^{\widehat{\pi}, V} $, $\pi = \widehat{\pi}$, and taking $V = V^{\widehat{\pi},\ror}$ which obeys  $V^{\widehat{\pi},\ror} = r_{\widehat{\pi}} + \gamma \Pv^{\widehat{\pi}, V} V^{\widehat{\pi},\ror}$, the term $\cD_4$ can be controlled similar to  \eqref{eq:tv-first-C4} as follows:
\begin{align}
\cD_4 &\leq   8\sqrt{\frac{\log(\frac{54SAN^2}{\delta})   }{\gamma^3(1-\gamma)^2 \max\{1-\gamma, \ror\} N }} 1. \label{eq:tv-first-D4}
\end{align}
\item For $\cD_5$, it is observed that 
\begin{align}
\cD_5 &= 2\sqrt{\frac{\log(\frac{54SAN^2}{(1-\gamma)\delta})}{N}} \Big(I - \gamma \Pv^{\widehat{\pi}, V} \Big)^{-1}  \sqrt{  \mathrm{Var}_{\Pv^{\widehat{\pi}, V} }(\widehat{V}^{\widehat{\pi}, \ror } -V^{\widehat{\pi}, \ror }) } \notag \\
& \leq 2\sqrt{\frac{\log(\frac{54SAN^2}{\delta})}{(1-\gamma)^2N}} \left\|V^{\widehat{\pi},\ror} - \widehat{V}^{\widehat{\pi}, \ror } \right\|_\infty 1. \label{eq:tv-first-D5}
\end{align}

\item Next, observing that $\cD_6$ and $\cD_7$ are almost the same as the terms $\cD_2$ (controlled in \eqref{eq:tv-first-D2}) and $\cD_3$ (controlled in \eqref{eq:tv-first-D3}) in \eqref{eq:tv-v-hat-3-main}, it is easily verified that they can be controlled as follows
\begin{align}
\cD_6  &\leq  4\sqrt{\frac{\gamma \varepsilon_{\mathsf{opt}} \log(\frac{54SAN^2}{(1-\gamma)\delta})}{(1-\gamma)^4N}} 1,  \qquad \qquad
\cD_7   \leq 4\sqrt{\frac{\log(\frac{54SAN^2}{(1-\gamma)\delta})}{\gamma^2 (1-\gamma)^2 \max\{1-\gamma, \ror\}  N} } 1.  \label{eq:tv-first-D6-D7}
\end{align} 
\end{itemize}
Then inserting  the results in \eqref{eq:tv-first-D4}, \eqref{eq:tv-first-D5}, and \eqref{eq:tv-first-D6-D7} back to \eqref{eq:tv-v-hat-3-main2} leads to
\begin{align}
&\Big(I - \gamma \Pv^{\widehat{\pi}, V} \Big)^{-1} \Big( \Phatv^{\widehat{\pi}, \widehat{V}} \widehat{V}^{\widehat{\pi}, \ror } - \Pv^{\widehat{\pi}, \widehat{V} } \widehat{V}^{\widehat{\pi}, \ror } \Big) \notag \\
& \leq \Bigg( \frac{8\log(\frac{54SAN^2 }{(1-\gamma)\delta})}{N(1-\gamma)^2} + \frac{2\gamma \varepsilon_{\mathsf{opt}} }{(1 -\gamma)^2} \Bigg) 1 + 8\sqrt{\frac{\log(\frac{54SAN^2}{\delta})   }{\gamma^3(1-\gamma)^2 \max\{1-\gamma, \ror\} N }} 1  \notag \\
& \quad   + 2\sqrt{\frac{\log(\frac{54SAN^2}{\delta})}{(1-\gamma)^2N}} \left\|V^{\widehat{\pi},\ror} - \widehat{V}^{\widehat{\pi}, \ror } \right\|_\infty 1    + 4\sqrt{\frac{\gamma \varepsilon_{\mathsf{opt}} \log(\frac{54SAN^2}{(1-\gamma)\delta})}{(1-\gamma)^4N}} 1 + 4\sqrt{\frac{\log(\frac{54SAN^2}{(1-\gamma)\delta})}{\gamma^2 (1-\gamma)^2 \max\{1-\gamma, \ror\}  N} } 1 \notag \\
& \leq 12\sqrt{\frac{2\log(\frac{8SAN^2}{(1-\gamma)\delta})  }{\gamma^3(1-\gamma)^2 \max\{1-\gamma, \ror\} N }} 1 +  \frac{14\log(\frac{54SAN^2}{(1-\gamma)\delta})}{N(1-\gamma)^2} 1 + 2\sqrt{\frac{\log(\frac{54SAN^2}{\delta})}{(1-\gamma)^2N}} \left\|V^{\widehat{\pi},\ror} - \widehat{V}^{\widehat{\pi}, \ror } \right\|_\infty 1, \label{eq:tv-v-hat-second-finish}
\end{align}
where the last inequality holds by letting $\varepsilon_{\mathsf{opt}} \leq  \frac{\log(\frac{54SAN^2}{(1-\gamma)\delta})}{\gamma N}$, which directly satisfies $\varepsilon_{\mathsf{opt}}  \leq \frac{1-\gamma}{\gamma}$ by letting $N \geq \frac{\log(\frac{54SAN^2}{\delta})}{1-\gamma}$. 

Finally, inserting \eqref{eq:tv-v-hat-first-finish} and \eqref{eq:tv-v-hat-second-finish} back to \eqref{eq:tv-v-pihat-two-terms} yields: taking $\varepsilon_{\mathsf{opt}} \leq \frac{\log(\frac{54SAN^2}{(1-\gamma)\delta})}{ \gamma N} $ and $N \geq \frac{16\log(\frac{54SAN^2}{\delta})}{(1-\gamma)^2}$, with probability at least $1-\delta$, one has
\begin{align}
&\left\|\widehat{V}^{\widehat{\pi}, \ror } - V^{\widehat{\pi}, \ror}\right\|_\infty \notag \\
&\leq \max\Big\{ 16 \sqrt{\frac{\log(\frac{54SAN^2}{(1-\gamma)\delta})}{\gamma^3 (1-\gamma)^2 \max\{1-\gamma, \ror\} N}}  + \frac{14\log(\frac{54SAN^2 }{(1-\gamma)\delta})}{N(1-\gamma)^2}    , \notag \\
& \quad 12\sqrt{\frac{2\log(\frac{8SAN^2}{(1-\gamma)\delta})  }{\gamma^3(1-\gamma)^2 \max\{1-\gamma, \ror\} N }}  +  \frac{14\log(\frac{54SAN^2}{(1-\gamma)\delta})}{N(1-\gamma)^2}  + 2\sqrt{\frac{\log(\frac{54SAN^2}{\delta})}{(1-\gamma)^2N}} \left\|V^{\widehat{\pi},\ror} - \widehat{V}^{\widehat{\pi}, \ror } \right\|_\infty  \Big\} \notag \\
& \leq 24 \sqrt{\frac{\log(\frac{54SAN^2}{(1-\gamma)\delta})}{\gamma^3 (1-\gamma)^2 \max\{1-\gamma, \ror\} N}}  + \frac{28\log(\frac{54SAN^2 }{(1-\gamma)\delta})}{N(1-\gamma)^2}. \label{eq:tv-pi-star-upper-final2}
\end{align}

\paragraph{Step 5: Summing up the results.}
Summing up the results in \eqref{eq:tv-pi-star-upper-final} and \eqref{eq:tv-pi-star-upper-final2} and inserting back to \eqref{eq:l1-decompose} complete the proof as follows: taking $\varepsilon_{\mathsf{opt}} \leq \frac{\log(\frac{54SAN^2}{(1-\gamma)\delta})}{ \gamma N}$ and $N \geq \frac{16\log(\frac{54SAN^2}{\delta})}{(1-\gamma)^2}$, with probability at least $1-\delta$,
\begin{align}
 \big\|V^{\star, \ror} - V^{\widehat{\pi}, \ror} 
 \big\|_\infty
	&\leq \big\|V^{\pi^\star, \ror} - \widehat{V}^{\pi^\star, \ror}\big\|_\infty + \frac{2\gamma \varepsilon_{\mathsf{opt}} }{1 -\gamma} + \big\|\widehat{V}^{\widehat{\pi}, \ror} - V^{\widehat{\pi}, \ror}\big\|_\infty \notag \\
	& \leq \frac{2\gamma \varepsilon_{\mathsf{opt}} }{1 -\gamma} + 160\sqrt{\frac{ \log(\frac{18SAN}{\delta})}{ (1-\gamma)^2 \max\{1-\gamma, \ror\} N} } + \frac{8\log(\frac{18SAN}{\delta}) }{(1-\gamma)^2N} \notag \\
	& \quad + 24 \sqrt{\frac{\log(\frac{54SAN^2}{(1-\gamma)\delta})}{\gamma^3 (1-\gamma)^2 \max\{1-\gamma, \ror\} N}}  + \frac{28\log(\frac{54SAN^2 }{(1-\gamma)\delta})}{N(1-\gamma)^2} \notag \\
	&\leq 184 \sqrt{\frac{\log(\frac{54SAN^2}{(1-\gamma)\delta})}{\gamma^3 (1-\gamma)^2 \max\{1-\gamma, \ror\} N}}  + \frac{36 \log(\frac{54SAN^2 }{(1-\gamma)\delta})}{N(1-\gamma)^2} \notag \\
	&\leq 1508 \sqrt{\frac{\log(\frac{54SAN^2}{(1-\gamma)\delta})}{ (1-\gamma)^2 \max\{1-\gamma, \ror\} N}}, 
\end{align}
where the last inequality holds by  $\gamma\geq \frac{1}{4}$ and $N \geq \frac{16 \log(\frac{54SAN^2}{\delta})}{(1-\gamma)^2}$.

\subsection{Proof of the auxiliary lemmas for Theorem~\ref{thm:l1-upper-bound}}\label{proof:thm:l1-upper-bound-2}

\subsubsection{Proof of Lemma~\ref{lemma:tv-key-value-range}}\label{proof:lemma:tv-key-value-range}
 
To begin, note that there at leasts exist one state $s_0$ for any $V^{\pi, \ror}$ such that $V^{\pi, \ror}(s_0) = \min_{s\in\cS}V^{\pi, \ror}(s)$.
With this in mind, for any policy $\pi$, one has by the definition in \eqref{eq:robust-value-def}  and the Bellman's equation \eqref{eq:bellman-equ-pi-infinite},
\begin{align*}
\max_{s\in\cS} V^{\pi, \ror}(s) & = \max_{s\in\cS} \mathbb{E}_{a \sim \pi(\cdot \mymid s)} \Big[r(s,a) + \gamma \inf_{\cP\in \cU^\ror(P_{s,a})} \cP V^{\pi, \ror} \Big] \notag \\
& \leq \max_{(s,a)\in \cS \times \cA} \Big( 1 + \gamma \inf_{\cP\in \cU^\ror(P_{s,a})} \cP V^{\pi, \ror} \Big)  ,
\end{align*}
 where the second line holds since the reward function $r(s,a) \in [0,1]$ for all $(s,a)\in \cS\times \cA$. To continue, note that for any $(s,a) \in\cS \times \cA$, there exists some $\widetilde{P}_{s,a} \in \mathbb{R}^{S}$ constructed by reducing the values of some elements of $P_{s,a}$ to obey $  P_{s,a} \geq \widetilde{P}_{s,a} \geq 0$ and $\sum_{s'} ( P_{s,a}(s') - \widetilde{P}_{s,a}(s') ) = \ror$. This implies $\widetilde{P}_{s,a} + \ror {e}_{s_0}^\top \in \cU^\ror(P_{s,a})$, where $e_{s_0}$ is the standard basis vector supported on $s_0$, since $\frac{1}{2} \| \widetilde{P}_{s,a} + \ror {e}_{s_0}^\top - P_{s,a}\|_1 \leq \frac{1}{2}\| \widetilde{P}_{s,a} - P_{s,a}\|_1 + \frac{\ror}{2} = \ror$. Consequently,
\begin{align}
\inf_{\cP\in \cU^\ror(P_{s,a})} \cP V^{\pi, \ror}  \leq \left( \widetilde{P}_{s,a} + \ror {e}_{s_0}^\top \right)V^{\pi, \ror}  &\leq \big\| \widetilde{P}_{s,a} \big\|_1 \big\| V^{\pi, \ror} \big\|_\infty +  \ror V^{\pi, \ror} (s_0) \notag \\
& \leq \left(1- \ror \right) \max_{s\in\cS} V^{\pi, \ror} (s) +  \ror \min_{s\in\cS} V^{\pi, \ror} (s),
\end{align}
where the second inequality holds by $\big\| \widetilde{P}_{s,a}\big\|_1 =  \sum_{s'} \widetilde{P}_{s,a}(s') =  - \sum_{s'} \left( P_{s,a}(s') - \widetilde{P}_{s,a}(s') \right) + \sum_{s'} P_{s,a}(s')  = 1-\ror$. Plugging this back to the previous relation gives
 \begin{align*}
\max_{s\in\cS} V^{\pi, \ror}(s)   &  \le    1 + \gamma \left(1- \ror \right) \max_{s\in\cS} V^{\pi, \ror}(s) + \gamma \ror \min_{s\in\cS} V^{\pi, \ror}(s) ,
\end{align*} 
which, by rearranging terms, immediately yields 
 \begin{align*}
\max_{s\in\cS} V^{\pi, \ror}(s)  & \leq \frac{1 + \gamma \ror \min_{s\in\cS} V^{\pi, \ror}(s) }{1- \gamma \left(1- \ror \right)}  
  \nonumber \\
&\leq \frac{1}{(1-\gamma) + \gamma \ror } + \min_{s\in\cS}V^{\pi, \ror}(s) \leq \frac{1}{\gamma \max\{ 1-\gamma, \sigma\}} + \min_{s\in\cS}V^{\pi, \ror}(s) .
\end{align*}

\subsubsection{Proof of Lemma~\ref{lemma:tv-key1}}\label{proof:lemma:tv-key1}
Observing that each row of $P_\pi$ belongs to $\Delta(S)$, it can be directly verified that each row of $(1-\gamma)\left(I - \gamma P_\pi \right)^{-1}$ falls into $\Delta(S)$. As a result,
\begin{align}
\left(I - \gamma P_\pi \right)^{-1} \sqrt{\mathrm{Var}_{P_\pi}(V^{\pi, P})} & = \frac{1}{1-\gamma}(1-\gamma)\left(I - \gamma P_\pi \right)^{-1}\sqrt{\mathrm{Var}_{P_\pi}(V^{\pi, P})} \nonumber \\
& \overset{\mathrm{(i)}}{\leq} \frac{1}{1-\gamma} \sqrt{(1-\gamma)\left(I - \gamma P_\pi \right)^{-1} \mathrm{Var}_{P_\pi}(V^{\pi, P}) } \nonumber \\
& = \sqrt{\frac{1}{1-\gamma}}\sqrt{ \sum_{t=0}^\infty \gamma^t \left(P_\pi \right)^t \mathrm{Var}_{P_\pi}(V^{\pi, P})}, \label{eq:bound-key}
\end{align}
where (i) holds by Jensen's inequality.

To continue, denoting the minimum value of $V$ as $V_{\min} = \min_{s\in\cS} V^{\pi, P}(s)$ and  $V' \defn   V^{\pi, P} - V_{\min} 1 $. We control $\mathrm{Var}_{P_\pi}(V^{\pi, P})$ as follows:
\begin{align}
	& \mathrm{Var}_{P_\pi}(V^{\pi, P})  \nonumber\\
	&\overset{\mathrm{(i)}}{=} \mathrm{Var}_{P_\pi}(V')  = P_\pi \left(V' \circ V'\right) - \big(P_\pi V'\big) \circ  \big(P_\pi V' \big) \nonumber \\
	& \overset{\mathrm{(ii)}}{=} P_\pi \left(V' \circ V'\right) - \frac{1}{\gamma^2} \left(V' - r_\pi + (1-\gamma)V_{\min} 1\right) \circ \left(V'- r_\pi + (1-\gamma)V_{\min} 1 \right) \nonumber \\
	& = P_\pi \left(V' \circ V'\right) - \frac{1}{\gamma^2} V' \circ V' + \frac{2}{\gamma^2} V' \circ \left(r_\pi - (1-\gamma)V_{\min} 1 \right) \notag \\
	& \quad -  \frac{1}{\gamma^2} \left(r_\pi - (1-\gamma)V_{\min} 1 \right) \circ \left(r_\pi - (1-\gamma)V_{\min} 1 \right) \nonumber \\
	& \leq P_\pi \left(V' \circ V'\right) - \frac{1}{\gamma} V' \circ V' + \frac{2}{\gamma^2} \|V'\|_\infty 1 ,\label{eq:variance-tight-bound}
\end{align}
where (i) holds by the fact that $\mathrm{Var}_{P_\pi}(V^{\pi, P} -b 1) = \mathrm{Var}_{P_\pi}(V^{\pi, P})  $ for any scalar $b$ and  $V^{\pi, P}\in\mathbb{R}^S$, (ii) follows from $V' = r_\pi + \gamma P_{\pi} V^{\pi, P} - V_{\min} 1 = r_\pi - (1-\gamma)V_{\min} 1 + \gamma P_{\pi} V'$, and the last line arises from $\frac{1}{\gamma^2} V' \circ V' \geq \frac{1}{\gamma} V' \circ V'$ and $\|r_\pi - (1-\gamma)V_{\min} 1\|_\infty \leq 1$.
Plugging \eqref{eq:variance-tight-bound} back to \eqref{eq:bound-key} leads to
\begin{align}
&\left(I - \gamma P_\pi \right)^{-1} \sqrt{\mathrm{Var}_{P_\pi}(V^{\pi, P})} 
	 \leq  \sqrt{\frac{1}{1-\gamma}}\sqrt{ \sum_{t=0}^\infty \gamma^t \left(P_\pi \right)^t \left(P_\pi \left(V' \circ V'\right) - \frac{1}{\gamma} V' \circ V' + \frac{2}{\gamma^2} \|V'\|_\infty 1\right )} \nonumber \\
	&  \overset{\mathrm{(i)}}{\leq} \sqrt{\frac{1}{1-\gamma}}\sqrt{ \bigg| \sum_{t=0}^\infty \gamma^t \left(P_\pi \right)^t \left(P_\pi \left(V' \circ V'\right) - \frac{1}{\gamma} V' \circ V' \right ) \bigg|}  + \sqrt{\frac{1}{1-\gamma}}\sqrt{ \sum_{t=0}^\infty \gamma^t \left(P_\pi \right)^t \frac{2}{\gamma^2} \|V'\|_\infty 1 } \nonumber \\
	& \leq \sqrt{\frac{1}{1-\gamma}}\sqrt{ \bigg| \bigg( \sum_{t=0}^\infty \gamma^{t} \left(P_\pi \right)^{t+1} - \sum_{t = 0}^\infty \gamma^{t-1} \left(P_\pi \right)^{t}   \bigg) \left(V' \circ V'\right) \bigg| } + \sqrt{ \frac{2 \|V'\|_\infty 1}{\gamma^2(1-\gamma)^2} } \nonumber \\
	& \overset{\mathrm{(ii)}}{\leq}  \sqrt{ \frac{\|V'\|_{\infty}^2 1 }{\gamma(1-\gamma)} } + \sqrt{ \frac{2 \|V'\|_\infty 1}{\gamma^2(1-\gamma)^2} } \nonumber \\
	& \leq \sqrt{\frac{8 \|V'\|_{\infty}1}{\gamma^2(1-\gamma)^2}},
\end{align}
where (i) holds by the triangle inequality, (ii) holds by following recursion, and the last inequality holds by $\|V'\|_\infty \leq \frac{1}{1-\gamma}$.

\subsubsection{Proof of Lemma~\ref{lemma:tv-dro-b-bound-star}}\label{proof:lemma:tv-dro-b-bound-star}

\paragraph{Step 1: controlling the point-wise concentration.}
We first consider a more general term w.r.t. any fixed (independent from $\widehat{P}^\no$) value vector $V$ obeying ${0} \leq V \leq \frac{1}{1-\gamma} 1$ and any policy $\pi$. Invoking Lemma~\ref{lemma:tv-dual-form} leads to that for any $(s,a)\in \cS\times \cA$, 
\begin{align}
 \left| \pmhat^{\pi, V}_{s,a} V  -  \pmin^{\pi, V}_{s,a} V \right|&\leq \Big| \max_{\alpha\in [\min_s V(s), \max_s V(s)]} \left\{\widehat{P}^{\no}_{s,a} \left[V\right]_{\alpha} - \ror \left(\alpha - \min_{s'}\left[V\right]_{\alpha}(s') \right)\right\}  \nonumber\\
	&\qquad - \max_{\alpha\in [\min_s V(s), \max_s V(s)]} \left\{P^{\no}_{s,a} \left[V\right]_{\alpha} - w\ror \left(\alpha - \min_{s'}\left[V\right]_{\alpha}(s') \right)\right\} \Big| \nonumber\\
	&\leq \max_{\alpha\in [\min_s V(s), \max_s V(s)]} \underbrace{\left| \left(P^{\no}_{s,a} - \widehat{P}^{\no}_{s,a} \right) [V]_\alpha\right|}_{=: g_{s,a}(\alpha, V)}, \label{eq:tv-t-that-gap}
\end{align}
where the last inequality holds by that the maximum operator is $1$-Lipschitz. 

 Then for a fixed $\alpha$ and any vector $V$ that is independent with $\widehat{P}^0$, using the Bernstein's inequality, one has with probability at least $1 - \delta$,
\begin{align} \label{eq:V-p-phat-gap-one-alpha-bernstein}
	g_{s,a}(\alpha, V) = \left| \left(P^{\no}_{s,a} - \widehat{P}^{\no}_{s,a} \right) [V]_\alpha\right|  & \leq \sqrt{\frac{2\log(\frac{2}{\delta})}{N}} \sqrt{\mathrm{Var}_{P^{\no}_{s,a}}([V]_\alpha)} +  \frac{2\log(\frac{2}{\delta})}{3N(1-\gamma)}  \nonumber \\
	&\leq \sqrt{\frac{2\log(\frac{2}{\delta})}{N}} \sqrt{\mathrm{Var}_{P^{\no}_{s,a}}(V)} +  \frac{2\log(\frac{2}{\delta})}{3N(1-\gamma)}.
\end{align}

\paragraph{Step 2: deriving the uniform concentration.}
To obtain the union bound, we first notice that $g_{s,a}(\alpha, V)$ is $1$-Lipschitz w.r.t. $\alpha$ for any $V$ obeying $\|V\|_\infty \leq \frac{1}{1-\gamma}$.  In addition, we can construct an $\varepsilon_1$-net $\cN_{\varepsilon_1}$ over $[0, \frac{1}{1-\gamma}]$ whose size satisfies $|\cN_{\varepsilon_1}| \leq \frac{3}{\varepsilon_1(1-\gamma)}$ \citep{vershynin2018high}. By the union bound and \eqref{eq:V-p-phat-gap-one-alpha-bernstein}, it holds  with probability at least $1 - \frac{\delta}{SA}$ that for all $\alpha \in \cN_{\varepsilon_1}$, 
\begin{equation} \label{eq:donkey}
g_{s,a}(\alpha, V) \leq  \sqrt{\frac{2\log(\frac{2SA|\cN_{\varepsilon_1}|}{\delta})}{N}} \sqrt{\mathrm{Var}_{P^{\no}_{s,a}}(V)} +  \frac{2\log(\frac{2SA|\cN_{\varepsilon_1}|}{\delta})}{3N(1-\gamma)} .
\end{equation}
Combined with \eqref{eq:tv-t-that-gap}, it yields that,
\begin{align}
  \left| \pmhat^{\pi, V}_{s,a} V  -  \pmin^{\pi, V}_{s,a} V \right| &\leq \max_{\alpha\in [\min_s V(s), \max_s V(s)]} \left| \left(P^{\no}_{s,a} - \widehat{P}^{\no}_{s,a} \right) [V]_\alpha\right| \nonumber \\
  & \overset{\mathrm{(i)}}{\leq} \varepsilon_1 + \sup_{\alpha\in \cN_{\varepsilon_1}} \left| \left(P^{\no}_{s,a} - \widehat{P}^{\no}_{s,a} \right) [V]_\alpha\right| \notag \\
  & \overset{\mathrm{(ii)}}{\leq} \varepsilon_1 + \sqrt{\frac{2\log(\frac{2SA|\cN_{\varepsilon_1}|}{\delta})}{N}} \sqrt{\mathrm{Var}_{P^{\no}_{s,a}}(V)} +  \frac{2\log(\frac{2SA|\cN_{\varepsilon_1}|}{\delta})}{3N(1-\gamma)} \label{eq:V-p-phat-gap-one-alpha-bernstein-union-N-net} \\
  &\overset{\mathrm{(iii)}}{\leq} \sqrt{\frac{2\log(\frac{2SA|\cN_{\varepsilon_1}|}{\delta})}{N}} \sqrt{\mathrm{Var}_{P^{\no}_{s,a}}(V)} +  \frac{\log(\frac{2SA|\cN_{\varepsilon_1}|}{\delta})}{N(1-\gamma)} \notag \\
  & \overset{\mathrm{(iv)}}{\leq} 2\sqrt{\frac{\log(\frac{18SAN}{\delta})}{N}} \sqrt{\mathrm{Var}_{P^{\no}_{s,a}}(V)} +  \frac{\log(\frac{18SAN}{\delta})}{N(1-\gamma)} \label{eq:V-p-phat-gap-one-alpha-bernstein-union} \\
  & \leq 2\sqrt{\frac{\log(\frac{18SAN}{\delta})}{N}} \|V\|_\infty  +  \frac{\log(\frac{18SAN}{\delta})}{N(1-\gamma)}   \notag \\
	&\leq 3\sqrt{\frac{\log(\frac{18SA N}{\delta})}{(1-\gamma)^2N}}  \label{eq:V-p-phat-gap-one-alpha-hoeffding-union}
\end{align}
where (i) follows from that the optimal $\alpha^\star$ falls into the $\varepsilon_1$-ball centered around some point inside $N_{\varepsilon_1}$ and $g_{s,a}(\alpha, V)$ is $1$-Lipschitz, (ii) holds by \eqref{eq:donkey}, (iii) arises from taking $\varepsilon_1 = \frac{\log(\frac{2SA|\cN_{\varepsilon_1}|}{\delta})}{3N(1-\gamma)}$, (iv) is verified by $|\cN_{\varepsilon_1}| \leq \frac{3}{\varepsilon_1(1-\gamma)} \leq 9N$, and the last inequality is due to the fact $\|V^{\star,\ror}\|_\infty \leq \frac{1}{1-\gamma}$ and letting $N \geq \log(\frac{18SA N}{\delta})$.

To continue, applying \eqref{eq:V-p-phat-gap-one-alpha-bernstein-union} and \eqref{eq:V-p-phat-gap-one-alpha-hoeffding-union} with $\pi = \pi^\star$ and $V = V^{\star, \ror }$ (independent with $\widehat{P}^0$) and taking the union bound over $(s,a)\in\cS\times \cA$ gives that with probability at least $1-\delta$, it holds simultaneously for all $(s,a)\in\cS \times \cA$ that 
\begin{align}
	\left| \widehat{P}^{\pi^\star,V}_{s,a} V^{\star, \ror } - P^{\pi^\star, V }_{s,a} V^{\star, \ror } \right| &\leq  2\sqrt{\frac{\log(\frac{ 18SAN}{\delta})}{N}} \sqrt{\mathrm{Var}_{P^{\no}_{s,a}}(V^{\star, \ror })} + \frac{\log(\frac{ 18SAN}{\delta})}{N(1-\gamma)} \notag \\
	&\leq 3\sqrt{\frac{\log(\frac{18SA N}{\delta})}{(1-\gamma)^2N}}. \label{eq:tv-v-star-sa}
\end{align}

By converting \eqref{eq:tv-v-star-sa} to the matrix form, one has with probability at least $1-\delta$, 
\begin{align}
	\left|\Phatv^{\pi^\star, V} V^{\pi^\star, \ror } - \Pv^{\pi^\star, V } V^{\pi^\star, \ror } \right| &\leq 2 \sqrt{\frac{\log(\frac{18SAN}{\delta})}{N}} \sqrt{\mathrm{Var}_{P^{\pi^\star}}(V^{\star, \ror })} + \frac{\log(\frac{ 18SAN}{\delta})}{N(1-\gamma)} 1 \notag \\
	&\leq 3\sqrt{\frac{\log(\frac{18SA N}{\delta})}{(1-\gamma)^2N}} 1.
	\label{eq:tv-v-star-sa-matrix}
\end{align}

\subsubsection{Proof of Lemma~\ref{lemma:tv-key0}}\label{proof:lemma:tv-key0}

Following the same argument as \eqref{eq:bound-key}, it follows
\begin{align}
\Big(I - \gamma \Phatv^{\pi^\star, V} \Big)^{-1} \sqrt{\mathrm{Var}_{\Phatv^{\pi^\star, V}}(V^{\star,\ror})}
& = \sqrt{\frac{1}{1-\gamma}}\sqrt{ \sum_{t=0}^\infty \gamma^t \left(\Phatv^{\pi^\star, V} \right)^t \mathrm{Var}_{\Phatv^{\pi^\star, V}}(V^{\star,\ror})}. \label{eq:bound-key-vstar}
\end{align}

To continue, we first focus on controlling $\mathrm{Var}_{\Phatv^{\pi^\star, V}}(V^{\star,\ror})$. Towards this, denoting the minimum value of $V^{\star,\ror}$ as $V_{\min} \defn \min_{s\in\cS} V^{\star,\ror}(s)$ and  $V' \defn   V^{\star,\ror} - V_{\min} 1 $, we arrive at (see the robust Bellman's consistency equation in \eqref{eq:r-bellman-matrix})
\begin{align}
V' = V^{\star,\ror} - V_{\min} 1 &= r_{\pi^\star} + \gamma \Pv^{\pi^\star, V} V^{\star,\ror} -  V_{\min} 1 \nonumber \\
&= r_{\pi^\star}  + \gamma \Phatv^{\pi^\star, V} V^{\star,\ror} + \gamma \Big( \Pv^{\pi^\star, V}- \Phatv^{\pi^\star, V} \Big) V^{\star,\ror} - V_{\min} 1\nonumber \\
&= r_{\pi^\star}  - (1-\gamma)V_{\min} 1  + \gamma \Phatv^{\pi^\star, V} V' + \gamma \Big( \Pv^{\pi^\star, V}- \Phatv^{\pi^\star, V} \Big) V^{\star,\ror} \nonumber \\
& = r_{\pi^\star}' + \gamma \Phatv^{\pi^\star, V} V' + \gamma \Big( \Pv^{\pi^\star, V}- \Phatv^{\pi^\star, V} \Big) V^{\star,\ror}, \label{eq:bellman-minus-vmin-vstar}
\end{align}
where the last line holds by letting $r_{\pi^\star}' \defn r_{\pi^\star}  - (1-\gamma)V_{\min} 1 \leq r_{\pi^\star}$.
With the above fact in hand, we control $\mathrm{Var}_{\Phatv^{\pi^\star, V}}(V^{\star,\ror})$ as follows:
\begin{align}
	\mathrm{Var}_{\Phatv^{\pi^\star, V}}(V^{\star,\ror})  &\overset{\mathrm{(i)}}{=} \mathrm{Var}_{\Phatv^{\pi^\star, V}}(V')  = \Phatv^{\pi^\star, V} \left(V' \circ V'\right) - \big(\Phatv^{\pi^\star, V} V'\big) \circ  \big(\Phatv^{\pi^\star, V} V' \big) \nonumber \\
	& \overset{\mathrm{(ii)}}{=} \Phatv^{\pi^\star, V} \left(V' \circ V'\right)  - \frac{1}{\gamma^2}\Big(V' - r_{\pi^\star}' - \gamma \Big( \Pv^{\pi^\star, V}- \Phatv^{\pi^\star, V} \Big) V^{\star,\ror} \Big)^{\circ 2} \nonumber \\
	& = \Phatv^{\pi^\star, V} \left(V' \circ V'\right) - \frac{1}{\gamma^2} V' \circ V' + \frac{2}{\gamma^2} V' \circ \Big(r_{\pi^\star}' + \gamma \Big( \Pv^{\pi^\star, V}- \Phatv^{\pi^\star, V} \Big) V^{\star,\ror}\Big) \nonumber \\
	& \quad -  \frac{1}{\gamma^2} \Big(r_{\pi^\star}' + \gamma \Big( \Pv^{\pi^\star, V}- \Phatv^{\pi^\star, V} \Big) V^{\star,\ror}\Big)^{\circ 2} \nonumber \\
	& \overset{\mathrm{(iii)}}{\leq} \Phatv^{\pi^\star, V} \left(V' \circ V'\right) - \frac{1}{\gamma} V' \circ V' + \frac{2}{\gamma^2} \|V'\|_\infty 1 + \frac{2}{\gamma} \|V'\|_\infty \Big| \Big( \Pv^{\pi^\star, V}- \Phatv^{\pi^\star, V} \Big) V^{\star,\ror}\Big| \label{eq:variance-tight-bound-vstar-repeat} \\
	& \leq \Phatv^{\pi^\star, V} \left(V' \circ V'\right) - \frac{1}{\gamma} V' \circ V' + \frac{2}{\gamma^2} \|V'\|_\infty 1 + \frac{6}{\gamma} \|V'\|_\infty  \sqrt{\frac{ \log(\frac{18SAN}{\delta})}{(1-\gamma)^2N}} 1, \label{eq:variance-tight-bound-vstar}
\end{align}
where (i) holds by the fact that $\mathrm{Var}_{P_\pi}(V -b 1) = \mathrm{Var}_{P_\pi}(V)$ for any scalar $b$ and  $V\in\mathbb{R}^S$, (ii) follows from \eqref{eq:bellman-minus-vmin-vstar}, (iii) arises from $\frac{1}{\gamma^2} V' \circ V' \geq \frac{1}{\gamma} V' \circ V'$ and $ - 1 \leq r_{\pi^\star}  - (1-\gamma)V_{\min} 1 =r_{\pi^\star}'\leq  r_{\pi^\star} \leq 1$, and the last inequality holds by Lemma~\ref{lemma:tv-dro-b-bound-star}.  

Plugging \eqref{eq:variance-tight-bound-vstar} into \eqref{eq:bound-key-vstar} leads to
\begin{align}
 & \Big(I - \gamma \Phatv^{\pi^\star, V} \Big)^{-1} \sqrt{\mathrm{Var}_{\Phatv^{\pi^\star, V}}(V^{\star,\ror})}  \notag \\
 & \leq \sqrt{\frac{1}{1-\gamma}}\sqrt{ \sum_{t=0}^\infty \gamma^t \left(\Phatv^{\pi^\star, V} \right)^t  \Bigg( \Phatv^{\pi^\star, V} \left(V' \circ V'\right) - \frac{1}{\gamma} V' \circ V' + \frac{2}{\gamma^2} \|V'\|_\infty 1 + \frac{6}{\gamma} \|V'\|_\infty  \sqrt{\frac{ \log(\frac{18SAN}{\delta})}{(1-\gamma)^2N}} 1 \Bigg) } \nonumber \\
 &  \overset{\mathrm{(i)}}{\leq} \sqrt{ \frac{1}{1-\gamma}} \sqrt{ \bigg| \sum_{t=0}^\infty \gamma^t \left(\Phatv^{\pi^\star, V} \right)^t \bigg( \Phatv^{\pi^\star, V} \left(V' \circ V'\right) - \frac{1}{\gamma} V' \circ V' \bigg) \bigg| } \nonumber \\
 &\qquad + \sqrt{\frac{1}{1-\gamma}}\sqrt{ \sum_{t=0}^\infty \gamma^t \left(\Phatv^{\pi^\star, V} \right)^t \Bigg( \frac{2}{\gamma^2} \|V'\|_\infty 1 + \frac{6}{\gamma} \|V'\|_\infty  \sqrt{\frac{ \log(\frac{18SAN}{\delta})}{(1-\gamma)^2N}} 1 \Bigg) } \notag \\
 & \leq \sqrt{ \frac{1}{1-\gamma}} \sqrt{   \bigg| \sum_{t=0}^\infty \gamma^t \left(\Phatv^{\pi^\star, V} \right)^t  \bigg[\Phatv^{\pi^\star, V} \left(V' \circ V'\right) - \frac{1}{\gamma} V' \circ V' \bigg] \bigg|    } + \sqrt{\frac{\left(2+ 6\sqrt{\frac{ \log(\frac{18SAN}{\delta})}{(1-\gamma)^2N}} \right) \|V'\|_\infty }{(1-\gamma)^2 \gamma^2}} 1, \label{eq:vstar-bound-key-recursive}
\end{align}
where (i) holds by the triangle inequality.
Therefore, the remainder of the proof shall focus on the first term, which follows
\begin{align}
	&  \bigg| \sum_{t=0}^\infty \gamma^t \Big(\Phatv^{\pi^\star, V} \Big)^t \Big( \Phatv^{\pi^\star, V} \left(V' \circ V'\right) - \frac{1}{\gamma} V' \circ V' \Big)  \bigg| \notag \\
	& = \bigg|  \bigg( \sum_{t=0}^\infty \gamma^t \Big(\Phatv^{\pi^\star, V} \Big)^{t+1} - \sum_{t=0}^\infty \gamma^{t-1} \Big(\Phatv^{\pi^\star, V} \Big)^{t}\bigg)\left(V' \circ V'\right) \bigg|  
	\leq \frac{1}{\gamma}\|V'\|_{\infty}^2 1 \label{eq:vstar-termI}
\end{align}
by recursion.
Inserting \eqref{eq:vstar-termI} back to \eqref{eq:vstar-bound-key-recursive} leads to
\begin{align}
&\bigg(I - \gamma \Phatv^{\pi^\star, V} \bigg)^{-1} \sqrt{\mathrm{Var}_{\Phatv^{\pi^\star, V}}(V^{\star,\ror})}   \notag\\
& \leq \sqrt{\frac{\|V'\|_{\infty}^2}{\gamma(1-\gamma)}} 1 +   3\sqrt{\frac{\Big(1+ \sqrt{\frac{ \log(\frac{18SAN}{\delta})}{(1-\gamma)^2N}} \Big) \|V'\|_\infty }{(1-\gamma)^2 \gamma^2}} 1  \nonumber \\
& \leq 4\sqrt{\frac{\Big(1+\sqrt{\frac{ \log(\frac{18SAN}{\delta})}{(1-\gamma)^2N}} \Big) \|V'\|_\infty }{(1-\gamma)^2 \gamma^2}} 1 \leq 4\sqrt{\frac{\Big(1+\sqrt{\frac{ \log(\frac{18SAN}{\delta})}{(1-\gamma)^2N}} \Big)  }{\gamma^3 (1-\gamma)^2  \max\{1-\gamma, \ror\}}} 1 \leq  4\sqrt{\frac{\Big(1+\sqrt{\frac{\log(\frac{18SAN}{\delta})}{(1-\gamma)^2N}} \Big) }{\gamma^3 (1-\gamma)^3 }} 1, \label{eq:tv-v-max-insert}
\end{align}
where the penultimate inequality follows from applying Lemma~\ref{lemma:tv-key-value-range} with $P = P^\no$ and $\pi = \pi^\star$:
\begin{align*}
\|V'\|_\infty = \max_{s\in\cS}V^{\star,\ror}(s) - \min_{s\in\cS} V^{\star,\ror}(s) \leq \frac{1}{\gamma \max\{1-\gamma, \ror\}}.
\end{align*}

\subsubsection{Proof of Lemma~\ref{lemma:dro-b-bound-infinite-loo-l1}}\label{proof:lemma:dro-b-bound-infinite-loo-l1}

To begin with, for any $(s,a) \in\cS\times \cA$, invoking the results in \eqref{eq:tv-t-that-gap}, we have
\begin{align}
 &\left| \pmhat^{\widehat{\pi}, \widehat{V}}_{s,a} \widehat{V}^{\widehat{\pi}, \ror }  -  \pmin^{\widehat{\pi}, \widehat{V}}_{s,a} \widehat{V}^{\widehat{\pi}, \ror }  \right| \leq \max_{\alpha\in [\min_s \widehat{V}^{\widehat{\pi}, \ror }(s), \max_s \widehat{V}^{\widehat{\pi}, \ror }(s)]} \left| \left(P^{\no}_{s,a} - \widehat{P}^{\no}_{s,a} \right) \big[\widehat{V}^{\widehat{\pi}, \ror }\big]_\alpha\right| \nonumber \\
  &\overset{\mathrm{(i)}}{\leq} \max_{\alpha\in [\min_s \widehat{V}^{\widehat{\pi}, \ror }(s), \max_s \widehat{V}^{\widehat{\pi}, \ror }(s)]}  \left(\left| \left(P^{\no}_{s,a} - \widehat{P}^{\no}_{s,a} \right) \big[\widehat{V}^{\star, \ror } \big]_\alpha\right| + \left| \left(P^{\no}_{s,a} - \widehat{P}^{\no}_{s,a} \right) \left( \big[\widehat{V}^{\widehat{\pi}, \ror } \big]_\alpha -  \big[\widehat{V}^{\star, \ror } \big]_\alpha \right) \right| \right) \nonumber \\
  & \leq \max_{\alpha\in [\min_s \widehat{V}^{\widehat{\pi}, \ror }(s), \max_s \widehat{V}^{\widehat{\pi}, \ror }(s)]} \Big(\left| \left(P^{\no}_{s,a} - \widehat{P}^{\no}_{s,a} \right) \big[ \widehat{V}^{\star, \ror } \right]_\alpha\big| +  \left\|P^{\no}_{s,a} - \widehat{P}^{\no}_{s,a} \right\|_1 \left\| \big[\widehat{V}^{\widehat{\pi}, \ror } \big]_\alpha -  \big[\widehat{V}^{\star, \ror } \right]_\alpha \big\|_\infty  \Big)  \nonumber \\
  & \overset{\mathrm{(ii)}}{\leq} \max_{\alpha\in [\min_s \widehat{V}^{\widehat{\pi}, \ror }(s), \max_s \widehat{V}^{\widehat{\pi}, \ror }(s)]} \left| \left(P^{\no}_{s,a} - \widehat{P}^{\no}_{s,a} \right) \big[\widehat{V}^{\star, \ror } \big]_\alpha\right| +  2\left\|\widehat{V}^{\widehat{\pi}, \ror }- \widehat{V}^{\star, \ror } \right\|_\infty   \nonumber \\
  & \overset{\mathrm{(iii)}}{\leq} \max_{\alpha\in [\min_s \widehat{V}^{\widehat{\pi}, \ror }(s), \max_s \widehat{V}^{\widehat{\pi}, \ror }(s)]}  \left| \left(P^{\no}_{s,a} - \widehat{P}^{\no}_{s,a} \right) \big[\widehat{V}^{\star, \ror }\big]_\alpha\right| + \frac{2\gamma \varepsilon_{\mathsf{opt}} }{1 -\gamma}, \label{eq:pihat-V-Vhat-gap}
\end{align}
where (i) holds by the triangle inequality, and (ii) follows from $\big\|P^{\no}_{s,a} - \widehat{P}^{\no}_{s,a} \big\|_1 \leq 2 $ and $ \big\|\big[\widehat{V}^{\widehat{\pi}, \ror }\big]_\alpha - \big[\widehat{V}^{\star, \ror }\big]_\alpha \big\|_\infty  \leq \big\|\widehat{V}^{\widehat{\pi}, \ror }- \widehat{V}^{\star, \ror } \big\|_\infty$, and (iii) follows from \eqref{eq:opt-error}.

To control $\left| \left(P^{\no}_{s,a} - \widehat{P}^{\no}_{s,a} \right) \big[\widehat{V}^{\star, \ror }\big]_\alpha\right|$ in \eqref{eq:pihat-V-Vhat-gap} for any given $\alpha \in \big[0,\frac{1}{1-\gamma}\big]$, and tame the dependency between $\widehat{V}^{\star, \ror}$ and $\widehat{P}^\no$, we resort to the following leave-one-out argument motivated by \citep{agarwal2020model,li2024settling,shi2022distributionally}. Specifically, we first construct a set of auxiliary RMDPs which simultaneously have the desired statistical  independence between robust value functions and the estimated nominal transition kernel, and are minimally different from the original RMDPs under consideration. Then we control the term of interest associated with these auxiliary RMDPs and show the value is close to the target quantity for the desired RMDP. The process is divided into several steps as below.

\paragraph{Step 1: construction of auxiliary RMDPs with deterministic empirical nominal transitions.}
Recall that we target the empirical infinite-horizon robust MDP $\widehat{\cM}_{\mathsf{rob}}$ with the nominal transition kernel $\widehat{P}^\no$. Towards this, we can construct an auxiliary robust MDP $\widehat{\cM}^{s,u}_{\mathsf{rob}}$ for each state $s$ and any non-negative scalar $u\geq 0$, so that it is the same as $\widehat{\cM}_{\mathsf{rob}}$ except for the transition properties in state $s$. In particular, we define the nominal transition kernel and reward function of $\widehat{\cM}^{s,u}_{\mathsf{rob}}$ as $P^{s,u}$ and
$r^{s,u}$, which are expressed as follows
 \begin{align}\label{eq:auxiliary-P-infinite}
  \begin{cases}
    P^{s,u}(s'\mymid s,a ) = \ind(s' = s) & \qquad \qquad \text{for all } (s', a) \in \cS\times \cA,  \\
    P^{s,u}(\cdot \mymid \widetilde{s} ,a ) = \widehat{P}^\no(\cdot \mymid \widetilde{s} ,a) & \qquad \qquad \text{for all } (\widetilde{s} ,a) \in \cS\times \cA \text{ and } \widetilde{s}  \neq s,
  \end{cases}
\end{align}
and
\begin{align}
\begin{cases}\label{eq:auxiliary-r-infinite}
    r^{s,u}(s,a) = u & \qquad \qquad \qquad  \text{for all } a \in \cA,  \\
    r^{s,u}(\widetilde{s},a) = r(\widetilde{s},a) & \qquad \qquad \qquad \text{for all } (\widetilde{s} ,a) \in \cS\times \cA \text{ and } \widetilde{s}  \neq s.
  \end{cases}
\end{align}
It is evident that the nominal transition probability at state $s$ of the auxiliary $\widehat{\cM}^{s,u}_{\mathsf{rob}}$, i.e. it never leaves state $s$ once entered. 
This useful property removes the randomness of $\widehat{P}^\no_{s,a}$ for all $a\in\cA$ in state $s$, which  will be leveraged later. 
 
Correspondingly, the robust Bellman operator $\widehat{\cT}^{\sigma}_{s,u}(\cdot)$ associated with the RMDP $\widehat{\cM}^{s,u}_{\mathsf{rob}}$ is defined as
\begin{align}\label{eq:aux-operator-auxiliary}
  \forall (\tilde{s},a)\in \cS\times \cA :\quad  \widehat{\cT}^{\sigma}_{s,u}(Q)(\tilde{s},a) = r^{s,u}(\tilde{s}, a)  + \gamma \inf_{ \cP \in \unb^{\sigma}(P^{s,u}_{\tilde{s},a})} \cP V, \qquad \text{with } V(\tilde{s}) = \max_a Q(\tilde{s},a).
\end{align}
\paragraph{Step 2: fixed-point equivalence between $\widehat{\cM}_{\mathsf{rob}}$ and the auxiliary RMDP $\widehat{\cM}^{s,u}_{\mathsf{rob}}$.}
Recall that $\widehat{Q}^{\star,\ror}$ is the unique fixed point of $\that^\ror(\cdot)$ with the corresponding robust value $\widehat{V}^{\star,\ror}$. We assert that the corresponding robust value function $\widehat{V}^{\star,\ror}_{s,u^\star}$ obtained from the fixed point of $\widehat{\cT}^{\sigma}_{s,u}(\cdot)$ aligns with the robust value function $\widehat{V}^{\star,\ror}$ derived from $\that^\ror(\cdot)$, as long as we choose $u$ in the following manner:
\begin{align}\label{eq:def-u-star}
 u^\star \defn u^{\star}(s) = \widehat{V}^{\star,\ror}(s) - \gamma \inf_{ \cP \in \unb^{\sigma}(e_{s})} \cP \widehat{V}^{\star,\ror}.
\end{align} 
where $e_s$ is the $s$-th standard basis vector in $\mathbb{R}^S$.
Towards verifying this, we shall break our arguments in two different cases.
\begin{itemize}

\item
{\bf For state $s$}: One has for any $a \in  \cA$:
\begin{align}
   r^{s,u^\star}(s, a)  + \gamma \inf_{ \cP \in \unb^{\sigma}(P^{s,u^\star}_{s,a})} \cP \widehat{V}^{\star,\ror}
  &= u^\star  + \gamma \inf_{ \cP \in \unb^{\sigma}(e_{s})} \cP \widehat{V}^{\star,\ror} \notag\\
  &  = \widehat{V}^{\star,\ror}(s) - \gamma \inf_{ \cP \in \unb^{\sigma}(e_{s})} \cP \widehat{V}^{\star,\ror} + \gamma \inf_{ \cP \in \unb^{\sigma}(e_{s})} \cP \widehat{V}^{\star,\ror}  = \widehat{V}^{\star,\ror}(s),
\end{align}
where the first equality follows from the definition of $P^{s,u^\star}_{s,a}$ in \eqref{eq:auxiliary-P-infinite}, and the second equality follows from plugging in the definition of $u^\star$ in \eqref{eq:def-u-star}.

\item {\bf For state $s' \neq s$}: It is easily verified that for all $a\in\cA$,
\begin{align}\label{eq:aux-opt-Q}
  r^{s,u^\star}(s', a)  + \gamma \inf_{ \cP \in \unb^{\sigma}(P^{s,u^\star}_{s',a})} \cP \widehat{V}^{\star,\ror}  
  &= r(s', a)  + \gamma \inf_{ \cP \in \unb^{\sigma}(\widehat{P}^\no_{s',a})} \cP \widehat{V}^{\star,\ror} \nonumber \\
  &= \widehat{\cT}^{\sigma}(\widehat{Q}^{\star,\ror})(s',a) =\widehat{Q}^{\star,\ror}(s',a),
\end{align}
where the first equality follows from the definitions in \eqref{eq:auxiliary-r-infinite} and \eqref{eq:auxiliary-P-infinite}, and the last line arises from the definition of the robust Bellman operator in \eqref{eq:robust_bellman_empirical}, and that $\widehat{Q}^{\star,\ror}$ is the fixed point of $\widehat{\cT}^{\sigma}(\cdot)$ (see Lemma~\ref{lem:contration-of-T}). 
 
\end{itemize}

Combining the facts in the above two cases, we establish that there exists a fixed point $\widehat{Q}^{\star,\ror}_{s,u^\star}$ of the operator $\widehat{\cT}^{\sigma}_{s,u^\star}(\cdot)$ by taking
\begin{align}
  \begin{cases}
    \widehat{Q}^{\star,\ror}_{s,u^\star}(s,a) = \widehat{V}^{\star,\ror}(s) & \qquad \qquad \qquad \text{for all } a \in \cA, \\
    \widehat{Q}^{\star,\ror}_{s,u^\star}(s',a) = \widehat{Q}^{\star,\ror}(s',a) & \qquad \qquad \qquad \text{for all } s' \neq s \text{ and } a \in \cA.
  \end{cases}\label{eq:auxiliary-mdp-Q-star}
\end{align}
Consequently, we confirm the existence of a fixed point of the operator $\widehat{\cT}^{\sigma}_{s,u^\star}(\cdot)$. In addition, its corresponding value function $\widehat{V}^{\star,\ror}_{s,u^\star}$ also coincides with $\widehat{V}^{\star,\ror}$.  Note that the corresponding facts between $\widehat{\cM}_{\mathsf{rob}}$ and $\widehat{\cM}_{\mathsf{rob}}^{s,u}$ in Step 1 and step 2 holds in fact for any uncertainty set.

\paragraph{Step 3: building an $\varepsilon$-net for all reward values $u$.}
It is easily verified that
\begin{align}\label{eq:chi2-def-u-star-bound}
  0 \leq u^\star \leq \widehat{V}^{\star,\ror}(s) \leq \frac{1}{1-\gamma}.
\end{align}
We can construct a $\cN_{\varepsilon_2}$-net over the interval $\big[0, \frac{1}{1-\gamma}\big]$, where the size is bounded by $|\cN_{\varepsilon_2}| \leq \frac{3}{\varepsilon_2(1-\gamma)}$ \citep{vershynin2018high}.
Following the same arguments in the proof of Lemma~\ref{lem:contration-of-T}, we can demonstrate that for each $u\in \cN_{\varepsilon_2}$, there exists a unique fixed point $\widehat{Q}^{\star,\ror}_{s,u}$ of the operator $\widehat{\cT}^{\sigma}_{s,u}(\cdot)$, which satisfies $  {0} \leq \widehat{Q}^{\star,\ror}_{s,u} \leq \frac{1}{1-\gamma} \cdot 1$. Consequently, the corresponding robust value function also satisfies $\left\|\widehat{V}^{\star,\ror}_{s,u} \right\|_\infty \leq \frac{1}{1-\gamma}$.

By the definitions in \eqref{eq:auxiliary-P-infinite} and  \eqref{eq:auxiliary-r-infinite}, we observe that for all $u\in \cN_{\varepsilon_2}$,  $\widehat{\cM}_{\mathsf{rob}}^{s,u}$ is statistically independent from $\widehat{P}^\no_{s,a}$. This independence indicates that $[\widehat{V}^{\star,\ror}_{s,u}]_\alpha$ and $\widehat{P}^\no_{s,a}$ are independent for a fixed $\alpha$. With this in mind, invoking the fact in \eqref{eq:V-p-phat-gap-one-alpha-bernstein-union} and \eqref{eq:V-p-phat-gap-one-alpha-hoeffding-union} and taking the union bound over all $(s,a, \alpha) \in\cS\times \cA \times \cN_{\varepsilon_1}$, $u\in \cN_{\varepsilon_2}$ yields that, with probability at least $1-\delta$, it holds for all $(s,a,u) \in \cS\times \cA \times  \cN_{\varepsilon_2}$ that   
\begin{align}\label{eq:approx-gap-union-bound-n1-n2}
  \max_{\alpha\in[0,1/(1-\gamma)]} \left| \left(P^{\no}_{s,a} - \widehat{P}^{\no}_{s,a} \right) \big[\widehat{V}^{\star,\ror}_{s,u}\big]_\alpha \right| &\leq \varepsilon_2 + 2\sqrt{\frac{\log(\frac{18SAN|N_{\varepsilon_2}|} {\delta})}{N}} \sqrt{\mathrm{Var}_{P^{\no}_{s,a}}(\widehat{V}^{\star,\ror}_{s,u})} +  \frac{2\log(\frac{18SAN|N_{\varepsilon_2}|}{\delta})}{3N(1-\gamma)} \notag \\
  & \leq \varepsilon_2 + 3\sqrt{\frac{\log(\frac{18SAN|N_{\varepsilon_2}|}{\delta})}{(1-\gamma)^2N}},
\end{align}
where the last inequality holds by the fact $\mathrm{Var}_{P^{\no}_{s,a}}(\widehat{V}^{\star,\ror}_{s,u}) \leq \|\widehat{V}^{\star,\ror}_{s,u}\|_\infty \leq \frac{1}{1-\gamma}$ and letting $N \geq \log \left(\frac{18SAN|N_{\varepsilon_2}|}{\delta} \right)$.

\paragraph{Step 4: uniform concentration.}
Recalling that $u^\star \in \big[0, \frac{1}{1-\gamma} \big]$ (see \eqref{eq:chi2-def-u-star-bound}), we can always find some $\overline{u} \in N_{\varepsilon_2}$ such that $|\overline{u} - u^\star| \leq \varepsilon_2$.  Consequently, plugging in the operator $\widehat{\cT}^{\sigma}_{s,u}(\cdot)$ in \eqref{eq:aux-operator-auxiliary} yields
\begin{align*}
  \forall Q\in \mathbb{R}^{SA} :\quad \Big\| \widehat{\cT}^{\sigma}_{s,\overline{u}}(Q) - \widehat{\cT}^{\sigma}_{s,u^\star}(Q) \Big\|_\infty = |\overline{u} - u^\star| \leq \varepsilon_2
\end{align*}

With this in mind, we observe that the fixed points of $\widehat{\cT}^{\sigma}_{s,\overline{u}}(\cdot)$ and $\widehat{\cT}^{\sigma}_{s,u^{\star}}(\cdot)$ obey
\begin{align*}
  \left\| \widehat{Q}^{\star,\ror}_{s,\overline{u}} -  \widehat{Q}^{\star,\ror}_{s,u^\star}\right\|_\infty &= \left\| \widehat{\cT}^{\sigma}_{s,\overline{u}}(\widehat{Q}^{\star,\ror}_{s,\overline{u}}) - \widehat{\cT}^{\sigma}_{s,u^\star}(\widehat{Q}^{\star,\ror}_{s, u^\star}) \right\|_\infty \notag \\
  &\leq \left\| \widehat{\cT}^{\sigma}_{s,\overline{u}}(\widehat{Q}^{\star,\ror}_{s,\overline{u}}) - \widehat{\cT}^{\sigma}_{s,\overline{u}}(\widehat{Q}^{\star,\ror}_{s, u^\star}) \right\|_\infty + \left\| \widehat{\cT}^{\sigma}_{s,\overline{u}}(\widehat{Q}^{\star,\ror}_{s, u^\star}) - \widehat{\cT}^{\sigma}_{s,u^\star}(\widehat{Q}^{\star,\ror}_{s, u^\star}) \right\|_\infty \notag \\
  & \leq \gamma \left\| \widehat{Q}^{\star,\ror}_{s,\overline{u}} -  \widehat{Q}^{\star,\ror}_{s,u^\star}\right\|_\infty + \varepsilon_2,
\end{align*}
where the last inequality holds by the fact that $\widehat{\cT}^{\sigma}_{s,u}(\cdot)$ is a $\gamma$-contraction. It directly indicates that
\begin{align} \label{eq:auxiliary-value-gap}
  \left\| \widehat{Q}^{\star,\ror}_{s,\overline{u}} -  \widehat{Q}^{\star,\ror}_{s,u^\star}\right\|_\infty  \leq \frac{\varepsilon_2}{(1-\gamma)}
\quad\mbox{and}\quad
  \left\| \widehat{V}^{\star,\ror}_{s,\overline{u}} -  \widehat{V}^{\star,\ror}_{s,u^\star}\right\|_\infty \leq \left\| \widehat{Q}^{\star,\ror}_{s,\overline{u}} -  \widehat{Q}^{\star,\ror}_{s,u^\star}\right\|_\infty \leq \frac{\varepsilon_2}{(1-\gamma)}.
\end{align}

Armed with the above facts, to control the first term in \eqref{eq:pihat-V-Vhat-gap}, invoking the identity $\widehat{V}^{\star,\ror} = \widehat{V}^{\star,\ror}_{s,u^\star}$ established in Step 2 gives that: for all $(s,a)\in\cS \times \cA$,
\begin{align}
&\max_{\alpha\in [\min_s \widehat{V}^{\widehat{\pi}, \ror }(s), \max_s \widehat{V}^{\widehat{\pi}, \ror }(s)]} \left| \left(P^{\no}_{s,a} - \widehat{P}^{\no}_{s,a} \right) [\widehat{V}^{\star, \ror }]_\alpha\right|   \notag \\
& \leq  \max_{\alpha\in[0,1/(1-\gamma)]}  \left| \left(P^{\no}_{s,a} - \widehat{P}^{\no}_{s,a} \right) [\widehat{V}^{\star,\ror}]_\alpha \right| = \max_{\alpha\in[0,1/(1-\gamma)]} \left| \left(P^{\no}_{s,a} - \widehat{P}^{\no}_{s,a} \right) [\widehat{V}^{\star,\ror}_{s,u^{\star}}]_\alpha \right| \nonumber \\
&\overset{\mathrm{(i)}}{\leq} \max_{\alpha\in[0,1/(1-\gamma)]} \left\{ \left| \left(P^{\no}_{s,a} - \widehat{P}^{\no}_{s,a} \right) [\widehat{V}^{\star,\ror}_{s,\overline{u}}]_\alpha \right| + \left| \left(P^{\no}_{s,a} - \widehat{P}^{\no}_{s,a} \right) \left([\widehat{V}^{\star,\ror}_{s,\overline{u}}]_\alpha  - [\widehat{V}^{\star,\ror}_{s, u^{\star}}]_\alpha \right) \right|  \right\} \nonumber \\
& \overset{\mathrm{(ii)}}{\leq} \max_{\alpha\in[0,1/(1-\gamma)]} \left| \left(P^{\no}_{s,a} - \widehat{P}^{\no}_{s,a} \right) [\widehat{V}^{\star,\ror}_{s,\overline{u}}]_\alpha \right|  + \frac{2\varepsilon_2}{(1-\gamma)} \nonumber \\
& \overset{\mathrm{(iii)}}{\leq}  \frac{2\varepsilon_2}{(1-\gamma)} + \varepsilon_2 + 2\sqrt{\frac{\log(\frac{18SAN|N_{\varepsilon_2}|} {\delta})}{N}} \sqrt{\mathrm{Var}_{P^{\no}_{s,a}}(\widehat{V}^{\star,\ror}_{s,u})} +  \frac{2\log(\frac{18SAN|N_{\varepsilon_2}|}{\delta})}{3N(1-\gamma)} \nonumber \\
&  \leq   \frac{3\varepsilon_2}{(1-\gamma)} + 2\sqrt{\frac{\log(\frac{18SAN|N_{\varepsilon_2}|} {\delta})}{N}} \sqrt{\mathrm{Var}_{P^{\no}_{s,a}}(\widehat{V}^{\star,\ror})} +  \frac{2\log(\frac{18SAN|N_{\varepsilon_2}|}{\delta})}{3N(1-\gamma)}  \notag \\
& \qquad + 2\sqrt{\frac{\log(\frac{18SAN|N_{\varepsilon_2}|} {\delta})}{N}} \sqrt{\left| \mathrm{Var}_{P^{\no}_{s,a}}(\widehat{V}^{\star,\ror}) - \mathrm{Var}_{P^{\no}_{s,a}}(\widehat{V}^{\star,\ror}_{s, \overline{u}}) \right|} \notag \\
&\overset{\mathrm{(iv)}}{\leq}   \frac{3\varepsilon_2}{(1-\gamma)} + 2\sqrt{\frac{\log(\frac{18SAN|N_{\varepsilon_2}|} {\delta})}{N}} \sqrt{\mathrm{Var}_{P^{\no}_{s,a}}(\widehat{V}^{\star,\ror})} +  \frac{2\log(\frac{18SAN|N_{\varepsilon_2}|}{\delta})}{3N(1-\gamma)}   + 2\sqrt{\frac{2 \varepsilon_2 \log(\frac{18SAN|N_{\varepsilon_2}|} {\delta})}{N(1-\gamma)^2}} \notag \\
& \leq 2\sqrt{\frac{\log(\frac{54SAN^2} {(1-\gamma)\delta})}{N}} \sqrt{\mathrm{Var}_{P^{\no}_{s,a}}(\widehat{V}^{\star,\ror})} +  \frac{8\log(\frac{54SAN^2 }{(1-\gamma)\delta})}{N(1-\gamma)} \label{eq:l1-Loo-summary} \\
& \leq 10\sqrt{\frac{\log(\frac{54SAN^2} {(1-\gamma)\delta})}{(1 -\gamma)^2 N}}, \label{eq:vhat-loo-hoeffindg-final1}
\end{align}
where (i) holds by  the triangle inequality, (ii) arises from (the last inequality holds by \eqref{eq:auxiliary-value-gap})
\begin{align}
\left| \left(P^{\no}_{s,a} - \widehat{P}^{\no}_{s,a} \right) \left([\widehat{V}^{\star,\ror}_{s,\overline{u}}]_\alpha  - [\widehat{V}^{\star,\ror}_{s, u^{\star}}]_\alpha \right) \right| &\leq \left\| P^{\no}_{s,a} - \widehat{P}^{\no}_{s,a} \right\|_1 \left\| [\widehat{V}^{\star,\ror}_{s,\overline{u}}]_\alpha  - [\widehat{V}^{\star,\ror}_{s, u^{\star}}]_\alpha \right\|_\infty  \notag \\
& \leq 2 \left\| \widehat{V}^{\star,\ror}_{s,\overline{u}}  - \widehat{V}^{\star,\ror}_{s, u^{\star}} \right\|_\infty \leq \frac{2\varepsilon_2}{(1-\gamma)}, 
\end{align}
(iii) follows from \eqref{eq:approx-gap-union-bound-n1-n2},
(iv) can be verified by applying Lemma~\ref{eq:tv-auxiliary-lemma} with \eqref{eq:auxiliary-value-gap}. Here, the penultimate inequality holds by letting $\varepsilon_2 = \frac{\log(\frac{18SAN|N_{\varepsilon_2}|} {\delta})}{N}$, which leads to $|N_{\varepsilon_2}| \leq \frac{3}{\varepsilon_2(1-\gamma)} \leq \frac{3N}{1-\gamma}$, and the last inequality holds by the fact $\mathrm{Var}_{P^{\no}_{s,a}}(\widehat{V}^{\star,\ror}) \leq \|\widehat{V}^{\star,\ror}\|_\infty \leq \frac{1}{1-\gamma}$ and letting $N \geq \log \left(\frac{54SAN^2 }{(1-\gamma)\delta} \right)$.

\paragraph{Step 5: finishing up.}
Inserting \eqref{eq:l1-Loo-summary}  and \eqref{eq:vhat-loo-hoeffindg-final1} back into \eqref{eq:pihat-V-Vhat-gap} and combining with \eqref{eq:vhat-loo-hoeffindg-final1} give that with probability at least $1-\delta$, 
\begin{align}
\left| \pmhat^{\widehat{\pi}, \widehat{V}}_{s,a} \widehat{V}^{\widehat{\pi}, \ror }  -  \pmin^{\widehat{\pi}, \widehat{V}}_{s,a} \widehat{V}^{\widehat{\pi}, \ror }  \right| &\leq \max_{\alpha\in [\min_s \widehat{V}^{\widehat{\pi}, \ror }(s), \max_s \widehat{V}^{\widehat{\pi}, \ror }(s)]} \left| \left(P^{\no}_{s,a} - \widehat{P}^{\no}_{s,a} \right) [\widehat{V}^{\star, \ror }]_\alpha\right| + \frac{2\gamma \varepsilon_{\mathsf{opt}} }{1 -\gamma} \notag \\
&\leq  \max_{\alpha\in[0,1/(1-\gamma)]}  \left| \left(P^{\no}_{s,a} - \widehat{P}^{\no}_{s,a} \right) [\widehat{V}^{\star,\ror}]_\alpha \right|  + \frac{2\gamma \varepsilon_{\mathsf{opt}} }{1 -\gamma} \nonumber \\
&\leq 2\sqrt{\frac{\log(\frac{54SAN^2} {(1-\gamma)\delta})}{N}} \sqrt{\mathrm{Var}_{P^{\no}_{s,a}}(\widehat{V}^{\star,\ror})} +  \frac{8\log(\frac{54SAN^2 }{(1-\gamma)\delta})}{N(1-\gamma)} + \frac{2\gamma \varepsilon_{\mathsf{opt}} }{1 -\gamma} \notag\\
& \leq 10\sqrt{\frac{\log(\frac{54SAN^2} {(1-\gamma)\delta})}{(1 -\gamma)^2 N}} + \frac{2\gamma \varepsilon_{\mathsf{opt}} }{1 -\gamma}  \label{eq:vhat-loo-hoeffindg-final2}
\end{align}
holds for all $(s,a)\in\cS\times \cA$.

Finally, we complete the proof by compiling everything into the matrix form as follows:
\begin{align}
\bigg| \Phatv^{\widehat{\pi}, \widehat{V}} \widehat{V}^{\widehat{\pi}, \ror }  -  \Pv^{\widehat{\pi}, \widehat{V}} \widehat{V}^{\widehat{\pi}, \ror } \bigg| &\leq 2\sqrt{\frac{\log(\frac{54SAN^2} {(1-\gamma)\delta})}{N}} \sqrt{\mathrm{Var}_{P^{\no}_{s,a}}(\widehat{V}^{\star,\ror})} 1 +  \frac{8\log(\frac{54SAN^2 }{(1-\gamma)\delta})}{N(1-\gamma)} 1 + \frac{2\gamma \varepsilon_{\mathsf{opt}} }{1 -\gamma} 1\notag\\
& \leq 10\sqrt{\frac{\log(\frac{54SAN^2} {(1-\gamma)\delta})}{(1 -\gamma)^2 N}} 1 + \frac{2\gamma \varepsilon_{\mathsf{opt}} }{1 -\gamma} 1. \label{eq:vhat-loo-final}
\end{align}

\subsubsection{Proof of Lemma~\ref{lemma:tv-vhat-key1}} \label{proof:lemma:tv-vhat-key1}

The proof can be achieved by directly applying the same routine as Appendix~\ref{proof:lemma:tv-key0}. Towards this, similar to \eqref{eq:bound-key-vstar}, we arrive at
\begin{align}
\Big(I - \gamma \Pv^{\widehat{\pi}, \widehat{V}} \Big)^{-1}  \sqrt{\mathrm{Var}_{\Pv^{\widehat{\pi}, \widehat{V}}}(\widehat{V}^{\widehat{\pi}, \ror })} \leq \sqrt{\frac{1}{1-\gamma}}\sqrt{ \sum_{t=0}^\infty \gamma^t \Big(\Pv^{\widehat{\pi}, \widehat{V}} \Big)^t \mathrm{Var}_{\Pv^{\widehat{\pi}, \widehat{V}}} (\widehat{V}^{\widehat{\pi}, \ror })}. \label{eq:bound-key-vhat}
\end{align}

To control $\mathrm{Var}_{\Pv^{\widehat{\pi}, \widehat{V}}} (\widehat{V}^{\widehat{\pi}, \ror })$, we denote the minimum value of $\widehat{V}^{\widehat{\pi}, \ror }$ as $V_{\min} = \min_{s\in\cS} \widehat{V}^{\widehat{\pi}, \ror }(s)$ and $V' \defn  \widehat{V}^{\widehat{\pi}, \ror } - V_{\min} 1$. By the same argument as \eqref{eq:variance-tight-bound-vstar-repeat}, we arrive at 
\begin{align}
	&\mathrm{Var}_{\Pv^{\widehat{\pi}, \widehat{V}}} (\widehat{V}^{\widehat{\pi}, \ror })  \notag \\
	& \leq \Pv^{\widehat{\pi}, \widehat{V}}  \left(V' \circ V'\right) - \frac{1}{\gamma} V' \circ V' + \frac{2}{\gamma^2} \|V'\|_\infty 1 + \frac{2}{\gamma} \|V'\|_\infty \left| \left(  \Phatv^{\widehat{\pi}, \widehat{V} }- \Pv^{\widehat{\pi}, \widehat{V} } \right) \widehat{V}^{\widehat{\pi}, \ror } \right| \nonumber \\
	& \leq \Pv^{\widehat{\pi}, \widehat{V}} \left(V' \circ V'\right) - \frac{1}{\gamma} V' \circ V' + \frac{2}{\gamma^2} \|V'\|_\infty 1 + \frac{2}{\gamma} \|V'\|_\infty \Bigg(10\sqrt{\frac{\log(\frac{54SAN^2} {(1-\gamma)\delta})}{(1 -\gamma)^2 N}}   + \frac{2\gamma \varepsilon_{\mathsf{opt}} }{1 -\gamma} \Bigg) 1, \label{eq:variance-tight-bound-vhat}
\end{align}
where the last inequality makes use of Lemma~\ref{lemma:dro-b-bound-infinite-loo-l1}.
Plugging \eqref{eq:variance-tight-bound-vhat} back into \eqref{eq:bound-key-vhat} leads to
\begin{align}
\Big(I - \gamma \Pv^{\widehat{\pi}, \widehat{V}} \Big)^{-1}  \sqrt{\mathrm{Var}_{\Pv^{\widehat{\pi}, \widehat{V}}}(\widehat{V}^{\widehat{\pi}, \ror })}
 &  \overset{\mathrm{(i)}}{\leq}  \sqrt{ \frac{1}{1-\gamma}} \sqrt{ \bigg| \sum_{t=0}^\infty \gamma^t \left(\Pv^{\widehat{\pi}, \widehat{V}} \right)^t \Big( \Pv^{\widehat{\pi}, \widehat{V}}  \left(V' \circ V'\right) - \frac{1}{\gamma} V' \circ V' \Big) \bigg|  }  \notag \\
 &\quad + \sqrt{\frac{1}{(1-\gamma)^2 \gamma^2}\bigg(2 + 20\sqrt{\frac{\log(\frac{54SAN^2} {(1-\gamma)\delta})}{(1 -\gamma)^2 N}} + \frac{2\gamma \varepsilon_{\mathsf{opt}} }{1 -\gamma}  \bigg) \|V'\|_\infty } 1 \notag \\
 & \overset{\mathrm{(ii)}}{\leq} \sqrt{\frac{\|V'\|_{\infty}^2}{\gamma(1-\gamma)}} 1 + \sqrt{\frac{\bigg(2 + 20\sqrt{\frac{\log(\frac{54SAN^2} {(1-\gamma)\delta})}{(1 -\gamma)^2 N}} + \frac{2\gamma \varepsilon_{\mathsf{opt}} }{1 -\gamma}  \bigg) \|V'\|_\infty }{(1-\gamma)^2 \gamma^2}} 1  \notag \\
 & \overset{\mathrm{(iii)}}{\leq} \sqrt{\frac{\|V'\|_{\infty}^2}{\gamma(1-\gamma)}} 1 + \sqrt{\frac{24 \|V'\|_\infty }{(1-\gamma)^2 \gamma^2}} 1  \leq  6\sqrt{\frac{\|V'\|_\infty }{(1-\gamma)^2 \gamma^2}} 1  ,\label{eq:vstar-bound-key-recursive-vhat}
\end{align}
where (i) arises from following the routine of \eqref{eq:vstar-bound-key-recursive}, (ii) holds by repeating the argument of \eqref{eq:vstar-termI}, (iii) follows by taking $ N \geq\frac{\log(\frac{54SAN^2}{(1-\gamma)\delta})}{(1-\gamma)^2}$ and $\varepsilon_{\mathsf{opt}}\leq \frac{1-\gamma}{\gamma}$, and the     last inequality holds by $\|V'\|_\infty \leq \|V^{\star,\ror}\|_\infty \leq \frac{1}{1-\gamma}$.

Finally, applying Lemma~\ref{lemma:tv-key-value-range} with $P = \widehat{P}^\no$ and $\pi = \widehat{\pi}$ yields
\begin{align*}
\|V'\|_\infty \leq \max_{s\in\cS}\widehat{V}^{\widehat{\pi}, \ror }(s) - \min_{s\in\cS} \widehat{V}^{\widehat{\pi}, \ror }(s) \leq \frac{1}{\gamma \max\{1-\gamma, \ror\}},
\end{align*}
which can be inserted into \eqref{eq:vstar-bound-key-recursive-vhat} and gives
\begin{align*}
\left(I - \gamma \Pv^{\widehat{\pi}, \widehat{V}} \right)^{-1}  \sqrt{\mathrm{Var}_{\Pv^{\widehat{\pi}, \widehat{V}}}(\widehat{V}^{\widehat{\pi}, \ror })} \leq 6\sqrt{\frac{1 }{\gamma^3 (1-\gamma)^2 \max\{1-\gamma, \ror\} }} 1 \leq  6\sqrt{\frac{1}{(1-\gamma)^3 \gamma^2}} 1.
\end{align*}


\section{Proof of the lower bound with TV distance: Theorem~\ref{thm:l1-lower-bound}} \label{proof:thm:l1-lower-bound-2}

\subsection{Construction of the hard problem instances} \label{sec:tv-lower-bound-instance}

First, note that we shall use the same MDPs defined in section~\ref{proof:thm:l1-lower-bound} as follows
\begin{align*}
   \left\{ \cM_\phi=
    \left(\mathcal{S}, \mathcal{A}, P^{\phi}, r, \gamma \right) 
    \mymid \phi = \{0,1\}
    \right\}.
\end{align*}
In particular, we shall keep the structure of the transition kernel in \eqref{eq:Ph-construction-lower-infinite}, reward function in \eqref{eq:rh-construction-lower-bound-infinite} and initial state distribution in \eqref{eq:rho-defn-infinite-LB}, while $p$ and $\Delta$ shall be specified according to TV distance case.
  
\paragraph{Uncertainty set of the transition kernels.}
Recalling the uncertainty set assumed throughout this section is defined as $\cU^{\ror}(P^\phi)$ with TV distance:
\begin{align}
\unb^{\ror}(P^\phi) \defn \cU^{\ror}_{\mathsf{TV}}(P^\phi) = \otimes \; \cU^{\ror}_{\mathsf{TV}}(P^\phi_{s,a}),\qquad &\cU^{\ror}_{\mathsf{TV}}(P^\phi_{s,a}) \defn \Big\{ P'_{s,a} \in \Delta(\cS): \frac{1}{2}\left\| P'_{s,a} - P^\phi_{s,a}\right\|_1 \leq \ror \Big\}, \label{eq:tv-ball-infinite-P-recall1}
\end{align}
where $P^{\phi}_{s,a} \defn P^{\phi}(\cdot \mymid s,a)$ is defined similar to \eqref{eq:defn-P-sa}.
In addition, without loss of generality, we recall the radius $\ror \in (0, 1-c_0]$ with $0< c_0 < 1$. With the uncertainty level in hand, taking $c_1 \defn \frac{c_0}{2}$,  $p$ and $\Delta$ which determines the instances obey
\begin{align}\label{eq:p-q-defn-infinite2}
    p = \left(1 + c_1 \right) \max\{ 1-\gamma, \sigma\} \qquad \text{and} \qquad \Delta \leq c_1 \max\{ 1-\gamma, \sigma\},
\end{align}
which ensure $ 0 \leq p \leq 1 $ as follows:
\begin{align}\label{eq:tv-1-p-bound}
\left(1 + c_1 \right)\ror  & \leq 1- c_0 + c_1 \ror \leq 1-\frac{c_0}{2} < 1,  \qquad 
\left(1+ c_1 \right) (1-\gamma) \leq \frac{3}{2} (1-\gamma) \leq \frac{3}{4} < 1.
\end{align}
Consequently, applying \eqref{eq:p-q-defn-infinite} directly leads to
\begin{align}\label{eq:infinite-p-q-bound}
    p \geq q \geq  \max\{ 1-\gamma, \sigma\}.
\end{align}

To continue, for any $(s,a,s')\in\cS\times \cA \times \cS$, we denote the infimum probability of moving to the next state $s'$ associated with any perturbed transition kernel $P_{s,a} \in \unb^{\ror}(P^{\phi}_{s,a})$ as  
\begin{align}\label{eq:infinite-lw-def-p-q}
\underline{P}^{\phi}(s' \mymid s,a) &\defn \inf_{P_{s,a} \in \unb^{\ror}(P^{\phi}_{s,a})} P(s'  \mymid s,a) = \max \{P(s'  \mymid s,a) - \ror, 0\},
\end{align}
where the last equation can be easily verified by the definition of  $\cU^{\ror}(P^\phi)$ in \eqref{eq:tv-ball-infinite-P-recall1}.
As shall be seen, the transition from state $0$ to state $1$ plays an important role in the analysis, for convenience, we denote
\begin{align}\label{eq:infinite-lw-p-q-perturb-inf}
\underline{p} &\defn \underline{P}^{\phi}(1 \mymid 0,\phi) = p - \ror ,\qquad \underline{q}  \defn \underline{P}^{\phi}(1  \mymid 0, 1-\phi) = q - \ror,
\end{align}
which follows from the fact that $p\geq q\geq \ror $ in \eqref{eq:infinite-p-q-bound}.

\paragraph{Robust value functions and robust optimal policies.}
To proceed, we are ready to derive the corresponding robust value functions, identify the optimal policies, and characterize the optimal values. For any MDP $\cM_\phi$ with the above uncertainty set, we denote $\pi^{\star}_\phi$ as the optimal policy, and the robust value function of any policy $\pi$ (resp.~the optimal policy $\pi^{\star}_\phi$) as $V^{\pi,\ror}_\phi$ (resp.~$V^{\star,\ror}_\phi$). Then, we introduce the following lemma which describes some important properties of the robust (optimal) value functions and optimal policies. The proof is postponed to Appendix~\ref{proof:lem:l1-lb-value}.

\begin{lemma}\label{lem:l1-lb-value}
For any $\phi = \{0,1\}$ and any policy $\pi$, the robust value function obeys
\begin{align}
    V^{\pi, \ror}_\phi(0) = \frac{ \gamma \big( z_{\phi}^{\pi} - \ror \big)  }{ (1-\gamma) \bigg( 1 + \frac{\gamma \left( z_{\phi}^{\pi} - \ror \right)}{1-\gamma \left(1-\ror \right)} \bigg) \left( 1-\gamma \left(1- \ror \right) \right)},
    \label{eq:infinite-lemma-value-0-pi}
\end{align}
where $z_{\phi}^{\pi}$ is defined as
\begin{align}
z_{\phi}^{\pi} \defn p\pi(\phi\mymid 0) + q \pi(1-\phi \mymid 0).\label{eq:infinite-x-h}
\end{align}
In addition, the robust optimal value functions and the robust optimal policies satisfy
\begin{subequations}
    \label{eq:l1-value-lemma}
\begin{align}
    V_\phi^{\star,\sigma}(0) &= \frac{ \gamma \left( p  - \ror \right)}{ (1-\gamma) \left( 1 + \frac{\gamma \left( p - \ror \right)}{1-\gamma \left(1-\ror \right)} \right) \left(1-\gamma \left(1-\ror \right) \right)}, \\
    \pi_\phi^{\star }(\phi \mymid s) &= 1, \qquad  \qquad \text{ for } s \in\cS.
\end{align}
\end{subequations}

\end{lemma}

\subsection{Establishing the minimax lower bound} \label{sec:tv-lower-bound-final-pipeline}
Note that our goal is to control the quantity w.r.t. any policy estimator $\widehat{\pi}$ based on the chosen initial distribution $\varphi$ in \eqref{eq:rho-defn-infinite-LB} and the dataset consisting of $N$ samples over each state-action pair generated from the nominal transition kernel $P^\phi$, which gives
\begin{align*}
  \big\langle \varphi, V^{\star, \sigma}_{\phi} - V^{\widehat{\pi}, \sigma}_{\phi} \big\rangle   = V^{\star,\ror}_\phi(0) - V^{\widehat{\pi},\ror}_\phi(0).
\end{align*}

\paragraph{Step 1: converting the goal to estimate $\phi$.} 
We make the following useful claim which shall be verified in Appendix~\ref{proof:tv-lower-diff-control}: With $\varepsilon \leq \frac{c_1}{32(1-\gamma)}$, 
letting 
\begin{align}\label{eq:Delta-chosen}
    \Delta  = 32  (1-\gamma) \max\{ 1 -\gamma, \sigma\} \varepsilon \leq c_1 \max\{ 1-\gamma, \sigma\} 
\end{align}
which satisfies \eqref{eq:p-q-defn-infinite2},
it leads to that for any policy $\widehat{\pi}$, 
\begin{align}
    \big\langle \varphi, V^{\star, \sigma}_{\phi} - V^{\widehat{\pi}, \sigma}_{\phi} \big\rangle  \geq 2\varepsilon \big(1-\widehat{\pi}(\phi\mymid 0)\big). \label{eq:tv-Value-0-recursive}
\end{align}

With this connection established between the policy $\widehat{\pi}$ and its sub-optimality gap as depicted in \eqref{eq:tv-Value-0-recursive}, we can now proceed to build an estimate for $\phi$. Here, we denote $\mathbb{P}_\phi$ as the probability distribution when the MDP is $\mathcal{M}_\phi$, where $\phi$ can take on values in the set $\{0,1\}$.

Let's assume momentarily that an estimated policy $\widehat{\pi}$ achieves
\begin{align}
    \mathbb{P}_\phi \big\{ \big\langle \varphi, V^{\star, \sigma}_{\phi} - V^{\widehat{\pi}, \sigma}_{\phi} \big\rangle  \leq \varepsilon\big\} \geq \frac{7}{8},
    \label{eq:assumption-theta-small-LB-finite}
\end{align}
then in view of \eqref{eq:tv-Value-0-recursive}, 
we necessarily have $\widehat{\pi}(\phi\mymid 0) \geq \frac{1}{2}$ with probability at least $\frac{7}{8}$.
With this in mind, we are motivated to construct the following estimate $\widehat{\phi}$ for $\phi\in \{0,1\}$: 
\begin{align}
    \widehat{\phi}=\arg\max_{a\in \{0,1\}} \, \widehat{\pi}(a\mymid 0),
    \label{eq:defn-theta-hat-inf-LB}
\end{align}
which obeys
\begin{align}
    \mathbb{P}_{\phi}\big\{ \widehat{\phi} = \phi \big\} 
    \geq \mathbb{P}_{\phi}\big\{ \widehat{\pi}(\phi \mymid 0) > 1/2 \big\} \geq \frac{7}{8}. 
    \label{eq:P-theta-accuracy-inf}
\end{align}
Subsequently, our aim is to demonstrate that \eqref{eq:P-theta-accuracy-inf} cannot occur without an adequate number of samples, which would in turn contradict \eqref{eq:tv-Value-0-recursive}.

\paragraph{Step 2: probability of error in testing two hypotheses.}
Equipped with the aforementioned groundwork, we can now delve into differentiating between the two hypotheses $\phi \in\{ 0, 1\}$. 
To achieve this, we consider the concept of minimax probability of error, defined as follows:
\begin{equation}
    p_{\mathrm{e}} \coloneqq \inf_{\psi}\max \big\{ \mathbb{P}_{0}(\psi \neq 0), \, \mathbb{P}_{1}(\psi \neq 1) \big\} . \label{eq:error-prob-two-hypotheses-finite-LB}
\end{equation}
Here, the infimum is taken over all possible tests $\psi$ constructed from the samples generated from the nominal transition kernel $P^\phi$.

Moving forward, let us  denote $\mu_{\phi}$ (resp.~$\mu_{\phi}(s)$) as the distribution of a sample tuple $(s_i, a_i, s_i')$ under the nominal transition kernel $P^\phi$ associated with $\mathcal{M}_\phi$ and the samples are generated independently. Applying standard results from \citet[Theorem~2.2]{tsybakov2009introduction} and  the additivity of the KL divergence (cf.~\citet[Page~85]{tsybakov2009introduction}), we obtain 
\begin{align}
p_{\mathrm{e}} &  \geq \frac{1}{4}\exp\Big(- NSA \cdot \mathsf{KL} \big(\mu_{0}\parallel \mu_{1} \big) \Big)\nonumber\\
    & = \frac{1}{4}\exp\Big\{- N  \Big( \mathsf{KL}\big(P^0(\cdot\mymid 0, 0)\parallel P^1(\cdot\mymid0,0)\big)+\mathsf{KL}\big(P^0(\cdot\mymid 0,1)\parallel P^1(\cdot\mymid0,1)\big) \Big)\Big\},
    \label{eq:finite-remainder-KL}
\end{align}
where the last inequality holds by observing that 
\begin{align}
\mathsf{KL} \big(\mu_{0}\parallel \mu_{1} \big) & = \frac{1}{SA}\sum_{s,a,s'} \mathsf{KL}\big(P^0(s' \mymid s, a)\parallel P^1( s' \mymid s, a)\big)   \notag \\
    & =   \frac{1}{SA}\sum_{a\in\{0,1\}}\mathsf{KL}\big(P^{0}(\cdot\mymid0,a)\parallel P^{1}(\cdot\mymid0,a)\big), \notag 
\end{align}
Here, the last equality holds by the fact that $P^{0}(\cdot\mymid s,a)$ and $P^{1}(\cdot\mymid s,a)$ only differ when $s=0$.

Now, our focus shifts towards bounding the terms involving the KL divergence in \eqref{eq:finite-remainder-KL}.
Given $p \geq q \geq \max\{ 1-\gamma, \sigma\}$ (cf.~\eqref{eq:infinite-p-q-bound}), applying \citet[Lemma~2.7]{tsybakov2009introduction} gives
\begin{align}
\mathsf{KL}\big(P^{0}(\cdot \mymid 0, 1)\parallel P^{1}(\cdot \mymid 0, 1)\big) & =\mathsf{KL}\left(p\parallel q \right)  \leq \frac{(p-q)^2}{(1-p)p}  \overset{\mathrm{(i)}}{=} \frac{\Delta^2}{p(1-p)} \notag\\
    & \overset{\mathrm{(ii)}}{=} \frac{ 1024  (1-\gamma)^2 \max\{ 1 -\gamma, \sigma\}^2 \varepsilon^2 }{p(1-p)} \notag \\
    & \leq  \frac{ 1024 (1-\gamma)^2 \max\{ 1 -\gamma, \sigma\} \varepsilon^2 }{1-p} \leq \frac{4096}{c_1} (1-\gamma)^2 \max\{ 1 -\gamma, \sigma\} \varepsilon^2,
    \label{eq:finite-KL-bounded}
\end{align}
where (i) stems from the definition in \eqref{eq:p-q-defn-infinite}, (ii) follows by the expression of $\Delta$ in \eqref{eq:Delta-chosen}, and the last inequality arises from $1- q \geq 1-p \geq \frac{c_0}{4} $ (see \eqref{eq:tv-1-p-bound}).

Note that it can be shown that $\mathsf{KL}\big(P^{0}(\cdot \mymid 0, 0)\parallel P^{1}(\cdot \mymid 0, 0)\big)$ can be upper bounded in a same manner. 
Substituting \eqref{eq:finite-KL-bounded} back into \eqref{eq:finite-remainder-KL} 
demonstrates that: if the sample size is selected as
\begin{align}\label{eq:finite-sample-N-condition}
    N \leq \frac{c_1 \log 2}{ 8192 (1-\gamma)^2 \max\{ 1 -\gamma, \sigma\} \varepsilon^2},
\end{align}
then one necessarily has
\begin{align}
    p_{\mathrm{e}} &\geq \frac{1}{4}\exp\bigg\{- N \frac{8192}{c_1} (1-\gamma)^2 \max\{ 1 -\gamma, \sigma\} \varepsilon^2 \bigg\}  \geq \frac{1}{8}, \label{eq:pe-LB-13579-inf}
\end{align}

\paragraph{Step 3: putting the results together.} 
Lastly, suppose that there exists an estimator $\widehat{\pi}$ such that
\[
    \mathbb{P}_0 \big\{  \big\langle \varphi, V_0^{\star, \sigma} - V_0^{\widehat{\pi}, \sigma} \big\rangle > \varepsilon  \big\} < \frac{1}{8}
    \qquad \text{and} \qquad
    \mathbb{P}_1 \big\{  \big\langle \varphi, V_1^{\star,\sigma} - V_1^{\widehat{\pi},\sigma} \big\rangle > \varepsilon  \big\} < \frac{1}{8}.
\]
According to Step 1, the estimator $\widehat{\phi}$ defined in \eqref{eq:defn-theta-hat-inf-LB} must satisfy
\[
    \mathbb{P}_0\big(\widehat{\phi} \neq 0\big) < \frac{1}{8} 
    \qquad \text{and} \qquad
    \mathbb{P}_1\big(\widehat{\phi} \neq 1\big) < \frac{1}{8}.
\]
However, this cannot occur under the sample size condition \eqref{eq:finite-sample-N-condition} to avoid contradiction with \eqref{eq:pe-LB-13579-inf}. Thus, we have completed the proof.

\subsection{Proof of the auxiliary facts}

\subsubsection{Proof of Lemma~\ref{lem:l1-lb-value}}\label{proof:lem:l1-lb-value}

\paragraph{Deriving the robust value function over different states.}
For any $\cM_\phi$ with $\phi\in\{0,1 \}$, we first characterize the robust value function of any policy $\pi$ over different states. Before proceeding, we denote the minimum of the robust value function over states as below:
\begin{equation}
    V_{\phi, \min}^{\pi, \sigma}  \defn \min_{s\in \cS} V_{\phi}^{\pi,\sigma}(s).
\end{equation}
Clearly, there exists at least one state $s_{\phi, \min}^{\pi}$ that satisfies $ V_{\phi}^{\pi,\sigma}(s_{\phi, \min}^{\pi}) = V_{\phi, \min}^{\pi, \sigma}$.

With this in mind, it is easily observed that for any policy $\pi$, the robust value function at state $s=1$ obeys
\begin{align}
  V_\phi^{\pi,\ror}(1) &= \mathbb{E}_{a\sim \pi(\cdot  \mymid 1)} \bigg[r(1,a) + \gamma \inf_{ \cP \in \unb^{\sigma}(P^{\phi}_{1,a})}\cP V_\phi^{\pi,\ror} \bigg] \notag \\
  &\overset{\mathrm{(i)}}{=} 1 + \gamma \mathbb{E}_{a\sim \pi(\cdot  \mymid 1)} \left[\underline{P}^{\phi}(1 \mymid 1,a) V_\phi^{\pi,\ror}(1) \right] + \gamma \ror  V_{\phi, \min}^{\pi, \sigma} \overset{\mathrm{(ii)}}{=} 1 + \gamma (1-\ror) V_\phi^{\pi,\ror}(1) + \gamma \ror  V_{\phi, \min}^{\pi, \sigma}, \label{eq:tv-s-value1}
\end{align}
where (i) holds by $r(1,a)=1$ for all $a\in\cA'$ and \eqref{eq:infinite-lw-def-p-q}, and (ii) follows from $P^{\phi}(1\mymid 1,a)=1$ for all $a\in\cA'$.

Similarly, for any $s\in\{2,3,\cdots, S-1\}$, we have
  \begin{align}
   V_\phi^{\pi,\ror}(s) &= 0 + \gamma \mathbb{E}_{a\sim \pi(\cdot  \mymid s)} \left[ \underline{P}^{\phi}(1 \mymid s, a) V_\phi^{\pi,\ror}(1)\right] +  \gamma \ror  V_{\phi, \min}^{\pi, \sigma} \nonumber \\
  & = \gamma \left(1-\ror \right) V_\phi^{\pi,\ror}(1) + \gamma \ror  V_{\phi, \min}^{\pi, \sigma},  \label{eq:tv-s-other-value1}
\end{align}
since $r(s,a)=0$ for all $s\in\{2,3,\cdots, S-1\}$ and the definition in  \eqref{eq:infinite-lw-def-p-q}.

Finally, we move onto compute $V_\phi^{\pi,\ror}(0)$, the robust value function at state $0$ associated with any policy $\pi$. First, it obeys
\begin{align}
 V_\phi^{\pi,\ror}(0) &= \mathbb{E}_{a \sim \pi(\cdot \mymid 0)} \bigg[ r(0,a) + \gamma \inf_{ \cP \in \unb^{\sigma}(P^{\phi}_{0,a})}  \cP V^{\pi,\sigma}_\phi\bigg] \nonumber \\
  & = 0 +  \gamma\pi(\phi \mymid 0) \inf_{ \cP \in \unb^{\sigma}(P^{\phi}_{0,\phi})}  \cP V^{\pi,\sigma}_\phi +  \gamma\pi(1 - \phi \mymid 0)  \inf_{ \cP \in \unb^{\sigma}(P^{\phi}_{0, 1- \phi})}  \cP V^{\pi,\sigma}_{\phi}. \label{eq:tv-s0-value-def}
\end{align}
Recall the transition kernel defined in \eqref{eq:Ph-construction-lower-infinite} and the fact about the uncertainty set over state $0$ in \eqref{eq:infinite-lw-p-q-perturb-inf}, it is easily verified that the following probability vector $P_1\in\Delta(\cS)$ obeys $P_1 \in \unb^{\ror}(P^{\phi}_{0, \phi})$, which is defined as
\begin{align}
    P_1(0) = 1-p + \ror \ind \left(0=s_{\phi, \min}^{\pi}  \right), \qquad P_1(1 ) = \underline{p} = p -\ror, \nonumber \\
    P_1(s ) = \ror \ind \left(s = s_{\phi, \min}^{\pi} \right), \qquad \forall s\in\{2,3,\cdots, S - 1\},
\end{align}
where $\underline{p} = p -\ror$ due to \eqref{eq:infinite-lw-p-q-perturb-inf}. 
Similarly, the following probability vector $P_2\in\Delta(\cS)$ also falls into the  uncertainty  set $\unb^{\ror}(P^{\phi}_{0, 1-\phi})$:
\begin{align}
    P_2(0 ) = 1-q + \ror \ind \left(0 = s_{\phi, \min}^{\pi}\right), \qquad P_2(1 ) = \underline{q} = q -\ror, \nonumber \\
    P_2(s) = \ror \ind \left(0 =s_{\phi, \min}^{\pi} \right) \qquad \forall s\in\{2,3,\cdots, S-1\}.
\end{align}
It is noticed that $P_0$ and $P_1$ defined above are the worst-case perturbations, since the probability mass at state $1$ will be moved to the state with the least value. 
Plugging the above facts about $P_1 \in \unb^{\ror}(P^{\phi}_{0, \phi})$ and $P_2 \in \unb^{\ror}(P^{\phi}_{0, 1-\phi})$ into \eqref{eq:tv-s0-value-def}, we arrive at
\begin{align}
    V_\phi^{\pi,\ror}(0)  
    &\leq \gamma \pi(\phi \mymid 0) P_1 V^{\pi,\sigma}_\phi + \gamma\pi(1 - \phi \mymid 0)  P_2 V^{\pi,\sigma}_{\phi} \notag \\
    & = \gamma \pi(\phi \mymid 0)\Big[ \left(p-\ror \right) V_\phi^{\pi,\sigma}(1) + \left(1- p\right) V_\phi^{\pi,\sigma}(0) + \ror  V_{\phi, \min}^{\pi, \sigma}\Big]  \notag \\
    &\qquad +  \gamma \pi(1-\phi \mymid 0)\Big[ \left(q-\ror \right) V_{\phi}^{\pi,\sigma}(1) + \left(1-q \right) V_{\phi}^{\pi,\sigma}(0) + \ror  V_{\phi, \min}^{\pi, \sigma}\Big] \notag\\
    & \overset{\mathrm{(i)}}{=} \gamma  \left(z_{\phi}^{\pi}- \ror \right) V_\phi^{\pi,\ror}(1) + \gamma \ror  V_{\phi, \min}^{\pi, \sigma} + \gamma (1-z_{\phi}^{\pi})V_\phi^{\pi,\ror}(0) \label{eq:tv-s-value2-pre},
\end{align}
where the last equality holds by the definition of $z_{\phi}^{\pi}$ in \eqref{eq:infinite-x-h}. To continue, recursively applying \eqref{eq:tv-s-value2-pre}  yields
\begin{align}
 V_\phi^{\pi,\ror}(0)   
 & \leq \gamma  \left(z_{\phi}^{\pi}- \ror \right) V_\phi^{\pi,\ror}(1) + \gamma \ror  V_{\phi, \min}^{\pi, \sigma} + \gamma (1-z_{\phi}^{\pi}) \Big[ \gamma  \left(z_{\phi}^{\pi}- \ror \right) V_\phi^{\pi,\ror}(1) + \gamma \ror  V_{\phi, \min}^{\pi, \sigma} + \gamma (1-z_{\phi}^{\pi})V_\phi^{\pi,\ror}(0)\Big] \nonumber \\
 & \overset{\mathrm{(i)}}{\leq} \gamma  \left(z_{\phi}^{\pi}- \ror \right) V_\phi^{\pi,\ror}(1) + \gamma \ror  V_{\phi, \min}^{\pi, \sigma} + \gamma (1-z_{\phi}^{\pi}) \Big[ \gamma  z_{\phi}^{\pi} V_\phi^{\pi,\ror}(1)  + \gamma (1-z_{\phi}^{\pi})V_\phi^{\pi,\ror}(0)\Big] \nonumber \\
 &\leq ... \nonumber \\
 & \leq \gamma  \left(z_{\phi}^{\pi}- \ror \right) V_\phi^{\pi,\ror}(1) + \gamma \ror  V_{\phi, \min}^{\pi, \sigma} +  \gamma z_{\phi}^{\pi} \sum_{t=1}^\infty \gamma^t(1-z_{\phi}^{\pi})^t V_\phi^{\pi,\ror}(1) + \lim_{t\rightarrow \infty} \gamma^t (1-z_{\phi}^{\pi})^t V_\phi^{\pi,\ror}(0)  \nonumber \\
 &\overset{\mathrm{(ii)}}{\leq} \gamma  \left(z_{\phi}^{\pi}- \ror \right) V_\phi^{\pi,\ror}(1) + \gamma \ror  V_{\phi, \min}^{\pi, \sigma} + \gamma(1-z_{\phi}^{\pi})\frac{\gamma z_{\phi}^{\pi}}{ 1- \gamma(1-z_{\phi}^{\pi})} V_\phi^{\pi,\ror}(1) + 0 \nonumber \\
 & < \gamma  \left(z_{\phi}^{\pi} - \ror \right) V_\phi^{\pi,\ror}(1) + \gamma \ror  V_{\phi, \min}^{\pi, \sigma} + \gamma(1- z_{\phi}^{\pi})V_\phi^{\pi,\ror}(1) \nonumber \\
 &= \gamma  \left(1- \ror \right) V_\phi^{\pi,\ror}(1) + \gamma \ror  V_{\phi, \min}^{\pi, \sigma}  \label{eq:tv-s-value2},
 \end{align}
 where (i) uses $ V_{\phi, \min}^{\pi, \sigma} \leq V_\phi^{\pi,\ror}(1)$, (ii) follows from $\gamma(1-z_{\phi}^{\pi})<1$, and the penultimate line follows from the trivial fact that $\frac{\gamma z_{\phi}^{\pi}}{ 1- \gamma(1-z_{\phi}^{\pi})} < 1$.

Combining \eqref{eq:tv-s-value1}, \eqref{eq:tv-s-other-value1},  and \eqref{eq:tv-s-value2}, we have that for any policy $\pi$, 
\begin{align}\label{eq:tv-value-order}
    V_{\phi}^{\pi, \sigma}(0) = V_{\phi, \min}^{\pi, \sigma},
\end{align}
which directly leads to
\begin{align}
    V_\phi^{\pi,\ror}(1) &= 1 + \gamma \left(1-\ror \right) V_\phi^{\pi,\ror}(1) + \gamma \ror  V_{\phi, \min}^{\pi, \sigma} = \frac{1 + \gamma \ror  V_{\phi}^{\pi, \sigma}(0) }{1-\gamma \left(1-\ror \right)}. \label{eq:tv-s2-value-pi}
\end{align}
Let's now return to the characterization of $V_{\phi}^{\pi, \sigma}(0)$. In view of \eqref{eq:tv-value-order}, the equality in \eqref{eq:tv-s-value2-pre} holds, and we have 
\begin{align}
   V_\phi^{\pi,\ror}(0) 
    & = \gamma \left( z_{\phi}^{\pi} - \ror \right) V_{\phi}^{\pi,\sigma}(1) + \gamma \left( 1- z_{\phi}^{\pi} + \ror \right)V_{\phi}^{\pi,\sigma}(0) \nonumber \\
    & \overset{\mathrm{(i)}}{=} \gamma \left( z_{\phi}^{\pi} - \ror \right) \frac{1 + \gamma \ror  V_{\phi}^{\pi, \sigma}(0) }{1-\gamma \left(1-\ror \right)} + \gamma \left( 1- z_{\phi}^{\pi} + \ror \right)V_{\phi}^{\pi,\sigma}(0)  \nonumber \\
    & = \frac{\gamma \left( z_{\phi}^{\pi} - \ror \right)}{1-\gamma \left(1-\ror \right)} + \gamma\left( 1 + \left( z_{\phi}^{\pi} - \ror \right)\frac{\gamma \ror - \left(1-\gamma \left(1-\ror \right)\right)}{1-\gamma \left(1-\ror \right)}\right) V_{\phi}^{\pi,\sigma}(0) \nonumber\\
    & = \frac{\gamma \left( z_{\phi}^{\pi} - \ror \right)}{1-\gamma \left(1-\ror \right)} + \gamma\left( 1 - \frac{(1-\gamma) \big( z_{\phi}^{\pi} - \ror \big)}{1-\gamma \left(1-\ror \right)}\right) V_{\phi}^{\pi,\sigma}(0),\nonumber
    \end{align}
    where  (i) arises from \eqref{eq:tv-s2-value-pi}.
Solving this relation gives 
\begin{align}
   V_\phi^{\pi,\ror}(0)   & = \frac{\frac{\gamma \left( z_{\phi}^{\pi} - \ror \right)}{1-\gamma \left(1-\ror \right)}}{ (1-\gamma) \bigg( 1 + \frac{\gamma \left( z_{\phi}^{\pi} - \ror \right)}{1-\gamma \left(1-\ror \right)} \bigg)} .\label{eq:tv-s0-value-final}
\end{align}

\paragraph{The optimal robust policy and optimal robust value function.}
We move on to characterize the robust optimal policy and its corresponding robust value function. To begin with, denoting
\begin{align}
z \defn  \frac{\gamma \big( z_{\phi}^{\pi} - \ror \big)}{1-\gamma \left(1-\ror \right)},
\end{align}
we rewrite \eqref{eq:tv-s0-value-final} as
\begin{align*}
     V_\phi^{\pi,\ror}(0)   = \frac{z}{(1-\gamma)(1+z)} =: f(z).
\end{align*}
Plugging in the fact that $z_{\phi}^{\pi} \geq q \geq \ror > 0$ in \eqref{eq:infinite-p-q-bound}, it follows that
$z   >0$.
So for any $z>0$, the derivative of $f(z)$ w.r.t. $z$ obeys  
\begin{align}
    \frac{(1-\gamma)(1+z) - (1-\gamma)z}{(1-\gamma)^2(1+z)^2} = \frac{1}{(1-\gamma)(1+z)^2} >0.
\end{align}
Observing that $f(z)$ is increasing in $z$, $z$ is increasing in $z_{\phi}^{\pi}$, and $z_{\phi}^{\pi}$ is also increasing in $\pi(\phi \mymid 0)$ (see the fact $p\geq q$ in \eqref{eq:infinite-p-q-bound}),  the optimal policy in state $0$ thus obeys
\begin{equation}
    \pi_\phi^{\star}(\phi \mymid 0) = 1 \label{eq:infinite-lb-optimal-policy}.
\end{equation}
Considering that the action does not influence the state transition for all states $s>0$, without loss of generality, we choose the robust optimal policy to obey
\begin{align}\label{eq:infinite-lower-optimal-pi}
    \forall s>0: \quad \pi_\phi^\star(\phi\mymid s) = 1.
\end{align}

Taking $\pi = \pi^\star_\phi$, we complete the proof by showing that the corresponding robust optimal robust value function at state $0$ as follows:
\begin{align}
    V_\phi^{\star,\ror}(0) = \frac{\frac{\gamma \left( z_{\phi}^{\pi^\star}  - \ror \right)}{1-\gamma \left(1-\ror \right)}}{ (1-\gamma) \left( 1 + \frac{\gamma \left( z_{\phi}^{\pi^\star}  - \ror \right)}{1-\gamma \left(1-\ror \right)} \right)} = \frac{\frac{\gamma \left( p  - \ror \right)}{1-\gamma \left(1-\ror \right)}}{ (1-\gamma) \left( 1 + \frac{\gamma \left( p - \ror \right)}{1-\gamma \left(1-\ror \right)} \right) }.
\end{align}

\subsubsection{Proof of the claim~\eqref{eq:tv-Value-0-recursive}}\label{proof:tv-lower-diff-control}
Plugging in the definition of $\varphi$, we arrive at that for any policy $\pi$,
\begin{align}
\big\langle \varphi, V^{\star, \sigma}_{\phi} - V^{\pi, \sigma}_{\phi} \big\rangle  = V^{\star, \sigma}_{\phi}(0) - V^{\pi, \sigma}_{\phi}(0) 
& = \frac{\frac{\gamma \left( p - z_{\phi}^{\pi} \right)}{1-\gamma \left(1-\ror \right)} }{(1-\gamma) \left( 1 + \frac{\gamma \left( p - \ror \right)}{1-\gamma \left(1-\ror \right)} \right) \left( 1 + \frac{\gamma \left( z_{\phi}^{\pi}  - \ror \right)}{1-\gamma \left(1-\ror \right)} \right)}, \label{eq:tv-value-gap}
\end{align}
which follows from applying \eqref{eq:infinite-lemma-value-0-pi} and basic calculus.
Then, we proceed to control the above term in two cases separately in terms of the uncertainty level $\sigma$.
\begin{itemize}
    \item When $\sigma\in (0, 1-\gamma]$. Then regarding the important terms in \eqref{eq:tv-value-gap}, we observe that
    \begin{align}\label{eq:tv-denominator1}
       1-\gamma < 1-\gamma \left(1-\ror \right) &\leq 1-\gamma \left(1- (1-\gamma) \right)  = (1-\gamma)(1+\gamma) \leq 2 (1-\gamma),
    \end{align}
    which directly leads to 
    \begin{align} 
     \frac{\gamma \big( z_{\phi}^{\pi}  - \ror \big)}{1-\gamma \left(1-\ror \right)} \overset{\mathrm{(i)}}{\leq} \frac{\gamma \left( p  - \ror \right)}{1-\gamma \left(1-\ror \right)} \leq \frac{\gamma  c_1(1-\gamma) }{1-\gamma \left(1-\ror \right)} \overset{\mathrm{(ii)}}{<} c_1 \gamma  ,\label{eq:tv-denominator2}
     \end{align}
     where (i) holds by $z_{\phi}^{\pi}<p$,  and (ii) is due to \eqref{eq:tv-denominator1}.
     Inserting \eqref{eq:tv-denominator1} and \eqref{eq:tv-denominator2}  back into \eqref{eq:tv-value-gap}, we arrive at
    \begin{align}
    \big\langle \varphi, V^{\star, \sigma}_{\phi} - V^{\pi, \sigma}_{\phi} \big\rangle  & \geq \frac{\frac{\gamma \left( p - z_{\phi}^{\pi} \right)}{2 (1-\gamma)}}{ (1-\gamma) (1+ c_1 \gamma)^2} \geq \frac{\gamma \big( p - z_{\phi}^{\pi} \big)}{ 8  (1-\gamma)^2 } \nonumber \\
    & = \frac{\gamma \left( p - q\right) \big(1-\pi(\phi\mymid 0)\big) }{ 8  (1-\gamma)^2 } = \frac{\gamma \Delta \big(1-\pi(\phi\mymid 0)\big) }{ 8  (1-\gamma)^2 } \geq 2\varepsilon \big(1-\pi(\phi\mymid 0)\big),
    \end{align}
    where the last inequality holds by setting ($\gamma \geq 1/2$) 
    \begin{align}
        \Delta = 32  (1-\gamma)^2\varepsilon.
    \end{align}
  Finally, it is easily verified that  
    \begin{align*}
        \varepsilon \leq \frac{c_1}{32(1-\gamma)} \quad \Longrightarrow  \quad
\Delta \leq c_1 (1-\gamma).
\end{align*}

    \item When $\sigma \in (1-\gamma, 1 - c_1 ]$. Regarding \eqref{eq:tv-value-gap}, we observe that
    \begin{align}\label{eq:tv-r2-denominator1}
       \gamma \ror < 1-\gamma \left(1-\ror \right) &=1-\gamma  + \gamma \ror \leq (1+\gamma)\sigma \leq 2\sigma,
    \end{align}
    which directly leads to 
    \begin{align} 
     \frac{\gamma \big( z_{\phi}^{\pi}  - \ror \big)}{1-\gamma \left(1-\ror \right)} \leq \frac{\gamma \left( p  - \ror \right)}{1-\gamma \left(1-\ror \right)} \leq \frac{\gamma c_1 \sigma }{1-\gamma \left(1-\ror \right)} \overset{\mathrm{(i)}}{<} c_1,  \label{eq:tv-r2-denominator2}
     \end{align}
     where (i) holds by \eqref{eq:tv-r2-denominator1}. Inserting \eqref{eq:tv-r2-denominator1} and \eqref{eq:tv-r2-denominator2}  back into \eqref{eq:tv-value-gap}, we arrive at
    \begin{align}
    \big\langle \varphi, V^{\star, \sigma}_{\phi} - V^{\pi, \sigma}_{\phi} \big\rangle  & \geq \frac{\frac{\gamma \left( p - z_{\phi}^{\pi} \right)}{2\sigma } } { (1-\gamma) (1+ c_1)^2} \geq \frac{\gamma \left( p - z_{\phi}^{\pi} \right)}{ 8  (1-\gamma) \sigma }   = \frac{\gamma \left( p - q\right) \big(1-\pi(\phi\mymid 0)\big) }{ 8   (1-\gamma) \sigma } \nonumber \\
    & = \frac{\gamma \Delta \big(1-\pi(\phi\mymid 0)\big) }{ 8   (1-\gamma) \sigma } \geq 2\varepsilon \big(1-\pi(\phi\mymid 0)\big),
    \end{align}
    where the last inequality holds by letting ($\gamma \geq 1/2$) 
    \begin{align}
        \Delta = 32  (1-\gamma) \sigma \varepsilon  .
    \end{align}
  Finally, it is easily verified that  
    \begin{align}
        \varepsilon \leq \frac{c_1}{32(1-\gamma)} \quad \Longrightarrow \quad \Delta \leq c_1 \ror.
    \end{align}

\end{itemize}

\subsubsection{Proof of Lemma~\ref{lemma:chi2-key0}}\label{proof:lemma:chi2-key0}

The proof follows the same routine as that of Lemma~\ref{lemma:tv-key0}. Taking the same pipeline as that in \ref{proof:lemma:tv-key0}, similar to \eqref{eq:variance-tight-bound-vstar-repeat}, we have \begin{align}
  \mathrm{Var}_{\Phatv^{\pi^\star, V}}(V^{\star,\ror})  
  & \leq \Phatv^{\pi^\star, V} \left(V' \circ V'\right) - \frac{1}{\gamma} V' \circ V' + \frac{2}{\gamma^2} \|V'\|_\infty 1 + \frac{2}{\gamma} \|V'\|_\infty \Big| \Big( \Pv^{\pi^\star, V}- \Phatv^{\pi^\star, V} \Big) V^{\star,\ror}\Big| \notag\\
  & \leq \Phatv^{\pi^\star, V} \left(V' \circ V'\right) - \frac{1}{\gamma} V' \circ V' + \frac{2}{\gamma^2} \|V'\|_\infty 1 + \frac{4}{\gamma} \|V'\|_\infty (1 + 2\sqrt{\ror}) \sqrt{\frac{\log(\frac{18SA N}{\delta})}{(1-\gamma)^2N}}  1, 
\end{align}
where the last inequality holds by Lemma~\ref{lemma:p-phat-V-gap-l2}.

Plugging the above results leads to
\begin{align}
 & \Big(I - \gamma \Phatv^{\pi^\star, V} \Big)^{-1} \sqrt{\mathrm{Var}_{\Phatv^{\pi^\star, V}}(V^{\star,\ror})}  \notag \\
 & \leq \sqrt{\frac{1}{1-\gamma}}\sqrt{ \sum_{t=0}^\infty \gamma^t \left(\Phatv^{\pi^\star, V} \right)^t  \Bigg( \Phatv^{\pi^\star, V} \left(V' \circ V'\right) - \frac{1}{\gamma} V' \circ V' + \frac{2}{\gamma^2} \|V'\|_\infty 1 + \frac{4}{\gamma} \|V'\|_\infty (1 + 2\sqrt{\ror}) \sqrt{\frac{\log(\frac{18SA N}{\delta})}{(1-\gamma)^2N}}  1 \Bigg) } \nonumber \\
 &  \overset{\mathrm{(i)}}{\leq} \sqrt{ \frac{1}{1-\gamma}} \sqrt{ \bigg| \sum_{t=0}^\infty \gamma^t \left(\Phatv^{\pi^\star, V} \right)^t \bigg( \Phatv^{\pi^\star, V} \left(V' \circ V'\right) - \frac{1}{\gamma} V' \circ V' \bigg) \bigg| } \nonumber \\
 &\qquad + \sqrt{\frac{1}{1-\gamma}}\sqrt{ \sum_{t=0}^\infty \gamma^t \left(\Phatv^{\pi^\star, V} \right)^t \Bigg( \frac{2}{\gamma^2} \|V'\|_\infty 1 + \frac{4}{\gamma} \|V'\|_\infty (1 + 2\sqrt{\ror}) \sqrt{\frac{\log(\frac{18SA N}{\delta})}{(1-\gamma)^2N}}  1 \Bigg) } \notag \\
 & \leq \sqrt{ \frac{1}{1-\gamma}} \sqrt{   \bigg| \sum_{t=0}^\infty \gamma^t \left(\Phatv^{\pi^\star, V} \right)^t  \bigg[\Phatv^{\pi^\star, V} \left(V' \circ V'\right) - \frac{1}{\gamma} V' \circ V' \bigg] \bigg|    } + \sqrt{\frac{\Big(2+ 4(1 + 2\sqrt{\ror}) \sqrt{\frac{ \log(\frac{18SAN}{\delta})}{(1-\gamma)^2N}} \Big) \|V'\|_\infty }{(1-\gamma)^2 \gamma^2}} 1 \notag \\
 & \overset{\mathrm{(ii)}}{\leq} \sqrt{\frac{\|V'\|_{\infty}^2}{\gamma(1-\gamma)}} 1 + \sqrt{\frac{\Big(2+ 4(1 + 2\sqrt{\ror}) \sqrt{\frac{ \log(\frac{18SAN}{\delta})}{(1-\gamma)^2N}} \Big) \|V'\|_\infty }{(1-\gamma)^2 \gamma^2}} 1 \notag \\
 &\leq 2\sqrt{\frac{\Big(2 + 4(1 + 2\sqrt{\ror}) \sqrt{\frac{ \log(\frac{18SAN}{\delta})}{(1-\gamma)^2N}} \Big)  }{(1-\gamma)^3 \gamma^2}} 1 
\end{align}
where (i) holds by the triangle inequality, (ii) follows from \eqref{eq:vstar-termI}, and the last inequality is obtained by the fact $\|V'\|_{\infty}\leq \frac{1}{1-\gamma}$.


\section{Proof of the upper bound with $\chi^2$ divergence: Theorem~\ref{thm:l2-upper-bound}}\label{proof:thm:chi2-upper-bound}

The proof of Theorem~\ref{thm:l2-upper-bound} mainly follows the structure of the proof of Theorem~\ref{thm:l1-upper-bound} in Appendix~\ref{proof:thm:l1-upper-bound}. Throughout this section, for any nominal transition kernel $P$, the uncertainty set is taken as (see \eqref{eq:chi-squared-distance})
\begin{align}\label{eq:consider-chi2}
\unb^{\ror}(P) = \cU^{\ror}_{\chi^2}(P) \defn \otimes \; \cU^{\ror}_{\chi^2}(P_{s,a}),\quad &\cU^{\ror}_{\chi^2}(P_{s,a}) \defn \Big\{ P'_{s,a} \in \Delta (\cS): \sum_{s'\in\cS} \frac{(P'(s' \mymid s,a) - P(s' \mymid s,a))^2}{P(s' \mymid s,a)} \leq \ror \Big\}.
\end{align} 

\subsection{Proof of Theorem~\ref{thm:l2-upper-bound}} 

In order to control the performance gap $\left\|V^{\star, \ror}- V^{\widehat{\pi}, \ror} \right\|_\infty$,
recall the error decomposition in \eqref{eq:l1-decompose}: as long as the iteration number $T \geq \log(\frac{1}{(1-\gamma) \varepsilon_{\mathsf{opt}}}) $,
\begin{align}
V^{\star, \ror} - V^{\widehat{\pi}, \ror} 
	\leq \left(V^{\pi^\star, \ror} - \widehat{V}^{\pi^\star, \ror}\right) + \frac{2\gamma \varepsilon_{\mathsf{opt}} }{1 -\gamma} {1} + \left(\widehat{V}^{\widehat{\pi}, \ror} - V^{\widehat{\pi}, \ror}\right), \label{eq:l2-decompose}
\end{align}
where $\varepsilon_{\mathsf{opt}}$ (cf.~\eqref{eq:opt-error}) shall be specified later (which justifies Remark~\ref{remark:chi2-upper}). To further control \eqref{eq:l2-decompose}, we  bound the remaining two terms separately.

\subsubsection{Controlling $\big\|\widehat{V}^{\pi^\star, \ror } - V^{ \pi^\star, \ror}\big\|_\infty$}
Towards this, recall the bound in \eqref{eq:tv-v-star-two-terms} which holds for any uncertainty set:
\begin{align} 
\big\|\widehat{V}^{\pi^\star, \ror } - V^{ \pi^\star, \ror}\big\|_\infty &\leq \gamma \max\Big\{ 
\Big\| \Big(I - \gamma \Phatv^{\pi^\star, \widehat{V}}\Big)^{-1} \Big( \Phatv^{\pi^\star, V} V^{\pi^\star, \ror } - \Pv^{\pi^\star, V } V^{\pi^\star, \ror } \Big) \Big\|_\infty, \nonumber\\
	&\qquad\qquad  \Big\|\Big(I - \gamma \Phatv^{\pi^\star, V} \Big)^{-1} \Big( \Phatv^{\pi^\star, V} V^{\pi^\star, \ror } - \Pv^{\pi^\star, V } V^{\pi^\star, \ror } \Big) \Big\|_\infty \Big\}. \label{eq:chi2-v-star-two-terms}
\end{align}

To control the main term $\Phatv^{\pi^\star, V} V^{\pi^\star, \ror } - \Pv^{\pi^\star, V } V^{\pi^\star, \ror }$ in \eqref{eq:chi2-v-star-two-terms}, we first introduce an important lemma whose proof is postponed to Appendix~\ref{proof:lemma:p-phat-V-gap-l2}.
 \begin{lemma}\label{lemma:p-phat-V-gap-l2}
Consider any $\ror>0$ and the uncertainty set $\unb^{\ror}(\cdot) \defn \cU^{\ror}_{\chi^2}(\cdot)$. For any $\delta \in (0,1)$, one has with probability at least $1-\delta$, 
\begin{align*} 
\left| \Phatv^{\pi^\star, V} V^{\pi^\star,\ror}  -  \Pv^{\pi^\star, V} V^{\pi^\star,\ror}  \right|_\infty &\leq  2\sqrt{\frac{\log(\frac{18SAN}{\delta})}{N}} \sqrt{\mathrm{Var}_{P^{\pi^\star}}(V^{\star,\ror})} +  \frac{\log(\frac{18SAN}{\delta})}{N(1-\gamma)} 1 + 4\sqrt{\frac{ \sigma\log(\frac{24SAN }{\delta})}{(1-\gamma)^2 N}} 1\notag \\
	&\leq  \sqrt{\frac{2\log(\frac{18SA N}{\delta})}{(1-\gamma)^2N}} 1 + 4\sqrt{\frac{ \sigma\log(\frac{24SAN }{\delta})}{(1-\gamma)^2 N}} 1.
\end{align*}
\end{lemma}

\paragraph{Step 1: controlling the first term in \eqref{eq:chi2-v-star-two-terms}.} 
Armed with the above lemma, now we control the first term on the right hand side of \eqref{eq:chi2-v-star-two-terms} as follows:
\begin{align}
&\Big(I - \gamma \Phatv^{\pi^\star, V} \Big)^{-1} \Big( \Phatv^{\pi^\star, V} V^{\pi^\star, \ror } - \Pv^{\pi^\star, V } V^{\pi^\star, \ror } \Big) \nonumber \\
& \overset{\mathrm{(i)}}{\leq} \Big(I - \gamma \Phatv^{\pi^\star, V} \Big)^{-1} \Big| \Phatv^{\pi^\star, V} V^{\pi^\star, \ror } - \Pv^{\pi^\star, V } V^{\pi^\star, \ror } \Big|_{\infty} \nonumber \\
& \overset{\mathrm{(ii)}}{\leq} \Big(I - \gamma \Phatv^{\pi^\star, V} \Big)^{-1} \bigg(2\sqrt{\frac{\log(\frac{18SAN}{\delta})}{N}} \sqrt{\mathrm{Var}_{P^{\pi^\star}}(V^{\star, \ror })} +  \frac{\log(\frac{18SAN}{\delta})}{N(1-\gamma)} 1 +  4\sqrt{\frac{ \sigma\log(\frac{24SAN }{\delta})}{(1-\gamma)^2 N}} 1\bigg) \notag \\
& = 2\Big(I - \gamma \Phatv^{\pi^\star, V} \Big)^{-1} \sqrt{\frac{\log(\frac{18SAN}{\delta})}{N}} \sqrt{\mathrm{Var}_{P^{\pi^\star}}(V^{\star, \ror })} + \bigg( \frac{\log(\frac{18SAN}{\delta})}{N(1-\gamma)}  +  4\sqrt{\frac{ \sigma\log(\frac{24SAN }{\delta})}{(1-\gamma)^2 N}} \bigg) \Big(I - \gamma \Phatv^{\pi^\star, V} \Big)^{-1} 1 \notag \\
&\leq \bigg( \frac{\log(\frac{18SAN}{\delta})}{N(1-\gamma)}  +  4\sqrt{\frac{ \sigma\log(\frac{24SAN }{\delta})}{(1-\gamma)^2 N}} \bigg) \Big(I - \gamma \Phatv^{\pi^\star, V} \Big)^{-1} 1 + \underbrace{2\sqrt{\frac{\log(\frac{18SAN}{\delta})}{N}} \Big(I - \gamma \Phatv^{\pi^\star, V} \Big)^{-1} \sqrt{\mathrm{Var}_{\Phatv^{\pi^\star, V} }(V^{\star, \ror })}}_{=: \chip_1}  \notag \\
& \quad + \underbrace{ 2\sqrt{\frac{\log(\frac{18SAN}{\delta})}{N}} \Big(I - \gamma \Phatv^{\pi^\star, V} \Big)^{-1} \sqrt{\left| \mathrm{Var}_{\widehat{P}^{\pi^\star}}(V^{\star, \ror }) - \mathrm{Var}_{\Phatv^{\pi^\star, V} }(V^{\star, \ror }) \right|} }_{=: \chip_2} \nonumber \\
& \quad + \underbrace{2\sqrt{\frac{\log(\frac{18SAN}{\delta})}{N}} \Big(I - \gamma \Phatv^{\pi^\star, V} \Big)^{-1}   \Big(\sqrt{\mathrm{Var}_{P^{\pi^\star}}(V^{\star, \ror })} - \sqrt{\mathrm{Var}_{\widehat{P}^{\pi^\star}}(V^{\star, \ror })} \Big)}_{=: \chip_3}, \label{eq:chi2-v-star-3-main}
\end{align}
where (i) holds by $\Big(I - \gamma \Phatv^{\pi^\star, V} \Big)^{-1} \geq 0$, (ii) follows from Lemma~\ref{lemma:p-phat-V-gap-l2}, and the last inequality can be obtained similarly as \eqref{eq:tv-v-star-3-main}.

We shall control the three terms $\chip_1, \chip_2, \chip_3$ in \eqref{eq:chi2-v-star-3-main} separately.

\begin{itemize}
	\item Consider $\chip_1$. We first introduce the following lemma, whose proof can be found in Appendix~\ref{proof:lemma:chi2-key0}.

	\begin{lemma}\label{lemma:chi2-key0}
Consider any $\delta \in (0,1)$. With probability at least  $1-\delta$, one has
\begin{align*}
&\Big(I - \gamma \Phatv^{\pi^\star, V} \Big)^{-1}\sqrt{\mathrm{Var}_{\Phatv^{\pi^\star, V} }(V^{\star, \ror })}\leq   2\sqrt{\frac{\Big(2 + 4(1 + 2\sqrt{\ror}) \sqrt{\frac{ \log(\frac{18SAN}{\delta})}{(1-\gamma)^2N}} \Big)  }{(1-\gamma)^3 \gamma^2}} 1.
\end{align*}
\end{lemma}
Applying Lemma~\ref{lemma:chi2-key0} leads to
\begin{align}
\chip_1 & =  2 \sqrt{\frac{\log(\frac{18SAN}{\delta})}{N}} \Big(I - \gamma \Phatv^{\pi^\star, V} \Big)^{-1} \sqrt{\mathrm{Var}_{\Phatv^{\pi^\star, V} }(V^{\star, \ror })} \nonumber \\
& \leq 4\sqrt{\frac{ \log(\frac{18SAN}{\delta})}{\gamma^2 (1-\gamma)^3 N} \bigg(2+ 4(1 + 2\sqrt{\ror})\sqrt{\frac{ \log(\frac{18SAN}{\delta})}{(1-\gamma)^2N}} \bigg)} 1. \label{eq:chi2-first-C1}
\end{align}

\item Consider $\chip_2$. For all $(s,a)\in \cS\times \cA$, $P_{s,a}\in \Delta(\cS)$, and $\widetilde{P}_{s,a} \in \cU^{\ror}(P_{s,a})$:
\begin{align}
  \big|\mathsf{Var}_{\widetilde{P}_{s,a}}(V^{\star,\ror}) - \mathsf{Var}_{P_{s,a}}(V^{\star,\ror}) \big|  \leq  \big\|\widetilde{P}_{s,a} - P_{s,a} \big\|_1 \big\|V^{\star,\ror}\big\|_{\infty}^2 \leq   \frac{\sqrt{\ror}}{(1-\gamma)^2}, \label{eq:chi2-vmax-sigma-big}
\end{align}
where the last inequality holds by the fact that $\big\|\widetilde{P}_{s,a} - P_{s,a} \big\|_1 \leq \sqrt{\rho_{\chi^2}(\widetilde{P}_{s,a},P_{s,a})}$, and $\big\|V'\big\|_{\infty} \leq \frac{1}{1-\gamma}$. Applying the above relation and following the same routine in \eqref{eq:tv-first-C2} give
\begin{align}
	\chip_2 &\leq 2\sqrt{\frac{\log(\frac{18SAN}{\delta})}{N}} \Big(I - \gamma \Phatv^{\pi^\star, V} \Big)^{-1}  \sqrt{\left\| \mathrm{Var}_{\widehat{P}^\no }(V^{\star, \ror }) - \mathrm{Var}_{\widehat{P}^{\pi^\star, V} }(V^{\star, \ror }) \right\|_\infty 1 }   \notag \\
	& \leq 2\sqrt{\frac{\log(\frac{18SAN}{\delta})}{N}} \Big(I - \gamma \Phatv^{\pi^\star, V} \Big)^{-1} \sqrt{\frac{\sqrt{\ror}}{(1-\gamma)^2}} 1 = 2\sqrt{\frac{\sqrt{\ror}\log(\frac{18SAN}{\delta})}{(1-\gamma)^4N}} 1, \label{eq:chi2-first-C2}
\end{align}
where the last equality uses $  \Big(I - \gamma \Phatv^{\pi^\star, V} \Big)^{-1}  1  = \frac{1}{1-\gamma}$ (cf.~\eqref{eq:tv-first-C0}).

\item Consider $\chip_3$. Applying Lemma~\ref{lemma:tv-key2} with $\pi = \pi^\star$ and $V = V^{\star,\ror}$ leads to
\begin{align*}
	\sqrt{\mathrm{Var}_{P^{\pi^\star}}(V^{\star, \ror })} - \sqrt{\mathrm{Var}_{\widehat{P}^{\pi^\star}}(V^{\star, \ror })} \leq \sqrt{\frac{2\|V^{\star,\ror}\|_{\infty}^2\log(\frac{2SA}{\delta})}{N}} 1 ,
\end{align*}
which can be plugged in to verify similar to \eqref{eq:tv-first-C3} as
\begin{align}
\chip_3  \leq \frac{4\log(\frac{18SAN}{\delta}) }{(1-\gamma)^2N} 1. \label{eq:chi2-first-C3}
\end{align}

\end{itemize}

Finally, inserting the results in \eqref{eq:chi2-first-C1}, \eqref{eq:chi2-first-C2}, and \eqref{eq:chi2-first-C3} back into \eqref{eq:chi2-v-star-3-main} gives  
\begin{align}
&\Big(I - \gamma \Phatv^{\pi^\star, V} \Big)^{-1} \Big( \Phatv^{\pi^\star, V} V^{\pi^\star, \ror } - \Pv^{\pi^\star, V } V^{\pi^\star, \ror } \Big) \leq \bigg( \frac{\log(\frac{18SAN}{\delta})}{N(1-\gamma)^2}  +  4\sqrt{\frac{ \sigma\log(\frac{24SAN }{\delta})}{(1-\gamma)^4 N}} \bigg)  1  \notag \\
&\quad + 4\sqrt{\frac{ \log(\frac{18SAN}{\delta})}{\gamma^2 (1-\gamma)^3 N} \bigg(2+ 4(1 + 2\sqrt{\ror})\sqrt{\frac{ \log(\frac{18SAN}{\delta})}{(1-\gamma)^2N}} \bigg)} 1 +  \frac{4\log(\frac{18SAN}{\delta}) }{(1-\gamma)^2N} 1 + 2\sqrt{\frac{\sqrt{\ror}\log(\frac{18SAN}{\delta})}{(1-\gamma)^4N}} 1 \notag \\
& \leq \frac{5\log(\frac{18SAN}{\delta}) }{(1-\gamma)^2N} 1 + 4\sqrt{\frac{ \sigma\log(\frac{24SAN }{\delta})}{(1-\gamma)^4 N}} 1 + 2\sqrt{\frac{\sqrt{\ror}\log(\frac{18SAN}{\delta})}{(1-\gamma)^4N}} 1 \notag \\
&\quad +  4\sqrt{\frac{ \log(\frac{18SAN}{\delta})}{\gamma^2 (1-\gamma)^3 N} \bigg(2+ 4(1 + 2\sqrt{\ror}) \bigg)} 1\notag \\
&  \leq  48\sqrt{\frac{ \log(\frac{24SAN }{\delta})}{(1-\gamma)^3 N}} \left(1+\frac{\sigma^{1/2} + \sigma^{1/4}}{\sqrt{1-\gamma}}\right) 1 + \frac{5\log(\frac{18SAN}{\delta}) }{(1-\gamma)^2N} 1,  \label{eq:chi2-v-star-first-finish}
\end{align}
where the last inequality holds by the fact $\gamma \geq \frac{1}{4}$ and letting $N \geq \frac{\log(\frac{SAN}{\delta})}{(1-\gamma)^2}$.

\paragraph{Step 2: bounding the second term in \eqref{eq:chi2-v-star-two-terms}.}
Applying Lemma~\ref{lemma:p-phat-V-gap-l2} to the second term on the right hand side of \eqref{eq:chi2-v-star-two-terms} leads to
\begin{align}
&\Big(I - \gamma \Phatv^{\pi^\star, \widehat{V}} \Big)^{-1} \Big( \Phatv^{\pi^\star, V} V^{\pi^\star, \ror } - \Pv^{\pi^\star, V } V^{\pi^\star, \ror } \Big) \nonumber \\
& \leq \Big(I - \gamma \Phatv^{\pi^\star, \widehat{V}} \Big)^{-1}   \bigg(2\sqrt{\frac{\log(\frac{18SAN}{\delta})}{N}} \sqrt{\mathrm{Var}_{P^{\pi^\star}}(V^{\star, \ror })} +  \frac{\log(\frac{18SAN}{\delta})}{N(1-\gamma)} 1 +  4\sqrt{\frac{ \sigma\log(\frac{24SAN }{\delta})}{(1-\gamma)^2 N}} 1\bigg)   \nonumber \\
& \leq  \bigg( \frac{\log(\frac{18SAN}{\delta})}{N(1-\gamma)}  +  4\sqrt{\frac{ \sigma\log(\frac{24SAN }{\delta})}{(1-\gamma)^2 N}} \bigg)  \Big(I - \gamma \Phatv^{\pi^\star, \widehat{V}}  \Big)^{-1} 1  \notag \\
&\quad + \underbrace{ 2\sqrt{\frac{\log(\frac{18SAN}{\delta})}{N}} \Big(I - \gamma \Phatv^{\pi^\star, \widehat{V}} \Big)^{-1} \sqrt{\mathrm{Var}_{\Phatv^{\pi^\star, \widehat{V}}  }(\widehat{V}^{\pi^\star, \ror })}}_{=: \chip_4}  \nonumber \\
& \quad + \underbrace{ 2\sqrt{\frac{\log(\frac{18SAN}{\delta})}{N}} \Big(I - \gamma \Phatv^{\pi^\star, \widehat{V}}  \Big)^{-1} \left(\sqrt{ \mathrm{Var}_{\Phatv^{\pi^\star, \widehat{V}}  }(V^{\pi^\star, \ror }- \widehat{V}^{\pi^\star, \ror }) } \right)}_{=: \chip_5} \nonumber \\
& \quad + \underbrace{ 2\sqrt{\frac{\log(\frac{18SAN}{\delta})}{N}} \Big(I - \gamma \Phatv^{\pi^\star, \widehat{V}}  \Big)^{-1}\left(\sqrt{\left| \mathrm{Var}_{\widehat{P}^{\pi^\star}}(V^{\star, \ror }) - \mathrm{Var}_{\Phatv^{\pi^\star, \widehat{V}} }(V^{\star, \ror }) \right|} \right)}_{ =: \chip_6} \nonumber \\
& \quad + \underbrace{ 2\sqrt{\frac{\log(\frac{18SAN}{\delta})}{N}} \Big(I - \gamma \Phatv^{\pi^\star, \widehat{V}}  \Big)^{-1} \left(\sqrt{\mathrm{Var}_{P^{\pi^\star}}(V^{\star, \ror })} - \sqrt{\mathrm{Var}_{\widehat{P}^{\pi^\star}}(V^{\star, \ror })} \right)}_{=: \chip_7}. \label{eq:chi2-v-star-second}
\end{align}

We now control the above terms separately.
\begin{itemize}

\item Applying Lemma~\ref{lemma:tv-key1} and in view of \eqref{eq:tv-first-C0},
 the term $\cF_4$ in \eqref{eq:chi2-v-star-second} can be controlled similarly to \eqref{eq:tv-first-C4} as follows:
\begin{align}
\chip_4 &= 2\sqrt{\frac{\log(\frac{18SAN}{\delta})}{N}} \Big(I - \gamma \Phatv^{\pi^\star, \widehat{V}} \Big)^{-1} \sqrt{\mathrm{Var}_{\Phatv^{\pi^\star, \widehat{V}}  }(\widehat{V}^{\pi^\star, \ror })} \leq  8\sqrt{\frac{\log(\frac{18SAN}{\delta})   }{\gamma^2(1-\gamma)^3  N }} 1.\label{eq:chi2-first-C4}
\end{align}

\item In view of \eqref{eq:tv-first-C0}, we have 
\begin{align}
\chip_5 &= 2\sqrt{\frac{\log(\frac{18SAN}{\delta})}{N}} \Big(I - \gamma \Phatv^{\pi^\star, \widehat{V}}  \Big)^{-1}  \sqrt{ \mathrm{Var}_{\Phatv^{\pi^\star, \widehat{V}}  }(V^{\pi^\star, \ror } - \widehat{V}^{\pi^\star, \ror }) }   \notag \\
& \leq 2\sqrt{\frac{\log(\frac{18SAN}{\delta})}{(1-\gamma)^2N}} \left\|V^{\star,\ror} - \widehat{V}^{\pi^\star, \ror } \right\|_\infty 1. \label{eq:chi2-first-C5}
\end{align}

\item Then, it is easily verified that $\cF_6$ can be controlled similarly to \eqref{eq:chi2-first-C2} as follows:
\begin{align}
\chip_6 & \leq 2\sqrt{\frac{\sqrt{\ror}\log(\frac{18SAN}{\delta})}{(1-\gamma)^4N}} 1. \label{eq:chi2-first-C6}
\end{align}

\item Similarly, $\chip_7$ can be controlled the same as \eqref{eq:chi2-first-C3} shown below:
\begin{align}
\chip_7  & \leq \frac{4\log(\frac{18SAN}{\delta}) }{(1-\gamma)^2N} 1. \label{eq:chi2-first-C7}
\end{align}
\end{itemize}

Plugging in the results in \eqref{eq:chi2-first-C4}, \eqref{eq:chi2-first-C5}, \eqref{eq:chi2-first-C6}, and \eqref{eq:chi2-first-C7}  to \eqref{eq:chi2-v-star-second} gives
\begin{align}
&\Big(I - \gamma \Phatv^{\pi^\star, \widehat{V}} \Big)^{-1} \Big( \Phatv^{\pi^\star, V} V^{\pi^\star, \ror } - \Pv^{\pi^\star, V } V^{\pi^\star, \ror } \Big) \nonumber \\
&  \leq \bigg( \frac{\log(\frac{18SAN}{\delta})}{N(1-\gamma)^2}  +  4\sqrt{\frac{ \sigma\log(\frac{24SAN }{\delta})}{(1-\gamma)^4 N}} \bigg)  1 + 8\sqrt{\frac{\log(\frac{18SAN}{\delta})   }{\gamma^2(1-\gamma)^3 N }} 1  \notag \\
& \qquad \qquad+ 2\sqrt{\frac{\log(\frac{18SAN}{\delta})}{(1-\gamma)^2N}} \left\|V^{\star,\ror} - \widehat{V}^{\pi^\star, \ror } \right\|_\infty 1 + 2\sqrt{\frac{\sqrt{\ror}\log(\frac{18SAN}{\delta})}{(1-\gamma)^4N}} 1  + \frac{4\log(\frac{18SAN}{\delta}) }{(1-\gamma)^2N} 1 \notag \\
& \leq  2\sqrt{\frac{\log(\frac{18SAN}{\delta})}{(1-\gamma)^2N}} \left\|V^{\star,\ror} - \widehat{V}^{\pi^\star, \ror } \right\|_\infty 1 + \frac{5\log(\frac{18SAN}{\delta}) }{(1-\gamma)^2N} 1 + 32\sqrt{\frac{\log(\frac{24SAN}{\delta})   }{(1-\gamma)^3 N }} \left(1+\frac{\sigma^{1/2} + \sigma^{1/4}}{\sqrt{1-\gamma}}\right) 1, \label{eq:chi2-v-star-second-finish}
\end{align}
where the last inequality follows from the assumption $\gamma \geq \frac{1}{4}$.

Finally, inserting \eqref{eq:chi2-v-star-first-finish} and \eqref{eq:chi2-v-star-second-finish} back to \eqref{eq:chi2-v-star-two-terms} yields
\begin{align}
&\left\|\widehat{V}^{\pi^\star, \ror } - V^{ \pi^\star, \ror}\right\|_\infty \notag \\
 &\leq \max\Bigg\{ 48\sqrt{\frac{ \log(\frac{24SAN }{\delta})}{(1-\gamma)^3 N}} \left(1+\frac{\sigma^{1/2} + \sigma^{1/4}}{\sqrt{1-\gamma}}\right) + \frac{5\log(\frac{18SAN}{\delta}) }{(1-\gamma)^2N} , \notag \\
& \qquad\quad 2\sqrt{\frac{\log(\frac{18SAN}{\delta})}{(1-\gamma)^2N}} \left\|V^{\star,\ror} - \widehat{V}^{\pi^\star, \ror } \right\|_\infty  + \frac{5\log(\frac{18SAN}{\delta}) }{(1-\gamma)^2N}  + 32\sqrt{\frac{\log(\frac{24SAN}{\delta})   }{(1-\gamma)^3 N }} \left(1+\frac{\sigma^{1/2} + \sigma^{1/4}}{\sqrt{1-\gamma}}\right)    \Bigg\} \notag \\
& \leq 96\sqrt{\frac{ \log(\frac{24SAN}{\delta})}{ (1-\gamma)^3  N} }\left(1+\frac{\sigma^{1/2} + \sigma^{1/4}}{\sqrt{1-\gamma}}\right) + \frac{10\log(\frac{18SAN}{\delta}) }{(1-\gamma)^2N}, \label{eq:chi2-pi-star-upper-final}
\end{align}
where the last inequality holds by taking $N \geq \frac{16\log(\frac{SAN}{\delta})}{(1-\gamma)^2}$ and rearranging terms.

\subsubsection{Controlling $\|\widehat{V}^{\widehat{\pi}, \ror } - V^{ \widehat{\pi}, \ror}\|_\infty$}

Recall the bound in \eqref{eq:tv-v-pihat-two-terms} which holds for any uncertainty set:
\begin{align} 
\big\|\widehat{V}^{\widehat{\pi}, \ror } - V^{\widehat{\pi}, \ror}\big\|_\infty &\leq \gamma \max\Big\{ 
	\Big\| \Big(I - \gamma \Pv^{\widehat{\pi}, V} \Big)^{-1} \Big( \Phatv^{\widehat{\pi}, \widehat{V}} \widehat{V}^{\widehat{\pi}, \ror } - \Pv^{\widehat{\pi}, \widehat{V} } \widehat{V}^{\widehat{\pi}, \ror } \Big) \Big\|_\infty, \nonumber\\
	&\qquad\qquad \Big\| \Big(I - \gamma \Pv^{\widehat{\pi}, \widehat{V}} \Big)^{-1} \Big( \Phatv^{\widehat{\pi}, \widehat{V}} \widehat{V}^{\widehat{\pi}, \ror } - \Pv^{\widehat{\pi}, \widehat{V} } \widehat{V}^{\widehat{\pi}, \ror } \Big) \Big\|_\infty \Big\}. \label{eq:chi2-v-pihat-two-terms}
\end{align}

To begin with, we introduce the following lemma which controls the main term on the right hand side of \eqref{eq:chi2-v-pihat-two-terms}, which is proved in Appendix~\ref{proof:lemma:dro-b-bound-infinite-loo-chi2}.

\begin{lemma}\label{lemma:dro-b-bound-infinite-loo-chi2}
Consider any $\delta \in (0,1)$. Taking $N \geq \log \left(\frac{54SAN^2}{(1-\gamma)\delta} \right)$, with probability at least $1- \delta$, one has 
\begin{align}\label{eq:dro-b-bound-infinite}
	&\Big| \Phatv^{\widehat{\pi}, \widehat{V}} \widehat{V}^{\widehat{\pi}, \ror } - \Pv^{\widehat{\pi}, \widehat{V} } \widehat{V}^{\widehat{\pi}, \ror }  \Big| \notag \\
	& \leq 2\sqrt{\frac{\log(\frac{54SAN^2} {(1-\gamma)\delta})}{N}} \sqrt{\mathrm{Var}_{P^{\widehat{\pi}}}(\widehat{V}^{\star,\ror})} 1 +  \frac{8\log(\frac{54SAN^2 }{(1-\gamma)\delta})}{N(1-\gamma)} 1+  6\sqrt{\frac{2 \sigma\log(\frac{36SAN^2 }{\delta})}{(1-\gamma)^2 N}} 1+ \frac{2\gamma \varepsilon_{\mathsf{opt}} + 4\sqrt{\sigma \varepsilon_{\mathsf{opt}} } }{1 -\gamma} 1\notag\\
& \leq 10\sqrt{\frac{\log(\frac{54SAN^2} {(1-\gamma)\delta})}{(1 -\gamma)^2 N}} 1 + 6\sqrt{\frac{2 \sigma\log(\frac{36SAN^2 }{\delta})}{(1-\gamma)^2 N}} 1 + \frac{2\gamma \varepsilon_{\mathsf{opt}} + 4\sqrt{\sigma \varepsilon_{\mathsf{opt}} } }{1 -\gamma} 1.
\end{align}
\end{lemma}
 
\paragraph{Step 3: controlling the first term in \eqref{eq:chi2-v-pihat-two-terms}.}  
Applying Lemma~\ref{lemma:dro-b-bound-infinite-loo-chi2} leads to
\begin{align}
	& \Big(I - \gamma \Pv^{\widehat{\pi}, \widehat{V}} \Big)^{-1} \Big( \Phatv^{\widehat{\pi}, \widehat{V}} \widehat{V}^{\widehat{\pi}, \ror } - \Pv^{\widehat{\pi}, \widehat{V} } \widehat{V}^{\widehat{\pi}, \ror } \Big) \nonumber \\
	&  \overset{\mathrm{(i)}}{\leq} \left(I - \gamma \Pv^{\widehat{\pi}, \widehat{V}} \right)^{-1} \left| \Phatv^{\widehat{\pi}, \widehat{V}} \widehat{V}^{\widehat{\pi}, \ror } - \Pv^{\widehat{\pi}, \widehat{V} } \widehat{V}^{\widehat{\pi}, \ror } \right| \nonumber \\
	&\leq  2\sqrt{\frac{\log(\frac{54SAN^2} {(1-\gamma)\delta})}{N}} \left(I - \gamma \Pv^{\widehat{\pi}, \widehat{V}} \right)^{-1}  \sqrt{\mathrm{Var}_{P^{\widehat{\pi} }}(\widehat{V}^{\star, \ror })} \notag \\
	& \quad  + \left(I - \gamma \pmin_Q^{\widehat{\pi}, V^{\widehat{\pi}}}\right)^{-1} \Bigg(\frac{8\log(\frac{54SAN^2 }{(1-\gamma)\delta})}{N(1-\gamma)} +  6\sqrt{\frac{2 \sigma\log(\frac{36SAN^2 }{\delta})}{(1-\gamma)^2 N}} + \frac{2\gamma \varepsilon_{\mathsf{opt}} + 4\sqrt{\sigma \varepsilon_{\mathsf{opt}} } }{1 -\gamma} \Bigg) 1 \notag \\
	& \overset{\mathrm{(ii)}}{\leq} \Bigg( \frac{8\log(\frac{54SAN^2 }{(1-\gamma)^2\delta})}{N(1-\gamma)} +  6\sqrt{\frac{2 \sigma\log(\frac{36SAN^2 }{\delta})}{(1-\gamma)^4 N}} + \frac{2\gamma \varepsilon_{\mathsf{opt}} + 4\sqrt{\sigma \varepsilon_{\mathsf{opt}} } }{(1 -\gamma)^2} \Bigg) 1 \notag \\
	&\quad +  \underbrace{ 2\sqrt{\frac{\log(\frac{54SAN^2} {(1-\gamma)\delta})}{N}} \left(I - \gamma \Pv^{\widehat{\pi}, \widehat{V}} \right)^{-1}  \sqrt{\mathrm{Var}_{\Pv^{\widehat{\pi}, \widehat{V}}}(\widehat{V}^{\widehat{\pi}, \ror })} }_{=: \cG_1 } \notag \\
	& \quad + \underbrace{ 2\sqrt{\frac{\log(\frac{54SAN^2} {(1-\gamma)\delta})}{N}} \left(I - \gamma \Pv^{\widehat{\pi}, \widehat{V}} \right)^{-1}  \sqrt{ \left|\mathrm{Var}_{\Pv^{\widehat{\pi}, \widehat{V}}}(\widehat{V}^{\star, \ror }) - \mathrm{Var}_{\Pv^{\widehat{\pi}, \widehat{V}}}(\widehat{V}^{\widehat{\pi}, \ror }) \right| } }_{=:  \cG_2 } \notag \\
	& \quad + \underbrace{ 2\sqrt{\frac{\log(\frac{54SAN^2} {(1-\gamma)\delta})}{N}} \left(I - \gamma \Pv^{\widehat{\pi}, \widehat{V}} \right)^{-1}  \sqrt{ \left|\mathrm{Var}_{P^{\widehat{\pi}}}(\widehat{V}^{\star, \ror }) - \mathrm{Var}_{\Pv^{\widehat{\pi}, \widehat{V}}}(\widehat{V}^{\star, \ror }) \right| } }_{=: \cG_3 }, \label{eq:tv-v-hat-3-main}
\end{align}
where (i) and (ii) hold by the fact that each row of $(1-\gamma)\left(I - \gamma \Pv^{\widehat{\pi}, \widehat{V}} \right)^{-1}$ is a probability vector that falls into $\Delta(\cS)$.

Therefore, the remainder of the proof will focus on controlling $\cG_1, \cG_2, \cG_3$ separately.

\begin{itemize}
\item For $\cG_1$, we introduce the following lemma, whose proof is postponed to \ref{proof:lemma:chi2-vhat-key1}.
\begin{lemma}\label{lemma:chi2-vhat-key1}
Consider any $\delta \in (0,1)$. Taking $ N \geq\frac{\log(\frac{54SAN^2}{(1-\gamma)\delta})}{(1-\gamma)^2}$ and $\varepsilon_{\mathsf{opt}}\leq \frac{1-\gamma}{\gamma}$, one has with probability at least $1-\delta$,
\begin{align*}
\Big(I - \gamma \Pv^{\widehat{\pi}, \widehat{V}} \Big)^{-1}  \sqrt{\mathrm{Var}_{\Pv^{\widehat{\pi}, \widehat{V}}}(\widehat{V}^{\widehat{\pi}, \ror })} &\leq 7\sqrt{\frac{1 }{(1-\gamma)^3 \gamma^2}}  + 6\sqrt{\frac{\sqrt{\sigma} }{(1-\gamma)^3 \gamma^2}} 1.
\end{align*}
\end{lemma}

Applying Lemma~\ref{lemma:chi2-vhat-key1} and \eqref{eq:tv-first-C0} to \eqref{eq:tv-v-hat-3-main} leads to
\begin{align}
\cG_1 &= 2\sqrt{\frac{\log(\frac{54SAN^2}{(1-\gamma)\delta})}{N}} \Big(I - \gamma \Pv^{\widehat{\pi}, \widehat{V}} \Big)^{-1}  \sqrt{\mathrm{Var}_{\Pv^{\widehat{\pi}, \widehat{V}}}(\widehat{V}^{\widehat{\pi}, \ror })} \nonumber \\
& \leq 14 \sqrt{\frac{\log(\frac{54SAN^2}{(1-\gamma)\delta})}{\gamma^2 (1-\gamma)^3 N}}  + 12 \sqrt{\frac{ \sqrt{\sigma}\log(\frac{54SAN^2}{(1-\gamma)\delta})}{\gamma^2 (1-\gamma)^3 N}}  1. \label{eq:chi2-first-D1}
\end{align}

\item Applying Lemma~\ref{eq:tv-auxiliary-lemma} with $\|\widehat{V}^{\star, \ror } - \widehat{V}^{\widehat{\pi}, \ror } \|_\infty \leq \frac{2\gamma \varepsilon_{\mathsf{opt}} }{1 -\gamma}$ and \eqref{eq:tv-first-C0}, $\cG_2$ can be controlled as 
\begin{align}
	\cG_2 &= 2\sqrt{\frac{\log(\frac{54SAN^2}{(1-\gamma)\delta})}{N}} \Big(I - \gamma \Pv^{\widehat{\pi}, \widehat{V}} \Big)^{-1}  \sqrt{ \left|\mathrm{Var}_{\Pv^{\widehat{\pi}, \widehat{V}}}(\widehat{V}^{\star, \ror }) - \mathrm{Var}_{\Pv^{\widehat{\pi}, \widehat{V}}}(\widehat{V}^{\widehat{\pi}, \ror }) \right| } \notag \\
	&\leq 4\sqrt{\frac{\log(\frac{54SAN^2}{(1-\gamma)\delta})}{N}} \Big(I - \gamma \Pv^{\widehat{\pi}, \widehat{V}} \Big)^{-1} \frac{\sqrt{\gamma \varepsilon_{\mathsf{opt}}} }{1-\gamma} \leq 4\sqrt{\frac{\gamma \varepsilon_{\mathsf{opt}} \log(\frac{54SAN^2}{(1-\gamma)\delta})}{(1-\gamma)^4N}} 1. \label{eq:chi2-first-D2}
\end{align}

\item $\cG_3$ can be controlled similar to $\cF_2$ in \eqref{eq:tv-first-C2} as follows:
\begin{align}
	\cG_3 & = 2\sqrt{\frac{\log(\frac{54SAN^2}{(1-\gamma)\delta})}{N}} \left(I - \gamma \Pv^{\widehat{\pi}, \widehat{V}} \right)^{-1}  \sqrt{ \left|\mathrm{Var}_{P^{\widehat{\pi}}}(\widehat{V}^{\star, \ror }) - \mathrm{Var}_{\Pv^{\widehat{\pi}, \widehat{V}}}(\widehat{V}^{\star, \ror }) \right| } \notag \\
	& \leq 2\sqrt{\frac{\log(\frac{54SAN^2}{\delta})}{N}} \Big(I - \gamma \Phatv^{\pi^\star, V} \Big)^{-1} \sqrt{\frac{\sqrt{\ror}}{(1-\gamma)^2}} 1 = 2\sqrt{\frac{\sqrt{\ror}\log(\frac{54SAN^2}{\delta})}{(1-\gamma)^4N}} 1. \label{eq:chi2-first-D3}
\end{align}
\end{itemize}
To proceed, summing up the results in \eqref{eq:chi2-first-D1},  \eqref{eq:chi2-first-D2}, and \eqref{eq:chi2-first-D3} and inserting them back to \eqref{eq:tv-v-hat-3-main} yields: taking $ N \geq\frac{\log(\frac{54SAN^2}{(1-\gamma)\delta})}{(1-\gamma)^2}$ and $\varepsilon_{\mathsf{opt}}\leq \frac{(1-\gamma)^2}{\gamma}$, with probability at least $1-\delta$,
\begin{align}
&\left(I - \gamma \Pv^{\widehat{\pi}, \widehat{V}} \right)^{-1} \left( \Phatv^{\widehat{\pi}, \widehat{V}} \widehat{V}^{\widehat{\pi}, \ror } - \Pv^{\widehat{\pi}, \widehat{V} } \widehat{V}^{\widehat{\pi}, \ror } \right) \nonumber \\
&  \leq  \Bigg( \frac{8\log(\frac{54SAN^2 }{(1-\gamma)^2\delta})}{N(1-\gamma)} +  6\sqrt{\frac{2 \sigma\log(\frac{36SAN^2 }{\delta})}{(1-\gamma)^4 N}} + \frac{2\gamma \varepsilon_{\mathsf{opt}} + 4\sqrt{\sigma \varepsilon_{\mathsf{opt}} } }{(1 -\gamma)^2} \Bigg) 1  \notag \\
& \quad + 14 \sqrt{\frac{\log(\frac{54SAN^2}{(1-\gamma)\delta})}{\gamma^2 (1-\gamma)^3 N}}  + 12 \sqrt{\frac{ \sqrt{\sigma}\log(\frac{54SAN^2}{(1-\gamma)\delta})}{\gamma^2 (1-\gamma)^3 N}}  1 +  4\sqrt{\frac{\gamma \varepsilon_{\mathsf{opt}} \log(\frac{54SAN^2}{(1-\gamma)\delta})}{(1-\gamma)^4N}} 1  + 2\sqrt{\frac{\sqrt{\ror}\log(\frac{54SAN^2}{\delta})}{(1-\gamma)^4N}} 1 \notag \\
&\leq 18 \left(1 + \frac{\sigma^{1/2} + \sigma^{1/4}}{\sqrt{1-\gamma}} \right) \sqrt{\frac{\log(\frac{54SAN^2}{(1-\gamma)\delta})}{\gamma^2 (1-\gamma)^3 N}} 1 + \frac{10\log(\frac{54SAN^2 }{(1-\gamma)\delta})}{N(1-\gamma)^2}  1, \label{eq:chi2-v-hat-first-finish}
\end{align}
where the last inequality holds by taking $\varepsilon_{\mathsf{opt}} \leq \min \left\{\frac{1-\gamma}{\gamma^3}, \frac{\log(\frac{54SAN^2}{(1-\gamma)\delta})}{ N} \right\} = \frac{\log(\frac{54SAN^2}{(1-\gamma)\delta})}{ N}$.

\paragraph{Step 4: bounding the second term in \eqref{eq:chi2-v-pihat-two-terms}.}
Towards this, applying Lemma~\ref{lemma:dro-b-bound-infinite-loo-chi2} leads to
\begin{align}
	&\Big(I - \gamma \Pv^{\widehat{\pi}, V} \Big)^{-1} \Big( \Phatv^{\widehat{\pi}, \widehat{V}} \widehat{V}^{\widehat{\pi}, \ror } - \Pv^{\widehat{\pi}, \widehat{V} } \widehat{V}^{\widehat{\pi}, \ror } \Big)  \leq \Big(I - \gamma \Pv^{\widehat{\pi}, V} \Big)^{-1} \Big| \Phatv^{\widehat{\pi}, \widehat{V}} \widehat{V}^{\widehat{\pi}, \ror } - \Pv^{\widehat{\pi}, \widehat{V} } \widehat{V}^{\widehat{\pi}, \ror } \Big| \nonumber \\
	&\leq  2\sqrt{\frac{\log(\frac{54SAN^2}{(1-\gamma)\delta})}{N}} \Big(I - \gamma \Pv^{\widehat{\pi}, V} \Big)^{-1}  \sqrt{\mathrm{Var}_{P^{\widehat{\pi} }}(\widehat{V}^{\star, \ror })} \notag \\
	&\quad + \Bigg(\frac{8\log(\frac{54SAN^2 }{(1-\gamma)\delta})}{N(1-\gamma)} +  6\sqrt{\frac{2 \sigma\log(\frac{36SAN^2 }{\delta})}{(1-\gamma)^2 N}} + \frac{2\gamma \varepsilon_{\mathsf{opt}} + 4\sqrt{\sigma \varepsilon_{\mathsf{opt}} } }{1 -\gamma} \Bigg) 1 \notag \\
	& \leq \Bigg(\frac{8\log(\frac{54SAN^2 }{(1-\gamma)\delta})}{N(1-\gamma)} +  6\sqrt{\frac{2 \sigma\log(\frac{36SAN^2 }{\delta})}{(1-\gamma)^2 N}} + \frac{2\gamma \varepsilon_{\mathsf{opt}} + 4\sqrt{\sigma \varepsilon_{\mathsf{opt}} } }{1 -\gamma} \Bigg) 1 \notag \\
	&\quad +  \underbrace{ 2\sqrt{\frac{\log(\frac{54SAN^2}{(1-\gamma)\delta})}{N}} \left(I - \gamma \Pv^{\widehat{\pi}, V} \right)^{-1}  \sqrt{\mathrm{Var}_{\Pv^{\widehat{\pi}, V} }(V^{\widehat{\pi}, \ror })} }_{ =: \cG_4 } \notag \\
	& \quad + \underbrace{ 2\sqrt{\frac{\log(\frac{54SAN^2}{(1-\gamma)\delta})}{N}} \Big(I - \gamma \Pv^{\widehat{\pi}, V} \Big)^{-1}  \sqrt{  \mathrm{Var}_{\Pv^{\widehat{\pi}, V} }(\widehat{V}^{\widehat{\pi}, \ror } -V^{\widehat{\pi}, \ror }) } }_{ =: \cG_5 } \notag \\
	& \quad + \underbrace{2 \sqrt{\frac{\log(\frac{54SAN^2}{(1-\gamma)\delta})}{N}} \Big(I - \gamma \Pv^{\widehat{\pi}, \widehat{V}} \Big)^{-1}  \sqrt{ \left|\mathrm{Var}_{\Pv^{\widehat{\pi}, V} }(\widehat{V}^{\star, \ror }) - \mathrm{Var}_{\Pv^{\widehat{\pi}, V} }(\widehat{V}^{\widehat{\pi}, \ror }) \right| } }_{ =: \cG_6 } \notag \\
	& \quad + \underbrace{ 2 \sqrt{\frac{\log(\frac{54SAN^2}{(1-\gamma)\delta})}{N}} \Big(I - \gamma \Pv^{\widehat{\pi}, \widehat{V}} \Big)^{-1}  \sqrt{ \left|\mathrm{Var}_{P^{\widehat{\pi}} }(\widehat{V}^{\star, \ror })  - \mathrm{Var}_{\Pv^{\widehat{\pi}, V} }(\widehat{V}^{\star, \ror })  \right| } }_{ =: \cG_7 }.  \label{eq:chi2-v-hat-3-main2}
\end{align}
We shall bound each of the terms separately.
\begin{itemize}
\item The term $\cG_4$ can be controlled similar to  \eqref{eq:chi2-first-C4} as follows:
\begin{align}
\cG_4 &\leq  8\sqrt{\frac{\log(\frac{18SAN}{\delta})   }{\gamma^2(1-\gamma)^3  N }} 1. \label{eq:chi2-first-G4}
\end{align}
\item For $\cG_5$, it is observed that 
\begin{align}
\cG_5 &= 2\sqrt{\frac{\log(\frac{54SAN^2}{(1-\gamma)\delta})}{N}} \Big(I - \gamma \Pv^{\widehat{\pi}, V} \Big)^{-1}  \sqrt{  \mathrm{Var}_{\Pv^{\widehat{\pi}, V} }(\widehat{V}^{\widehat{\pi}, \ror } -V^{\widehat{\pi}, \ror }) } \notag \\
& \leq 2\sqrt{\frac{\log(\frac{54SAN^2}{\delta})}{(1-\gamma)^2N}} \left\|V^{\widehat{\pi},\ror} - \widehat{V}^{\widehat{\pi}, \ror } \right\|_\infty 1. \label{eq:chi2-first-D5}
\end{align}

\item Observing that $\cG_6$ and $\cG_7$ are almost the same as the terms $\cG_2$ (controlled in \eqref{eq:chi2-first-D2}) and $\cG_3$ (controlled in \eqref{eq:chi2-first-D3}), it is easily verified that they can be controlled as follows
\begin{align}
\cG_6  &\leq  4\sqrt{\frac{\gamma \varepsilon_{\mathsf{opt}} \log(\frac{54SAN^2}{(1-\gamma)\delta})}{(1-\gamma)^4N}} 1,  \qquad \qquad
\cG_7   \leq 2\sqrt{\frac{\sqrt{\ror}\log(\frac{54SAN^2}{\delta})}{(1-\gamma)^4N}} 1 .  \label{eq:chi2-first-D6-D7}
\end{align} 
\end{itemize}
To continue,  inserting  the results in \eqref{eq:chi2-first-G4}, \eqref{eq:chi2-first-D5}, and \eqref{eq:chi2-first-D6-D7} back to \eqref{eq:chi2-v-hat-3-main2} leads to
\begin{align}
&\Big(I - \gamma \Pv^{\widehat{\pi}, V} \Big)^{-1} \Big( \Phatv^{\widehat{\pi}, \widehat{V}} \widehat{V}^{\widehat{\pi}, \ror } - \Pv^{\widehat{\pi}, \widehat{V} } \widehat{V}^{\widehat{\pi}, \ror } \Big) \notag \\
& \leq \Bigg(\frac{8\log(\frac{54SAN^2 }{(1-\gamma)\delta})}{N(1-\gamma)} +  6\sqrt{\frac{2 \sigma\log(\frac{36SAN^2 }{\delta})}{(1-\gamma)^2 N}} + \frac{2\gamma \varepsilon_{\mathsf{opt}} + 4\sqrt{\sigma \varepsilon_{\mathsf{opt}} } }{1 -\gamma} \Bigg) 1 +  8\sqrt{\frac{\log(\frac{18SAN}{\delta})   }{\gamma^2(1-\gamma)^3  N }} 1  \notag \\
& \quad   + 2\sqrt{\frac{\log(\frac{54SAN^2}{\delta})}{(1-\gamma)^2N}} \left\|V^{\widehat{\pi},\ror} - \widehat{V}^{\widehat{\pi}, \ror } \right\|_\infty 1   + 4\sqrt{\frac{\gamma \varepsilon_{\mathsf{opt}} \log(\frac{54SAN^2}{(1-\gamma)\delta})}{(1-\gamma)^4N}} 1 + 2\sqrt{\frac{\sqrt{\ror}\log(\frac{54SAN^2}{\delta})}{(1-\gamma)^4N}} 1 \notag \\
& \leq 16 \left(1 + \frac{\sigma^{1/2} + \sigma^{1/4}}{\sqrt{1-\gamma}} \right) \sqrt{\frac{\log(\frac{54SAN^2}{(1-\gamma)\delta})}{\gamma^2 (1-\gamma)^3 N}} 1  +  \frac{10\log(\frac{54SAN^2}{(1-\gamma)\delta})}{N(1-\gamma)^2} 1 + 2\sqrt{\frac{\log(\frac{54SAN^2}{\delta})}{(1-\gamma)^2N}} \left\|V^{\widehat{\pi},\ror} - \widehat{V}^{\widehat{\pi}, \ror } \right\|_\infty 1, \label{eq:chi2-v-hat-second-finish}
\end{align}
where the last inequality holds by letting $\varepsilon_{\mathsf{opt}} \leq  \frac{\log(\frac{54SAN^2}{(1-\gamma)\delta})}{\gamma N}$, which directly satisfies $\varepsilon_{\mathsf{opt}}  \leq \frac{1-\gamma}{\gamma^3}$ by letting $N \geq \frac{\log(\frac{54SAN^2}{\delta})}{1-\gamma}$. 

Finally, inserting \eqref{eq:chi2-v-hat-first-finish} and \eqref{eq:chi2-v-hat-second-finish} back to \eqref{eq:chi2-v-pihat-two-terms} yields: taking $\varepsilon_{\mathsf{opt}} \leq \frac{\log(\frac{54SAN^2}{(1-\gamma)\delta})}{ \gamma N} $ and $N \geq \frac{16\log(\frac{54SAN^2}{\delta})}{(1-\gamma)^2}$, with probability at least $1-\delta$, one has
\begin{align}
&\left\|\widehat{V}^{\widehat{\pi}, \ror } - V^{\widehat{\pi}, \ror}\right\|_\infty \notag \\
&\leq \max\Big\{18 \left(1 + \frac{\sigma^{1/2} + \sigma^{1/4}}{\sqrt{1-\gamma}} \right) \sqrt{\frac{\log(\frac{54SAN^2}{(1-\gamma)\delta})}{\gamma^2 (1-\gamma)^3 N}} + \frac{10\log(\frac{54SAN^2 }{(1-\gamma)\delta})}{N(1-\gamma)^2}     , \notag \\
& \quad 16 \left(1 + \frac{\sigma^{1/2} + \sigma^{1/4}}{\sqrt{1-\gamma}} \right) \sqrt{\frac{\log(\frac{54SAN^2}{(1-\gamma)\delta})}{\gamma^2 (1-\gamma)^3 N}}   +  \frac{10\log(\frac{54SAN^2}{(1-\gamma)\delta})}{N(1-\gamma)^2} + 2\sqrt{\frac{\log(\frac{54SAN^2}{\delta})}{(1-\gamma)^2N}} \left\|V^{\widehat{\pi},\ror} - \widehat{V}^{\widehat{\pi}, \ror } \right\|_\infty \Big\} \notag \\
& \leq 36 \left(1 + \frac{\sigma^{1/2} + \sigma^{1/4}}{\sqrt{1-\gamma}} \right) \sqrt{\frac{\log(\frac{54SAN^2}{(1-\gamma)\delta})}{\gamma^2 (1-\gamma)^3 N}} + \frac{20\log(\frac{54SAN^2 }{(1-\gamma)\delta})}{N(1-\gamma)^2}. \label{eq:chi2-pi-star-upper-final2}
\end{align}

\paragraph{Step 5: summing up the results.}
Inserting the results in \eqref{eq:chi2-pi-star-upper-final2} and \eqref{eq:chi2-pi-star-upper-final} back to \eqref{eq:l2-decompose} completes the proof as follows: taking $\varepsilon_{\mathsf{opt}} \leq \frac{\log(\frac{54SAN^2}{(1-\gamma)\delta})}{ \gamma N}$ and $N \geq \frac{16\log(\frac{54SAN^2}{\delta})}{(1-\gamma)^2}$, with probability at least $1-\delta$,
\begin{align}
 \big\|V^{\star, \ror} - V^{\widehat{\pi}, \ror} 
 \big\|_\infty
	&\leq \big\|V^{\pi^\star, \ror} - \widehat{V}^{\pi^\star, \ror}\big\|_\infty + \frac{2\gamma \varepsilon_{\mathsf{opt}} }{1 -\gamma} + \big\|\widehat{V}^{\widehat{\pi}, \ror} - V^{\widehat{\pi}, \ror}\big\|_\infty \notag \\
	& \leq \frac{2\gamma \varepsilon_{\mathsf{opt}} }{1 -\gamma} +  96\sqrt{\frac{ \log(\frac{24SAN}{\delta})}{ (1-\gamma)^3  N} }\left(1+\frac{\sigma^{1/2} + \sigma^{1/4}}{\sqrt{1-\gamma}}\right) + \frac{10\log(\frac{18SAN}{\delta}) }{(1-\gamma)^2N} \notag \\
	& \quad + 36 \left(1 + \frac{\sigma^{1/2} + \sigma^{1/4}}{\sqrt{1-\gamma}} \right) \sqrt{\frac{\log(\frac{54SAN^2}{(1-\gamma)\delta})}{\gamma^2 (1-\gamma)^3 N}} + \frac{20\log(\frac{54SAN^2 }{(1-\gamma)\delta})}{N(1-\gamma)^2} \notag \\
	&\leq 132 \left(1 + \frac{\sigma^{1/2} + \sigma^{1/4}}{\sqrt{1-\gamma}} \right) \sqrt{\frac{\log(\frac{54SAN^2}{(1-\gamma)\delta})}{\gamma^2 (1-\gamma)^3 N}}  + \frac{22 \log(\frac{54SAN^2 }{(1-\gamma)\delta})}{N(1-\gamma)^2} \notag \\
	&\leq 550 \left(1 + \frac{\sigma^{1/2} + \sigma^{1/4}}{\sqrt{1-\gamma}} \right) \sqrt{\frac{\log(\frac{54SAN^2}{(1-\gamma)\delta})}{\gamma^2 (1-\gamma)^3 N}}, 
\end{align}
where the last inequality holds by  $\gamma\geq \frac{1}{4}$ and $N \geq \frac{16 \log(\frac{54SAN^2}{\delta})}{(1-\gamma)}$.

\subsection{Proof of the auxiliary lemmas}
\subsubsection{Proof of Lemma~\ref{lemma:p-phat-V-gap-l2}}\label{proof:lemma:p-phat-V-gap-l2}

Without loss of generality, we focus on a more general form that considers any fixed deterministic policy $\pi$.
\paragraph{Step 1: controlling the point-wise concentration.}
Consider any fixed policy $\pi$ and the corresponding robust value vector $V \defn V^{\pi,\ror}$ (independent from $\widehat{P}^\no$).
Invoking Lemma~\ref{lem:dual-vi-l2-norm} leads to that for any $(s,a)\in \cS\times \cA$,
\begin{align}
 &\left| \pmhat^{\pi, V}_{s,a} V^{\pi,\ror}  -  \pmin^{\pi, V}_{s,a} V^{\pi,\ror} \right| \notag \\
 &=\bigg|\max_{\alpha\in\left[\min_s V(s), \max_s V(s)\right]} \left\{ P^\no_{s,a} [V]_\alpha - \sqrt{\sigma \mathsf{Var}_{P^\no_{s,a} }\left([V]_\alpha\right) }  \right\}  \nonumber\\
	&\qquad \qquad- \max_{\alpha\in\left[\min_s V(s), \max_s V(s)\right]} \left\{ \widehat{P}^\no_{s,a} [V]_\alpha - \sqrt{\sigma \mathsf{Var}_{\widehat{P}^\no_{s,a} }\left([V]_\alpha\right) } \right\} \bigg| \nonumber\\
	&\leq \max_{\alpha\in\left[\min_s V(s), \max_s V(s)\right]} \left| \left(P^{\no}_{s,a} - \widehat{P}^{\no}_{s,a} \right) [V]_\alpha +  \sqrt{\sigma \mathsf{Var}_{\widehat{P}^\no_{s,a} }\left([V]_\alpha\right) } - \sqrt{\sigma \mathsf{Var}_{P^\no_{s,a} }\left([V]_\alpha\right) } \right| \nonumber \\
	& \leq  \max_{\alpha\in\left[\min_s V(s), \max_s V(s)\right]} \left| \left(P^{\no}_{s,a} - \widehat{P}^{\no}_{s,a} \right) [V]_\alpha  \right| + \nonumber \\
	&\qquad + \max_{\alpha\in\left[\min_s V(s), \max_s V(s)\right]} \sqrt{\sigma}  \left| \sqrt{  \mathsf{Var}_{\widehat{P}^\no_{s,a} }\left([V]_\alpha\right) } - \sqrt{  \mathsf{Var}_{P^\no_{s,a} }\left([V]_\alpha\right) } \right| , \label{eq:l2-t-that-gap}
\end{align}
where the first inequality follows by that the maximum operator is $1$-Lipschitz, and the second inequality follows from the triangle inequality.
Observing that the first term in \eqref{eq:l2-t-that-gap} is exactly the same as \eqref{eq:tv-t-that-gap}, recalling the fact in \eqref{eq:V-p-phat-gap-one-alpha-hoeffding-union} directly leads to: with probability at least $1-\delta$, 
\begin{align}
\max_{\alpha\in [\min_s V(s), \max_s V(s)]} \left| \left(P^{\no}_{s,a} - \widehat{P}^{\no}_{s,a} \right) [V]_\alpha\right| & \leq 2\sqrt{\frac{\log(\frac{18SAN}{\delta})}{N}} \sqrt{\mathrm{Var}_{P^{\no}_{s,a}}(V)} +  \frac{\log(\frac{18SAN}{\delta})}{N(1-\gamma)} \notag \\
& \leq  2\sqrt{\frac{\log(\frac{18SA N}{\delta})}{(1-\gamma)^2N}}  \label{eq:chi2-V-p-star-first}
\end{align}
holds for all $(s,a)\in\cS\times\cA$.
Then the remainder of the proof focuses on controlling the second term in \eqref{eq:l2-t-that-gap}.
\paragraph{Step 2: controlling the second term in \eqref{eq:l2-t-that-gap}.}
For any given $(s,a) \in\cS\times \cA$ and fixed $\alpha\in [0,\frac{1}{1-\gamma}]$, applying the concentration inequality \citep[Lemma~6]{panaganti2021sample} with $\|[V]_\alpha\|_\infty\leq \frac{1}{1-\gamma}$, we arrive at
\begin{align}
	  \left| \sqrt{  \mathsf{Var}_{\widehat{P}^\no_{s,a} }\left([V]_\alpha\right) } - \sqrt{  \mathsf{Var}_{P^\no_{s,a} }\left([V]_\alpha\right) } \right| \leq \sqrt{\frac{2  \log(\frac{2}{\delta})}{(1-\gamma)^2 N}} \label{eq:chi2-v-star-pointwise}
\end{align}
holds with probability at least $1-\delta$.
To obtain a uniform bound, we first observe the follow lemma proven in Appendix~\ref{proof:lem:chi2-alpha-loo}.

 \begin{lemma}\label{lem:chi2-alpha-loo}
For any $V$ obeying $\|V\|_\infty \leq \frac{1}{1-\gamma}$, the function $J_{s,a}(\alpha, V): = \Big| \sqrt{  \mathsf{Var}_{\widehat{P}^\no_{s,a} }\left([V]_\alpha\right) } - \sqrt{  \mathsf{Var}_{P^\no_{s,a} }\left([V]_\alpha\right) } \Big| $ w.r.t. $\alpha$  obeys
\begin{align*}
\left|J_{s,a}(\alpha_1, V) - J_{s,a}(\alpha_2, V) \right| \leq 4\sqrt{\frac{ |\alpha_1 - \alpha_2|}{1-\gamma}}.
\end{align*}

\end{lemma} 

 In addition, we can construct an $\varepsilon_3$-net $\cN_{\varepsilon_3}$ over $[0, \frac{1}{1-\gamma}]$ whose size is $|\cN_{\varepsilon_3}| \leq \frac{3}{\varepsilon_3(1-\gamma)}$ \citep{vershynin2018high}.
Armed with the above, we can derive the uniform bound over $\alpha \in [\min_s V(s), \max_s V(s)] \subset [0, 1/(1-\gamma)]$: with probability at least $1 - \frac{\delta}{SA}$, it holds that for any $(s,a)\in \cS \times \cA$,
\begin{align}
  &\max_{\alpha\in\left[\min_s V(s), \max_s V(s)\right]} \left| \sqrt{  \mathsf{Var}_{\widehat{P}^\no_{s,a} }\left([V]_\alpha\right) } - \sqrt{  \mathsf{Var}_{P^\no_{s,a} }\left([V]_\alpha\right) } \right| \nonumber \\
  &\leq \max_{\alpha\in\left[0, 1/(1-\gamma)\right]} \left| \sqrt{  \mathsf{Var}_{\widehat{P}^\no_{s,a} }\left([V]_\alpha\right) } - \sqrt{  \mathsf{Var}_{P^\no_{s,a} }\left([V]_\alpha\right) } \right| \nonumber \\
  & \overset{\mathrm{(i)}}{\leq} 4\sqrt{\frac{ \varepsilon_3}{1-\gamma}}  + \sup_{\alpha\in \cN_{\varepsilon_3}} \left| \sqrt{  \mathsf{Var}_{\widehat{P}^\no_{s,a} }\left([V]_\alpha\right) } - \sqrt{  \mathsf{Var}_{P^\no_{s,a} }\left([V]_\alpha\right) } \right|\nonumber  \\
  & \overset{\mathrm{(ii)}}{\leq} 4\sqrt{\frac{ \varepsilon_3}{1-\gamma}}  + \sqrt{\frac{2  \log(\frac{2SA|N_{\varepsilon_3}| }{\delta})}{(1-\gamma)^2 N}} \nonumber \\
  &\overset{\mathrm{(iii)}}{\leq} 2\sqrt{\frac{2  \log(\frac{2SA|N_{\varepsilon_3}| }{\delta})}{(1-\gamma)^2 N}} \leq 2\sqrt{\frac{2  \log(\frac{24SAN }{\delta})}{(1-\gamma)^2 N}}  ,\label{eq:chi2-V-p-phat-gap-one-alpha-bernstein-union}
\end{align}
where (i) holds by the property of $N_{\varepsilon_3}$, (ii) follows from \eqref{eq:chi2-v-star-pointwise}, (iii) arises from taking $\varepsilon_3 = \frac{\log(\frac{2SA|\cN_{\varepsilon_3}|}{\delta})}{8N(1-\gamma)}$, and the last inequality is verified by $|\cN_{\varepsilon_3}| \leq \frac{3}{\varepsilon_3(1-\gamma)} \leq 24N$.

Inserting \eqref{eq:chi2-V-p-star-first} and \eqref{eq:chi2-V-p-phat-gap-one-alpha-bernstein-union} back to \eqref{eq:l2-t-that-gap} and taking the union bound over $(s,a)\in\cS\times \cA$, we arrive at that for all $(s,a)\in\cS\times \cA$, with probability at least $1- {\delta}$,
\begin{align*}
\left| \pmhat^{\pi, V}_{s,a} V  -  \pmin^{\pi, V}_{s,a} V \right| &\leq \max_{\alpha\in\left[\min_s V(s), \max_s V(s)\right]} \left| \left(P^{\no}_{s,a} - \widehat{P}^{\no}_{s,a} \right) [V]_\alpha  \right| + \nonumber \\
	&\qquad + \max_{\alpha\in\left[\min_s V(s), \max_s V(s)\right]}\left| \sqrt{\sigma \mathsf{Var}_{\widehat{P}^\no_{s,a} }\left([V]_\alpha\right) } - \sqrt{\sigma \mathsf{Var}_{P^\no_{s,a} }\left([V]_\alpha\right) } \right| \nonumber \\
	& \leq 2\sqrt{\frac{\log(\frac{18SAN}{\delta})}{N}} \sqrt{\mathrm{Var}_{P^{\no}_{s,a}}(V)} +  \frac{\log(\frac{18SAN}{\delta})}{N(1-\gamma)} + 4\sqrt{\frac{ \sigma\log(\frac{24SAN }{\delta})}{(1-\gamma)^2 N}} \notag \\
	&\leq  \sqrt{\frac{2\log(\frac{18SA N}{\delta})}{(1-\gamma)^2N}} + 4\sqrt{\frac{ \sigma\log(\frac{24SAN }{\delta})}{(1-\gamma)^2 N}}.
\end{align*}

Finally, recalling the matrix form leads to: with probability at least $1-\delta$,
\begin{align*}
\left| \Phatv^{\pi, V} V -  \Pv^{\pi, V} V\right|_\infty 
& \leq  2\sqrt{\frac{\log(\frac{18SAN}{\delta})}{N}} \sqrt{\mathrm{Var}_{P^{\pi}}(V)} +  \frac{\log(\frac{18SAN}{\delta})}{N(1-\gamma)} 1 + 4\sqrt{\frac{ \sigma\log(\frac{24SAN }{\delta})}{(1-\gamma)^2 N}} 1\notag \\
	&\leq  \sqrt{\frac{2\log(\frac{18SA N}{\delta})}{(1-\gamma)^2N}} 1 + 4\sqrt{\frac{ \sigma\log(\frac{24SAN }{\delta})}{(1-\gamma)^2 N}} 1.
\end{align*}

Applying the above results with $\pi = \pi^\star$ and $V = V^{\star, \ror}$ completes the proof.

\subsubsection{Proof of Lemma~\ref{lemma:dro-b-bound-infinite-loo-chi2}} \label{proof:lemma:dro-b-bound-infinite-loo-chi2}
\paragraph{Step 1: decomposing the term of interest.}
The proof follows the routine of the proof of  Lemma~\ref{lemma:dro-b-bound-infinite-loo-l1} in Appendix~\ref{proof:lemma:dro-b-bound-infinite-loo-l1}. To begin with, for any $(s,a)\in\cS\times\cA$, following the same arguments of \eqref{eq:l2-t-that-gap} yields

\begin{align}
&\left| \pmhat^{\widehat{\pi}, \widehat{V}}_{s,a} \widehat{V}^{\widehat{\pi}, \ror}  -  \pmin^{\widehat{\pi}, \widehat{V} }_{s,a} \widehat{V}^{\widehat{\pi}, \ror }  \right| \leq \max_{\alpha\in\left[\min_s \widehat{V}^{\widehat{\pi}, \ror }(s), \max_s \widehat{V}^{\widehat{\pi}, \ror } (s)\right]} \left| \left(P^{\no}_{s,a} - \widehat{P}^{\no}_{s,a} \right) \big[\widehat{V}^{\widehat{\pi}, \ror}\big]_\alpha  \right| + \nonumber \\
	&\qquad +  \max_{\alpha\in\left[\min_s \widehat{V}^{\widehat{\pi}, \ror }(s), \max_s \widehat{V}^{\widehat{\pi}, \ror } (s)\right]} \sqrt{\sigma } \left| \sqrt{  \mathsf{Var}_{\widehat{P}^\no_{s,a} }\left( \big[\widehat{V}^{\widehat{\pi}, \ror}\big]_\alpha \right) } - \sqrt{  \mathsf{Var}_{P^\no_{s,a} }\Big( \big[\widehat{V}^{\widehat{\pi}, \ror}\big]_\alpha \Big) } \right|. \label{eq:l2-t-that-gap-vhat}
\end{align}

Invoking the fact in the first line of \eqref{eq:pihat-V-Vhat-gap} and \eqref{eq:vhat-loo-hoeffindg-final2} (for proving Lemma~\ref{lemma:dro-b-bound-infinite-loo-l1}), the first term in \eqref{eq:l2-t-that-gap-vhat} obeys
\begin{align}
 &\max_{\alpha\in\left[\min_s \widehat{V}^{\widehat{\pi}, \ror }(s), \max_s \widehat{V}^{\widehat{\pi}, \ror } (s)\right]} \left| \left(P^{\no}_{s,a} - \widehat{P}^{\no}_{s,a} \right) \big[\widehat{V}^{\widehat{\pi}, \ror}\big]_\alpha  \right| \notag \\
 & \leq \max_{\alpha\in[0,1/(1-\gamma)]} \left| \left(P^{\no}_{s,a} - \widehat{P}^{\no}_{s,a} \right) \big[\widehat{V}^{\widehat{\pi}, \ror }\big]_\alpha\right| \notag \\
& \leq  2\sqrt{\frac{\log(\frac{54SAN^2} {(1-\gamma)\delta})}{N}} \sqrt{\mathrm{Var}_{P^{\no}_{s,a}}(\widehat{V}^{\star,\ror})} +  \frac{8\log(\frac{54SAN^2 }{(1-\gamma)\delta})}{N(1-\gamma)} + \frac{2\gamma \varepsilon_{\mathsf{opt}} }{1 -\gamma} \notag\\
& \leq 10\sqrt{\frac{\log(\frac{54SAN^2} {(1-\gamma)\delta})}{(1 -\gamma)^2 N}} + \frac{2\gamma \varepsilon_{\mathsf{opt}} }{1 -\gamma},    \label{eq:chi2-vhat-first-term-final}
\end{align}
by letting $N \geq \log \left(\frac{54SAN^2}{(1-\gamma)\delta} \right)$.
The remainder of the proof will focus on controlling the second term  of \eqref{eq:l2-t-that-gap-vhat}.
\paragraph{Step 2: controlling the second term  of \eqref{eq:l2-t-that-gap-vhat}.} Towards this, we recall the auxiliary robust MDP $\widehat{\cM}_{\mathsf{rob}}^{s,u}$ defined in Appendix~\ref{proof:lemma:dro-b-bound-infinite-loo-l1}. Taking the uncertainty set $\cU^{\ror}(\cdot) \defn \cU_{\chi^2}^{\ror}(\cdot)$ for both $\widehat{\cM}_{\mathsf{rob}}^{s,u}$ and $\widehat{\cM}_{\mathsf{rob}}$, we recall the corresponding robust Bellman operator $\widehat{\cT}^{\sigma}_{s,u}(\cdot)$ in \eqref{eq:aux-operator-auxiliary} and the following definition in \eqref{eq:def-u-star}
\begin{align}\label{eq:chi2-def-u-star}
 u^\star \defn \widehat{V}^{\star,\ror}(s) - \gamma \inf_{ \cP \in \unb^{\sigma}(e_{s})} \cP \widehat{V}^{\star,\ror}.
\end{align} 
Following the arguments in Appendix~\ref{proof:lemma:dro-b-bound-infinite-loo-l1}, it can be verified that there exists a unique fixed point $\widehat{Q}^{\star,\ror}_{s,u}$ of the operator $\widehat{\cT}^{\sigma}_{s,u}(\cdot)$, which satisfies $ 0\leq \widehat{Q}^{\star,\ror}_{s,u} \leq \frac{1}{1-\gamma}    {1}$. In addition, the corresponding robust value function coincides with that of the operator $\widehat{\cT}^{\sigma}(\cdot)$, i.e., $\widehat{V}^{\star,\ror}_{s,u}  = \widehat{V}^{\star,\ror} $.

We recall the $\cN_{\varepsilon_2}$-net over $\left[0, \frac{1}{1-\gamma}\right]$ whose size obeying $|\cN_{\varepsilon_2}| \leq \frac{3}{\varepsilon_2(1-\gamma)}$ \citep{vershynin2018high}. Then for all $u\in \cN_{\varepsilon_2}$ and a fixed $\alpha$,  $\widehat{\cM}_{\mathsf{rob}}^{s,u}$ is statistically independent from $\widehat{P}^\no_{s,a}$, which indicates the independence between $[\widehat{V}^{\star,\ror}_{s,u}]_\alpha$ and $\widehat{P}^\no_{s,a}$. With this in mind, invoking the fact in \eqref{eq:chi2-V-p-phat-gap-one-alpha-bernstein-union} and taking the union bound over all $(s,a) \in\cS\times \cA$ and $u\in \cN_{\varepsilon_2}$ yields that, with probability at least $1-\delta$,
\begin{align}\label{eq:chi2-approx-gap-union-bound-n1-n2}
  \max_{\alpha\in\left[0, 1/(1-\gamma)\right]} \bigg| \sqrt{  \mathsf{Var}_{\widehat{P}^\no_{s,a} }\left([\widehat{V}^{\star,\ror}_{s,u}]_\alpha\right) } - \sqrt{  \mathsf{Var}_{P^\no_{s,a} }\left([\widehat{V}^{\star,\ror}_{s,u}]_\alpha\right) } \bigg|  &\leq 2\sqrt{\frac{2  \log(\frac{24SAN |\cN_{\varepsilon_2}|}{\delta})}{(1-\gamma)^2 N}}  
\end{align}
holds for all $(s,a, u) \in \cS\times \cA \times  \cN_{\varepsilon_2}$.

To continue, we decompose the main part of the second term in \eqref{eq:l2-t-that-gap-vhat} as follows:
\begin{align}
  &   \max_{\alpha\in\left[\min_s \widehat{V}^{\widehat{\pi}, \ror }(s), \max_s \widehat{V}^{\widehat{\pi}, \ror } (s)\right]}  \bigg| \sqrt{  \mathsf{Var}_{\widehat{P}^\no_{s,a} }\left( \big[\widehat{V}^{\widehat{\pi}, \ror}\big]_\alpha \right) } - \sqrt{  \mathsf{Var}_{P^\no_{s,a} }\left( \big[\widehat{V}^{\widehat{\pi}, \ror}\big]_\alpha \right) } \bigg| \nonumber \\
& \leq  \max_{\alpha\in\left[0, 1/(1-\gamma)\right]}  \bigg| \sqrt{  \mathsf{Var}_{\widehat{P}^\no_{s,a} }\left( \big[\widehat{V}^{\widehat{\pi}, \ror}\big]_\alpha \right) } - \sqrt{  \mathsf{Var}_{P^\no_{s,a} }\left( \big[\widehat{V}^{\widehat{\pi}, \ror}\big]_\alpha \right) } \bigg| \nonumber \\
& \overset{\mathrm{(i)}}{\leq} \max_{\alpha\in\left[0, 1/(1-\gamma)\right]}  \bigg| \sqrt{  \mathsf{Var}_{\widehat{P}^\no_{s,a} }\left( \big[\widehat{V}^{\star, \ror}\big]_\alpha \right) } - \sqrt{  \mathsf{Var}_{P^\no_{s,a} }\left( \big[\widehat{V}^{\star, \ror}\big]_\alpha \right) } \bigg| \nonumber \\
& \quad +  \max_{\alpha\in\left[0, 1/(1-\gamma)\right]} \bigg[ \sqrt{  \left|\mathsf{Var}_{\widehat{P}^\no_{s,a} }\left( \big[\widehat{V}^{\widehat{\pi}, \ror}\big]_\alpha \right)   - \mathsf{Var}_{\widehat{P}^\no_{s,a} }\left( \big[\widehat{V}^{\star, \ror}\big]_\alpha \right)  \right| } \notag \\
&\quad + \sqrt{  \left|\mathsf{Var}_{P^\no_{s,a} }\left( \big[\widehat{V}^{\widehat{\pi}, \ror}\big]_\alpha \right)   - \mathsf{Var}_{P^\no_{s,a} }\left( \big[\widehat{V}^{\star, \ror}\big]_\alpha \right)  \right| } \bigg] \notag \\
& \overset{\mathrm{(ii)}}{\leq}  \max_{\alpha\in\left[0, 1/(1-\gamma)\right]}  \bigg| \sqrt{  \mathsf{Var}_{\widehat{P}^\no_{s,a} }\left( \big[\widehat{V}^{\star, \ror}\big]_\alpha \right) } - \sqrt{  \mathsf{Var}_{P^\no_{s,a} }\left( \big[\widehat{V}^{\star, \ror}\big]_\alpha \right) } \bigg| \notag \\
&\quad + \max_{\alpha\in\left[0, 1/(1-\gamma)\right]}  2\sqrt{ \frac{2 }{(1-\gamma)} \left\| \big[\widehat{V}^{\widehat{\pi}, \ror}\big]_\alpha - \big[\widehat{V}^{\star, \ror}\big]_\alpha \right\|_\infty} \notag \\
& \leq \max_{\alpha\in\left[0, 1/(1-\gamma)\right]}  \left| \sqrt{  \mathsf{Var}_{\widehat{P}^\no_{s,a} }\left( \big[\widehat{V}^{\star, \ror}\big]_\alpha \right) } - \sqrt{  \mathsf{Var}_{P^\no_{s,a} }\left( \big[\widehat{V}^{\star, \ror}\big]_\alpha \right) } \right| + 4\sqrt{ \frac{  \varepsilon_{\mathsf{opt}} }{(1-\gamma)^2}}, \label{eq:chi2-Loo-summary-vhat-initial}
\end{align}
where (i) holds by the triangle inequality, (ii) arises from applying Lemma~\ref{eq:tv-auxiliary-lemma}, and the last inequality holds by \eqref{eq:opt-error}.

Armed with the above facts, invoking the identity $\widehat{V}^{\star,\ror} = \widehat{V}^{\star,\ror}_{s,u^\star}$ leads to that for all $(s,a)\in\cS \times \cA$, with probability at least $1-\delta$,
\begin{align}
&  \max_{\alpha\in\left[0, 1/(1-\gamma)\right]}  \bigg| \sqrt{  \mathsf{Var}_{\widehat{P}^\no_{s,a} }\left( \big[\widehat{V}^{\star, \ror}\big]_\alpha \right) } - \sqrt{  \mathsf{Var}_{P^\no_{s,a} }\left( \big[\widehat{V}^{\star, \ror}\big]_\alpha \right) } \bigg| \notag \\
& = \max_{\alpha\in\left[0, 1/(1-\gamma)\right]}  \left| \sqrt{  \mathsf{Var}_{\widehat{P}^\no_{s,a} }\left( \left[\widehat{V}^{\star,\ror}_{s,u^\star}\right]_\alpha \right) } - \sqrt{  \mathsf{Var}_{P^\no_{s,a} }\left( \left[\widehat{V}^{\star,\ror}_{s,u^\star}\right]_\alpha \right) } \right|  \notag \\
&\overset{\mathrm{(i)}}{\leq} \max_{\alpha\in\left[0, 1/(1-\gamma)\right]}  \left| \sqrt{  \mathsf{Var}_{\widehat{P}^\no_{s,a} }\left( \left[\widehat{V}^{\star,\ror}_{s, \overline{u} }\right]_\alpha \right) } - \sqrt{  \mathsf{Var}_{P^\no_{s,a} }\left( \left[\widehat{V}^{\star,\ror}_{s,\overline{u} }\right]_\alpha \right) } \right|  \nonumber \\
& \quad + \max_{\alpha\in\left[0, 1/(1-\gamma)\right]} \bigg[ \sqrt{  \left|\mathsf{Var}_{\widehat{P}^\no_{s,a} }\left( \left[\widehat{V}^{\star,\ror}_{s, u^\star }\right]_\alpha \right)   - \mathsf{Var}_{\widehat{P}^\no_{s,a} }\left( \left[\widehat{V}^{\star,\ror}_{s, \overline{u} }\right]_\alpha \right)  \right| } \notag \\
&\quad + \sqrt{  \left|\mathsf{Var}_{P^\no_{s,a} }\left( \left[\widehat{V}^{\star,\ror}_{s, u^\star }\right]_\alpha \right)   - \mathsf{Var}_{P^\no_{s,a} }\left( \left[\widehat{V}^{\star,\ror}_{s, \overline{u} }\right]_\alpha \right)  \right| } \bigg] \notag \\
& \overset{\mathrm{(ii)}}{\leq}  \max_{\alpha\in\left[0, 1/(1-\gamma)\right]}  \left| \sqrt{  \mathsf{Var}_{\widehat{P}^\no_{s,a} }\left( \left[\widehat{V}^{\star,\ror}_{s, \overline{u} }\right]_\alpha \right) } - \sqrt{  \mathsf{Var}_{P^\no_{s,a} }\left( \left[\widehat{V}^{\star,\ror}_{s,\overline{u} }\right]_\alpha \right) } \right| + 4\sqrt{\frac{ \varepsilon_2}{(1-\gamma)}} \nonumber \\
& \overset{\mathrm{(iii)}}{\leq}  2\sqrt{\frac{2  \log(\frac{24SAN |\cN_{\varepsilon_2}|}{\delta})}{(1-\gamma)^2 N}}    + 4\sqrt{\frac{ \varepsilon_2}{(1-\gamma)}} \nonumber \\
& \leq 6\sqrt{\frac{2  \log(\frac{36SAN^2}{\delta})}{(1-\gamma)^2 N}} , \label{eq:chi2-Loo-summary-vhat}
\end{align}
where (i) holds by   the triangle inequality, (ii) arises from applying Lemma~\ref{eq:tv-auxiliary-lemma} and the fact $\left\| \widehat{V}^{\star,\ror}_{s,\overline{u}} -  \widehat{V}^{\star,\ror}_{s,u^\star}\right\|_\infty \leq  \frac{\varepsilon_2}{(1-\gamma)}$ (see \eqref{eq:auxiliary-value-gap}), (iii) follows from \eqref{eq:chi2-approx-gap-union-bound-n1-n2}, and the last inequality holds by letting $\varepsilon_2 = \frac{2\log(\frac{24SAN |N_{\varepsilon_2}|} {\delta})}{(1-\gamma)N}$, which leads to $|N_{\varepsilon_2}| \leq \frac{3}{\varepsilon_2(1-\gamma)} \leq \frac{3N}{2}$.

In summary, inserting \eqref{eq:chi2-Loo-summary-vhat} back to \eqref{eq:chi2-Loo-summary-vhat-initial} leads to with probability at least $1-\delta$,
\begin{align}
  &   \max_{\alpha\in\left[\min_s \widehat{V}^{\widehat{\pi}, \ror }(s), \max_s \widehat{V}^{\widehat{\pi}, \ror } (s)\right]}  \bigg| \sqrt{  \mathsf{Var}_{\widehat{P}^\no_{s,a} }\left( \big[\widehat{V}^{\widehat{\pi}, \ror}\big]_\alpha \right) } - \sqrt{  \mathsf{Var}_{P^\no_{s,a} }\left( \big[\widehat{V}^{\widehat{\pi}, \ror}\big]_\alpha \right) } \bigg| \nonumber \\ 
& \leq 6\sqrt{\frac{2 \sigma\log(\frac{36SAN^2 }{\delta})}{(1-\gamma)^2 N}} + 4\sqrt{ \frac{\ror \varepsilon_{\mathsf{opt}} }{(1-\gamma)^2}} \label{eq:chi2-vhat-K-sa}
\end{align}
holds for all $(s,a)\in\cS\times \cA$.

\paragraph{Step 4: finishing up.}
Inserting \eqref{eq:chi2-vhat-K-sa} and \eqref{eq:chi2-vhat-first-term-final} back to \eqref{eq:l2-t-that-gap-vhat}, we have for all $(s,a)\in \cS\ times \cA$, with probability at least $1-\delta$,
\begin{align}
 &\left| \pmhat^{\widehat{\pi}, \widehat{V}}_{s,a} \widehat{V}^{\widehat{\pi}, \ror}  -  \pmin^{\widehat{\pi}, \widehat{V} }_{s,a} \widehat{V}^{\widehat{\pi}, \ror }  \right|   \notag \\
& \leq 2\sqrt{\frac{\log(\frac{54SAN^2} {(1-\gamma)\delta})}{N}} \sqrt{\mathrm{Var}_{P^{\no}_{s,a}}(\widehat{V}^{\star,\ror})} +  \frac{8\log(\frac{54SAN^2 }{(1-\gamma)\delta})}{N(1-\gamma)} +  6\sqrt{\frac{2 \sigma\log(\frac{36SAN^2 }{\delta})}{(1-\gamma)^2 N}} + \frac{2\gamma \varepsilon_{\mathsf{opt}} + 4\sqrt{\sigma \varepsilon_{\mathsf{opt}} } }{1 -\gamma} \notag\\
& \leq 10\sqrt{\frac{\log(\frac{54SAN^2} {(1-\gamma)\delta})}{(1 -\gamma)^2 N}} + 6\sqrt{\frac{2 \sigma\log(\frac{36SAN^2 }{\delta})}{(1-\gamma)^2 N}} + \frac{2\gamma \varepsilon_{\mathsf{opt}} + 4\sqrt{\sigma \varepsilon_{\mathsf{opt}} } }{1 -\gamma}.  \label{eq:l2-t-that-gap-vhat-bounded}
\end{align}

Finally, recalling the matrix form in \eqref{eqn:ppivq}, taking $N \geq \log \left(\frac{54SAN^2}{(1-\gamma)\delta} \right)$, we complete the proof:
\begin{align}
&\Big| \Phatv^{\widehat{\pi}, \widehat{V}} \widehat{V}^{\widehat{\pi}, \ror } - \Pv^{\widehat{\pi}, \widehat{V} } \widehat{V}^{\widehat{\pi}, \ror }  \Big| \notag\\
&\leq 2\sqrt{\frac{\log(\frac{54SAN^2} {(1-\gamma)\delta})}{N}} \sqrt{\mathrm{Var}_{P^{\widehat{\pi}}}(\widehat{V}^{\star,\ror})} 1 +  \frac{8\log(\frac{54SAN^2 }{(1-\gamma)\delta})}{N(1-\gamma)} 1+  6\sqrt{\frac{2 \sigma\log(\frac{36SAN^2 }{\delta})}{(1-\gamma)^2 N}} 1+ \frac{2\gamma \varepsilon_{\mathsf{opt}} + 4\sqrt{\sigma \varepsilon_{\mathsf{opt}} } }{1 -\gamma} 1\notag\\
& \leq 10\sqrt{\frac{\log(\frac{54SAN^2} {(1-\gamma)\delta})}{(1 -\gamma)^2 N}} 1 + 6\sqrt{\frac{2 \sigma\log(\frac{36SAN^2 }{\delta})}{(1-\gamma)^2 N}} 1 + \frac{2\gamma \varepsilon_{\mathsf{opt}} + 4\sqrt{\sigma \varepsilon_{\mathsf{opt}} } }{1 -\gamma} 1. 
\end{align}

\subsubsection{Proof of Lemma~\ref{lemma:chi2-vhat-key1}}\label{proof:lemma:chi2-vhat-key1}

Following the proof pipeline of Lemma~\ref{lemma:tv-vhat-key1} in Appendix~\ref{proof:lemma:tv-vhat-key1}, we first recall \eqref{eq:bound-key-vstar}
\begin{align}
\Big(I - \gamma \Pv^{\widehat{\pi}, \widehat{V}} \Big)^{-1}  \sqrt{\mathrm{Var}_{\Pv^{\widehat{\pi}, \widehat{V}}}(\widehat{V}^{\widehat{\pi}, \ror })} \leq \sqrt{\frac{1}{1-\gamma}}\sqrt{ \sum_{t=0}^\infty \gamma^t \Big(\Pv^{\widehat{\pi}, \widehat{V}} \Big)^t \mathrm{Var}_{\Pv^{\widehat{\pi}, \widehat{V}}} (\widehat{V}^{\widehat{\pi}, \ror })}. \label{eq:bound-key-vhat-chi2}
\end{align}

Recall that we denote the minimum value of $\widehat{V}^{\widehat{\pi}, \ror }$ as $V_{\min} = \min_{s\in\cS} \widehat{V}^{\widehat{\pi}, \ror }(s)$ and $V' \defn  \widehat{V}^{\widehat{\pi}, \ror } - V_{\min} 1$. By the same argument as \eqref{eq:variance-tight-bound-vhat}, we arrive at 
\begin{align}
	&\mathrm{Var}_{\Pv^{\widehat{\pi}, \widehat{V}}} (\widehat{V}^{\widehat{\pi}, \ror })  \notag \\
	& \leq \Pv^{\widehat{\pi}, \widehat{V}}  \left(V' \circ V'\right) - \frac{1}{\gamma} V' \circ V' + \frac{2}{\gamma^2} \|V'\|_\infty 1 + \frac{2}{\gamma} \|V'\|_\infty \left| \left(  \Phatv^{\widehat{\pi}, \widehat{V} }- \Pv^{\widehat{\pi}, \widehat{V} } \right) \widehat{V}^{\widehat{\pi}, \ror } \right| \nonumber \\
	& \leq \Pv^{\widehat{\pi}, \widehat{V}} \left(V' \circ V'\right) - \frac{1}{\gamma} V' \circ V' + \frac{2}{\gamma^2} \|V'\|_\infty 1 \notag \\
	&\quad + \frac{2}{\gamma} \|V'\|_\infty \Bigg(10\sqrt{\frac{\log(\frac{54SAN^2} {(1-\gamma)\delta})}{(1 -\gamma)^2 N}}  + 6\sqrt{\frac{2 \sigma\log(\frac{36SAN^2 }{\delta})}{(1-\gamma)^2 N}}  + \frac{2\gamma \varepsilon_{\mathsf{opt}} + 4\sqrt{\sigma \varepsilon_{\mathsf{opt}} } }{1 -\gamma}  \Bigg) 1, \label{eq:variance-tight-bound-vhat-chi2}
\end{align}
where the last inequality follows by Lemma~\ref{lemma:dro-b-bound-infinite-loo-chi2}.
Plugging \eqref{eq:variance-tight-bound-vhat-chi2} back into \eqref{eq:bound-key-vhat-chi2} and following the routine of \eqref{eq:vstar-bound-key-recursive-vhat} leads to
\begin{align}
&\Big(I - \gamma \Pv^{\widehat{\pi}, \widehat{V}} \Big)^{-1}  \sqrt{\mathrm{Var}_{\Pv^{\widehat{\pi}, \widehat{V}}}(\widehat{V}^{\widehat{\pi}, \ror })} \notag\\
 &  \overset{\mathrm{(i)}}{\leq}  \sqrt{ \frac{1}{1-\gamma}} \sqrt{ \bigg| \sum_{t=0}^\infty \gamma^t \left(\Pv^{\widehat{\pi}, \widehat{V}} \right)^t \Big( \Pv^{\widehat{\pi}, \widehat{V}}  \left(V' \circ V'\right) - \frac{1}{\gamma} V' \circ V' \Big) \bigg|  }  \notag \\
 &\quad + \sqrt{\frac{1}{(1-\gamma)^2 \gamma^2}\bigg(2 + 20\sqrt{\frac{\log(\frac{54SAN^2} {(1-\gamma)\delta})}{(1 -\gamma)^2 N}}  + 12\sqrt{\frac{2 \sigma\log(\frac{36SAN^2 }{\delta})}{(1-\gamma)^2 N}}  + \frac{2\gamma \varepsilon_{\mathsf{opt}} + 8\sqrt{\sigma \varepsilon_{\mathsf{opt}} } }{1 -\gamma}   \bigg) \|V'\|_\infty } 1 \notag \\
 & \overset{\mathrm{(ii)}}{\leq} \sqrt{\frac{\|V'\|_{\infty}^2}{\gamma(1-\gamma)}} 1 + \sqrt{\frac{1}{(1-\gamma)^2 \gamma^2}\bigg(2 + 20\sqrt{\frac{\log(\frac{54SAN^2} {(1-\gamma)\delta})}{(1 -\gamma)^2 N}}  + 12\sqrt{\frac{2 \sigma\log(\frac{36SAN^2 }{\delta})}{(1-\gamma)^2 N}}  + \frac{2\gamma \varepsilon_{\mathsf{opt}} + 8\sqrt{\sigma \varepsilon_{\mathsf{opt}} } }{1 -\gamma}   \bigg) \|V'\|_\infty } 1 \notag \\
 & \overset{\mathrm{(iii)}}{\leq} \sqrt{\frac{\|V'\|_{\infty}^2}{\gamma(1-\gamma)}} 1 + \sqrt{\frac{32(1 + \sqrt{\sigma})\|V'\|_\infty }{(1-\gamma)^2 \gamma^2}} 1  \leq  7\sqrt{\frac{1 }{(1-\gamma)^3 \gamma^2}}  + 6\sqrt{\frac{\sqrt{\sigma} }{(1-\gamma)^3 \gamma^2}} 1,\label{eq:vstar-bound-key-recursive-vhat-chi2}
\end{align}
where (i) arises from following the routine of \eqref{eq:vstar-bound-key-recursive}, (ii) holds by repeating the argument of \eqref{eq:vstar-termI}, (iii) follows by taking $ N \geq\frac{\log(\frac{54SAN^2}{(1-\gamma)\delta})}{(1-\gamma)^2}$ and $\varepsilon_{\mathsf{opt}}\leq \frac{(1-\gamma)^2}{\gamma}$, and the     last inequality holds by $\|V'\|_\infty \leq \|V^{\star,\ror}\|_\infty \leq \frac{1}{1-\gamma}$.

\subsubsection{Proof of Lemma~\ref{lem:chi2-alpha-loo}} \label{proof:lem:chi2-alpha-loo}

For any $ 0 \leq \alpha_1, \alpha_2 \leq 1/(1-\gamma)$, one has
\begin{align}
	&|J_{s,a}(\alpha_1, V) - J_{s,a}(\alpha_2, V)| \notag \\
	&= \bigg| \left| \sqrt{  \mathsf{Var}_{\widehat{P}^\no_{s,a} }\left([V]_{\alpha_1}\right) } - \sqrt{  \mathsf{Var}_{P^\no_{s,a} }\left([V]_{\alpha_1}\right) } \right| - \left| \sqrt{  \mathsf{Var}_{\widehat{P}^\no_{s,a} }\left([V]_{\alpha_2}\right) } - \sqrt{  \mathsf{Var}_{P^\no_{s,a} }\left([V]_{\alpha_2}\right) } \right| \bigg| \notag \\
	& \overset{\mathrm{(i)}}{\leq} \left| \sqrt{  \mathsf{Var}_{\widehat{P}^\no_{s,a} }\left([V]_{\alpha_1}\right) } - \sqrt{  \mathsf{Var}_{P^\no_{s,a} }\left([V]_{\alpha_1}\right) } -  \sqrt{  \mathsf{Var}_{\widehat{P}^\no_{s,a} }\left([V]_{\alpha_2}\right) } + \sqrt{  \mathsf{Var}_{P^\no_{s,a} }\left([V]_{\alpha_2}\right) }\right| \notag \\
	& \leq \left| \sqrt{  \mathsf{Var}_{\widehat{P}^\no_{s,a} }\left([V]_{\alpha_1}\right) } - \sqrt{  \mathsf{Var}_{\widehat{P}^\no_{s,a} }\left([V]_{\alpha_2}\right) }\right| + \left| \sqrt{  \mathsf{Var}_{P^\no_{s,a} }\left([V]_{\alpha_1}\right) } - \sqrt{  \mathsf{Var}_{P^\no_{s,a} }\left([V]_{\alpha_2}\right) } \right| \notag \\
	& \overset{\mathrm{(ii)}}{\leq} \sqrt{  \mathsf{Var}_{\widehat{P}^\no_{s,a} }\left([V]_{\alpha_2}\right) -   \mathsf{Var}_{\widehat{P}^\no_{s,a} }\left([V]_{\alpha_1}\right) } + \sqrt{  \mathsf{Var}_{P^\no_{s,a} }\left([V]_{\alpha_2}\right) -   \mathsf{Var}_{P^\no_{s,a} }\left([V]_{\alpha_1}\right) }  \notag \\
& \overset{\mathrm{(iii)}}{\leq}   \sqrt{ \left| \widehat{P}^\no_{s,a} \left [ \left([V]_{\alpha_1}\right) \circ \left([V]_{\alpha_1}\right) - \left([V]_{\alpha_2}\right) \circ \left([V]_{\alpha_2}\right)\right] \right| +  \left| \widehat{P}^\no_{s,a}  \left([V]_{\alpha_1} + [V]_{\alpha_2}\right) \cdot \widehat{P}^\no_{s,a}  \left([V]_{\alpha_1} - [V]_{\alpha_2}\right)\right| }\notag \\
	& \quad +  \sqrt{\left| P^\no_{s,a} \left [ \left([V]_{\alpha_1}\right) \circ \left([V]_{\alpha_1}\right) - \left([V]_{\alpha_2}\right) \circ \left([V]_{\alpha_2}\right)\right] \right| +  \left| P^\no_{s,a}  \left([V]_{\alpha_1} + [V]_{\alpha_2}\right) \cdot P^\no_{s,a}  \left([V]_{\alpha_1} - [V]_{\alpha_2}\right)\right| } \notag \\
	& \leq  2  \sqrt{2(\alpha_1 +\alpha_2) |\alpha_1 - \alpha_2|} \leq  4\sqrt{\frac{ |\alpha_1 - \alpha_2|}{1-\gamma}}.
	\end{align}
Here, (i) holds by the fact $||x| - |y|| \leq |x-y|$ for all $x,y \in\mathbb{R}$, (ii) follows from the fact that $\sqrt{x} - \sqrt{y} \leq \sqrt{x-y}$ for any $x\geq y \geq 0$ and $\mathsf{Var}_{P }\left([V]_{\alpha_2}\right) \geq \mathsf{Var}_{P }\left([V]_{\alpha_1}\right)$ for any transition kernel $P\in \Delta(\cS)$, (iii) holds by the definition of $\mathrm{Var}_P(\cdot)$ defined in \eqref{eq:defn-variance}, and the last inequality arises from $ 0 \leq \alpha_1, \alpha_2 \leq 1/(1-\gamma)$.

\section{Proof of the lower bound with $\chi^2$ divergence: Theorem~\ref{thm:chi2-lower-bound}} \label{proof:thm:chi2-lower-bound}

To prove Theorem~\ref{thm:chi2-lower-bound}, we shall first construct some hard instances and then characterize the sample complexity requirements over these instances. The structure of the hard instances are the same as the ones used in the proof of Theorem~\ref{thm:l1-lower-bound}.

\subsection{Construction of the hard problem instances}
First, note that we shall use the same MDPs defined in section~\ref{proof:thm:l1-lower-bound} as follows
\begin{align*}
   \left\{ \cM_\phi=
    \left(\mathcal{S}, \mathcal{A}, P^{\phi}, r, \gamma \right) 
    \mymid \phi = \{0,1\}
    \right\}.
\end{align*}
In particular, we shall keep the structure of the transition kernel in \eqref{eq:Ph-construction-lower-infinite}, reward function in \eqref{eq:rh-construction-lower-bound-infinite} and initial state distribution in \eqref{eq:rho-defn-infinite-LB}, while $p$ and $\Delta$ shall be tailored to the $\chi^2$ divergence case.

\paragraph{Uncertainty set of the transition kernels.}
Recalling the uncertainty set associated with $\chi^2$ divergence in \eqref{eq:consider-chi2}, for any uncertainty level $\ror$, the uncertainty set throughout this section is defined as $\cU^{\ror}(P^\phi)$:
\begin{align}
\cU^{\ror}(P^\phi) &\defn \otimes \; \cU^{\ror}_{\mathsf{\chi^2}}(P^{\phi}_{s,a}),\qquad &\cU^{\ror}_{\mathsf{\chi^2}}(P^\phi_{s,a}) \defn \bigg\{ P_{s,a} \in \Delta(\cS): \sum_{s'\in \cS} \frac{\left(P(s' \mymid s,a) - P^\phi(s' \mymid s,a)\right)^2}{P^{\phi}(s' \mymid s,a)}  \leq \ror \bigg\}, \label{eq:chi2-ball-infinite-P-recall1}
\end{align}
where $\Delta(\cS)$ denote the simplex over the state space $\cS$.
Clearly, $\unb^{\ror}(P^{\phi}_{s,a}) = \{ P^{\phi}_{s,a} \}$ whenever the state transition is deterministic for $\chi^2$ divergence.
Here, $q$ and $\Delta$ (whose choice will be specified later in more detail) which determine the instances are specified as
\begin{align}\label{eq:chi2-p-q-defn-infinite2}
   0 &\leq q = \begin{cases}
      1-\gamma &\text{if } \ror \in \big( 0,\frac{1-\gamma}{4} \big) \\
       \frac{\ror}{1+\ror} &\text{if } \ror \in \big[ \frac{1-\gamma}{4} , \infty\big)
    \end{cases},  \qquad 
    p = q + \Delta  ,
\end{align}
and
\begin{align}\label{eq:chi2-delta-defn-infinite2}
0 &<\Delta \leq \begin{cases}
     \frac{1}{4}(1-\gamma) &\text{if } \ror \in \left( 0,\frac{1-\gamma}{4} \right ) \\
       \min \left\{\frac{1-\gamma}{\gamma(3+\sigma)},\, \frac{1}{2(1+\ror)}\right\}  &\text{if } \ror \in \left[ \frac{1-\gamma}{4}, \infty \right) \\
    \end{cases}. 
\end{align}
This directly ensures that 
$$p = \Delta + q \leq \max \left\{ \frac{\frac{1}{2} + \ror }{1+\ror}, \frac{5}{4}(1-\gamma) \right\}     \leq 1$$
since $\gamma \in \big[\frac{3}{4}, 1\big)$.

To continue, for any $(s,a,s')\in\cS\times \cA \times \cS$, we denote the infimum probability of moving to the next state $s'$ associated with any perturbed transition kernel $P_{s,a} \in \unb^{\ror}(P^{\phi}_{s,a})$ as
\begin{align}\label{eq:chi2-infinite-lw-def-p-q}
\underline{P}^{\phi}(s' \mymid s,a) &\defn \inf_{P_{s,a} \in \unb^{\ror}(P^{\phi}_{s,a})} P(s'  \mymid s,a).
\end{align}
In addition, we denote the transition from state $0$ to state $1$ as follows, which plays an important role in the analysis,
\begin{align}\label{eq:chi2-infinite-lw-p-q-perturb-inf}
\underline{p} &\defn \underline{P}^{\phi}(1 \mymid 0,\phi),\qquad \underline{q}  \defn \underline{P}^{\phi}(1  \mymid 0, 1-\phi).
\end{align}

Before continuing, we introduce some facts about $\underline{p}$ and $\underline{q}$ which are summarized as the following lemma; the proof is postponed to Appendix~\ref{proof:eq:chi2-under-p-range}.
\begin{lemma}\label{eq:chi2-under-p-range}
Consider any $\ror\in (0,\infty)$ and any $p,q,\Delta$ obeying \eqref{eq:chi2-p-q-defn-infinite2} and \eqref{eq:chi2-delta-defn-infinite2}, the following properties hold
\begin{align}
\begin{cases}
 \frac{1-\gamma}{2} < \underline{q} < 1- \gamma, \quad \underline{q} + \frac{3}{4}\Delta \leq \underline{p} \leq  \underline{q} + \Delta \leq \frac{5(1-\gamma)}{4} &\text{if } \ror \in \left(0, \frac{1-\gamma}{4} \right), \\
   \underline{q} =0, \quad \frac{\sigma + 1}{2}\Delta \leq \underline{p} \leq (3+ \ror) \Delta &\text{if } \ror \in \left[ \frac{1-\gamma}{4}, \infty \right).
 \end{cases}
 \end{align}
\end{lemma}

\paragraph{Value functions and optimal policies.}
Armed with above facts, we are positioned to derive the corresponding robust value functions, the optimal policies,  and its corresponding optimal robust value functions. 
For any RMDP $\cM_\phi$ with the uncertainty set defined in \eqref{eq:chi2-ball-infinite-P-recall1}, we denote the robust optimal policy as $\pi^{\star}_\phi$, the robust value function of any policy $\pi$ (resp.~the optimal policy $\pi^{\star}_\phi$) as $V^{\pi,\ror}_\phi$ (resp.~$V^{\star,\ror}_\phi$). 
The following lemma describes some key properties of the robust (optimal) value functions and optimal policies whose proof is postponed to Appendix~\ref{proof:lem:chi2-lb-value}.

\begin{lemma}\label{lem:chi2-lb-value}
For any $\phi = \{0,1\}$ and any policy $\pi$, one has
\begin{align}
    V^{\pi, \ror}_\phi(0) = \frac{\gamma z_{\phi}^{\pi} }{(1-\gamma) \left(1-\gamma \big( 1- z_{\phi}^{\pi}  \big) \right)},
    \label{eq:chi2-infinite-lemma-value-0-pi}
\end{align}
where $z_{\phi}^{\pi}$ is defined as
\begin{align}
z_{\phi}^{\pi} \defn \underline{p}\pi(\phi\mymid 0) + \underline{q} \pi(1-\phi \mymid 0).\label{eq:chi2-infinite-x-h}
\end{align}
In addition, the optimal value functions and the optimal policies obey 
\begin{subequations}
    \label{eq:chi2-value-lemma}
\begin{align}
    V_\phi^{\star,\sigma}(0) &= \frac{\gamma \underline{p} }{(1-\gamma) \left(1-\gamma \left( 1- \underline{p} \right) \right)}, \\
    \pi_\phi^{\star }(\phi \mymid s) &= 1, \qquad  \qquad \text{ for } s \in\cS.
\end{align}
\end{subequations}

\end{lemma}

\subsection{Establishing the minimax lower bound}
Our goal is to control the performance gap w.r.t. any policy estimator $\widehat{\pi}$ based on the generated dataset and the chosen initial distribution $\varphi$ in \eqref{eq:rho-defn-infinite-LB}, which gives
\begin{align}
  \big\langle \varphi, V^{\star, \sigma}_{\phi} - V^{\widehat{\pi}, \sigma}_{\phi} \big\rangle   = V^{\star,\ror}_\phi(0) - V^{\widehat{\pi},\ror}_\phi(0).
\end{align}

\paragraph{Step 1: converting the goal to estimate $\phi$.}
To achieve the goal, we first introduce the following fact which shall be verified in Appendix~\ref{proof:chi2-lower-diff-control}: given 

\begin{align}\label{eq:chi2-varepsilon-bound}
        \varepsilon \leq \frac{1}{768(1-\gamma)},
\end{align}
and choosing 
\begin{align}\label{eq:chi2-Delta-chosen}
 \Delta = \begin{cases}
18 (1-\gamma)^2 \varepsilon \quad &\text{if } \ror \in \left(0, \frac{1-\gamma}{4}\right), \\
  \frac{64\varepsilon(1-\gamma)^2}{3(1+\sigma)}. \quad &\text{if } \ror \in \left[ \frac{1-\gamma}{4}, \infty \right),
\end{cases} 
\end{align}
which satisfies the requirement of $\Delta$ in \eqref{eq:chi2-p-q-defn-infinite2},
it holds that for any policy $\widehat{\pi}$, 
\begin{align}
    \big\langle \varphi, V^{\star, \sigma}_{\phi} - V^{\widehat{\pi}, \sigma}_{\phi} \big\rangle  \geq 2\varepsilon \big(1-\widehat{\pi}(\phi\mymid 0)\big). \label{eq:chi2-Value-0-recursive}
\end{align}

\paragraph{Step 2: arriving at the final results.}
To continue, following the same definitions and argument in Appendix~\ref{sec:tv-lower-bound-final-pipeline}, we recall the minimax probability of the error and its property as follows:
\begin{align}
p_{\mathrm{e}} &  \geq  
\frac{1}{4}\exp\bigg\{- N  \Big( \mathsf{KL}\big(P^0(\cdot\mymid 0, 0)\parallel P^1(\cdot\mymid0,0)\big)+\mathsf{KL}\big(P^0(\cdot\mymid 0,1)\parallel P^1(\cdot\mymid0,1)\big) \Big)\bigg\},
    \label{eq:chi2-finite-remainder-KL}
\end{align}
then we can complete the proof by showing $p_{\mathrm{e}} \geq \frac{1}{8}$ given the bound for the sample size $N$.
In the following, we shall control the KL divergence terms in \eqref{eq:chi2-finite-remainder-KL} in three different cases.
\begin{itemize}
  \item
Case 1: $\sigma \in \left(0, \frac{1-\gamma}{4}\right)$. In this case, applying $\gamma \in[\frac{3}{4}, 1)$ yields  
\begin{align}\label{eq:chi2-1-q-q-bound1}
  1-q &> 1-p  = 1- q - \Delta > \gamma - \frac{1-\gamma}{4} > \frac{3}{4} - \frac{1}{16} > \frac{1}{2}, \notag \\
  p &\geq q = 1-\gamma.
\end{align}

Armed with the above facts, applying \citet[Lemma~2.7]{tsybakov2009introduction} yields
\begin{align}
\mathsf{KL}\big(P^{0}(\cdot \mymid 0, 1)\parallel P^{1}(\cdot \mymid 0, 1)\big) & =\mathsf{KL}\left(p\parallel q \right)  \leq \frac{(p-q)^2}{(1-p)p}  \overset{\mathrm{(i)}}{=} \frac{\Delta^2}{p(1-p)} \notag\\
    & \overset{\mathrm{(ii)}}{=} \frac{ 324 (1-\gamma)^4 \varepsilon^2 }{p(1-p)} \notag \\
    & \overset{\mathrm{(iii)}}{\leq}  648 (1-\gamma)^3 \varepsilon^2,
    \label{eq:chi2-finite-KL-bounded1}
\end{align}
where (i) follows from the definition in \eqref{eq:chi2-p-q-defn-infinite2}, (ii) holds by plugging in the expression of $\Delta$ in \eqref{eq:chi2-Delta-chosen}, and (iii) arises from \eqref{eq:chi2-1-q-q-bound1}. The same bound can be established for $\mathsf{KL}\big(P_1^{0}(\cdot \mymid 0, 0)\parallel P_1^{1}(\cdot \mymid 0, 0)\big)$.
Substituting \eqref{eq:chi2-finite-KL-bounded1} back into \eqref{eq:chi2-finite-remainder-KL} 
demonstrates that: if the sample size is chosen as
\begin{align}\label{eq:chi2-finite-sample-N-condition1}
    N \leq \frac{ \log 2}{ 1296 (1-\gamma)^3  \varepsilon^2},
\end{align}
then one necessarily has
\begin{align}
    p_{\mathrm{e}} &\geq \frac{1}{4}\exp\Big\{- N \cdot 1296 (1-\gamma)^3 \varepsilon^2 \Big\}  \geq \frac{1}{8}. \label{eq:chi2-pe-LB-13579-inf1}
\end{align}

\item Case 2: $\sigma \in \left[ \frac{1-\gamma}{4}, \infty  \right)$. Applying the facts of $\Delta$ in \eqref{eq:chi2-delta-defn-infinite2}, one has 
\begin{align} \label{eq:chi2-1-q-q-bound2}
  1-q &> 1-p  = 1- q - \Delta \geq \frac{1}{1+\sigma} - \frac{1}{2(1+\ror) } = \frac{1}{2(1+\sigma)},\notag \\
  p &\geq q = \frac{\ror}{1+\ror}.
\end{align}

Given  \eqref{eq:chi2-1-q-q-bound2}, applying \citet[Lemma~2.7]{tsybakov2009introduction} yields
\begin{align}
\mathsf{KL}\big(P^{0}(\cdot \mymid 0, 1)\parallel P^{1}(\cdot \mymid 0, 1)\big) & =\mathsf{KL}\left(p\parallel q \right)  \leq \frac{(p-q)^2}{(1-p)p}  \overset{\mathrm{(i)}}{=} \frac{\Delta^2}{p(1-p)} \notag\\
    & \overset{\mathrm{(ii)}}{=} \frac{\frac{4096\varepsilon^2(1-\gamma)^4}{9(1+\sigma)^2} }{p(1-p)} \notag \\
    & \overset{\mathrm{(iii)}}{\leq}  \frac{ \frac{4096\varepsilon^2(1-\gamma)^4}{9(1+\sigma)^2}  }{\frac{\ror}{2(1+\ror)^2}} \leq \frac{8192 (1-\gamma)^4  \varepsilon^2}{\ror},
    \label{eq:chi2-finite-KL-bounded2}
\end{align}
where (i) follows from the definition in \eqref{eq:chi2-p-q-defn-infinite2}, (ii) holds by plugging in the expression of $\Delta$ in \eqref{eq:chi2-Delta-chosen}, and (iii) arises from \eqref{eq:chi2-1-q-q-bound2}.
The same bound can be established for $\mathsf{KL}\big(P_1^{0}(\cdot \mymid 0, 0)\parallel P_1^{1}(\cdot \mymid 0, 0)\big)$.

Substituting \eqref{eq:chi2-finite-KL-bounded2} back into \eqref{eq:finite-remainder-KL} 
demonstrates that: if the sample size is chosen as
\begin{align}\label{eq:chi2-finite-sample-N-condition2}
    N \leq \frac{\ror \log 2}{ 16384 (1-\gamma)^4  \varepsilon^2},
\end{align}
then one necessarily has
\begin{align}
    p_{\mathrm{e}} &\geq \frac{1}{4}\exp\bigg\{-N \frac{16384 (1-\gamma)^4  \varepsilon^2}{\ror}  \bigg\} \geq \frac{1}{8}. \label{eq:chi2-pe-LB-13579-inf2}
\end{align}

\end{itemize}

\paragraph{Step 3: putting things together.} 
Finally, summing up the results in \eqref{eq:chi2-finite-sample-N-condition1} and \eqref{eq:chi2-finite-sample-N-condition2}, combined with the requirement in \eqref{eq:chi2-varepsilon-bound}, one has when 
\begin{align}
\varepsilon &\leq \frac{c_1}{1-\gamma},
\end{align}
taking
\begin{align}
N \leq c_2
\begin{cases}
 \frac{ 1}{  (1-\gamma)^3  \varepsilon^2} &\text{if } \ror \in \left(0, \frac{1-\gamma}{4}\right) \\
\frac{\ror }{ (1-\gamma)^4 \varepsilon^2} &\text{if } \ror \in \left[ \frac{1-\gamma}{4}, \infty \right)
\end{cases}
\end{align}
leads to $p_e \geq \frac{1}{8}$, for some universal constants $c_1, c_2>0$.

\subsection{Proof of the auxiliary facts}

We begin with some basic facts about the $\chi^2$ divergence defined in \eqref{eq:defn-KL-bernoulli} for any two Bernoulli distributions $\mathsf{Ber}(w)$ and $\mathsf{Ber}(x)$, denoted as
\begin{align} \label{eq:chi_sss}
f(w,x) \defn \chi^2( x \parallel w) = \frac{(w-x)^2}{w} + \frac{(1-w-(1-x))^2}{1-w} = \frac{(w-x)^2}{w(1-w)}.
\end{align}
For $x\in [0,w)$, it is easily verified that the partial derivative w.r.t. $x$ obeys $\frac{\partial f(w,x)}{\partial x} = \frac{2(x-w)}{w(1-w)}< 0$, implying that 
\begin{align}\label{eq:monotone-chi2-dis}
 \forall \; x_1 < x_2 \in [0,w), \qquad f(w, x_1) > f(w, x_2).
\end{align}
In other words, the $\chi^2$ divergence $f(w,x)$ increases as $x$ decreases from $w$ to $0$.

Next, we introduce the following function for any fixed $\ror\in(0,\infty)$ and any $x\in \left[ \frac{\ror}{1+\ror}, 1 \right)$:
\begin{align}\label{eq:defn-underline-p}
f_\ror(x) \defn \inf_{\{y: \chi^2(y \parallel x) \leq \ror , y\in[0,x] \} } y \overset{\mathrm{(i)}}{=} \max \left\{ 0, x - \sqrt{\ror x(1-x)} \right\} = x - \sqrt{\ror x(1-x)},
\end{align}
where (i) has been verified in \citet[Corollary~B.2]{yang2021towards}, and the last equality holds since $x\geq  \frac{\ror}{1+\ror}$.
The next lemma summarizes some useful facts about $f_\ror(\cdot)$, which again has been verified in \citet[Lemma~B.12 and Corollary~B.2]{yang2021towards}.  
\begin{lemma}\label{lem:chi2-property}
Consider any $\ror\in (0,\infty)$. For $x\in [\frac{\ror}{1+\ror},1)$, $f_\ror(x)$ is convex and differentiable, which obeys 
\begin{align*}
f_\ror'(x) = 1 + \frac{\sqrt{\ror}(2x-1)}{2\sqrt{x(1-x)}}.
\end{align*}
\end{lemma}

\subsubsection{Proof of Lemma~\ref{eq:chi2-under-p-range}} \label{proof:eq:chi2-under-p-range}

Let us control $\underline{q}$ and $\underline{p}$ respectively.

\paragraph{Step 1: controlling $\underline{q}$.}
We shall control $\underline{q}$ in different cases w.r.t. the uncertainty level $\ror$. 
\begin{itemize}
\item Case 1: $\ror \in \left(0, \frac{1-\gamma}{4} \right)$. In this case, recall that $q = 1-\gamma$ defined in \eqref{eq:chi2-p-q-defn-infinite2}, applying \eqref{eq:defn-underline-p} with $x = q$ leads to
\begin{align}\label{eq:q-min-sigma-small}
 1- \gamma = q > \underline{q} =  f_\ror(q)  = 1-\gamma - \sqrt{\ror \gamma (1-\gamma)} \geq  1-\gamma - \sqrt{\frac{1-\gamma}{4}\gamma (1-\gamma) } > \frac{1-\gamma}{2}.
\end{align}

\item Case 2: $\ror \in \left[ 1, \infty \right)$. Note that it suffices to treat $P^{\phi}_{0,1-\phi}$ as a Bernoulli distribution $\mathsf{Ber}(q)$ over states $1$ and $0$, since we do not allow transition to other states.
Recalling $q =  \frac{\sigma}{1+\sigma}$ in \eqref{eq:chi2-p-q-defn-infinite2} and noticing the fact that 
\begin{align}\label{eq:chi2-fq-0-sigma}
  f( q,0) &= \frac{q^2}{q} + \frac{(1-(1-q))^2}{1-q} = \frac{q}{(1-q)} = \ror,
\end{align}
one has the probability $\mathsf{Ber}(0)$ falls into the uncertainty set of $\mathsf{Ber}(q))$ of size $\ror$. As a result, recalling the definition \eqref{eq:chi2-infinite-lw-p-q-perturb-inf} leads to
\begin{align}\label{eq:chi2-underline-q-0}
  \underline{q} = \underline{P}^{\phi}(1  \mymid 0, 1-\phi) = 0,
\end{align}
since $\underline{q} \geq 0$.

\end{itemize}

\paragraph{Step 2: controlling $\underline{p}$.}
To characterize the value of $\underline{p}$, we also divide into several cases separately.
\begin{itemize}
\item Case 1: $\ror \in \left(0, \frac{1-\gamma}{4} \right)$. In this case, note that $p>q = 1-\gamma \geq \frac{\ror}{1+\ror}$. Therefore, applying that $f_\ror(\cdot)$ is convex and the form of its derivative in Lemma~\ref{lem:chi2-property}, one has
\begin{align}\label{eq:p-min-sigma-small}
  \underline{p} &= f_\ror(p) \geq f_\ror(q) + f_\ror'(q) (p-q) \notag \\
  &= \underline{q} + \Bigg(1 + \frac{\sqrt{\ror}(2q-1)}{2\sqrt{q(1-q)}} \Bigg)\Delta   \geq \underline{q} + \Bigg(1 - \frac{\sqrt{\frac{1-\gamma}{4}}(1-2(1-\gamma)) }{2 \sqrt{(1-\gamma) \gamma}}\Bigg)\Delta \geq \underline{q} + \frac{3\Delta}{4}.
\end{align}
Similarly, applying Lemma~\ref{lem:chi2-property} leads to
\begin{align}\label{eq:p-min-sigma-small-upper}
  \underline{p} &= f_\ror(p) \leq f_\ror(q) + f_\ror'(p) (p-q) \notag \\
  &= \underline{q} + \Bigg(1 - \frac{\sqrt{\ror}(1-2p)}{2\sqrt{p(1-p)}} \Bigg)\Delta   \leq \underline{q} + \Delta,
\end{align}
where the last inequality holds by $1-2p>0$ due to the fact $p = q + \Delta \leq \frac{5}{4}(1-\gamma) \leq \frac{5}{16} <\frac{1}{2}$ (cf. \eqref{eq:chi2-delta-defn-infinite2} and $\gamma \in [\frac{3}{4}, 1)$).
To sum up, given $\ror \in \left(0, \frac{1-\gamma}{4} \right)$, combined with \eqref{eq:q-min-sigma-small}, we arrive at 
\begin{align}\label{eq:p-min-sigma-small-sum}
 \underline{q} + \frac{3}{4}\Delta \leq \underline{p} \leq  \underline{q} + \Delta \leq \frac{5(1-\gamma)}{4},
\end{align}
where the last inequality holds by $\Delta \leq \frac{1}{4}(1-\gamma)$ (see~\eqref{eq:chi2-p-q-defn-infinite2}).

\item Case 2: $\ror \in \left[\frac{1-\gamma}{4}, \infty \right)$. We recall that $p = q + \Delta > q =\frac{\sigma}{1+\sigma}$ in \eqref{eq:chi2-p-q-defn-infinite2}. To derive the lower bound for $\underline{p}$ in \eqref{eq:chi2-infinite-lw-p-q-perturb-inf}, similar to \eqref{eq:p-min-sigma-small}, one has
\begin{align}\label{eq:p-min-sigma-large}
  \underline{p} &= f_\ror(p) \geq f_\ror(q) + f_\ror'(q) (p-q) \notag \\
  &= \underline{q} + \left(1 + \frac{\sqrt{\ror}(2q-1)}{2\sqrt{q(1-q)}} \right)\Delta  \notag \\
  &\overset{\mathrm{(i)}}{=}  0 + \left(1 + \frac{\sqrt{\ror} \frac{\ror -1}{1+\ror} }{2 \sqrt{\frac{\ror }{1+\ror} \frac{1}{1+\ror}  }}\right)\Delta = \left(1 + \frac{\sigma - 1}{2} \right) \Delta  = \left( \frac{\sigma + 1}{2} \right) \Delta,
\end{align}
where (i) follows from $q =\frac{\sigma}{1+\sigma}$ and  $\underline{q} = 0$ (see \eqref{eq:chi2-underline-q-0}). For the other direction,
similar to \eqref{eq:p-min-sigma-small-upper}, we have
\begin{align}
  \underline{p} &= f_\ror(p) \leq f_\ror(q) + f_\ror'(p) (p-q) = \underline{q} + \left(1 + \frac{\sqrt{\ror}(2p-1)}{2\sqrt{p(1-p)}} \right)\Delta \notag \\
  & \overset{\mathrm{(i)}}{=}   \left(1 + \frac{\sqrt{\ror}(2p-1)}{2\sqrt{p(1-p)}} \right)\Delta 
  \overset{\mathrm{(ii)}}{=}   \left(1 + \frac{\sqrt{\ror} \left( \frac{\ror-1}{1+\ror} + 2\Delta \right)}{2\sqrt{ \left(\frac{\ror}{1+\ror} + \Delta\right) \left(\frac{1}{1+\ror} - \Delta\right) }} \right)\Delta \notag \\
  &\overset{\mathrm{(iii)}}{\leq} \left(1 + \frac{\sqrt{\ror} (1+ 2\Delta) }{2 \sqrt{ \frac{\ror}{1+\ror} \cdot \frac{1}{2(1+\ror)}  }} \right)\Delta \overset{\mathrm{(iv)}}{\leq} \left(1 + (1+\ror) \left( 1 + \frac{1}{1+\ror}\right)\right)\Delta = (3+ \ror) \Delta, \label{eq:p-min-sigma-large-upper-region-all}
\end{align}
where (i) holds by $\underline{q} = 0$ (see \eqref{eq:chi2-underline-q-0}), (ii) follows from plugging in $p = q + \Delta  = \frac{\sigma}{1+\sigma} + \Delta $, and (iii) and (iv) arises from $\Delta = \min \left\{ \frac{1}{4}(1-\gamma), \frac{1}{2(1+\ror)}\right\} \leq 1$ in \eqref{eq:chi2-delta-defn-infinite2}. Combining \eqref{eq:p-min-sigma-large} and \eqref{eq:p-min-sigma-large-upper-region-all} yields
\begin{align}\label{eq:p-min-sigma-large-sum}
  \frac{\ror+1}{2} \Delta \leq \underline{p} \leq (3+ \ror) \Delta.
\end{align}

\end{itemize}

\paragraph{Step 3: combining all the results.}
Finally, summing up the results for both $\underline{q}$ (in \eqref{eq:q-min-sigma-small} and \eqref{eq:chi2-underline-q-0}) and $\underline{p}$ (in \eqref{eq:p-min-sigma-small-sum} and \eqref{eq:p-min-sigma-large-sum}), we arrive at the advertised bound.

\subsubsection{Proof of Lemma~\ref{lem:chi2-lb-value} } \label{proof:lem:chi2-lb-value}

\paragraph{The robust value function for any policy $\pi$.}
For any $\cM_\phi$ with $\phi\in\{0,1 \}$, we first characterize the robust value function of any policy $\pi$ over different states. 

Towards this, it is easily observed that for any policy $\pi$, the robust value functions at state $s=1$ or any $s\in\{2,3,\cdots, S-1\}$ obey 
\begin{subequations}\label{eq:chi2-s-value1}
\begin{align}
  &V_\phi^{\pi,\ror}(1) \overset{\mathrm{(i)}}{=} 1 + \gamma V_\phi^{\pi,\ror}(1) = \frac{1}{1-\gamma} \label{eq:chi2-s-value1-s1}
  \end{align} 
  and
  \begin{align}
  \forall s\in\{2,3,\cdots, S\}: \qquad   V_\phi^{\pi,\ror}(s) &\overset{\mathrm{(ii)}}{=} 0 + \gamma V_\phi^{\pi,\ror}(1) = \frac{\gamma}{1-\gamma},  \label{eq:chi2-s-value1-smore}
\end{align}
\end{subequations}
where (i) and (ii) is according to the facts that the transitions defined over states $s\geq 1$ in \eqref{eq:Ph-construction-lower-infinite} give only one possible next state $1$, leading to a non-random transition in the uncertainty set associated with $\chi^2$ divergence, and $r(1,a)=1$ for all $a\in\cA'$ and $r(s,a)=0$ holds all $(s,a) \in\{2,3,\cdots, S-1\} \times \cA$.

To continue, the robust value function at state $0$ with policy $\pi$ satisfies
\begin{align}
 V_\phi^{\pi,\ror}(0) &= \mathbb{E}_{a \sim \pi(\cdot \mymid 0)} \bigg[ r(0,a) + \gamma \inf_{ \cP \in \unb^{\sigma}(P^{\phi}_{0,a})}  \cP V^{\pi,\sigma}_\phi\bigg] \nonumber \\
  & = 0 +  \gamma\pi(\phi \mymid 0) \inf_{ \cP \in \unb^{\sigma}(P^{\phi}_{0,\phi})}  \cP V^{\pi,\sigma}_\phi +  \gamma\pi(1 - \phi \mymid 0)  \inf_{ \cP \in \unb^{\sigma}(P^{\phi}_{0, 1- \phi})}  \cP V^{\pi,\sigma}_{\phi} \label{eq:chi2-s0-value-def}  \\
  & \overset{\mathrm{(i)}}{\leq} \frac{\gamma}{1-\gamma} ,\label{eq:chi2-s0<s1}
\end{align}
where (i) holds by that $\|V^{\pi,\sigma}_{\phi}\|_\infty \leq \frac{1}{1 - \gamma}$.
Summing up the results in \eqref{eq:chi2-s-value1-smore} and \eqref{eq:chi2-s0<s1} leads to
\begin{align}
\forall s\in\{2,3,\cdots, S\}, \qquad V_\phi^{\pi,\ror}(1) > V_\phi^{\pi,\ror}(s) \geq V_\phi^{\pi,\ror}(0). \label{eq:chi2-value-state-order}
\end{align}
With the transition kernel in \eqref{eq:Ph-construction-lower-infinite} over state $0$ and the fact in \eqref{eq:chi2-value-state-order}, \eqref{eq:chi2-s0-value-def} can be rewritten as
\begin{align}
    V_\phi^{\pi,\ror}(0) &= \gamma\pi(\phi \mymid 0) \inf_{ \cP \in \unb^{\sigma}(P^{\phi}_{0,\phi})}  \cP V^{\pi,\sigma}_\phi +  \gamma\pi(1 - \phi \mymid 0)  \inf_{ \cP \in \unb^{\sigma}(P^{\phi}_{0, 1- \phi})}  \cP V^{\pi,\sigma}_{\phi} \nonumber\\
    & \overset{\mathrm{(i)}}{=} \gamma \pi(\phi \mymid 0)\Big[ \underline{p} V_\phi^{\pi,\sigma}(1) + \left(1- \underline{p}\right) V_\phi^{\pi,\sigma}(0) \Big]  +  \gamma \pi(1-\phi \mymid 0)\Big[ \underline{q} V_{\phi}^{\pi,\sigma}(1) + \left(1-\underline{q} \right) V_{\phi}^{\pi,\sigma}(0) \Big] \nonumber \\
    & \overset{\mathrm{(ii)}}{=} \gamma z_{\phi}^{\pi}  V_{\phi}^{\pi,\sigma}(1) + \gamma \left( 1- z_{\phi}^{\pi}  \right)V_{\phi}^{\pi,\sigma}(0) \nonumber \\
    & = \frac{\gamma z_{\phi}^{\pi} }{(1-\gamma) \Big(1-\gamma \big( 1- z_{\phi}^{\pi}  \big) \Big)},\label{eq:chi2-value-s0}
\end{align}
where (i) holds by the definition of $\underline{p}$ and $\underline{q}$ in \eqref{eq:chi2-infinite-lw-p-q-perturb-inf}, (ii) follows from the definition of $z_{\phi}^{\pi}$ in \eqref{eq:chi2-infinite-x-h}, and the last line holds by applying \eqref{eq:chi2-s-value1-s1} and solving the resulting linear equation for $V_{\phi}^{\pi,\sigma}(0)$.

\paragraph{Optimal policy and its optimal value function.} To continue, observing that $ V_\phi^{\pi,\ror}(0) =: f(z_{\phi}^{\pi})$ is increasing in $z_{\phi}^{\pi}$ since
 the derivative of $f(z_{\phi}^{\pi}) $ w.r.t. $z_{\phi}^{\pi}$ obeys
\begin{align*}
  f'(z_{\phi}^{\pi}) = \frac{\gamma (1-\gamma) \left(1-\gamma \big( 1- z_{\phi}^{\pi}  \big) \right) - \gamma^2  z_{\phi}^{\pi} (1-\gamma)}{(1-\gamma)^2 \left(1-\gamma \big( 1- z_{\phi}^{\pi}  \big) \right)^2} = \frac{\gamma}{\left(1-\gamma \big( 1- z_{\phi}^{\pi}  \big) \right)^2} > 0,
\end{align*} 
where the last inequality holds by $ 0 \leq z_{\phi}^{\pi} \leq 1$.
Further,
 $z_{\phi}^{\pi}$ is also increasing in $\pi(\phi \mymid 0)$ (see the fact $\underline{p}\geq \underline{q}$ in \eqref{eq:chi2-infinite-lw-p-q-perturb-inf}),  the optimal robust policy in state $0$ thus obeys
\begin{equation}
    \pi_\phi^{\star}(\phi \mymid 0) = 1 \label{eq:chi2-infinite-lb-optimal-policy}.
\end{equation}
Considering that the action does not influence the state transition for all states $s>0$, without loss of generality, we choose the optimal robust policy to obey
\begin{align}\label{eq:infinite-lower-optimal-pi}
    \forall s>0: \quad \pi_\phi^\star(\phi\mymid s) = 1.
\end{align}

Taking $\pi = \pi^\star_\phi$ and $z_{\phi}^{\pi^\star_\phi} = \underline{p}$ in \eqref{eq:chi2-value-s0}, we complete the proof by showing the corresponding optimal robust value function at state $0$ as follows:
\begin{align*}
   V_\phi^{\star,\ror}(0) & = \frac{\gamma z_{\phi}^{\pi^\star_\phi} }{(1-\gamma) \left(1-\gamma \left( 1- z_{\phi}^{\pi^\star_\phi}  \right) \right)} = \frac{\gamma \underline{p} }{(1-\gamma) \left(1-\gamma \left( 1- \underline{p} \right) \right)}.
\end{align*}

\subsubsection{Proof of the claim~\eqref{eq:chi2-Value-0-recursive}}\label{proof:chi2-lower-diff-control}

Plugging in the definition of $\varphi$, we arrive at that for any policy $\pi$,
\begin{align}
\big\langle \varphi, V^{\star, \sigma}_{\phi} - V^{\pi, \sigma}_{\phi} \big\rangle  &= V^{\star, \sigma}_{\phi}(0) - V^{\pi, \sigma}_{\phi}(0) \nonumber \\
&  \overset{\mathrm{(i)}}{=} \frac{\gamma \underline{p} }{(1-\gamma) \left(1-\gamma \big( 1- \underline{p} \big) \right)} - \frac{\gamma z_{\phi}^{\pi} }{(1-\gamma) \left(1-\gamma \big( 1- z_{\phi}^{\pi}  \big) \right)} \nonumber \\
& = \frac{\gamma \left( \underline{p} - z_{\phi}^{\pi}\right)}{\left(1-\gamma \big( 1- \underline{p} \big) \right)  \left(1-\gamma \big( 1- z_{\phi}^{\pi}  \big) \right)} \overset{\mathrm{(ii)}}{\geq} \frac{\gamma \left( \underline{p} - z_{\phi}^{\pi}\right)}{\left(1-\gamma \left( 1- \underline{p} \right) \right)^2} \overset{\mathrm{(iii)}}{=} \frac{\gamma (\underline{p}  -\underline{q} )\big(1-\pi(\phi\mymid 0)\big) }{\left(1-\gamma \left( 1- \underline{p} \right) \right)^2}, \label{eq:chi2-value-gap}
\end{align}
where (i) holds by applying Lemma~\ref{lem:chi2-lb-value}, (ii) arises from $z_{\phi}^\pi \leq \underline{p}$ (see the definition of $z_{\phi}^\pi$ in \eqref{eq:chi2-infinite-x-h} and the fact $\underline{p}\geq \underline{q} + \frac{3\Delta}{4}$ in \eqref{eq:chi2-infinite-lw-p-q-perturb-inf}), and (iii) follows from the definition of $z_{\phi}^{\pi}$ in \eqref{eq:chi2-infinite-x-h}.

To further control \eqref{eq:chi2-value-gap}, we consider it in two cases separately:
\begin{itemize}
  \item Case 1: $\ror \in \left(0, \frac{1-\gamma}{4} \right)$. In this case, applying Lemma~\ref{eq:chi2-under-p-range} to  \eqref{eq:chi2-value-gap} yields
  \begin{align}
  \big\langle \varphi, V^{\star, \sigma}_{\phi} - V^{\pi, \sigma}_{\phi} \big\rangle  & \geq \frac{\gamma (\underline{p}  -\underline{q} )\big(1-\pi(\phi\mymid 0)\big) }{\left(1-\gamma \left( 1- \underline{p} \right) \right)^2}  \geq  \frac{\gamma \frac{3\Delta}{4}  \big(1-\pi(\phi\mymid 0)\big) }{\left(1-\gamma \left( 1- \frac{5(1-\gamma)}{4} \right) \right)^2} \notag \\
  &\geq  \frac{\Delta \big(1-\pi(\phi\mymid 0)\big)}{9(1-\gamma)^2} = 2\varepsilon \big(1- {\pi}(\phi\mymid 0)\big), \label{eq:chi2-gap-delta}
\end{align}
where the penultimate inequality follows from $\gamma \geq 3/4$, and the last inequality holds by taking the specification of $\Delta$ in \eqref{eq:chi2-Delta-chosen} as follows:
\begin{align}
\Delta = 18 (1-\gamma)^2 \varepsilon.
\end{align}
  It is easily verified that taking
$        \varepsilon \leq \frac{1}{72(1-\gamma)}  $
as in \eqref{eq:chi2-varepsilon-bound} directly leads to meeting the requirement in \eqref{eq:chi2-delta-defn-infinite2}, i.e.,
$\Delta \leq  \frac{1}{4}(1-\gamma)$.

  \item Case 2: $\ror \in \left[ \frac{1-\gamma}{4}, \infty \right)$. Similarly, applying Lemma~\ref{eq:chi2-under-p-range} to  \eqref{eq:chi2-value-gap} gives
  \begin{align}
  \big\langle \varphi, V^{\star, \sigma}_{\phi} - V^{\pi, \sigma}_{\phi} \big\rangle  & \geq \frac{\gamma (\underline{p}  -\underline{q} )\big(1-\pi(\phi\mymid 0)\big) }{\left(1-\gamma \left( 1- \underline{p} \right) \right)^2 }  \geq  \frac{\gamma \frac{\sigma+1}{2}\Delta \big(1-\pi(\phi\mymid 0)\big) }{ \min\left\{ 1, \left(1-\gamma \left( 1- (3+\ror) \Delta \right) \right)^2 \right\} } \label{eq:chi2-gap-delta-large}
\end{align}
Before continuing, it can be verified that
  \begin{align}
  1-\gamma \left( 1- (3+\ror)\Delta \right) &= 1-\gamma + \gamma (3+\ror) \Delta \overset{\mathrm{(i)}}{\leq} 2(1-\gamma),\label{eq:chi2-gap-delta-large-middle}
\end{align}
where (i) is obtained by $\Delta \leq  \frac{1-\gamma}{\gamma(3+\sigma)}$ (see~\eqref{eq:chi2-delta-defn-infinite2}). Applying the above fact to \eqref{eq:chi2-gap-delta-large} gives
    \begin{align}
    \big\langle \varphi, V^{\star, \sigma}_{\phi} - V^{\pi, \sigma}_{\phi} \big\rangle  & \geq \frac{\gamma \frac{\sigma+1}{2}\Delta \big(1-\pi(\phi\mymid 0)\big) }{ \min \left\{1,  \left(1-\gamma \left( 1- (3+\ror) \Delta \right) \right)^2 \right\} }  \overset{\mathrm{(i)}}{\geq} \frac{3(\ror+1)\Delta \big(1-\pi(\phi\mymid 0)\big)}{32(1-\gamma)^2} \notag \\
    &   = 2\varepsilon \big(1- {\pi}(\phi\mymid 0)\big),
    \end{align}
where (i) holds by $\gamma \geq \frac{3}{4}$ and \eqref{eq:chi2-gap-delta-large}, and the last equality holds by the specification in \eqref{eq:chi2-Delta-chosen}:
\begin{align}
\Delta = \frac{64\varepsilon(1-\gamma)^2}{3(1+\sigma)}.
\end{align}
As a result, it is easily verified that  the requirement in \eqref{eq:chi2-delta-defn-infinite2} \begin{align}
\Delta  \leq \frac{1-\gamma}{\gamma(3+\sigma)}
\end{align}
is met if we let 
\begin{align}
        \varepsilon \leq \frac{1}{768(1-\gamma)}
\end{align}
as in \eqref{eq:chi2-varepsilon-bound}.

\end{itemize}
The proof is then completed by summing up the results in the above two cases.

\section{Proof for the offline setting}\label{proof:corollary}

 \subsection{Proof of the upper bounds: Corollary~\ref{thm:l1-upper-bound-offline} and Corollary~\ref{thm:l2-upper-bound-offline}}\label{proof:offline-upper}
 
As the proofs of Corollary~\ref{thm:l1-upper-bound-offline} and Corollary~\ref{thm:l2-upper-bound-offline} are similar, without loss of generality, we first focus on Corollary~\ref{thm:l1-upper-bound-offline} in the case of TV distance.

To begin with, suppose we have access to in total $N_{\mathsf{b}}$ independent sample tuples $\{s_i, a_i, a_i',r_i\}_{i=1}^{N_{\mathsf{b}}}$ from either the generative model or a historical dataset. We denote the number of samples generated based on the state-action pair $(s,a)$ as $N(s,a)$, i.e,
\begin{align}
	\forall (s,a)\in\cS\times \cA: \quad N(s,a) =  \sum\limits_{i=1}^{N_{\mathsf{b}}} \mathds{1} \big\{ s_i= s, a_i = a \big\}.
\end{align}
Then according to \eqref{eq:empirical-P-infinite}, we can construct an empirical nominal transition for \DRVI (Algorithm~\ref{alg:cvi-dro-infinite}).
\begin{align}
\forall (s,a)\in \cS\times \cA:\quad \widehat{P}^0(s'\mymid s,a) \defn  \frac{1}{N(s,a)} \sum\limits_{i=1}^{N(s,a)} \mathds{1} \big\{  s_i= s, a_i = a ,s_i'= s' \big\}. \label{eq:empirical-P-infinite-general}
\end{align}

Armed with the above estimate of nominal transition kernel, we introduce a slightly general version of Theorem~\ref{thm:l1-upper-bound}, which follows directly from the same proof routine in Appendix~\ref{proof:thm:l1-upper-bound}.

\begin{theorem}[Upper bound under TV distance]\label{thm:l1-upper-bound-general} 
	Let the uncertainty set be $\unb_\rho^\ror(\cdot) = \cU^{\ror}_{\mathsf{TV}}(\cdot)$, as specified by the TV distance \eqref{eq:tv-distance}. Consider any discount factor $\gamma \in \left[\frac{1}{4},1 \right)$, 
 uncertainty level $\ror\in (0,1)$, and $\delta \in (0,1)$. 
	Based on the empirical nominal transition kernel in \eqref{eq:empirical-P-infinite-general}, let $\widehat{\pi}$ be the output policy of Algorithm~\ref{alg:cvi-dro-infinite} after $T = C_1 \log \big( \frac{N_{\mathsf{b}}}{1-\gamma}\big)$ iterations. 
	Then with probability at least $1-\delta$, one has
\begin{align}
	\forall s\in\cS: \quad V^{\star, \ror}(s) - V^{\widehat{\pi}, \ror}(s) \leq \varepsilon
\end{align}
for any $\varepsilon \in \left(0, \sqrt{1/\max\{1-\gamma, \ror\}} \right]$,
as long as 
\begin{align}
	\forall (s,a) \in \cS\times \cA: \quad N(s,a) \geq  \frac{C_2 }{ (1-\gamma)^2 \max\{1-\gamma, \ror\} \varepsilon^2}\log\left(\frac{SAN_{\mathsf{b}}}{(1-\gamma)\delta}\right).
\end{align}
Here, $C_1, C_2>0$ are some large enough universal constants.  
\end{theorem}

Furthermore, we invoke a  fact derived from basic concentration inequalities \citep{li2024settling} as below.
\begin{lemma} 
	\label{lem:sample-split-infinite}
Consider any $\delta \in (0,1)$ and a dataset with $N_{\mathsf{b}}$ independent samples satisfying Assumption~\ref{assumption-offline}. With probability at least $1-\delta$, the quantities $\{N(s,a)\}$ obey
\begin{align}
	\max\Big\{N(s, a), \frac{2}{3}\infs \log\frac{N_{\mathsf{b}} }{\delta}\Big\} &\ge \frac{ N_{\mathsf{b}} \mu^{\mathsf{b}}(s, a) }{12} 
	\label{eq:samples-infinite-LB}
\end{align}
simultaneously for all $(s, a) \in \cS \times \cA$. 
\end{lemma}

Now we are ready to verify Corollary~\ref{thm:l1-upper-bound-offline}. Armed with a historical dataset $\cD^{\mathsf{b}}$ with $N_{\mathsf{b}}$ independent samples that obeys Assumption~\ref{assumption-offline}, one has with probability at least $1-\delta$,
 \begin{align}
 \forall (s,a)\in \cS\times \cA: \quad N(s,a) \geq \frac{N_{\mathsf{b}} \mu^{\mathsf{b}}(s,a)}{12} \geq \frac{N_{\mathsf{b}} \mu_{\min}}{12} \label{eq:n-sa-all-satisfied}
 \end{align}
 as long as $N_{\mathsf{b}} \geq \frac{8\log\frac{N_{\mathsf{b}} }{\delta}}{\mu_{\min}}  \geq \frac{8\log\frac{N_{\mathsf{b}} }{\delta}}{\mu^{\mathsf{b}}(s,a)}$ for all $(s,a)\in\cS \times \cA$.
Consequently, given $N_{\mathsf{b}} \geq \frac{8\log\frac{N_{\mathsf{b}} }{\delta}}{\mu_{\min}}$, applying Theorem~\ref{thm:l1-upper-bound-general} with the fact $N(s,a) \geq \frac{N_{\mathsf{b}} \mu_{\min}}{12}$ for all $(s,a)\in\cS\times \cA$ (see \eqref{eq:n-sa-all-satisfied}) directly leads to: \DRVI can achieve an $\varepsilon$-optimal policy as long as \begin{align}
   N(s,a) \geq \frac{N_{\mathsf{b}} \mu_{\min}}{12} \geq  \frac{C_2 }{ (1-\gamma)^2 \max\{1-\gamma, \ror\} \varepsilon^2}\log\left(\frac{SAN_{\mathsf{b}}}{(1-\gamma)\delta}\right),
\end{align}
namely
\begin{align}
 N_{\mathsf{b}}  \geq  \frac{C_3 }{\mu_{\min} (1-\gamma)^2 \max\{1-\gamma, \ror\} \varepsilon^2}\log\left(\frac{SAN_{\mathsf{b}}}{(1-\gamma)\delta}\right),
\end{align}
where $C_3$ is some large enough universal constant.
Note that the above inequality directly implies $N_{\mathsf{b}} \geq \frac{8\log\frac{N_{\mathsf{b}} }{\delta}}{\mu_{\min}}$.
This complete the proof of Corollary~\ref{thm:l1-upper-bound-offline}. The same argument holds for Corollary~\ref{thm:l2-upper-bound-offline}.

 \subsection{Proof of the lower bounds: Corollary~\ref{thm:l1-lower-bound-offline} and Corollary~\ref{thm:chi2-lower-bound-offline}}
Analogous to Appendix~\ref{proof:offline-upper}, without loss of generality, we firstly focus on verifying Corollary~\ref{thm:l1-lower-bound-offline}, where we use the TV distance to measure the uncertainty set.

We stick to the two hard instances $\cM_0$ and $\cM_1$ (i.e., $\cM_{\phi}$ with $\phi \in\{ 0, 1\}$) constructed in the proof for Theorem~\ref{thm:l1-lower-bound} (Appendix~\ref{sec:tv-lower-bound-instance}). Recall that the state space is defined as $\cS = \{0,1,2,\cdots, S-1\}$, where the corresponding action space for any state $s\in\{2,3,\cdots, S-1\}$ is $\mathcal{A} = \{0, 1, 2, \cdots, A-1\}$. For states $s=0$ or $s=1$, the action space is only $\cA' =\{0,1\}$. Hence, for a given factor $\mu_{\min} \in (0, \frac{1}{SA}]$, we can construct a historical dataset $\cD^{\mathsf{b}}$ with $N_{\mathsf{b}}$ samples such that the data coverage becomes the smallest over the state-action pairs $(0,0)$ and $(0,1)$, i.e.,
\begin{align}
\mu^{\mathsf{b}}(0,0) = \mu^{\mathsf{b}}(0,1) = \mu_{\min} \quad \text{and} \quad \mu^{\mathsf{b}}(s,a) = \frac{1 -2\mu_{\min} }{(S-2)A + 2},  \quad \forall s\in \{1,2,\cdots, S\}.
\end{align}

Armed with the above hard instance and historical dataset, we follow the proof procedure in Appendix~\ref{sec:tv-lower-bound-final-pipeline} to verify the corollary. Our goal is to distinguish between the two hypotheses $\phi \in\{ 0, 1\}$ by considering the minimax probability of error as follows:
\begin{equation}
    p_{\mathrm{e}} \coloneqq \inf_{\psi}\max \big\{ \mathbb{P}_{0}(\psi \neq 0), \, \mathbb{P}_{1}(\psi \neq 1) \big\}, \label{eq:error-prob-two-hypotheses-finite-LB-corollary}
\end{equation}
where the infimum is taken over all possible tests $\psi$ constructed from the samples in $\cD^{\mathsf{b}}$.

Recall that we denote $\mu_{\phi}$ (resp.~$\mu_{\phi}(s)$) as the distribution of a sample tuple $(s_i, a_i, s_i')$ under the nominal transition kernel $P^\phi$ associated with $\mathcal{M}_\phi$ and the samples are generated independently. Analogous to \eqref{eq:finite-remainder-KL}, one has
\begin{align}
p_{\mathrm{e}} &  \geq \frac{1}{4}\exp\Big(- N_{\mathsf{b}} \mathsf{KL} \big(\mu_{0}\parallel \mu_{1} \big) \Big)\nonumber\\
    & = \frac{1}{4}\exp\Big\{- N_{\mathsf{b}} \mu_{\min} \Big( \mathsf{KL}\big(P^0(\cdot\mymid 0, 0)\parallel P^1(\cdot\mymid0,0)\big)+\mathsf{KL}\big(P^0(\cdot\mymid 0,1)\parallel P^1(\cdot\mymid0,1)\big) \Big)\Big\},
    \label{eq:finite-remainder-KL-corollary}
\end{align}
where the last inequality holds by observing that 
\begin{align}
\mathsf{KL} \big(\mu_{0}\parallel \mu_{1} \big) & = \sum_{s,a,s'} \mu^{\mathsf{b}}(s,a) \mathsf{KL}\big(P^0(s' \mymid s, a)\parallel P^1( s' \mymid s, a)\big)   \notag \\
    & =   \sum_{a\in\{0,1\}} \mu^{\mathsf{b}}(0,a) \mathsf{KL}\big(P^{0}(\cdot\mymid0,a)\parallel P^{1}(\cdot\mymid0,a)\big) = \mu_{\min} \sum_{a\in\{0,1\}}  \mathsf{KL}\big(P^{0}(\cdot\mymid0,a)\parallel P^{1}(\cdot\mymid0,a)\big).
\end{align}
Here, the last line holds by the fact that $P^{0}(\cdot\mymid s,a)$ and $P^{1}(\cdot\mymid s,a)$ (associated with $\cM_0$ and $\cM_1$) only differ from each other in state-action pairs $(0,0)$ and $(0,1)$, each has a visitation density of $\mu_{\min}$. Consequently, following the same routine from \eqref{eq:finite-KL-bounded} to the end of Appendix~\ref{sec:tv-lower-bound-final-pipeline}, we applying \eqref{eq:finite-sample-N-condition} and \eqref{eq:pe-LB-13579-inf}
 with $N = N_{\mathsf{b}}\mu_{\min}$ and complete the proof by showing: if the sample size is selected as
\begin{align}\label{eq:finite-sample-N-condition-offline}
    N_{\mathsf{b}}\mu_{\min} = N \leq \frac{c_1 \log 2}{ 8192 (1-\gamma)^2 \max\{ 1 -\gamma, \sigma\} \varepsilon^2},
\end{align}
then one necessarily has 
\begin{align}
p_e = \inf_{\widehat{\pi}}\max\left\{ \mathbb{P}_{0}\big( V^{\star,\ror}(\varphi)-V^{\widehat{\pi}, \ror}(\varphi)>\varepsilon\big), \,
	\mathbb{P}_{1}\big( V^{\star,\ror }(\varphi) - V^{\widehat{\pi},\ror}(\varphi) >\varepsilon\big)\right\} \geq\frac{1}{8}.
	\end{align}

We can follow the same argument to complete the proof of Corollary~\ref{thm:chi2-lower-bound-offline}.

\end{document}